\documentclass[10pt,twocolumn,letterpaper]{article}

\usepackage{cvpr}
\usepackage{times}
\usepackage{epsfig}
\usepackage{graphicx}
\usepackage{amsmath}
\usepackage{amssymb}

\usepackage{microtype}

\usepackage{booktabs}
\usepackage[table]{xcolor}

\usepackage[pagebackref=true,breaklinks=true,letterpaper=true,colorlinks,bookmarks=false]{hyperref}

\cvprfinalcopy 

\def\cvprPaperID{1588} 

\newcommand{\bm}[1]{\boldsymbol{#1}}

\newcommand{\dv}{\ensuremath{\bm{d}}}
\newcommand{\gv}{\ensuremath{\bm{g}}}
\newcommand{\wv}{\ensuremath{\bm{w}}}
\newcommand{\xv}{\ensuremath{\bm{x}}}
\newcommand{\yv}{\ensuremath{\bm{y}}}

\newtheorem{lemma-ap}{Lemma}

\newtheorem{claim-ap}{Claim}

\makeatletter
\usepackage{xspace}
\def\@onedot{\ifx\@let@token.\else.\null\fi\xspace}
\DeclareRobustCommand\onedot{\futurelet\@let@token\@onedot}

\newcommand{\figref}[1]{Fig\onedot~\ref{#1}}
\newcommand{\equref}[1]{Eq\onedot~\eqref{#1}}
\newcommand{\secref}[1]{Sec\onedot~\ref{#1}}
\newcommand{\tabref}[1]{Tab\onedot~\ref{#1}}

\newcommand{\by}[2]{\ensuremath{#1 \! \times \! #2}}
\newcommand{\norm}[1]{\ensuremath{\lVert#1\rVert}}

\def\eg{\emph{e.g}\onedot} 
\def\ie{\emph{i.e}\onedot} 
 
 \def\vs{\emph{vs}\onedot}
\def\wrt{w.r.t\onedot} 
\def\etal{\emph{et al}\onedot}

\newcommand{\eat}[1]{}

\makeatother

\begin{document}

\title{Semantic Image Segmentation with Task-Specific Edge Detection Using CNNs and a Discriminatively Trained Domain Transform}

\author{
\begin{tabular}[t]{c c c}
Liang-Chieh Chen\footnotemark & Jonathan T. Barron, George Papandreou, Kevin Murphy & Alan L. Yuille \\
lcchen@cs.ucla.edu & \{barron, gpapan, kpmurphy\}@google.com  & yuille@stat.ucla.edu \\
& & alan.yuille@jhu.edu \\
\end{tabular}
}


\maketitle

\footnotetext{$^*$Work done in part during an internship at Google Inc.}


\begin{abstract}
Deep convolutional neural networks (CNNs) are the backbone of
state-of-art semantic image segmentation systems. Recent work has
shown that complementing CNNs with fully-connected conditional random
fields (CRFs) can significantly enhance their object localization
accuracy, yet dense CRF inference is computationally expensive. We
propose replacing the fully-connected CRF with domain transform (DT),
a modern edge-preserving filtering method in which the amount of
smoothing is controlled by a reference edge map. Domain transform
filtering is several times faster than dense CRF inference and we show
that it yields comparable semantic segmentation results, accurately
capturing object boundaries. Importantly, our formulation allows
learning the reference edge map from intermediate CNN features
instead of using the image gradient magnitude as in standard DT
filtering. This produces task-specific edges in an end-to-end
trainable system optimizing the target semantic segmentation quality.
\end{abstract}

\section{Introduction}

Deep convolutional neural networks (CNNs) are very effective in semantic
image segmentation, the task of assigning a semantic label to every
pixel in an image. Recently, it has been demonstrated that
post-processing the output of a CNN with a fully-connected CRF can
significantly increase segmentation accuracy near object
boundaries \cite{chen2014semantic}.

As explained in \cite{krahenbuhl2011efficient},
mean-field inference in the fully-connected CRF model amounts to
iterated application of the bilateral filter, a popular technique for
edge-aware filtering. This encourages pixels which are nearby in
position and in color to be assigned the same semantic label. In
practice, this produces semantic segmentation results which are well
aligned with object boundaries in the image.

One key impediment in adopting the fully-connected CRF is the rather
high computational cost of the underlying bilateral filtering
step. Bilateral filtering amounts to high-dimensional Gaussian
filtering in the 5-D bilateral (2-D position, 3-D color) space and is
expensive in terms of both memory and CPU time, even when advanced
algorithmic techniques are used.

In this paper, we propose replacing the fully-connected CRF and
its associated bilateral filtering with the domain
transform (DT) \cite{GastalOliveira2011DomainTransform}, an alternative edge-aware filter.
The recursive formulation of the domain transform amounts to adaptive
recursive filtering of a signal, where information is not
allowed to propagate across edges in some reference signal.
This results in an extremely efficient
scheme which is an order of magnitude faster than the fastest
algorithms for a bilateral filter of equivalent quality.

\begin{figure}
  \centering
  \includegraphics[width=0.98\columnwidth]{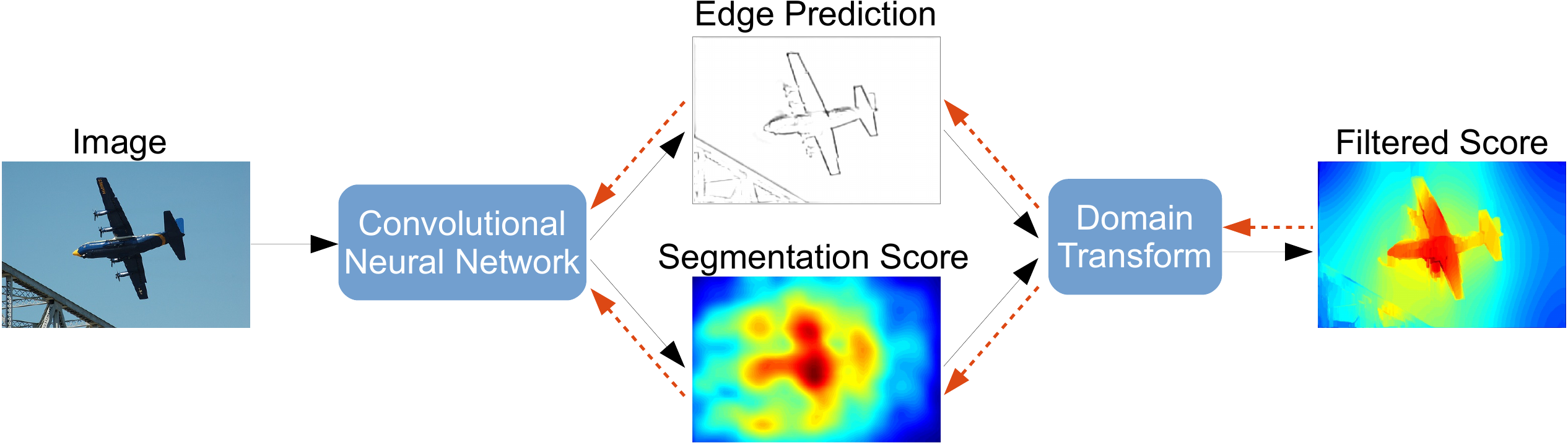}
  \caption{A single unified CNN produces both coarse semantic
    segmentation scores and an edge map, which respectively serve
    as input multi-channel image and reference edge to a domain
    transform edge-preserving filter. The resulting filtered semantic
    segmentation scores are well-aligned with the object
    boundaries. The full architecture is discriminatively trained by
    backpropagation (red dashed arrows) to optimize the target
    semantic segmentation.}
  \label{model_illustration}
\end{figure}

The domain transform can equivalently be seen as a recurrent neural
network (RNN). In particular, we show that the domain transform is a
special case of the recently proposed RNN with gated recurrent
units. This connection allows us to share insights, better
understanding two seemingly different methods,
as we explain in Section~\ref{sec:gru}.

The amount of smoothing in a DT is spatially modulated by a reference
edge map, which in the standard DT corresponds to image gradient
magnitude. Instead, we will learn the reference
edge map from intermediate layer features of the same CNN that
produces the semantic segmentation scores, as illustrated in
\figref{model_illustration}. Crucially, this allows us to learn a
task-specific edge detector tuned for semantic image segmentation in
an end-to-end trainable system.

We evaluate the performance of the proposed method on the challenging
PASCAL VOC 2012 semantic segmentation task. In this task, domain
transform filtering is several times faster than dense CRF inference,
while performing almost as well in terms of the mean intersection-over-union (mIOU) metric.
In addition, although we only trained for semantic segmentation,
the learned edge map performs competitively on the BSDS500 edge
detection benchmark.

\section{Related Work}

\paragraph{Semantic image segmentation} Deep Convolutional Neural
Networks (CNNs) \cite{lecun1989backpropagation} have demonstrated
excellent performance on the task of semantic image segmentation
\cite{dai2015boxsup, lin2015efficient, liu2015semantic}. However, due
to the employment of max-pooling layers and downsampling, the
output of these networks tend to have poorly localized object
boundaries. Several approaches have been adopted to handle this
problem. \cite{long2014fully, hariharan2014hypercolumns, chen2014semantic} proposed to extract features from the intermediate
layers of a deep network to better estimate the object
boundaries. Networks employing deconvolutional layers and unpooling
layers to recover the ``spatial invariance'' effect of max-pooling
layers have been proposed by \cite{zeiler2011adaptive, noh2015learning}. \cite{farabet2013learning,
  mostajabi2014feedforward} used super-pixel representation, which
essentially appeals to low-level segmentation methods for the task of
localization. The fully connected Conditional Random Field (CRF)
\cite{krahenbuhl2011efficient} has been applied to capture long range
dependencies between pixels in \cite{chen2014semantic,
  lin2015efficient, liu2015semantic, papandreou2015weakly}. Further
improvements have been shown in \cite{zheng2015conditional, schwing2015fully} when
backpropagating through the CRF to refine the segmentation CNN. 
In contrary, we adopt another approach based on the domain transform \cite{GastalOliveira2011DomainTransform}
and show that beyond refining the segmentation CNN, we can also jointly 
learn to  detect object boundaries, embedding task-specific edge 
detection into the proposed model.


\paragraph{Edge detection} The edge/contour detection task has a long history
\cite{konishi2003statistical, amfm_pami2011, dollar2013structured}, which
we will only briefly review. Recently, several works have
achieved outstanding performance on the edge detection task by employing
CNNs \cite{bertasius2014deepedge, bertasius2015high, ganin2014n,
  hwang2015pixel, shen2015deepcontour, xie2015holistically}. Our work
is most related to the ones by \cite{xie2015holistically, bertasius2015high, kokkinos2016pushing}. While Xie and Tu
\cite{xie2015holistically} also exploited features from the
intermediate layers of a deep network \cite{simonyan2014very} for edge
detection, they did not apply the learned edges for high-level tasks,
such as semantic image segmentation. On the other hand, Bertasius
\etal \cite{bertasius2015high} and Kokkinos \cite{kokkinos2016pushing} made use of their learned boundaries to
improve the performance of semantic image segmentation. However, the
boundary detection and semantic image segmentation are considered as
two separate tasks. They optimized the performance of boundary
detection instead of the performance of high level tasks. On the
contrary, we learn object boundaries in order to directly optimize the
performance of semantic image segmentation. 

\paragraph{Long range dependency} Recurrent neural networks (RNNs)
\cite{elman1990finding} with long short-term memory (LSTM) units
\cite{hochreiter1997long} or gated recurrent units (GRUs)
\cite{cho2014properties, chung2014empirical} have proven successful to
model the long term dependencies in sequential data (\eg, text and
speech). Sainath \etal \cite{sainath2015convolutional} have combined
CNNs and RNNs into one unified architecture for speech
recognition. Some recent work has attempted to model
{\it spatial} long range dependency with recurrent networks for
computer vision tasks \cite{graves2009offline, socher2012convolutional,
  pinheiro2014recurrent, byeon2015scene, visin2015renet}. Our
work, integrating CNNs and Domain Transform (DT) with recursive
filtering \cite{GastalOliveira2011DomainTransform}, bears a similarity
to ReNet \cite{visin2015renet}, which also performs recursive
operations both horizontally and vertically to capture long range
dependency within whole image. In this work, we show the relationship
between DT and GRU, and we also demonstrate the effectiveness of
exploiting long range dependency by DT for semantic image
segmentation. While \cite{vineet2014filter} has previously employed
the DT (for joint object-stereo labeling), we propose to backpropagate through both of
the DT inputs to jointly learn segmentation scores and edge maps in an
end-to-end trainable system. We show that these learned edge maps
bring significant improvements compared to standard image
gradient magnitude used by \cite{vineet2014filter} or earlier DT
literature \cite{GastalOliveira2011DomainTransform}.


\section{Proposed Model}

\begin{figure*}[!t]
  \centering
  \includegraphics[width=2\columnwidth]{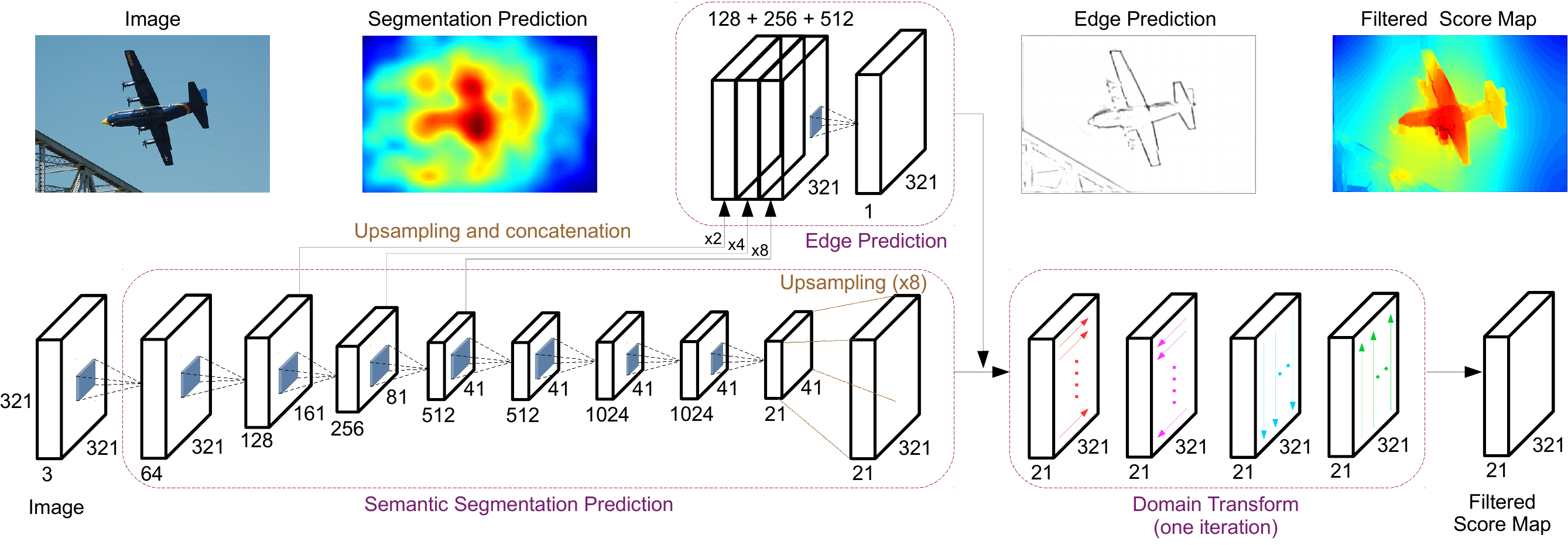}
  \caption{Our proposed model has three components: (1)
    DeepLab for semantic segmentation prediction, (2) EdgeNet
    for edge prediction, and (3) Domain Transform to accurately
    align segmentation scores with object boundaries. EdgeNet
    reuses features from intermediate DeepLab layers, resized
    and concatenated before edge prediction. Domain transform takes as
    input the raw segmentation scores and edge map, and recursively
    filters across rows and columns to produce the final filtered
    segmentation scores.}
  \label{model_illustration_detail}
\end{figure*}

\subsection{Model overview}
\label{sec:overview}
Our proposed model consists of three components, illustrated in
\figref{model_illustration_detail}. They are jointly trained
end-to-end to optimize the output semantic segmentation quality.

The first component that produces coarse semantic segmentation score
predictions is based on the publicly available DeepLab model,
\cite{chen2014semantic}, which modifies VGG-16 net
\cite{simonyan2014very} to be FCN \cite{long2014fully}. The model is
initialized from the VGG-16 ImageNet \cite{ILSVRC15} pretrained
model. We employ the DeepLab-LargeFOV variant of \cite{chen2014semantic},
which introduces zeros into the filters to enlarge its Field-Of-View,
which we will simply denote by DeepLab in the sequel.

We add a second component, which we refer to as EdgeNet. The EdgeNet
predicts edges by exploiting features from intermediate layers of
DeepLab. The features are resized to have the
same spatial resolution by bilinear interpolation before
concatenation. A convolutional layer with kernel size \by{1}{1} and
one output channel is applied to yield edge prediction. ReLU is used
so that the edge prediction is in the range of zero to infinity.

The third component in our system is the domain transform (DT),
which is is an edge-preserving filter that lends itself to very
efficient implementation by separable 1-D recursive filtering across
rows and columns. Though DT is traditionally used for graphics
applications \cite{GastalOliveira2011DomainTransform}, we use it to
filter the raw CNN semantic segmentation scores to be better aligned
with object boundaries, guided by the EdgeNet produced edge map. 

We review the standard DT in \secref{sec:dt}, we extend it to a fully
trainable system with learned edge detection in
\secref{sec:dt_learned}, and we discuss connections with the recently
proposed gated recurrent unit networks in \secref{sec:gru}.

\subsection{Domain transform with recursive filtering}
\label{sec:dt}

The domain transform takes two inputs: (1) The raw input signal $\xv$
to be filtered, which in our case corresponds to the coarse DCNN
semantic segmentation scores, and (2) a positive ``domain transform
density'' signal $\dv$, whose choice we discuss in detail in the
following section. The output of the DT is a filtered signal $\yv$.
We will use the recursive formulation of the DT due to its speed
and efficiency, though the filter can be applied via other techniques
\cite{GastalOliveira2011DomainTransform}.

For 1-D signals of length $N$, the output is computed by setting $y_1
= x_1$ and then recursively for $i=2,\dots,N$
\begin{equation}
\label{dt_rf}
y_i = (1 - w_i) x_i + w_i y_{i-1} \,.
\end{equation}
The weight $w_i$ depends on the domain transform density $d_i$
\begin{equation}
\label{dt_w_orig}
w_i = \exp \left(-\sqrt{2}d_i/\sigma_s \right) \,,
\end{equation}
where $\sigma_s$ is the standard deviation of the filter kernel over
the input's spatial domain.

Intuitively, the strength of the domain transform density $d_i \geq 0$
determines the amount of diffusion/smoothing by controlling the
relative contribution of the raw input signal $x_i$ to the filtered
signal value at the previous position $y_{i-1}$ when computing the
filtered signal at the current position $y_i$. The value of
$w_i \in (0,1)$ acts like a gate, which controls how much information
is propagated from pixel $i-1$ to $i$. We have full diffusion when
$d_i$ is very small, resulting into $w_i = 1$ and $y_i = y_{i-1}$. On
the other extreme, if $d_i$ is very large, then $w_i = 0$ and
diffusion stops, resulting in $y_i = x_i$.

Filtering by \equref{dt_rf} is asymmetric, since the current output
only depends on previous outputs. To overcome this asymmetry, we filter
1-D signals twice, first left-to-right, then right-to-left on
the output of the left-to-right pass.

Domain transform filtering for 2-D signals works in a separable
fashion, employing 1-D filtering sequentially along each signal
dimension. That is, a horizontal pass (left-to-right and
right-to-left) is performed along each row, followed by a vertical
pass (top-to-bottom and bottom-to-top) along each column. In practice,
$K > 1$ iterations of the two-pass 1-D filtering process can suppress
``striping'' artifacts resulting from 1-D filtering on 2-D
signals \cite[Fig.~4]{GastalOliveira2011DomainTransform}. We reduce
the standard deviation of the DT filtering kernel at each iteration,
requiring that the sum of total variances equals the desired variance
$\sigma_s^2$, following \cite[Eq.~14]{GastalOliveira2011DomainTransform}
\begin{equation}
\label{sigma_k}
\sigma_k = \sigma_s \sqrt{3} \frac{2^{K-k}}{\sqrt{4^K-1}}, \quad
k=1,\dots,K \,,
\end{equation} 
plugging $\sigma_k$ in place of $\sigma_s$ to compute the
weights $w_i$ by \equref{dt_w_orig} at the $k$-th iteration.

The domain transform density values $d_i$ are defined as
\begin{equation}
\label{distance}
d_i = 1 + g_i \frac{\sigma_s}{\sigma_r} \,,
\end{equation}
where $g_i \ge 0$ is the ``reference edge'', and $\sigma_r$ is the
standard deviation of the filter kernel over the reference edge map's
range. Note that the larger the value of $g_i$ is, the more confident
the model thinks there is a strong edge at pixel $i$, thus
inhibiting diffusion (\ie, $d_i \rightarrow \infty$ and $w_i = 0$).
The standard DT \cite{GastalOliveira2011DomainTransform} usually employs
the color image gradient
\begin{equation}
\label{eq:grad}
g_i = \sum_{c=1}^3 \norm{\nabla I_i^{(c)}} 
\end{equation}
but we show next that better results can be obtained by computing the
reference edge map by a learned DCNN.

\subsection{Trainable domain transform filtering}
\label{sec:dt_learned}

One novel aspect of our proposed approach is to backpropagate the
segmentation errors at the DT output $\yv$ through the DT onto its two
inputs. This allows us to use the DT as a layer in a CNN, thereby 
allowing us to jointly learn DCNNs that compute the coarse
segmentation score maps in $\xv$ and the reference edge map in
$\gv$.

\begin{figure}
  \centering
  \begin{tabular}{c c}
    \includegraphics[width=0.45\columnwidth]{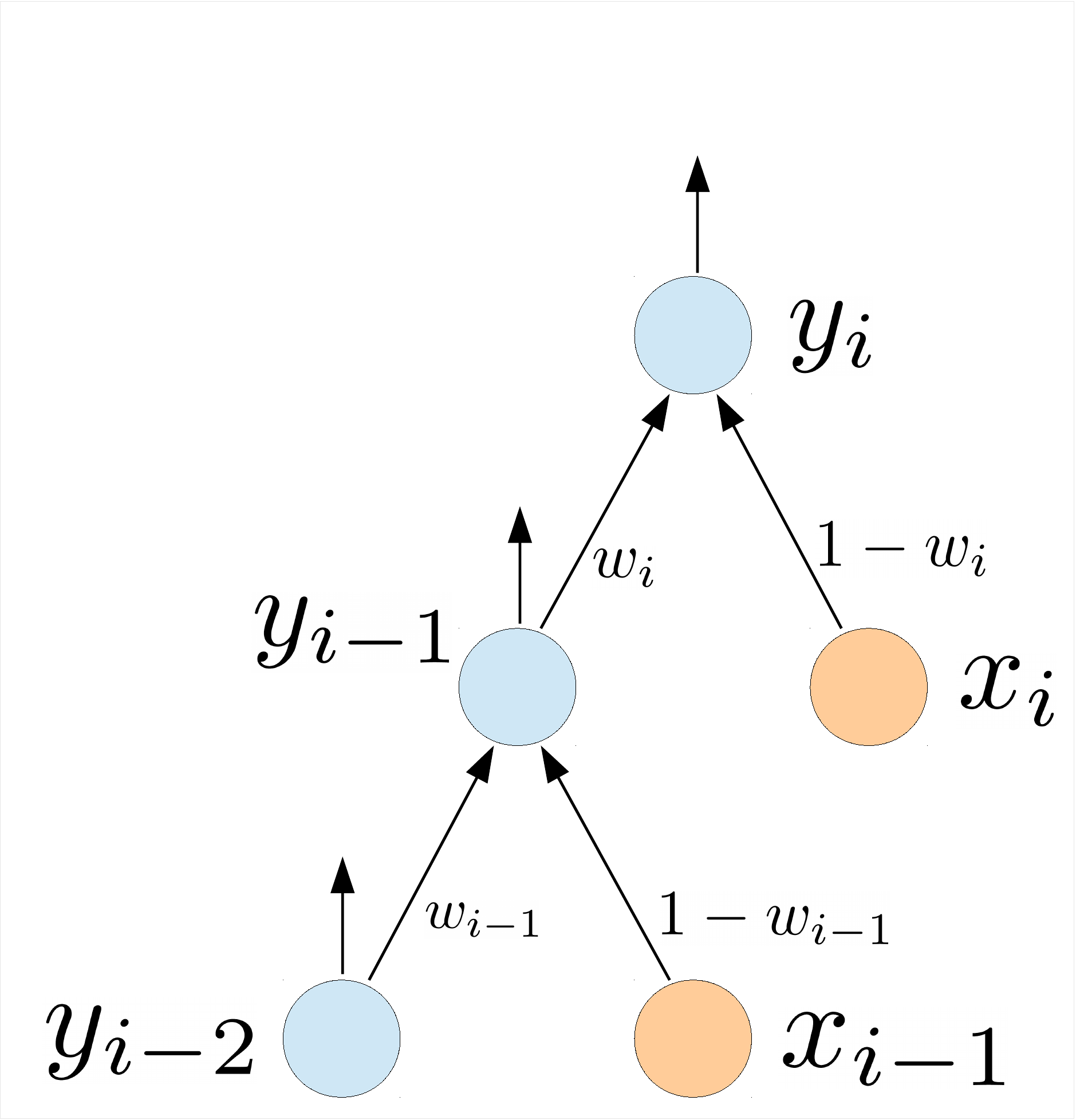} &
    \includegraphics[width=0.45\columnwidth]{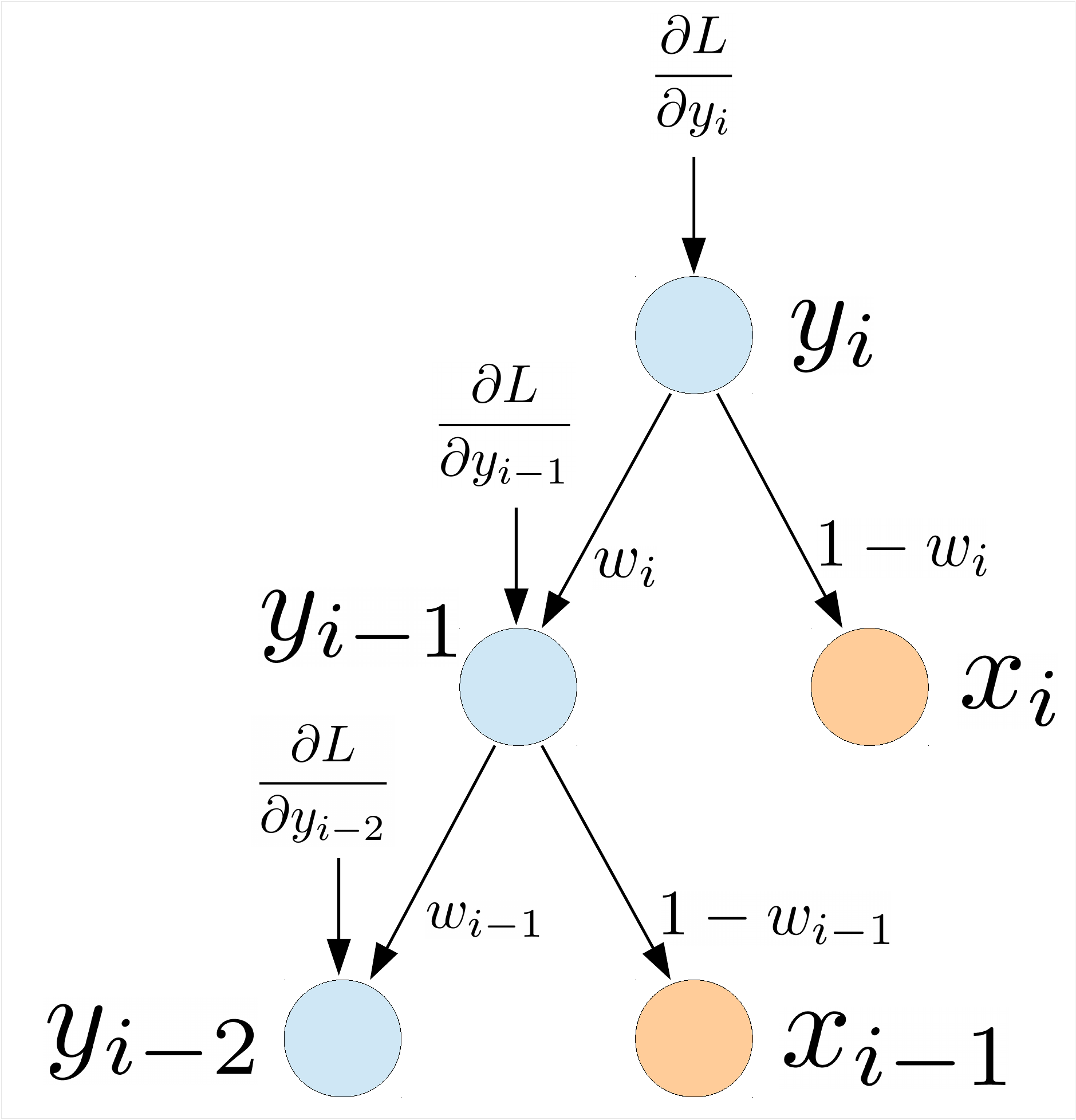} \\
    (a) & (b)
  \end{tabular}
  \caption{Computation tree for domain transform recursive filtering:
    (a) Forward pass. Upward arrows from $y_i$ nodes denote feeds to
    subsequent layers. (b) Backward pass, including contributions
    $\frac{\partial L}{\partial y_i}$ from subsequent layers.}
  \label{dt_illustration}
\end{figure}

We demonstrate how DT backpropagation works for the 1-D filtering
process of \equref{dt_rf}, whose forward pass is illustrated as
computation tree in \figref{dt_illustration}(a). We assume that each
node $y_i$ not only influences the following node $y_{i+1}$ but
also feeds a subsequent layer, thus also receiving gradient
contributions $\frac{\partial L}{\partial y_i}$ from that layer during
back-propagation. Similar to standard back-propagation in time, we
unroll the recursion of \equref{dt_rf} in reverse for $i = N,\dots,2$
as illustrated in \figref{dt_illustration}(b) to update the
derivatives with respect to $\yv$, and to also compute derivatives
with respect to $\xv$ and $\wv$,
\begin{eqnarray}
\frac{\partial L}{\partial x_i} & \gets & 
\left(1 - w_i \right) \frac{\partial L}{\partial
  y_{i}} \label{dt_backprop_3} 
\\
\frac{\partial L}{\partial w_i} & \gets & 
\frac{\partial L}{\partial w_i} + \left(y_{i-1} - x_i \right)
\frac{\partial L}{\partial y_i} 
\label{dt_backprop_1} \\
\frac{\partial L}{\partial y_{i-1}} & \gets & 
\frac{\partial L}{\partial y_{i-1}} + w_i \frac{\partial L}{\partial y_{i}} \,, \label{dt_backprop_2}
\end{eqnarray}
where $\frac{\partial L}{\partial x_i}$ and 
$\frac{\partial L}{\partial w_i}$ are initialized to 0 and
$\frac{\partial L}{\partial y_i}$ is initially set to the value sent
by the subsequent layer. Note that the weight $w_i$ is shared across
all filtering stages (\ie, left-to-right/right-to-left within
horizontal pass and top-to-bottom/bottom-to-top within vertical pass)
and $K$ iterations, with each pass contributing to the partial
derivative.

With these partial derivatives we can produce derivatives with respect to the reference
edge $g_i$. Plugging \equref{distance} into \equref{dt_w_orig}
yields
\begin{equation}
  \label{dt_w_full}
  w_i = \exp \left(-\frac{\sqrt{2}}{\sigma_k} \left(1 + g_i \frac{\sigma_s}{\sigma_r} \right) \right) \,.
\end{equation}
Then, by the chain rule, the derivative with respect to $g_i$ is
\begin{equation}
  \label{dt_backprop_4}
  \frac{\partial L}{\partial g_i} = -
  \frac{\sqrt{2}}{\sigma_k}
  \frac{\sigma_s}{\sigma_r} 
  w_i \frac{\partial L}{\partial w_i} \,.
\end{equation}
This gradient is then further propagated onto the deep convolutional
neural network that generated the edge predictions that were used as input to the DT.

\subsection{Relation to gated recurrent unit networks}
\label{sec:gru}

Equation~\ref{dt_rf} defines DT filtering as a recursive operation. It
is interesting to draw connections with other recent RNN
formulations. Here we establish a precise
connection with the gated recurrent unit (GRU) RNN architecture
\cite{cho2014properties} recently proposed for modeling sequential
text data. The GRU employs the update rule
\begin{equation}
  \label{gru}
  y_i = z_i \tilde{y}_i + (1 - z_i) y_{i-1} \,.
\end{equation}
Comparing with \equref{dt_rf}, we can relate the GRU's ``update gate'' $z_i$ and
``candidate activation'' $\tilde{y}_i$ with DT's weight and raw input
signal as follows: $z_i = 1 - w_i$ and $\tilde{y}_i = x_i$. 

The GRU update gate $z_i$ is defined as $z_i = \sigma(f_i)$, where
$f_i$ is an activation signal and $\sigma(t) = 1/(1+e^{-t})$.
Comparing with \equref{dt_w_full} yields a direct correspondence
between the DT reference edge map $g_i$ and the GRU activation $f_i$:
\begin{equation}
  \label{gru_dt}
  g_i = \frac{\sigma_r}{\sigma_s} \left(
  \frac{\sigma_k}{\sqrt{2}} \log(1 + e^{f_i}) - 1
  \right) \,.
\end{equation}

\section{Experimental Evaluation}

\subsection{Experimental Protocol}
\paragraph{Dataset} We evaluate the proposed method on the PASCAL VOC
2012 segmentation benchmark \cite{everingham2014pascal}, consisting of
20 foreground object classes and one background class. We augment the
training set from the annotations by \cite{hariharan2011semantic}. The
performance is measured in terms of pixel intersection-over-union
(IOU) averaged across the 21 classes.

\paragraph{Training} A two-step training process is employed. We first
train the DeepLab component and then we jointly fine-tune the
whole model. Specifically, we employ exactly the same setting as
\cite{chen2014semantic} to train DeepLab in the first
stage. In the second stage, we employ a small learning rate of
$10^{-8}$ for fine-tuning. The added convolutional layer of EdgeNet is
initialized with Gaussian variables with zero mean and standard
deviation of $10^{-5}$ so that in the beginning the EdgeNet predicts
no edges and it starts to gradually learn edges for semantic
segmentation. Total training time is 11.5 hours (10.5 and 1 hours
for each stage). 


\paragraph{Reproducibility} The proposed methods are implemented
by extending the Caffe framework \cite{jia2014caffe}. The code and
models are available at \url{http://liangchiehchen.com/projects/DeepLab.html}.


\subsection{Experimental Results}

We first explore on the validation set the
hyper-parameters in the proposed model, including (1) features for
EdgeNet, (2) hyper-parameters for domain transform (\ie, number of
iterations, $\sigma_s$, and $\sigma_r$). We also experiment with
different methods to generate edge prediction. After that, we analyze
our models and evaluate on the official test set.

\paragraph{Features for EdgeNet} The EdgeNet we employ exploits
intermediate features from DeepLab. We first investigate
which VGG-16 \cite{simonyan2014very} layers give better
performance with the DT hyper-parameters fixed. As shown in
\tabref{tab:edgenet_feature}, baseline DeepLab attains
$62.25\%$ mIOU on PASCAL VOC 2012 validation set. We start to exploit
the features from conv3\_3, which has receptive field size 40. The
size is similar to the patch size typically used for edge detection
\cite{dollar2013structured}. The resulting model achieves performance
of $65.64\%$, $3.4\%$ better than the baseline. When using features
from conv2\_2, conv3\_3, and conv4\_3, the performance can be further
improved to $66.03\%$. However, we do not observe any significant
improvement if we also exploit the features from conv1\_2 or
conv5\_3. We use features from conv2\_2, conv3\_3, and conv4\_3
in remaining experiments involving EdgeNet.

\begin{table}[!t]
  \centering
  \addtolength{\tabcolsep}{2.5pt}
  \begin{tabular}{ l | c }
    \toprule[0.2 em]
    Method & mIOU (\%) \\
    \toprule[0.2 em]
    Baseline: DeepLab & 62.25 \\
    \midrule
    conv3\_3 & 65.64 \\
    conv2\_2 + conv3\_3 & 65.75 \\
    conv2\_2 + conv3\_3 + conv4\_3 & 66.03 \\
    conv2\_2 + conv3\_3 + conv4\_3 + conv5\_3 & 65.94 \\
    conv1\_2 + conv2\_2 + conv3\_3 + conv4\_3 & 65.89 \\
    \bottomrule[0.1 em]
  \end{tabular}
  \caption{VOC 2012 val set. Effect of using features from different
    convolutinal layers for EdgeNet ($\sigma_s=100$ and $\sigma_r=1$
    for DT).}
  \label{tab:edgenet_feature}
\end{table}

\begin{figure}
  \centering
  \begin{tabular}{c c}
  \includegraphics[width=0.47\linewidth]{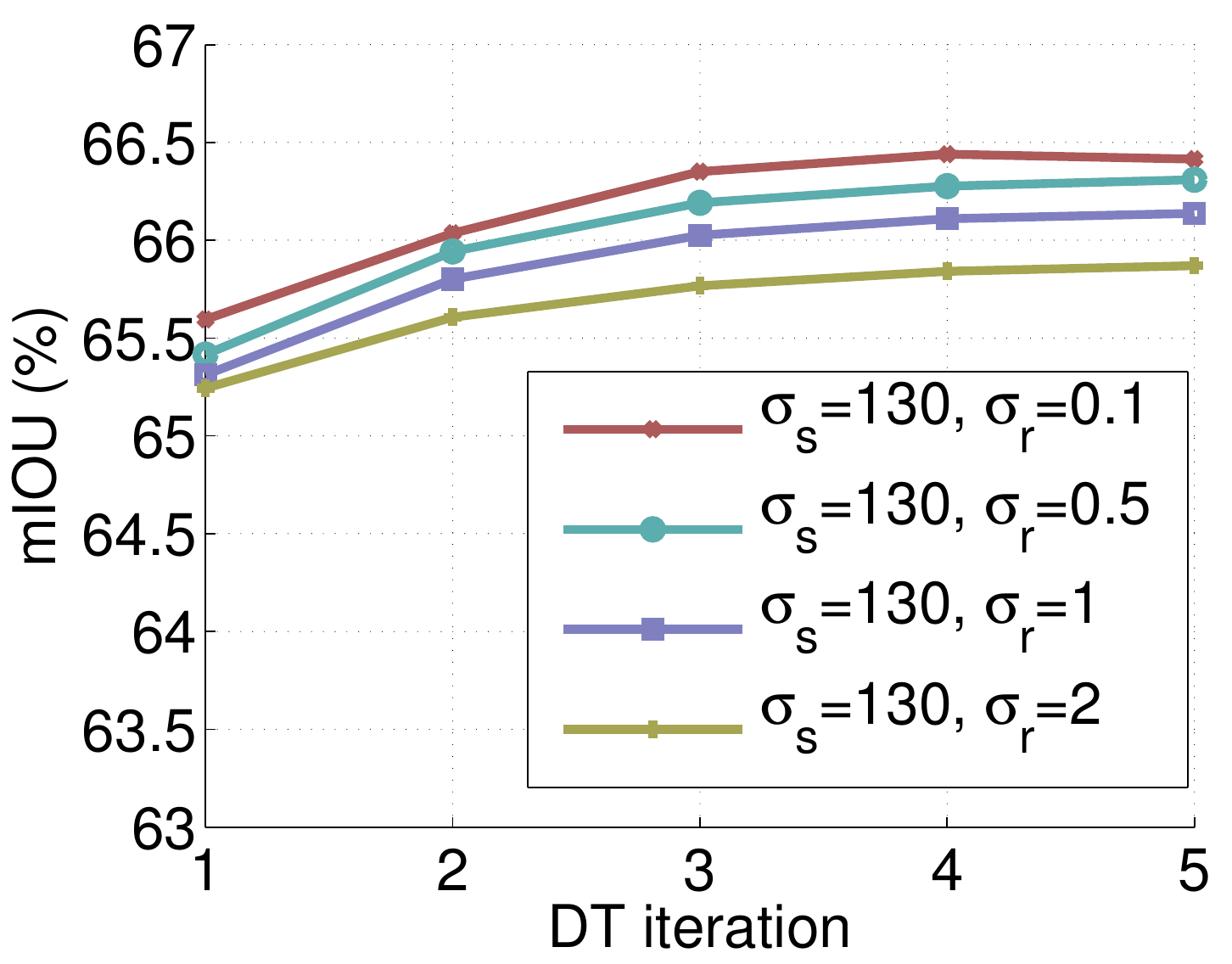} &
  \includegraphics[width=0.47\linewidth]{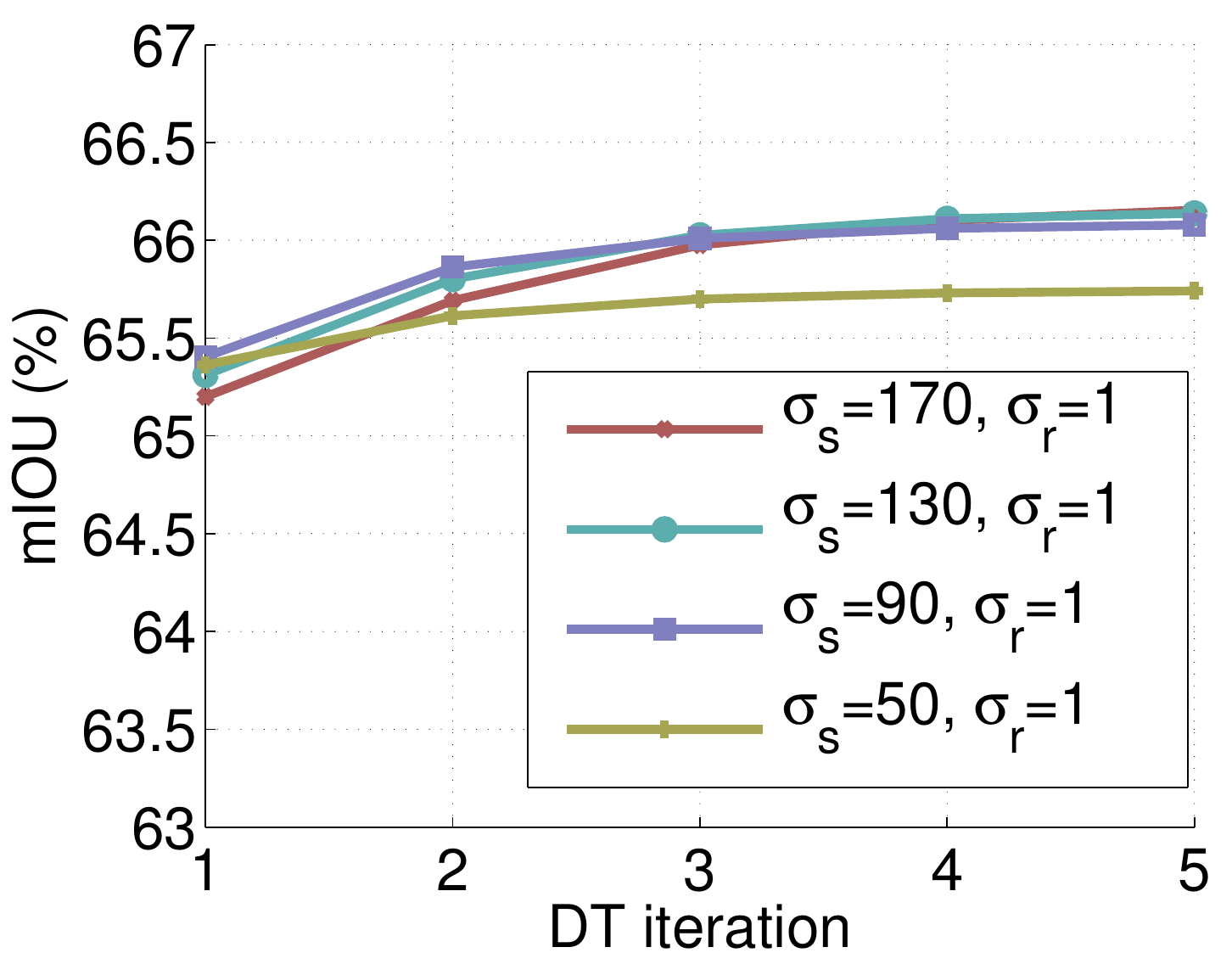} \\
  (a) & (b) 
  \end{tabular}
  \caption{VOC 2012 val set. Effect of varying number of iterations
    for domain transform: (a) Fix $\sigma_s$ and vary both $\sigma_r$
    and $K$ iterations. (b) Fix $\sigma_r$ and vary both $\sigma_s$
    and $K$ iterations.}
  \label{fig_dt_iterations}
\end{figure}

\paragraph{Number of domain transform iterations} Domain transform
requires multiple iterations of the two-pass 1-D filtering process to
avoid the ``striping'' effect
\cite[Fig.~4]{GastalOliveira2011DomainTransform}. We train the
proposed model with $K$ iterations for the domain transform, and
perform the same $K$ iterations during test. Since there are two more
hyper-parameters $\sigma_s$ and $\sigma_r$ (see \equref{dt_w_full}),
we also vary their values to investigate the effect of varying the $K$
iterations for domain transform. As shown in
\figref{fig_dt_iterations}, employing $K=3$ iterations for domain
transform in our proposed model is sufficient to reap most of the
gains for several different values of $\sigma_s$ and $\sigma_r$.

\begin{figure}
  \centering
  \begin{tabular}{c c}
  \includegraphics[width=0.47\linewidth]{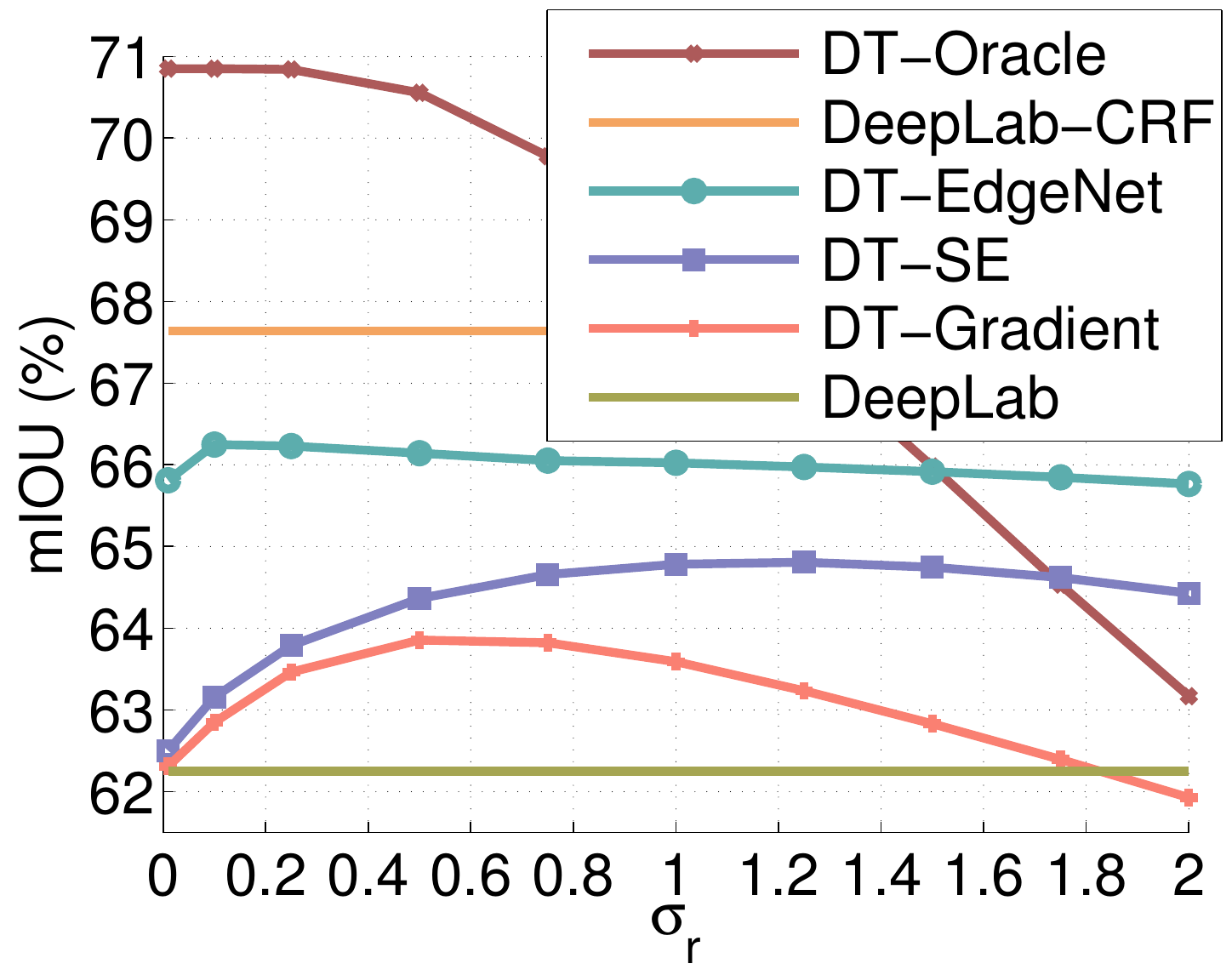} &
  \includegraphics[width=0.47\linewidth]{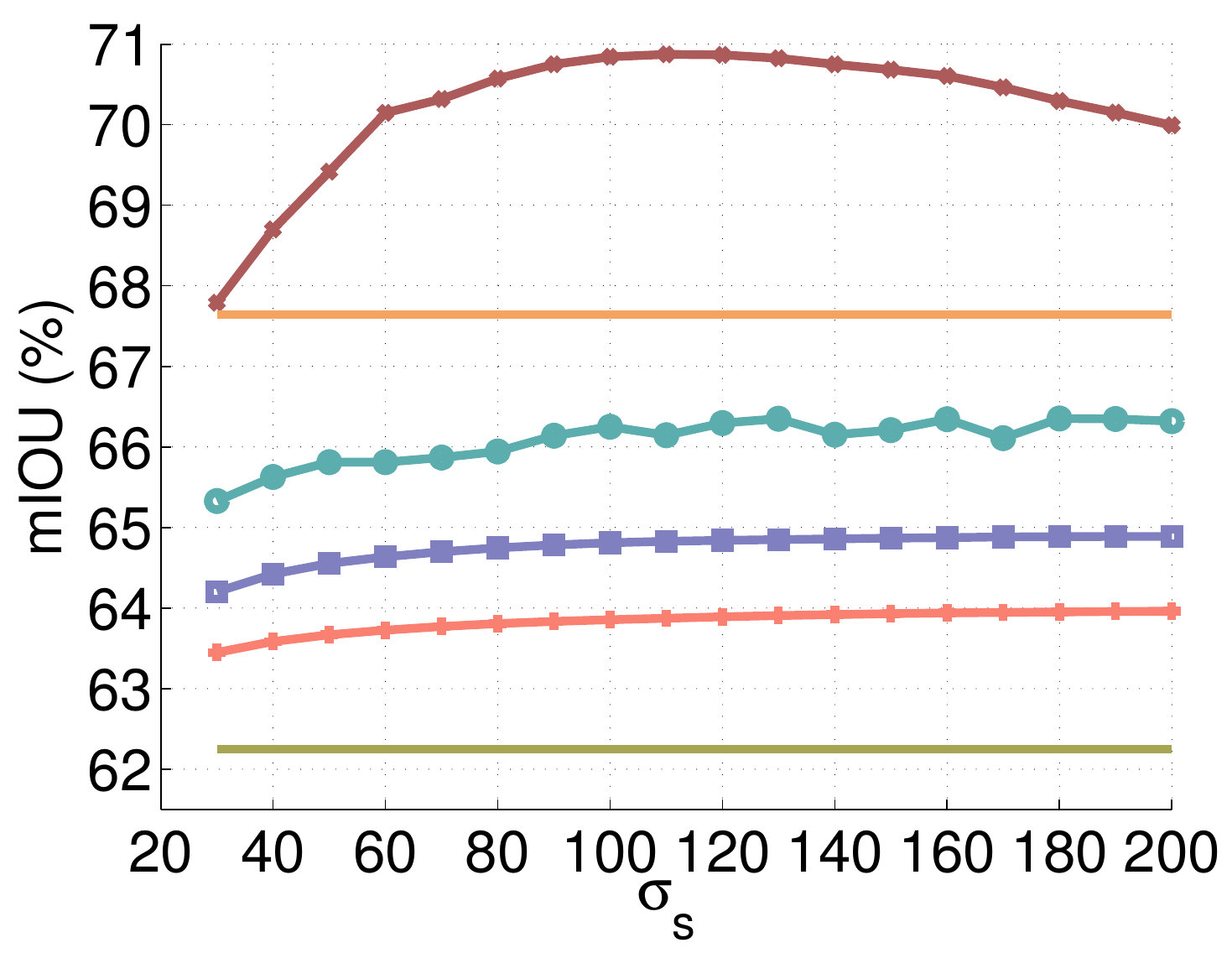} \\
  (a) & (b) 
  \end{tabular}
  \caption{VOC 2012 val set. Effect of varying $\sigma_s$ and
    $\sigma_r$. (a) Fix $\sigma_s=100$ and vary $\sigma_r$. (b) Use
    the best $\sigma_r$ from (a) and vary $\sigma_s$.}
  \label{fig:dt_vary_sigma}
\end{figure}

\paragraph{Varying domain transform $\sigma_s$, $\sigma_r$ and
  comparison with other edge detectors} We investigate the
effect of varying $\sigma_s$ and $\sigma_r$ for domain transform. We
also compare alternative methods to generate edge prediction for domain
transform: (1) DT-Oracle, where groundtruth object boundaries are
used, which serves as an upper bound on our method. (2) The proposed
DT-EdgeNet, where the edges are produced by EdgeNet. (3) DT-SE, where
the edges are found by Structured Edges (SE)
\cite{dollar2013structured}. (4) DT-Gradient, where the image (color)
gradient magnitude of \equref{eq:grad} is used as in standard domain
transform \cite{GastalOliveira2011DomainTransform}. We search for
optimal $\sigma_s$ and $\sigma_r$ for those methods. First, we fix
$\sigma_s=100$ and vary $\sigma_r$ in
\figref{fig:dt_vary_sigma}(a). We found that the performance of
DT-Oracle, DT-SE, and DT-Gradient are affected a lot by different
values of $\sigma_r$, since they are generated by other ``plugged-in''
modules (\ie, not jointly fine-tuned). We also show the performance of
baseline DeepLab and DeepLab-CRF which employs dense
CRF. We then fix the found optimal value of $\sigma_r$ and vary
$\sigma_s$ in \figref{fig:dt_vary_sigma}~(b). We found that as long as
$\sigma_s \geq 90$, the performance of DT-EdgeNet, DT-SE, and
DT-Gradient do not vary significantly. After finding optimal
values of $\sigma_r$ and $\sigma_s$ for each setting, we use them for
remaining experiments.

\begin{figure*}[!th]
  \centering
  \begin{tabular}{c c c c c}
    \includegraphics[height=0.09\linewidth]{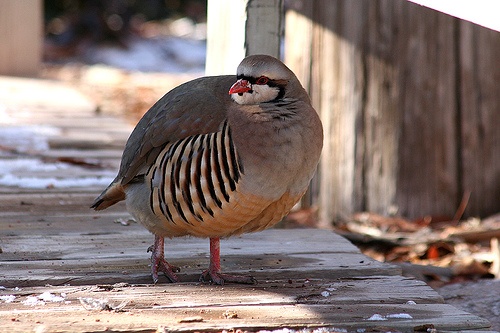} &
    \includegraphics[height=0.09\linewidth]{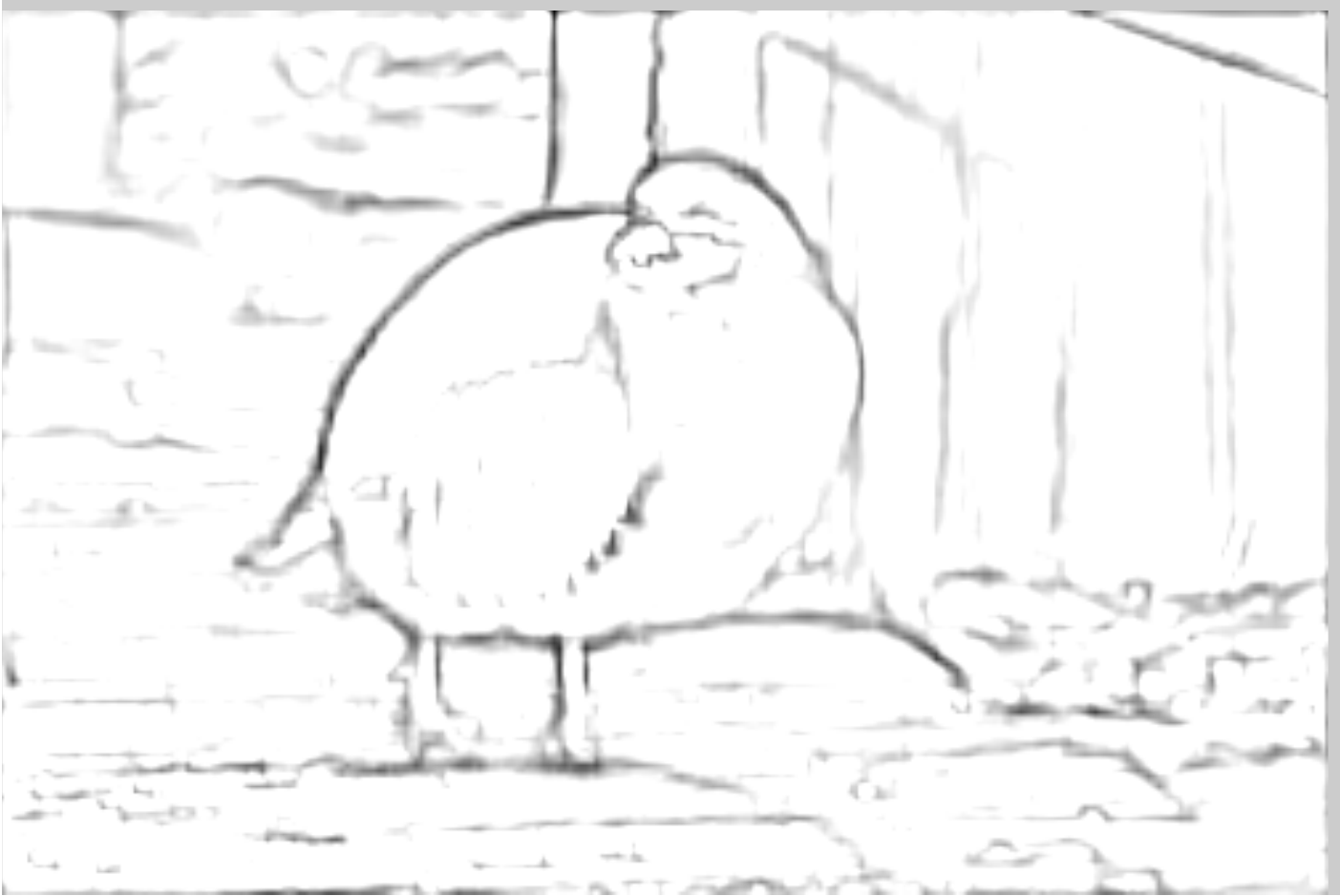} &
    \includegraphics[height=0.09\linewidth]{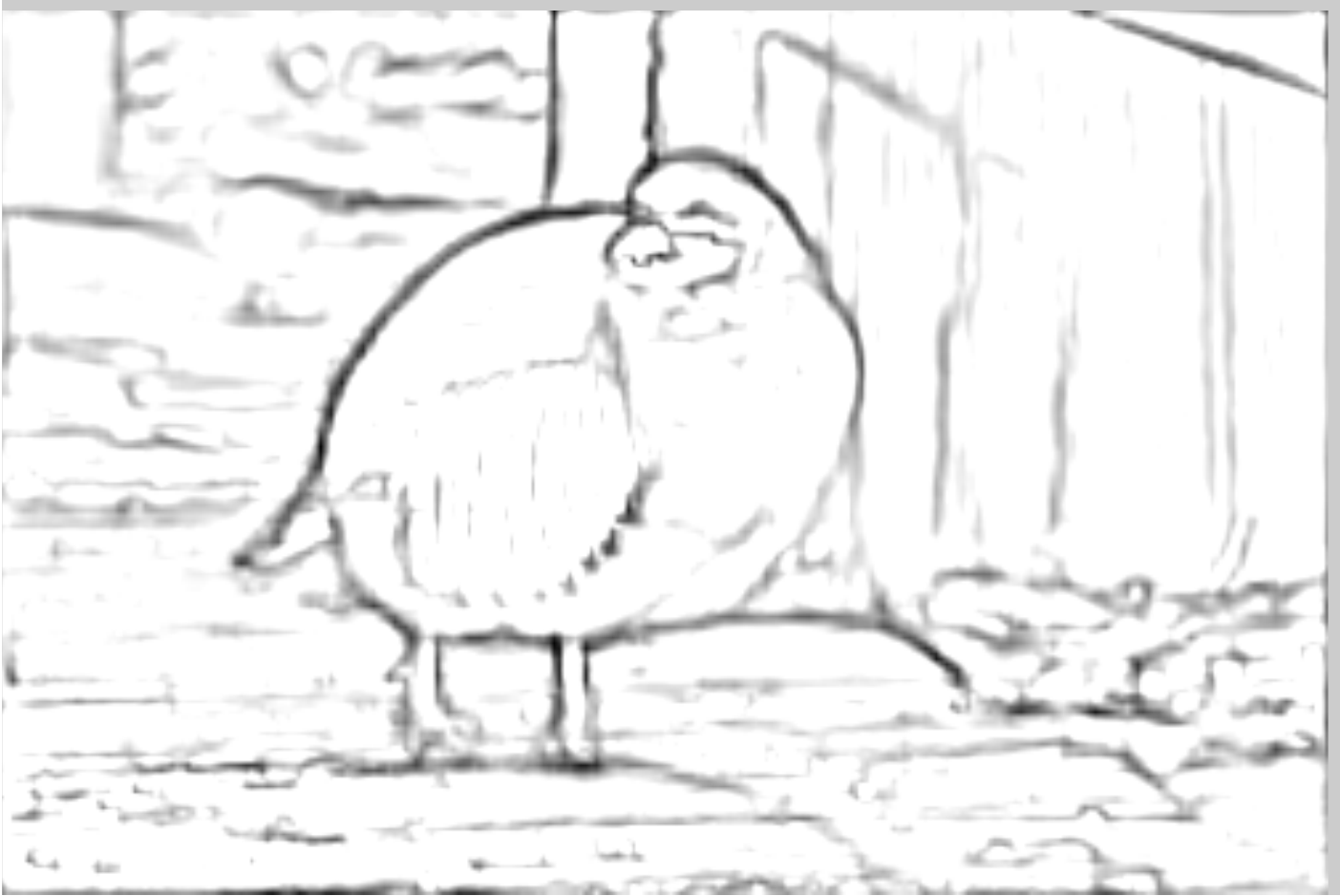} &
    \includegraphics[height=0.09\linewidth]{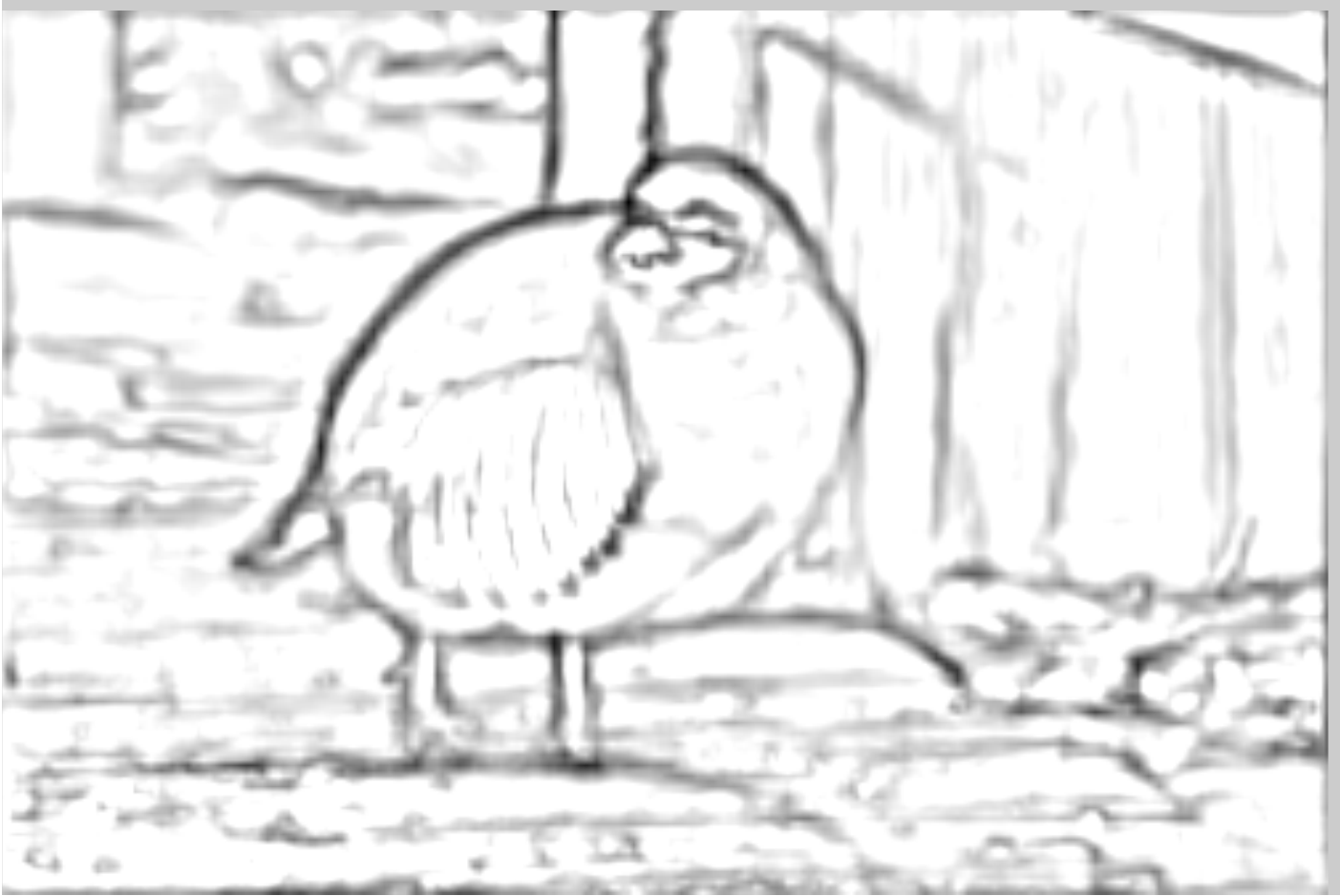} &
    \includegraphics[height=0.09\linewidth]{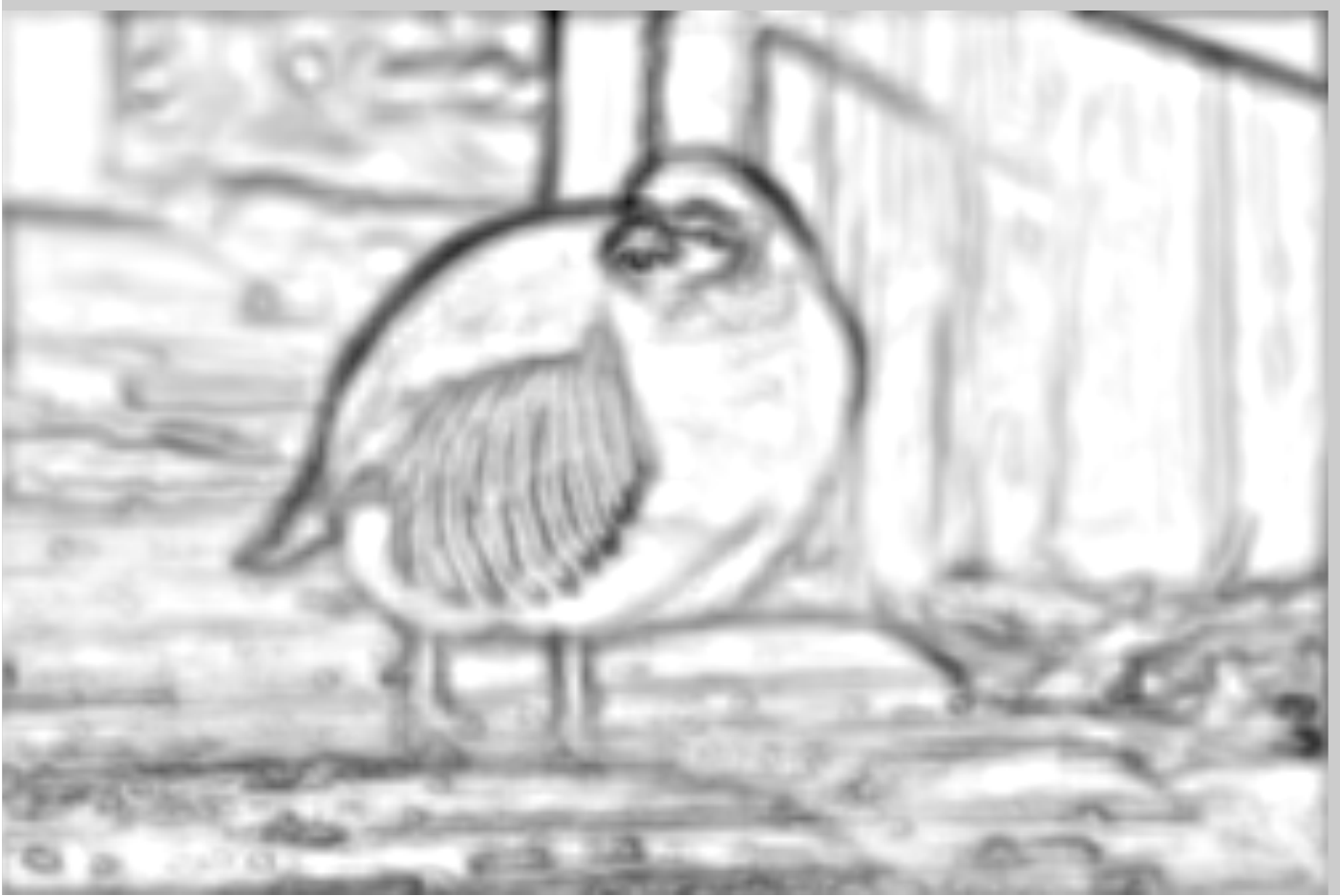} \\
    (a) Image &
    (b) $\sigma_s=100, \sigma_r=0.1$ &
    (c) $\sigma_s=100, \sigma_r=0.5$ &
    (d) $\sigma_s=100, \sigma_r=2$ &
    (e) $\sigma_s=100, \sigma_r=10$ \\
    \includegraphics[height=0.09\linewidth]{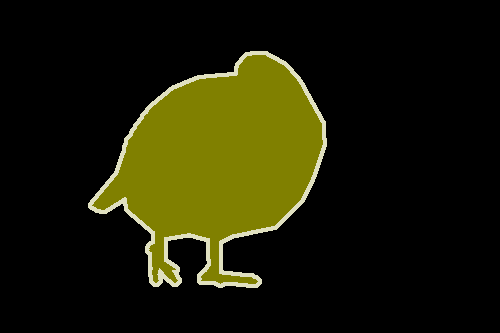} &
    \includegraphics[height=0.09\linewidth]{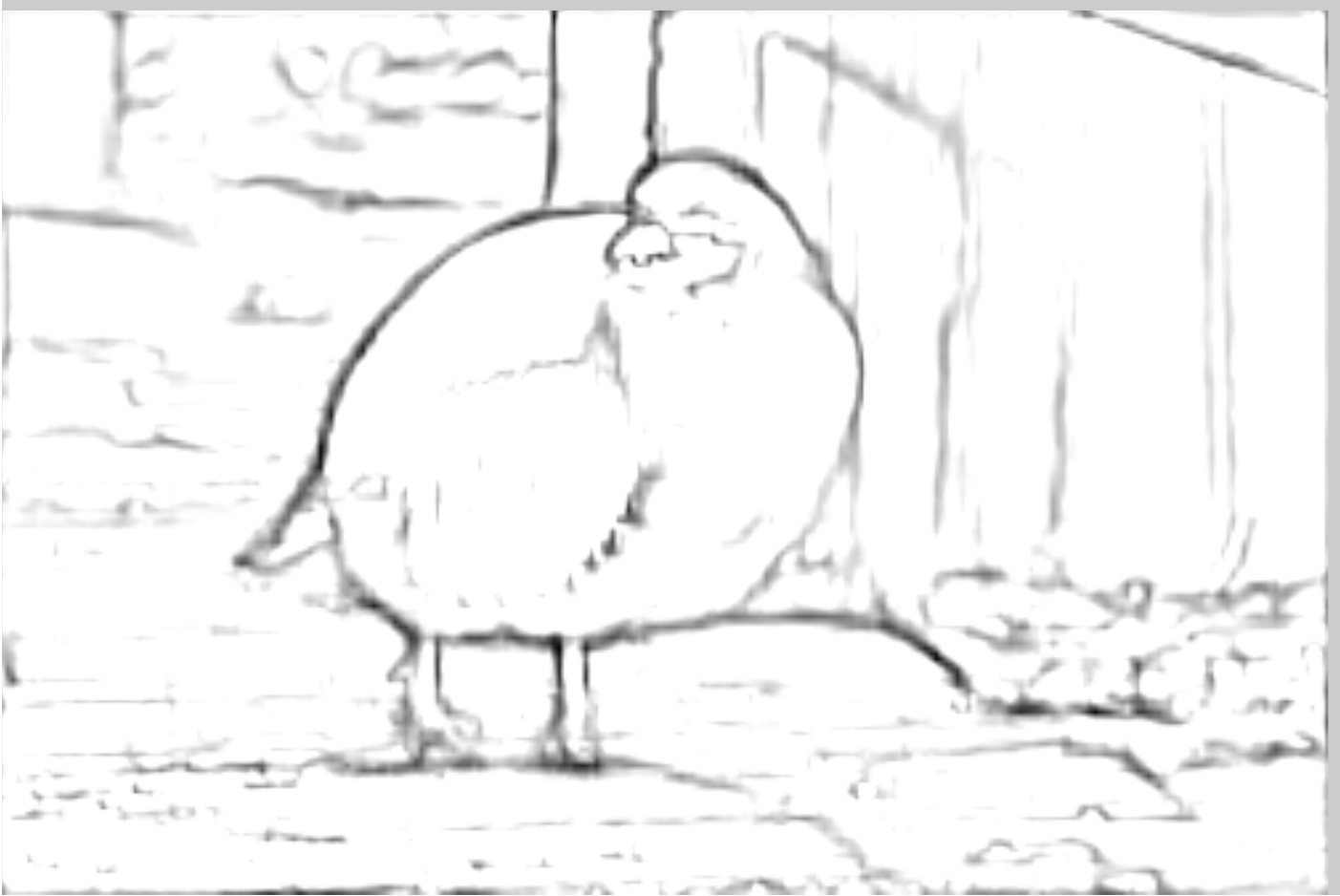} &
    \includegraphics[height=0.09\linewidth]{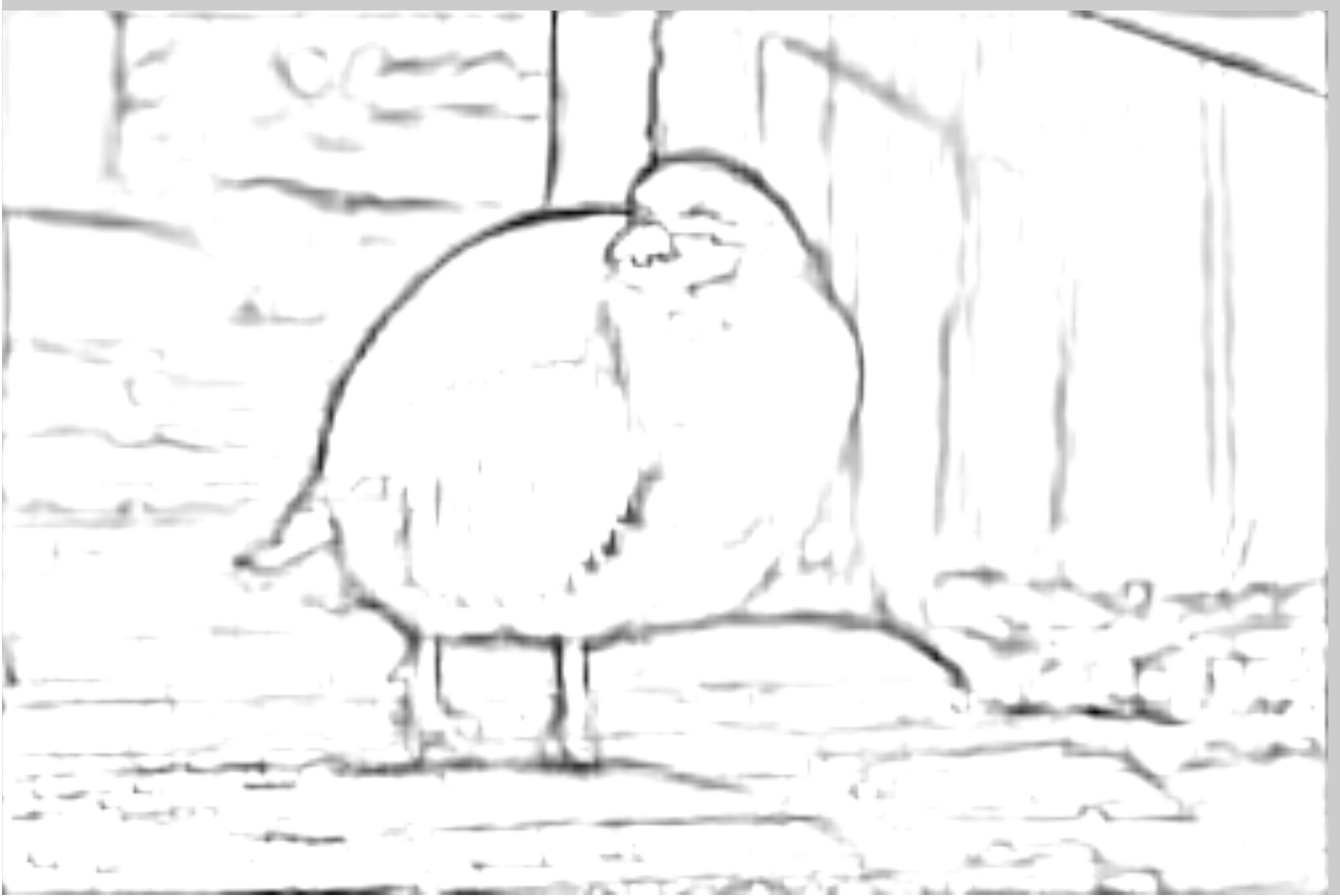} &
    \includegraphics[height=0.09\linewidth]{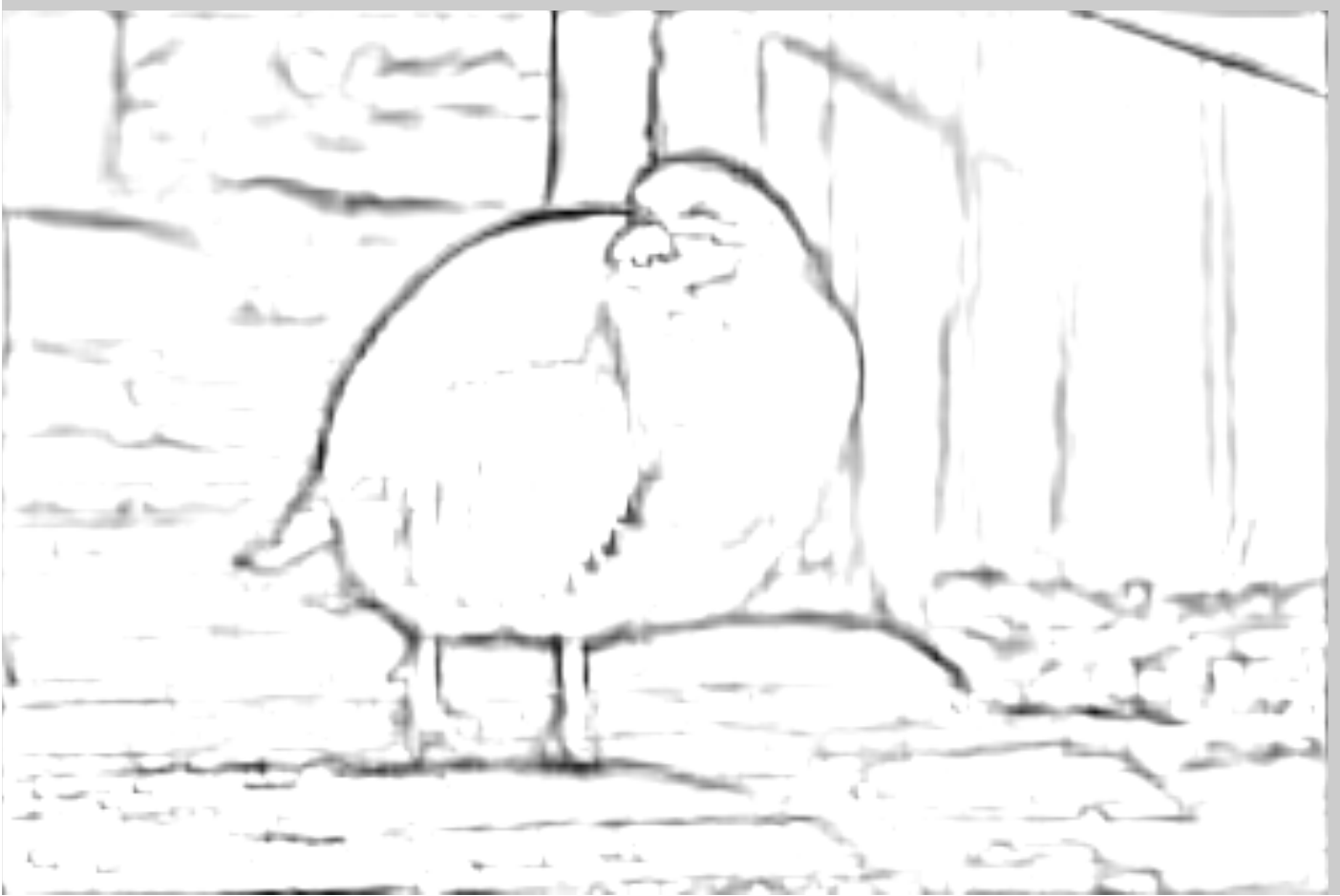} &
    \includegraphics[height=0.09\linewidth]{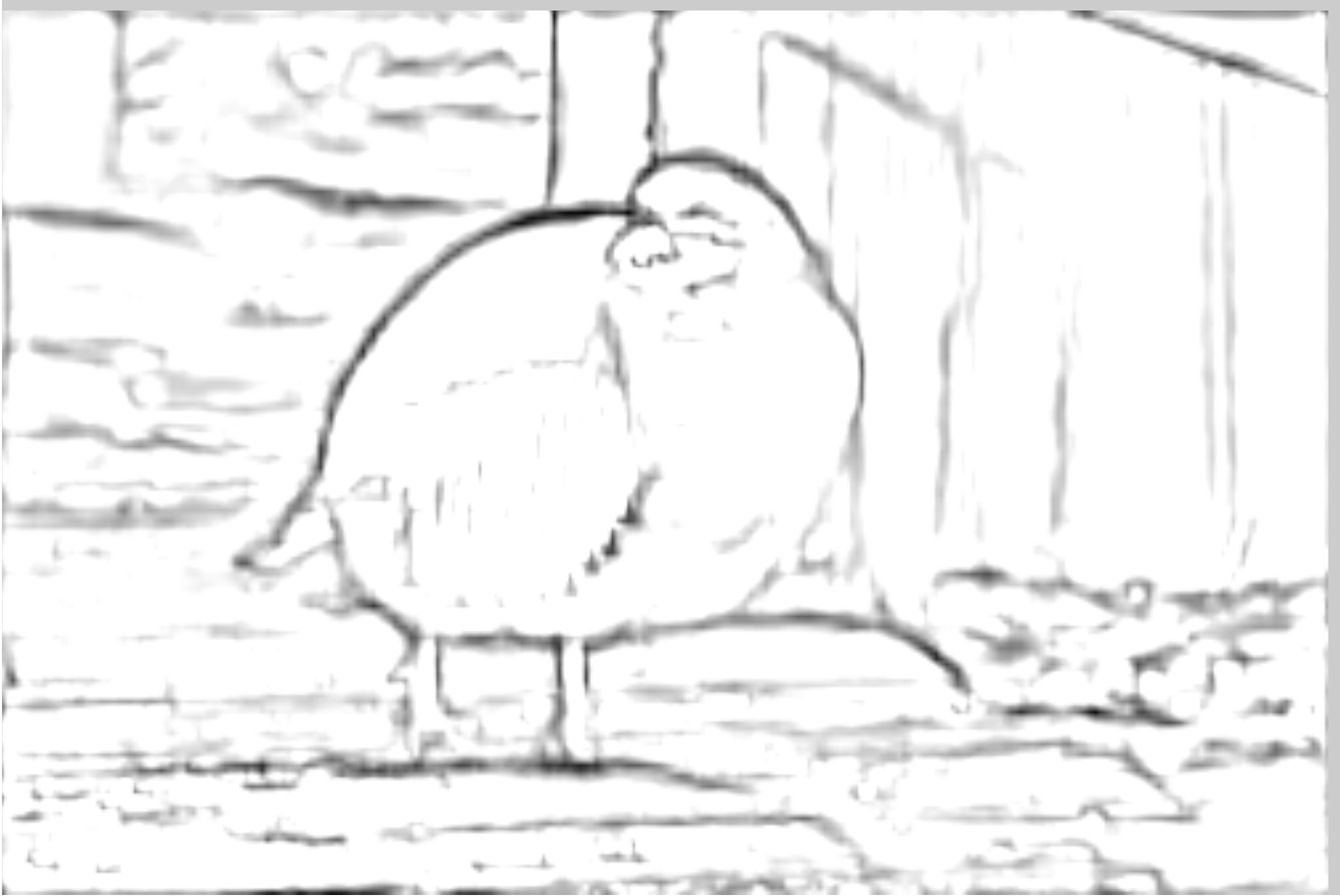} \\
    (f) Groundtruth & 
    (g) $\sigma_s=50, \sigma_r=0.1$ &
    (h) $\sigma_s=90, \sigma_r=0.1$ &
    (i) $\sigma_s=130, \sigma_r=0.1$ &
    (j) $\sigma_s=170, \sigma_r=0.1$ \\
    \end{tabular}
  \caption{Effect of varying domain transform's $\sigma_s$ and
    $\sigma_r$. First row: when $\sigma_s$ is fixed and $\sigma_r$
    increases, the EdgeNet starts to include more background
    edges. Second row: when $\sigma_r$ is fixed, varying $\sigma_s$
    has little effect on learned edges.}
  \label{fig:vis_vary_sigma}
\end{figure*}

We further visualize the edges learned by our DT-EdgeNet in
\figref{fig:vis_vary_sigma}. As shown in the first row, when
$\sigma_r$ increases, the learned edges start to include not only
object boundaries but also background textures, which degrades the
performance for semantic segmentation in our method (\ie, noisy edges
make it hard to propagate information between neighboring pixels). As
shown in the second row, varying $\sigma_s$ does not change the
learned edges a lot, as long as its value is large enough (\ie, $\geq
90$).

We show {\it val} set performance (with the best values of $\sigma_s$
and $\sigma_r$) for each method in \tabref{tab:models_valset}. The
method DT-Gradient improves over the baseline DeepLab by
1.7\%. While DT-SE is 0.9\% better than DT-Gradient, DT-EdgeNet
further enhances performance ($4.1\%$ over baseline). Even though
DT-EdgeNet is $1.2\%$ lower than DeepLab-CRF, it is several
times faster, as we discuss later. Moreover, we have found that
combining DT-EdgeNet and dense CRF yields the best performance ($0.8\%$
better than DeepLab-CRF). In this hybrid DT-EdgeNet+DenseCRF scheme
we post-process the DT filtered score maps in an extra fully-connected
CRF step.

\begin{table}
  \centering
  \addtolength{\tabcolsep}{2.5pt}
  \begin{tabular}{l | c }
    \toprule[0.2 em]
    Method & mIOU (\%) \\
    \toprule[0.2 em]
    DeepLab & 62.25 \\
    DeepLab-CRF & 67.64 \\
    \midrule
    DT-Gradient & 63.96 \\
    DT-SE & 64.89 \\
    DT-EdgeNet & 66.35 \\
    DT-EdgeNet + DenseCRF & 68.44 \\
    \midrule
    DT-Oracle & 70.88 \\
    \bottomrule[0.1 em]
  \end{tabular}
  \caption{Performance on PASCAL VOC 2012 val set.}
  \label{tab:models_valset}
\end{table}

\begin{figure}[!tbp]
\centering
\resizebox{\columnwidth}{!}{
  \begin{tabular} {c c}
    \raisebox{0.9cm} {
    \begin{tabular}{c c}
      \includegraphics[height=0.1\linewidth]{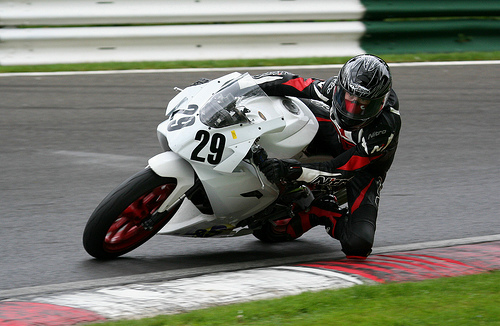} &
      \includegraphics[height=0.1\linewidth]{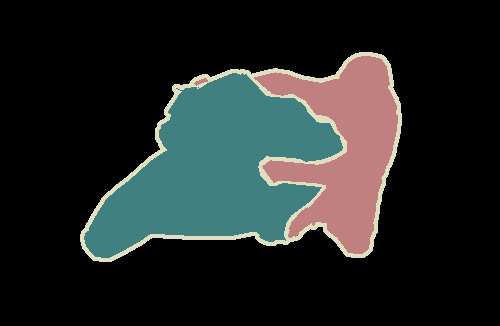} \\
      \includegraphics[height=0.1\linewidth]{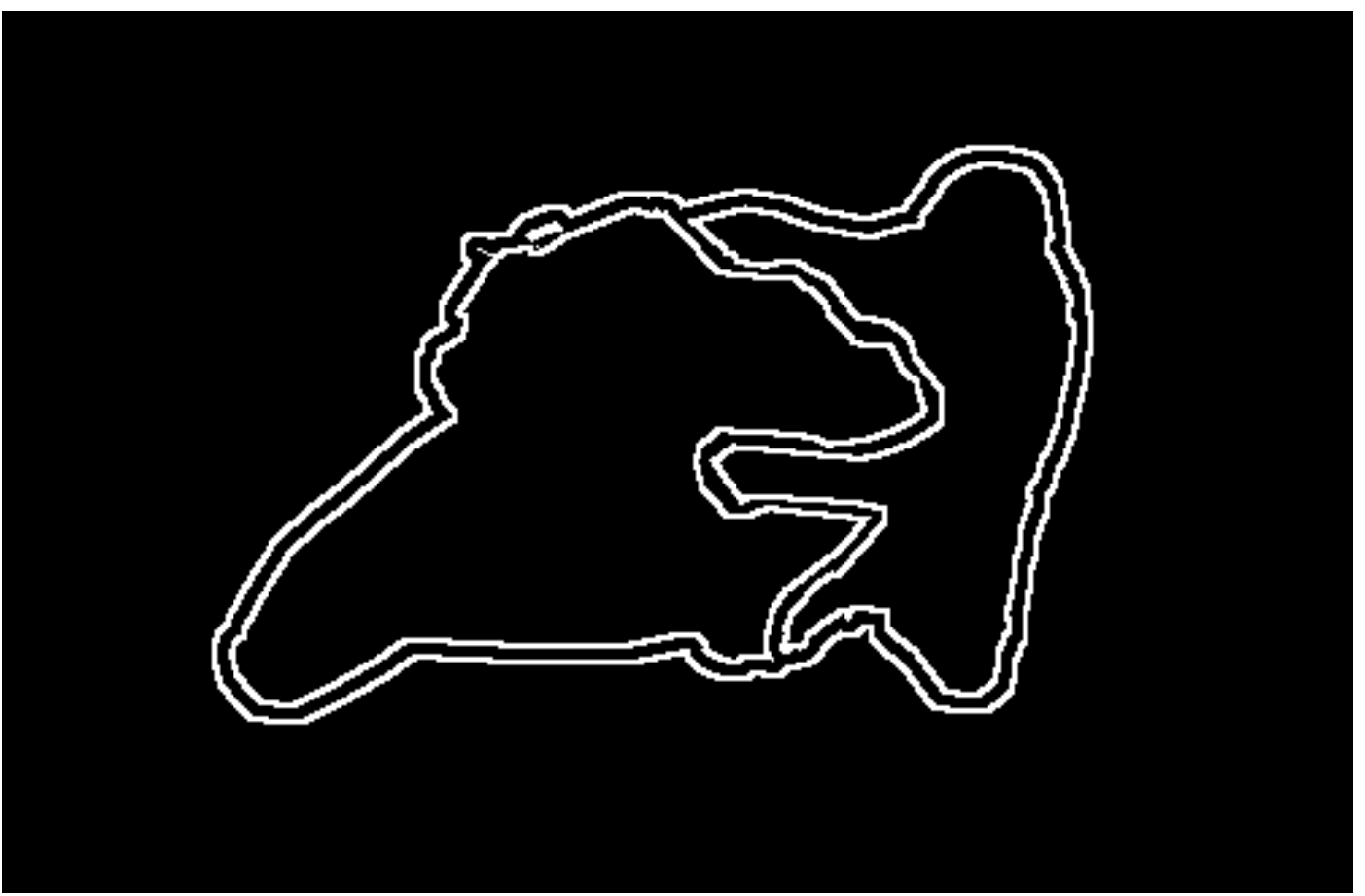} &
      \includegraphics[height=0.1\linewidth]{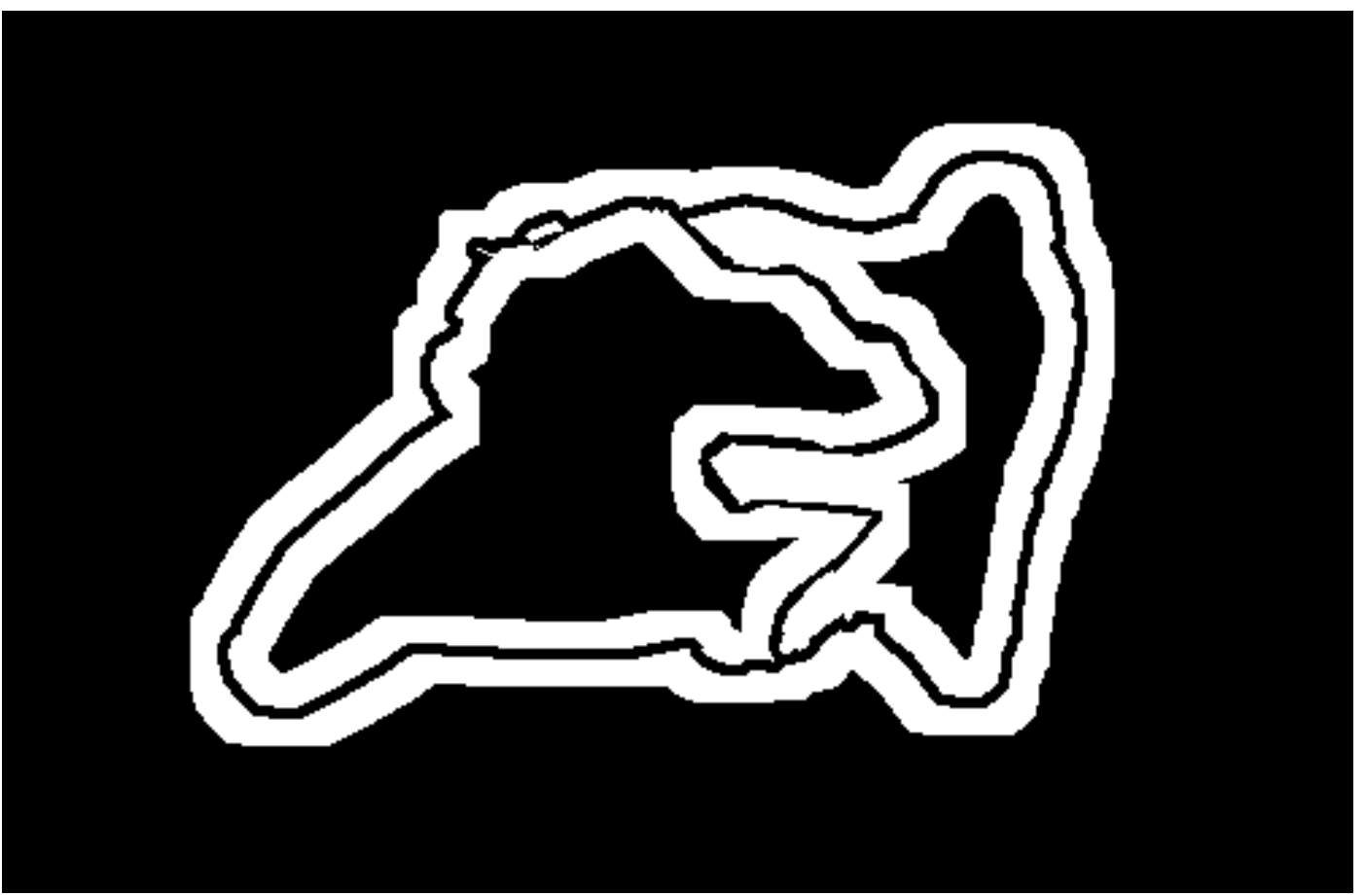} \\
    \end{tabular} } &
    \includegraphics[height=0.25\linewidth]{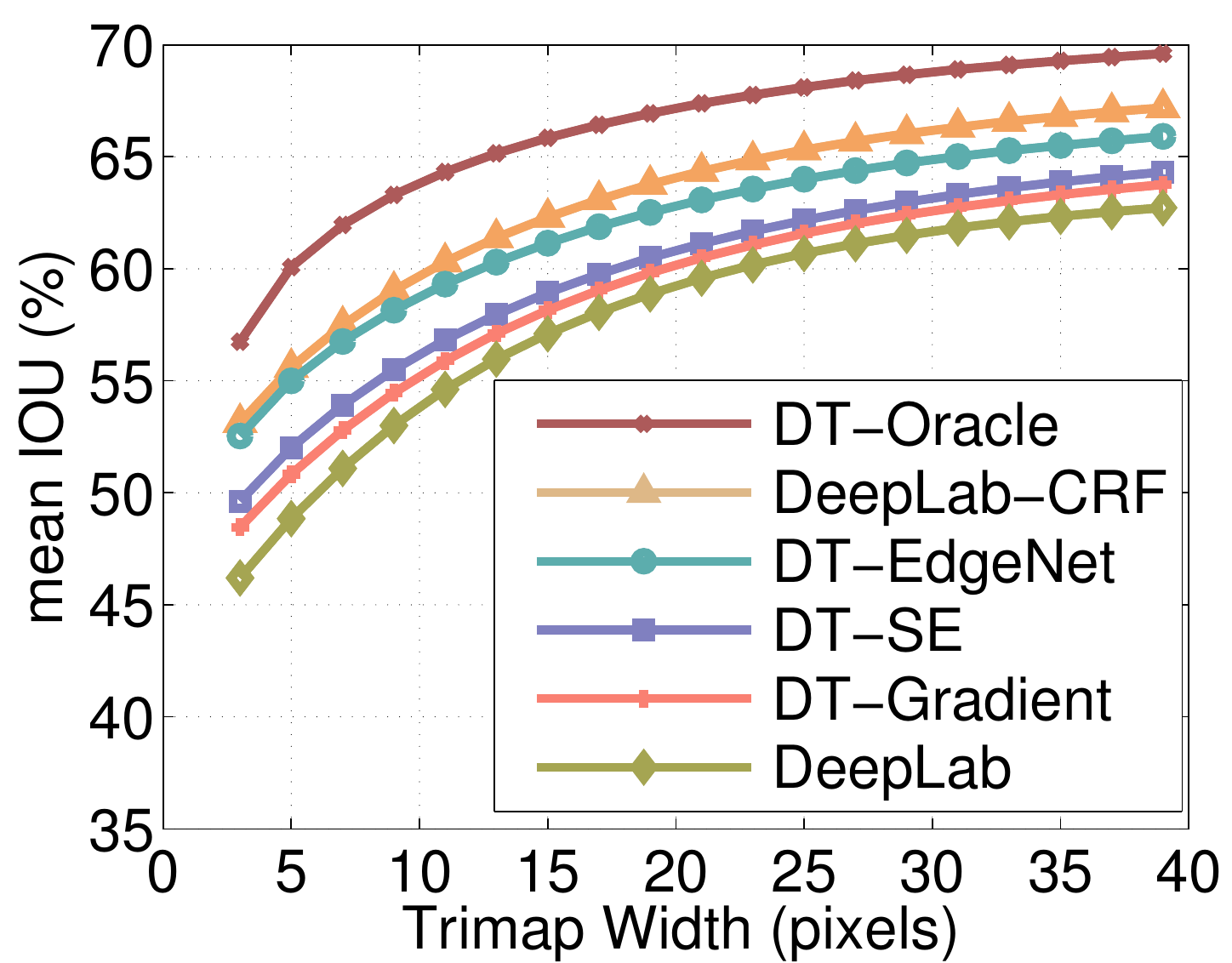} \\
    (a) & (b) \\
   \end{tabular}
}
  \caption{(a) Some trimap examples (top-left: image. top-right:
    ground-truth. bottom-left: trimap of 2 pixels. bottom-right:
    trimap of 10 pixels). (b) Segmentation result within a band around
    the object boundaries for the proposed methods (mean IOU).}  
  \label{fig:IOUBoundary}
\end{figure}

\paragraph{Trimap} Similar to \cite{kohli2009robust,
  krahenbuhl2011efficient, chen2014semantic}, we quantify the accuracy of the proposed
model near object boundaries. We use the ``void'' label annotated on
PASCAL VOC 2012 validation set. The annotations usually correspond to
object boundaries. We compute the mean IOU for the pixels that lie
within a narrow band (called trimap) of ``void'' labels, and vary the
width of the band, as shown in \figref{fig:IOUBoundary}.

\paragraph{Qualitative results} We show some semantic segmentation
results on PASCAL VOC 2012 val set in
\figref{fig:pascal_voc12_results}. DT-EdgeNet visually improves over
the baseline DeepLab and DT-SE. Besides, when comparing the
edges learned by Structured Edges and our EdgeNet, we found that
EdgeNet better captures the object exterior boundaries and responds
less than SE to interior edges. We also show failure cases in the
bottom two rows of \figref{fig:pascal_voc12_results}. The first is due
to the wrong predictions from DeepLab, and the second due to
the difficulty in localizing object boundaries with cluttered
background.

\paragraph{Test set results} After finding the best hyper-parameters,
we evaluate our models on the {\it test} set. As shown in the top of
\tabref{tab:testset}, DT-SE improves $2.7\%$ over the baseline
DeepLab, and DT-EdgeNet can further enhance the performance
to $69.0\%$ ($3.9\%$ better than baseline), which is $1.3\%$ behind
employing a fully-connected CRF as post-processing (\ie,
DeepLab-CRF) to smooth the results. However, if we also
incorporate a fully-connected CRF as post-processing to our model, we
can further increase performance to $71.2\%$.

\paragraph{Models pretrained with MS-COCO} We perform another
experiment with the stronger baseline of \cite{papandreou2015weakly},
where DeepLab is pretrained with the MS-COCO 2014 dataset
\cite{lin2014microsoft}. Our goal is to test if we can still obtain
improvements with the proposed methods over that stronger baseline. We
use the same optimal values of hyper-parameters as before, and report
the results on validation set in \tabref{tab:models_valset_coco}. We
still observe $1.6\%$ and $2.7\%$ improvement over the baseline by
DT-SE and DT-EdgeNet, respectively. Besides, adding a fully-connected
CRF to DT-EdgeNet can bring another $1.8\%$ improvement. We then evaluate
the models on test set in the bottom of \tabref{tab:testset}. Our best
model, DT-EdgeNet, improves the baseline DeepLab by $2.8\%$,
while it is $1.0\%$ lower than DeepLab-CRF. When combining
DT-EdgeNet and a fully-connected CRF, we achieve $73.6\%$ on the test
set. Note the gap between DT-EdgeNet and DeepLab-CRF becomes
smaller when stronger baseline is used.

\paragraph{Incorporating multi-scale inputs} State-of-art models on the PASCAL VOC 2012 leaderboard usually employ multi-scale features (either multi-scale inputs \cite{dai2015boxsup, lin2015efficient, chen2015attention} or features from intermediate layers of DCNN \cite{long2014fully, hariharan2014hypercolumns, chen2014semantic}). Motivated by this, we further combine our proposed discriminatively trained domain transform and the model of \cite{chen2015attention}, yielding $76.3\%$ performance on test set, $1.5\%$ behind current best models \cite{lin2015efficient} which jointly train CRF and DCNN \cite{chen2014learning}

\begin{figure}
  \centering
  \includegraphics[width=0.5\linewidth]{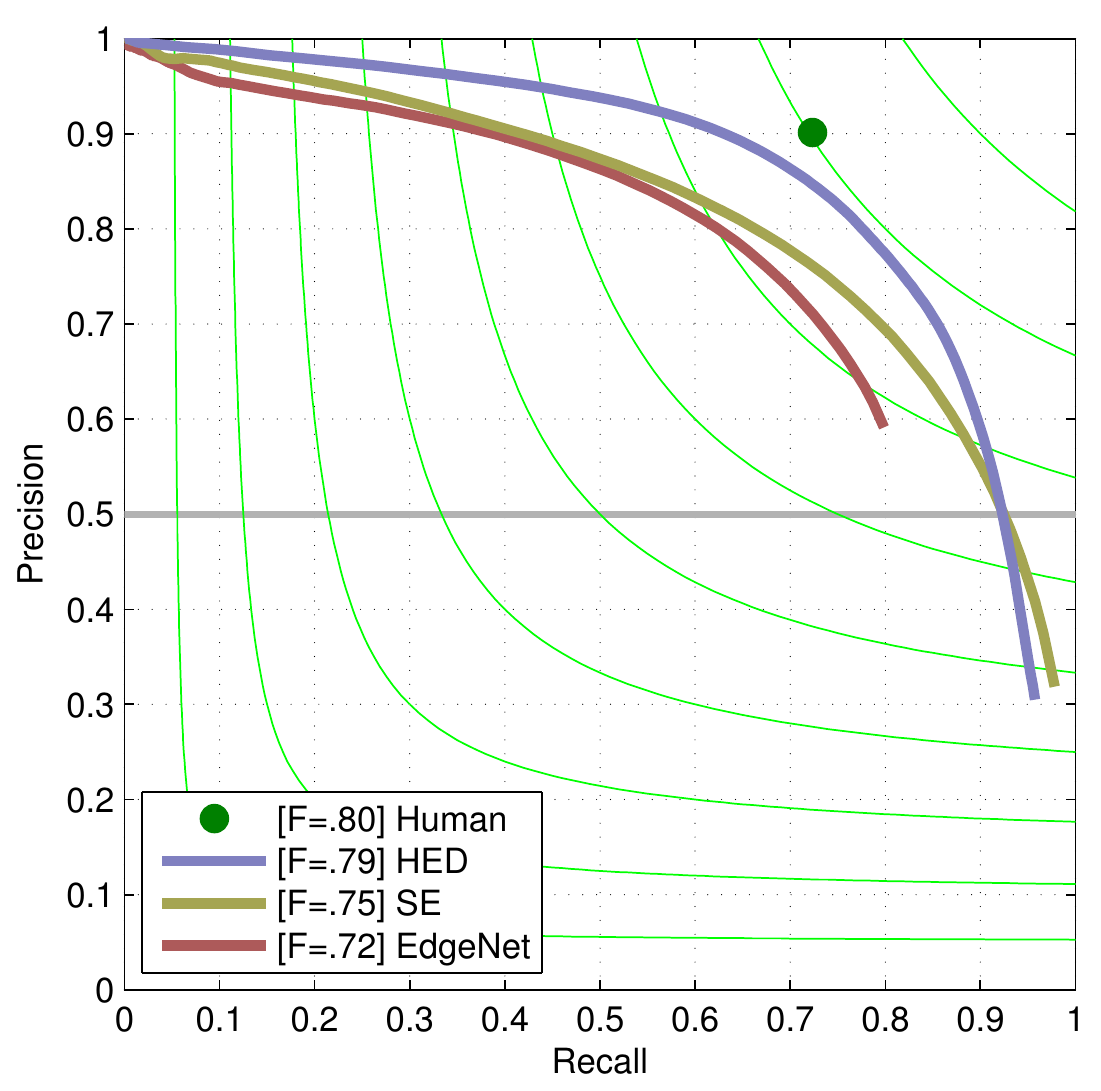}
  \caption{Evaluation of our learned EdgeNet on the test set of
    BSDS500. Note that our EdgeNet is only trained on PASCAL VOC 2012
    semantic segmentation task without edge supervision.}
  \label{fig:bsds500}
\end{figure}

\begin{figure*}
  \centering
  \begin{tabular}{c c c c c c}
    \includegraphics[height=0.1\linewidth]{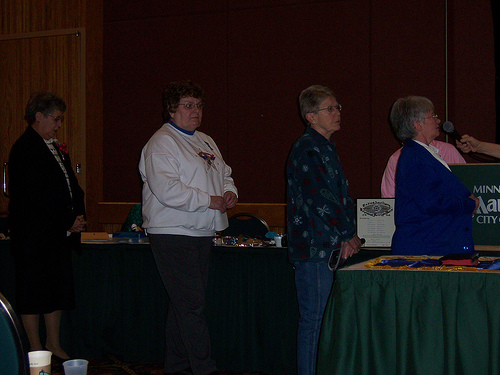} &
    \includegraphics[height=0.1\linewidth]{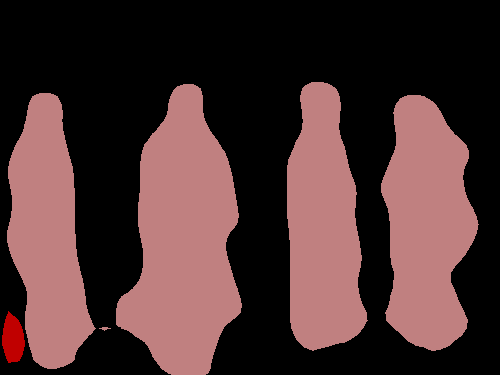} &
    \includegraphics[height=0.1\linewidth]{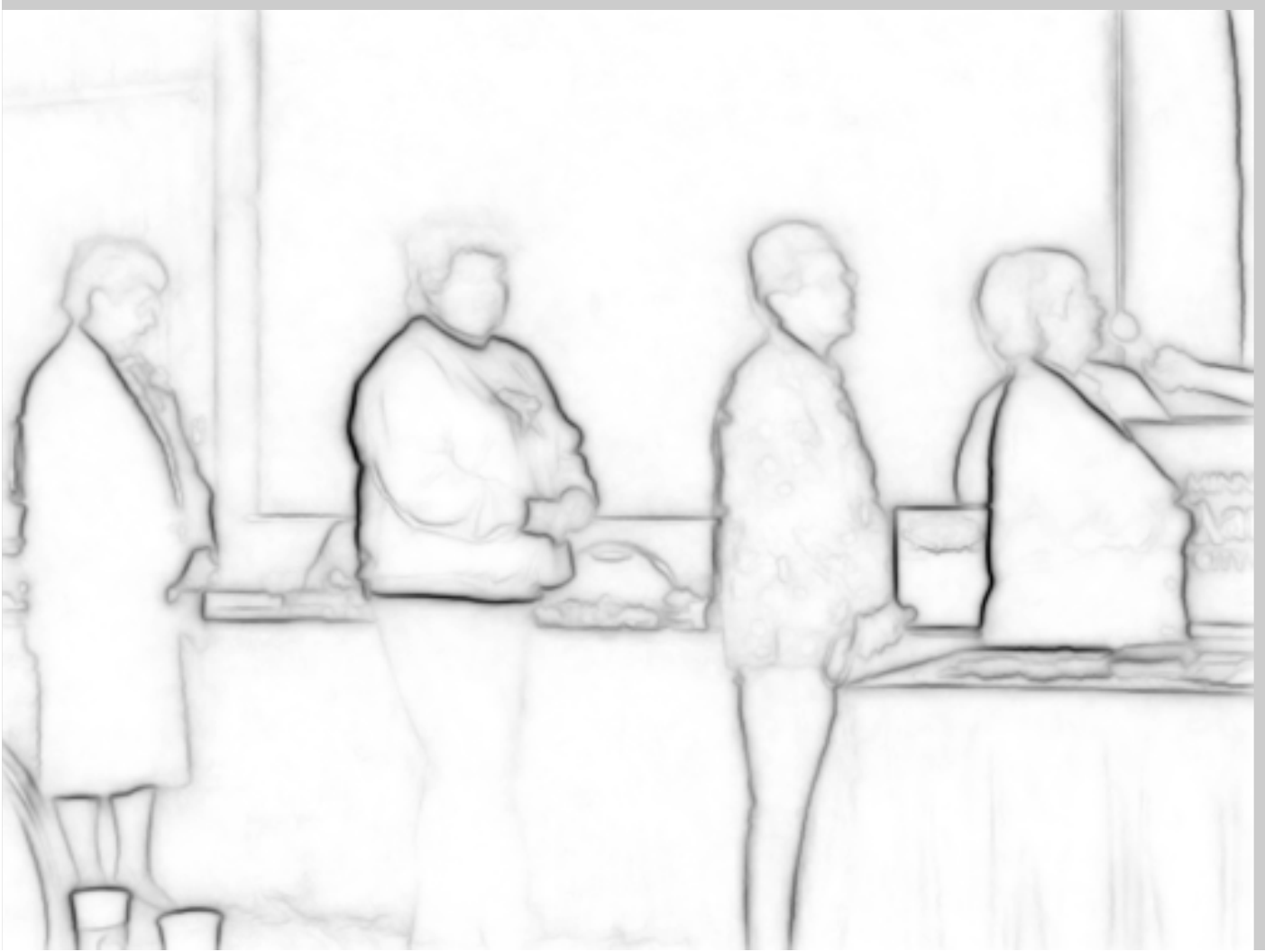} &
    \includegraphics[height=0.1\linewidth]{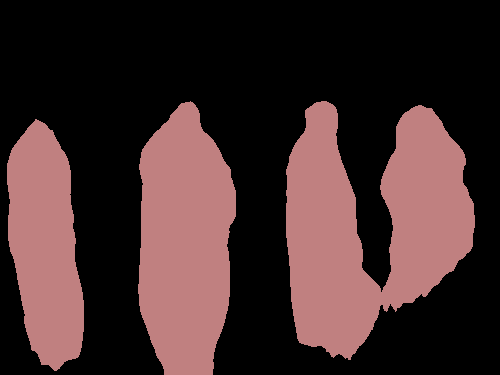} &
    \includegraphics[height=0.1\linewidth]{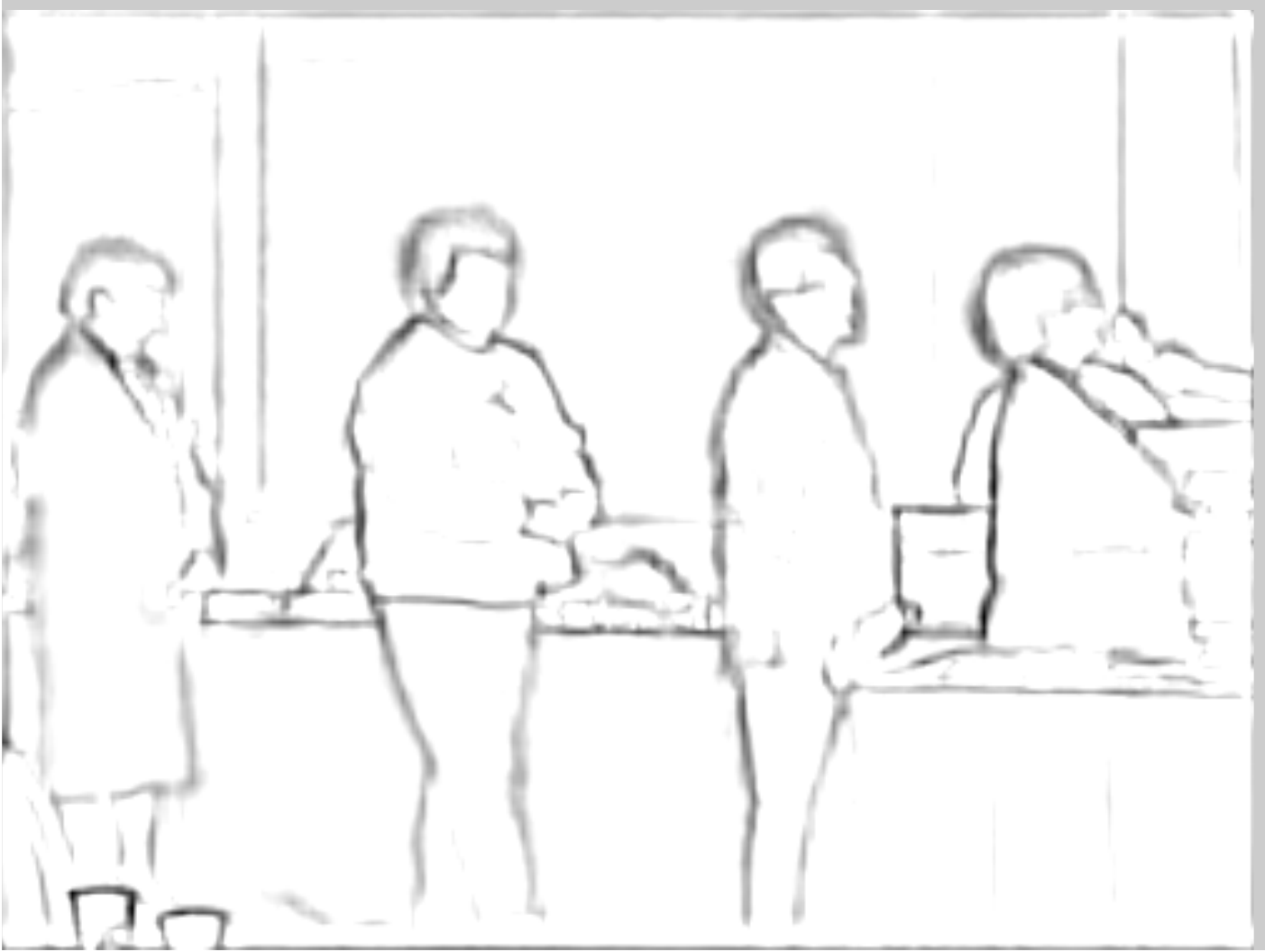} &
    \includegraphics[height=0.1\linewidth]{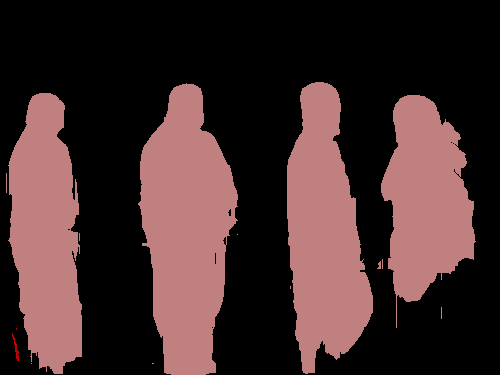} \\

    \includegraphics[height=0.1\linewidth]{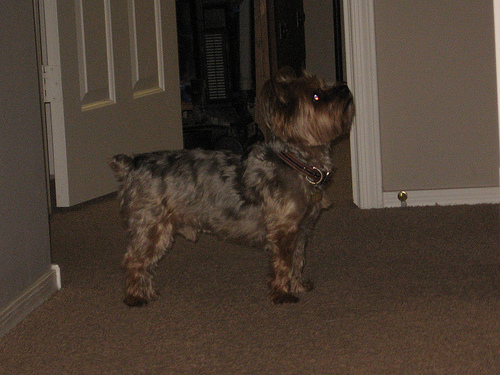} &
    \includegraphics[height=0.1\linewidth]{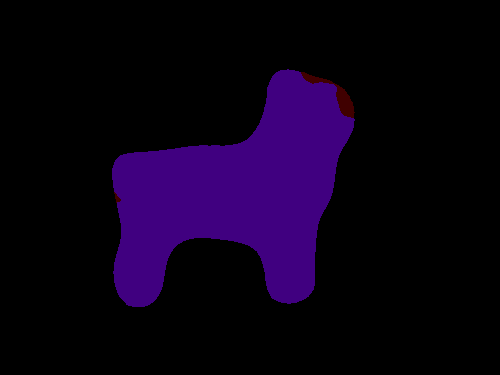} &
    \includegraphics[height=0.1\linewidth]{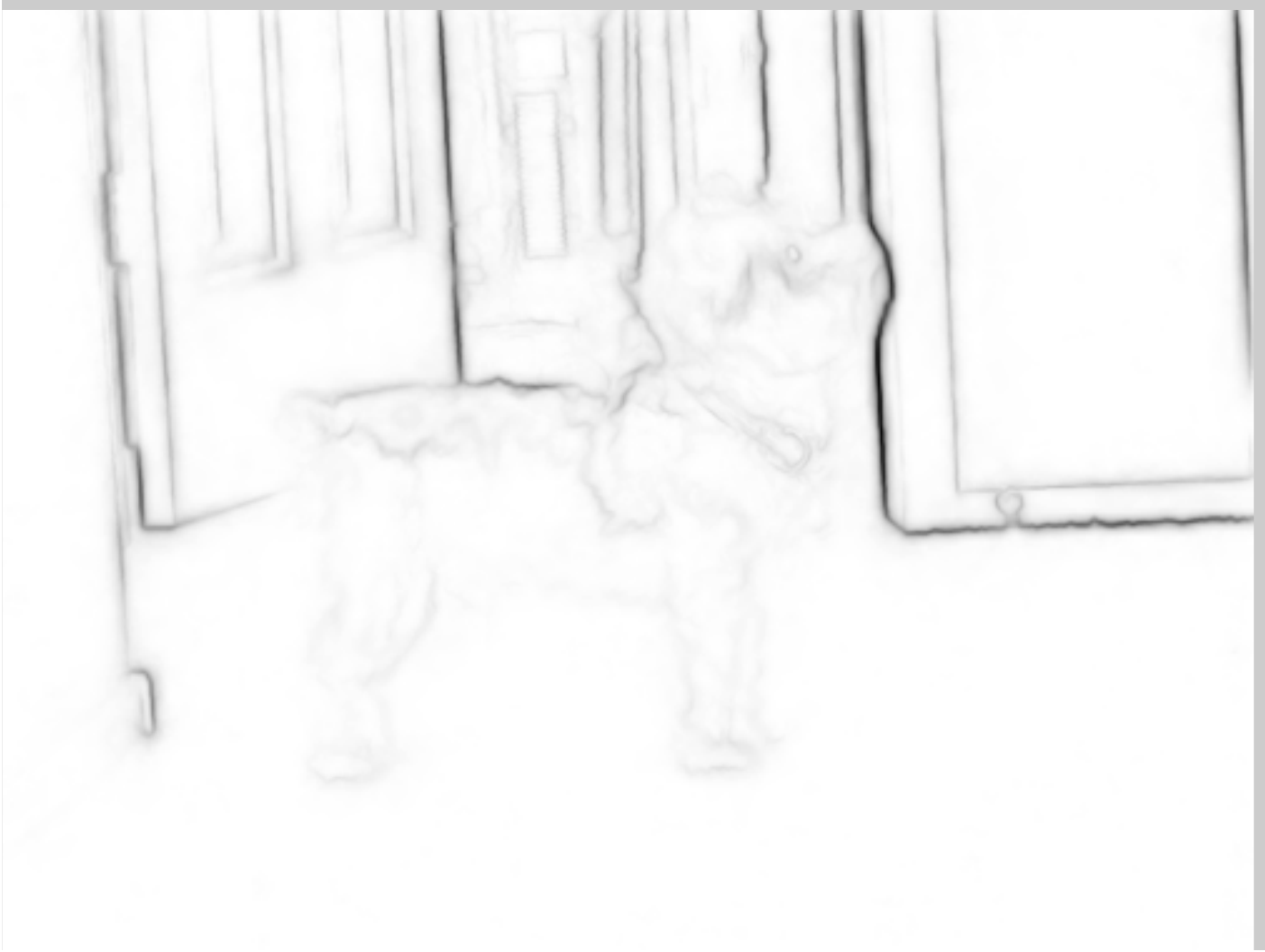} &
    \includegraphics[height=0.1\linewidth]{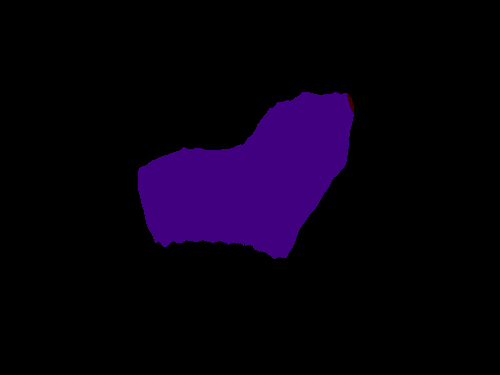} &
    \includegraphics[height=0.1\linewidth]{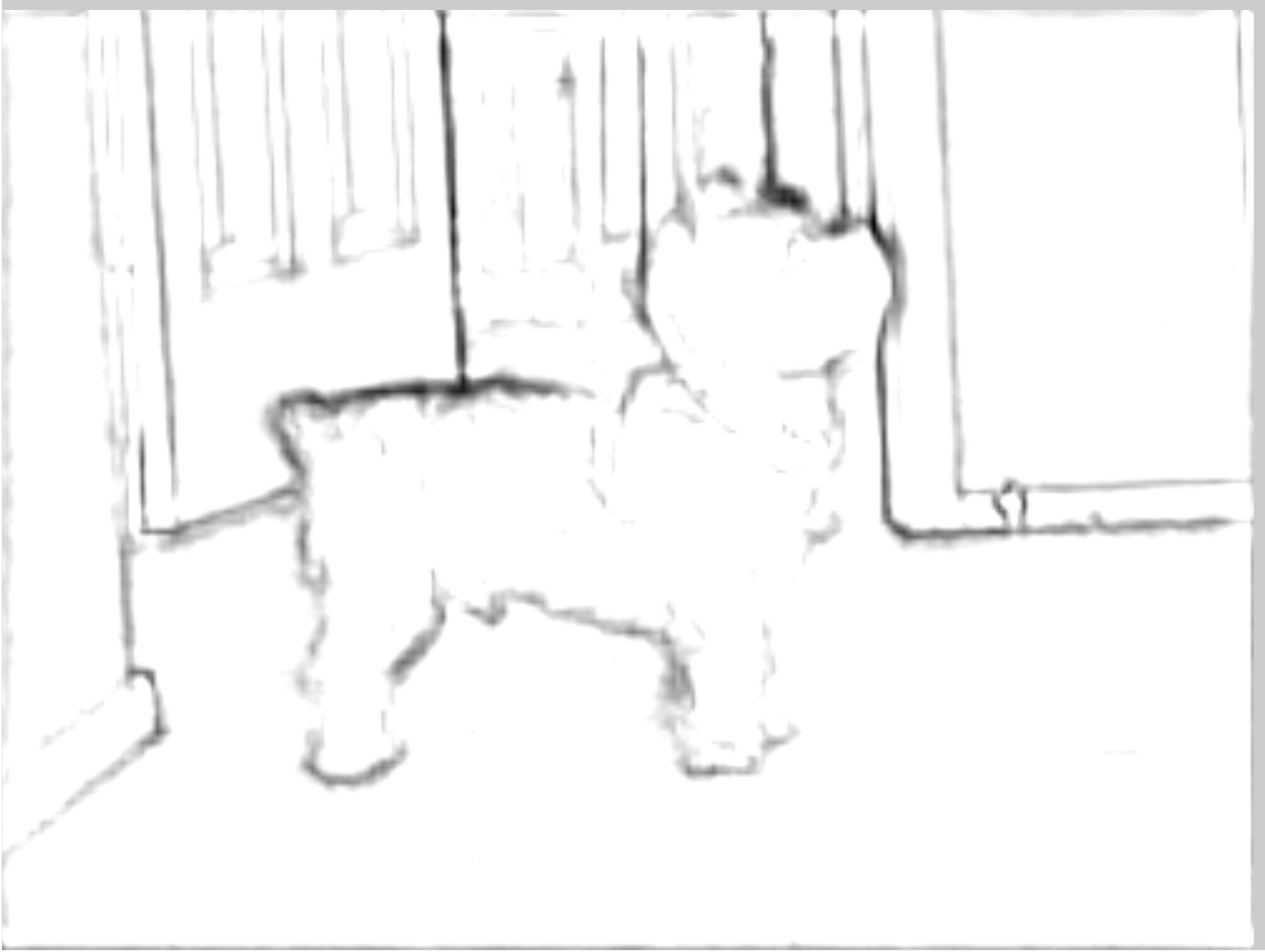} &
    \includegraphics[height=0.1\linewidth]{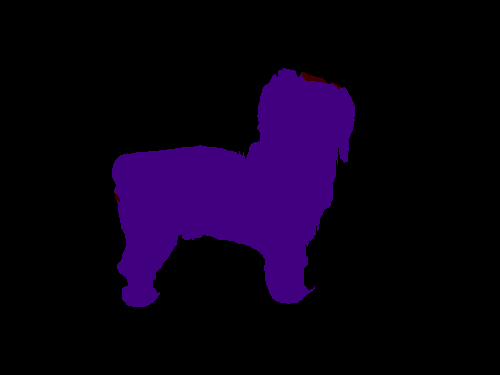} \\

    \includegraphics[height=0.1\linewidth]{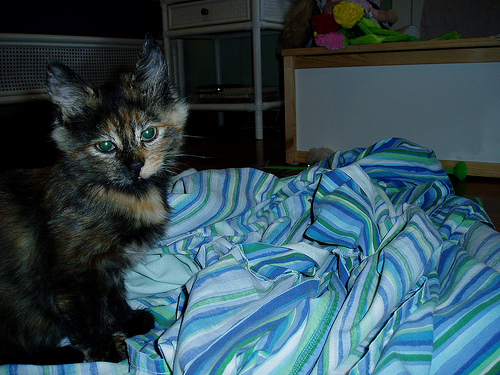} &
    \includegraphics[height=0.1\linewidth]{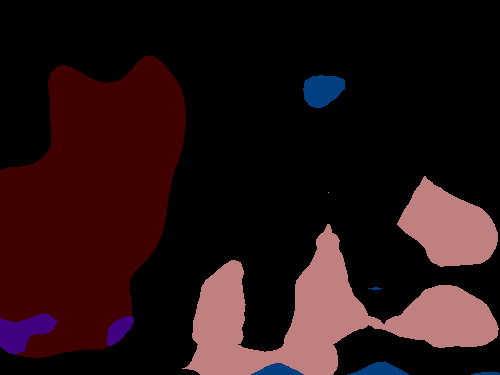} &
    \includegraphics[height=0.1\linewidth]{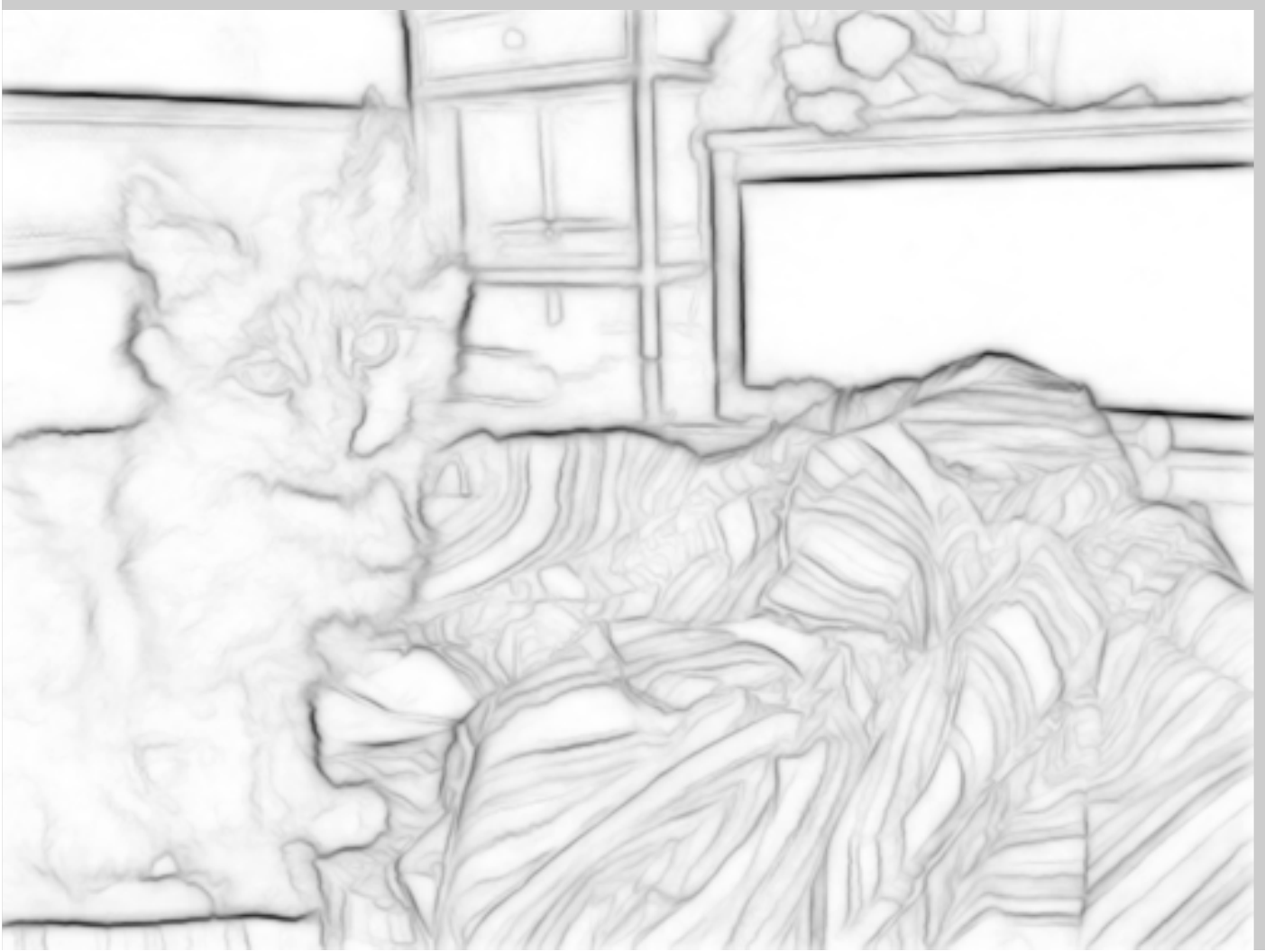} &
    \includegraphics[height=0.1\linewidth]{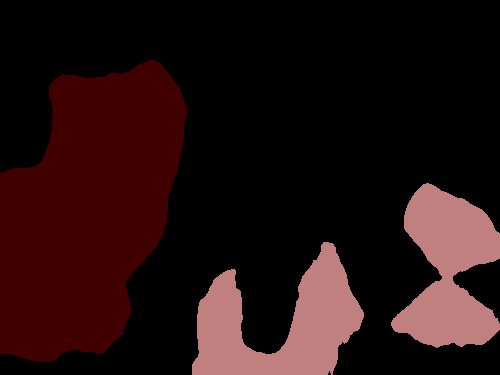} &
    \includegraphics[height=0.1\linewidth]{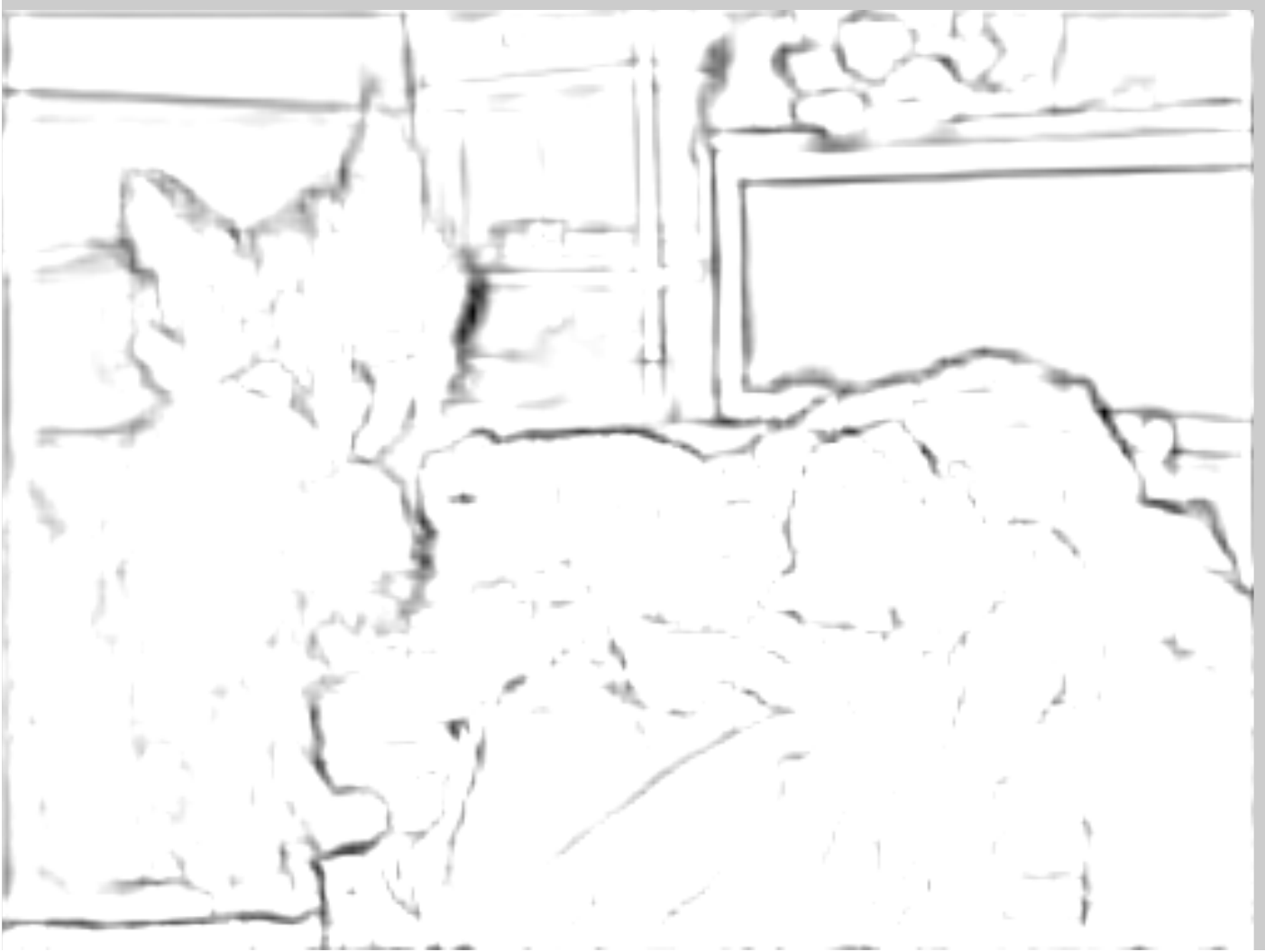} &
    \includegraphics[height=0.1\linewidth]{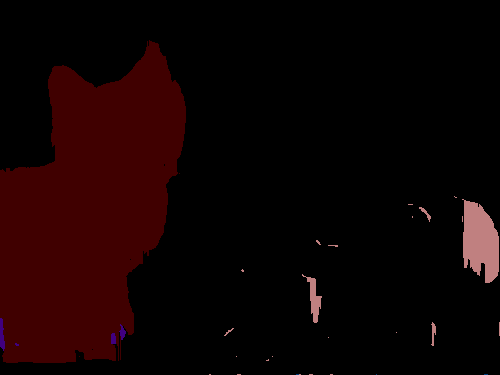} \\

    \includegraphics[height=0.105\linewidth]{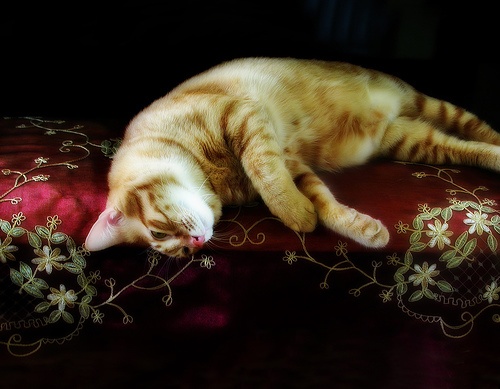} &
    \includegraphics[height=0.105\linewidth]{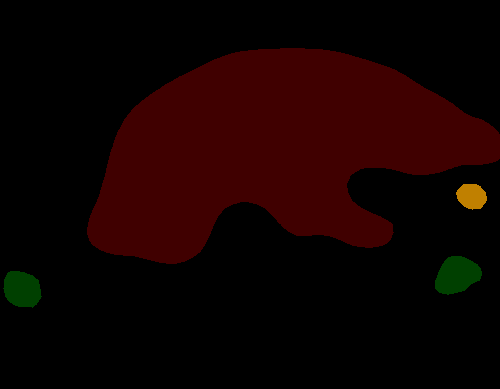} &
    \includegraphics[height=0.105\linewidth]{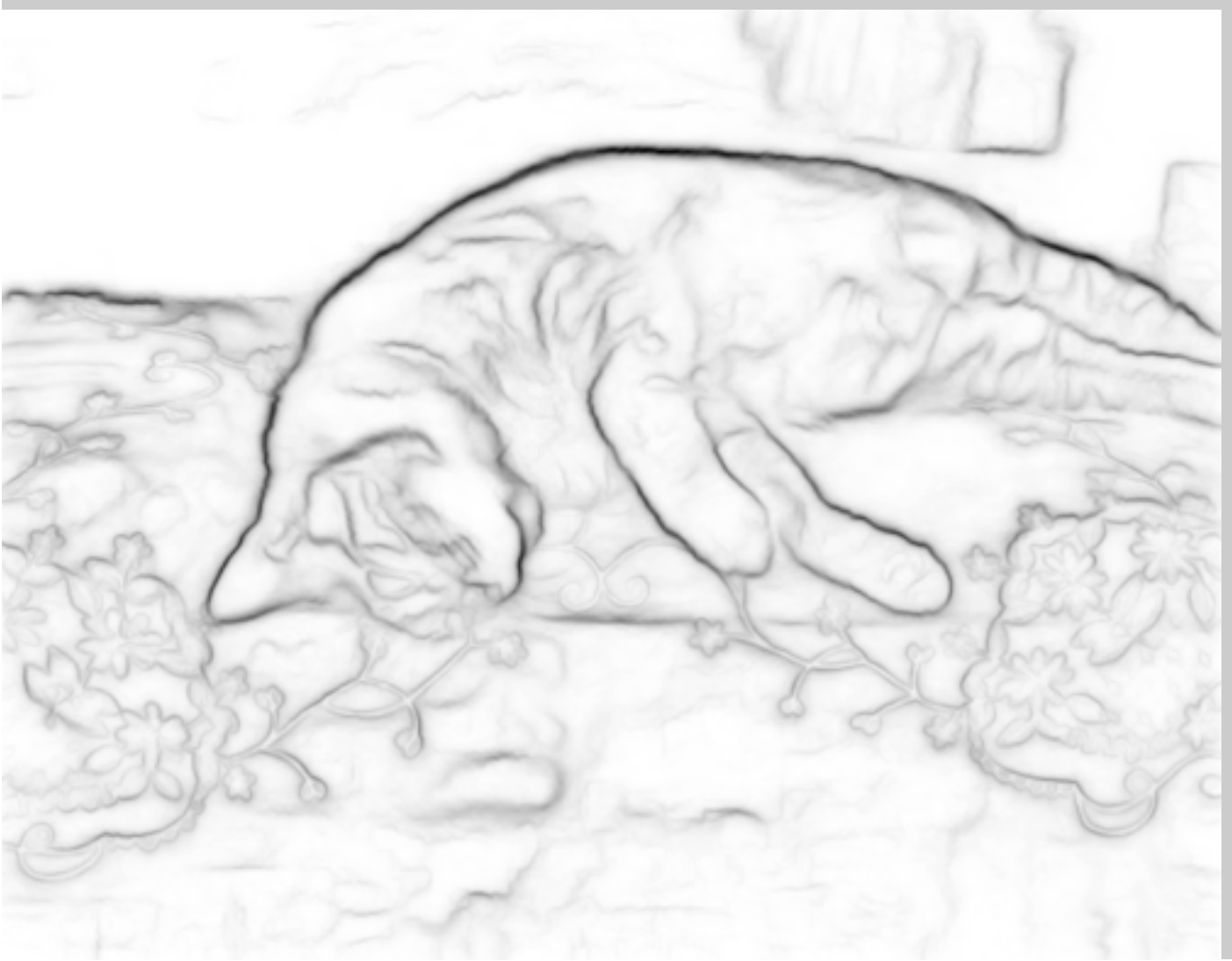} &
    \includegraphics[height=0.105\linewidth]{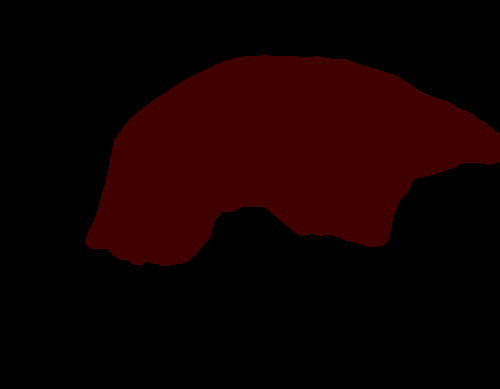} &
    \includegraphics[height=0.105\linewidth]{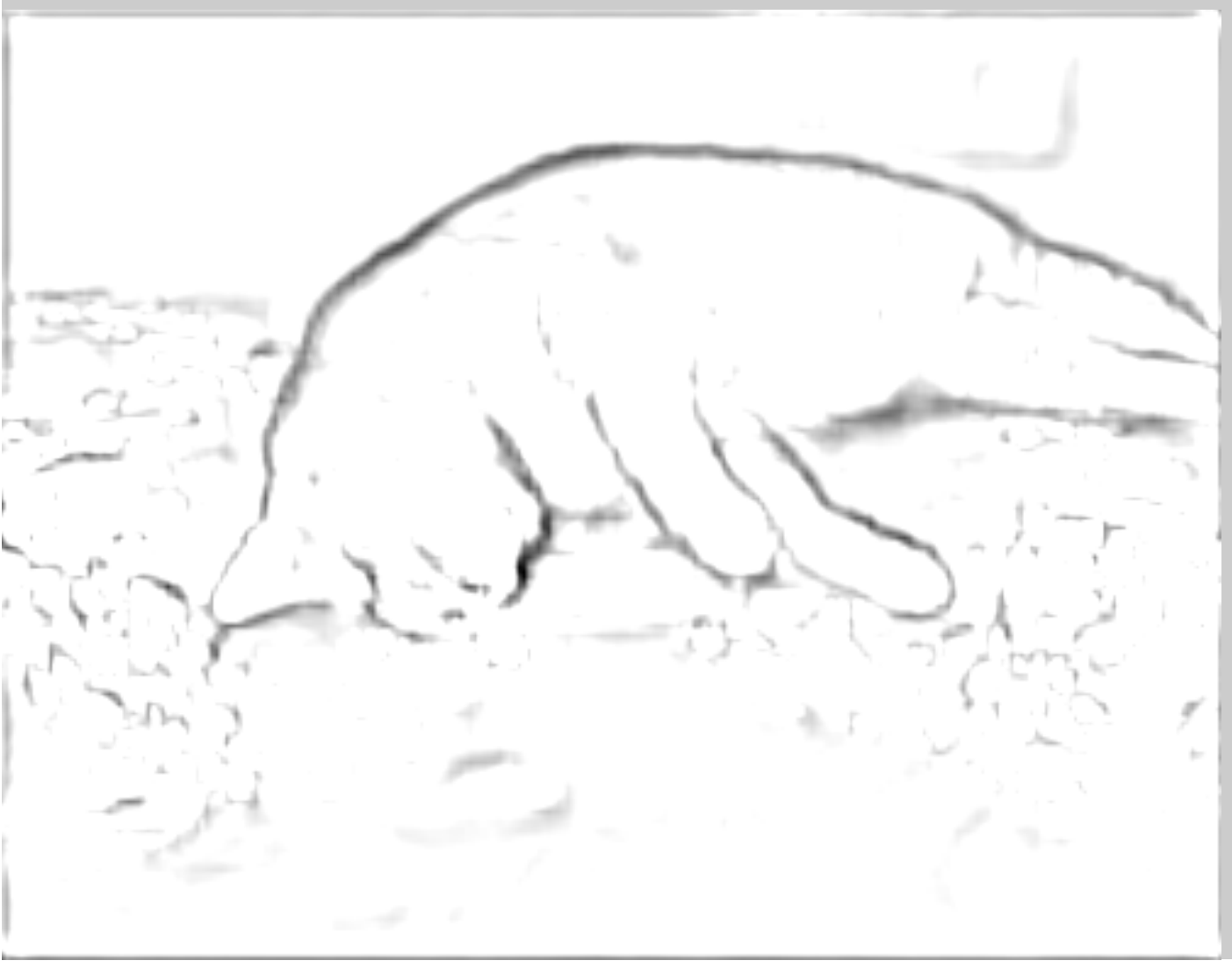} &
    \includegraphics[height=0.105\linewidth]{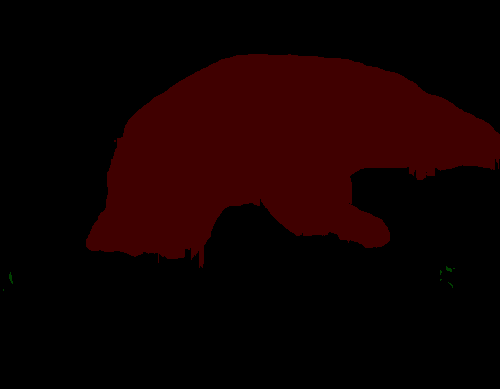} \\

    \includegraphics[height=0.058\linewidth]{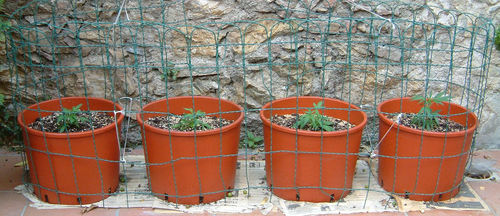} &
    \includegraphics[height=0.058\linewidth]{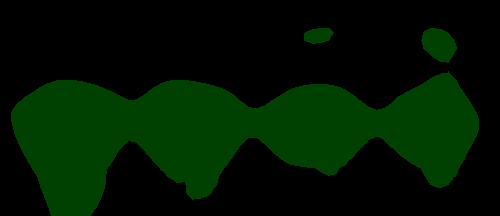} &
    \includegraphics[height=0.058\linewidth]{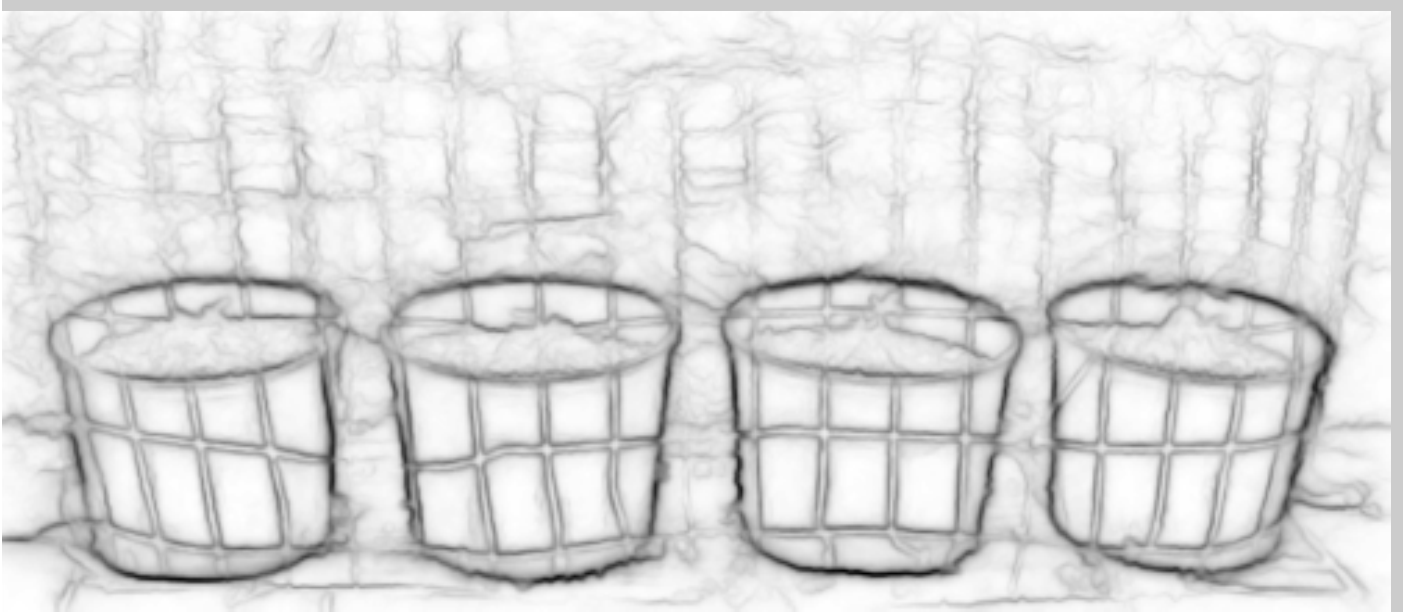} &
    \includegraphics[height=0.058\linewidth]{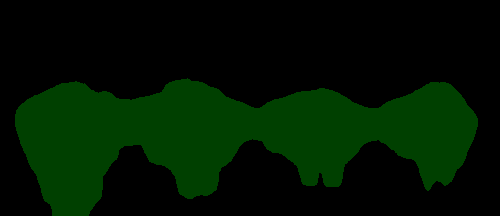} &
    \includegraphics[height=0.058\linewidth]{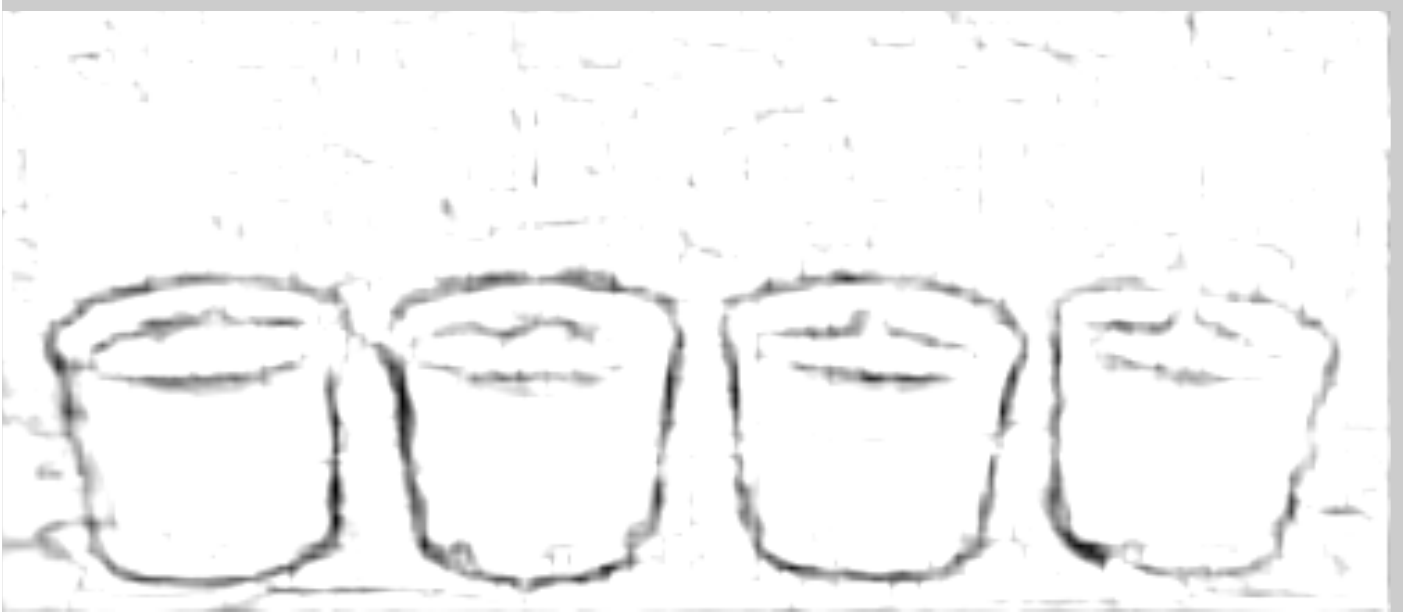} &
    \includegraphics[height=0.058\linewidth]{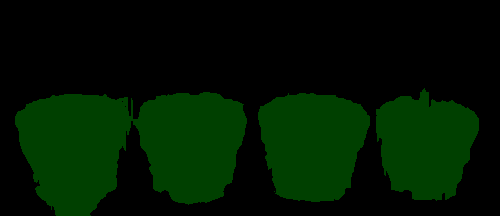} \\

    \includegraphics[height=0.1\linewidth]{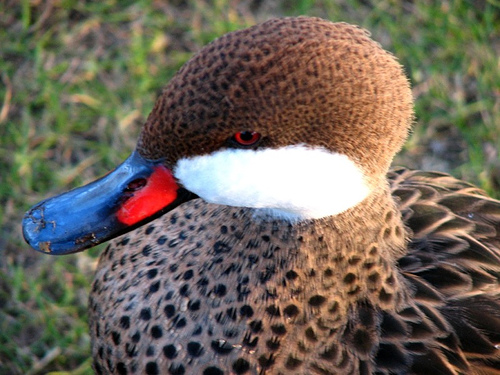} &
    \includegraphics[height=0.1\linewidth]{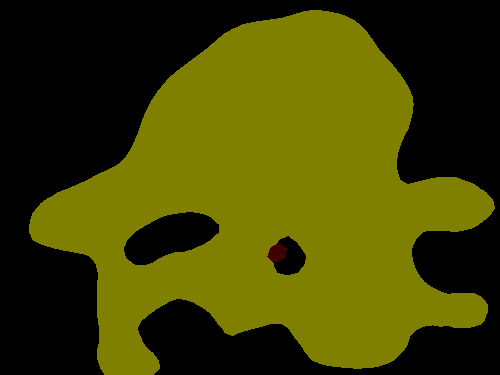} &
    \includegraphics[height=0.1\linewidth]{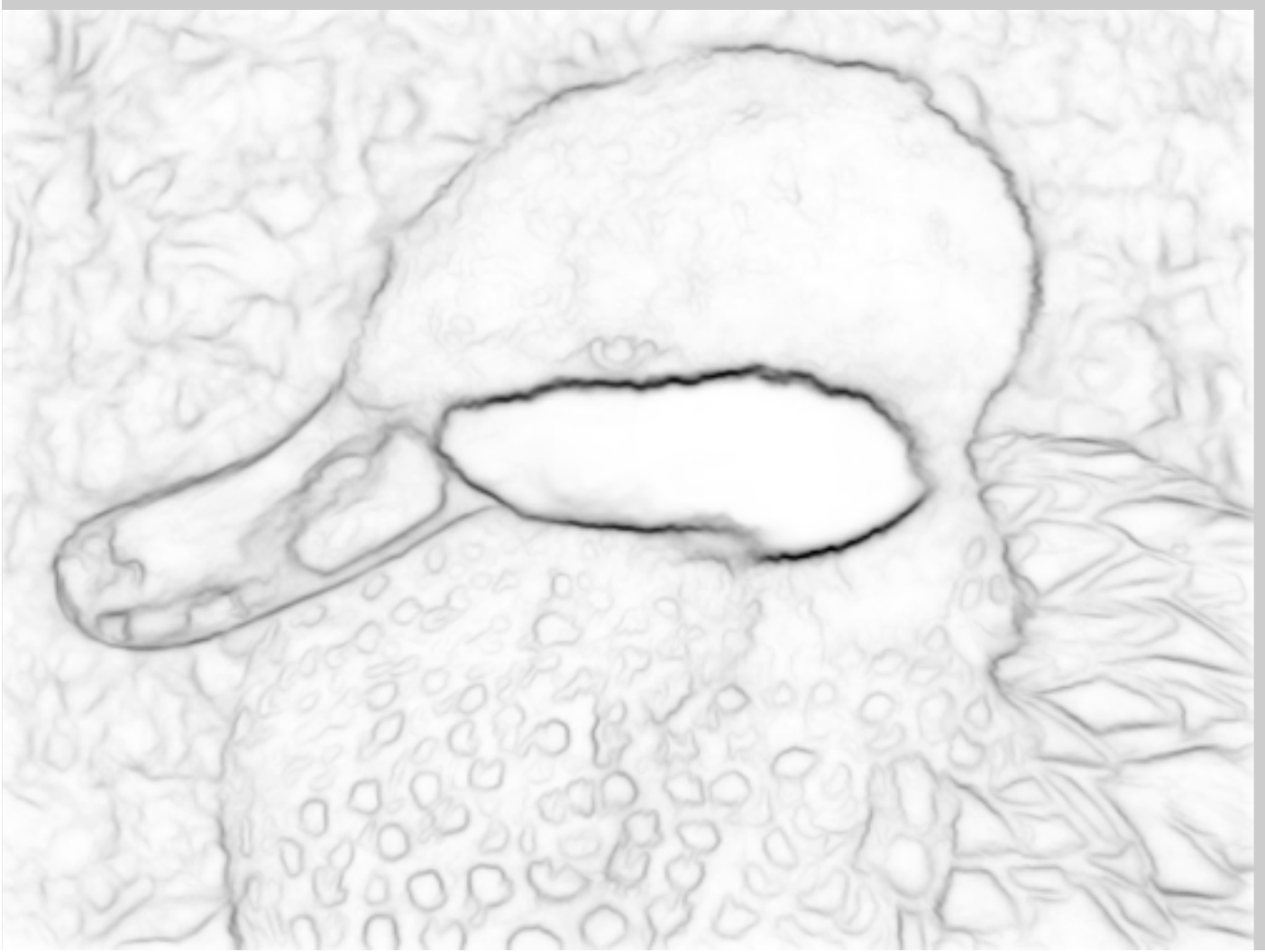} &
    \includegraphics[height=0.1\linewidth]{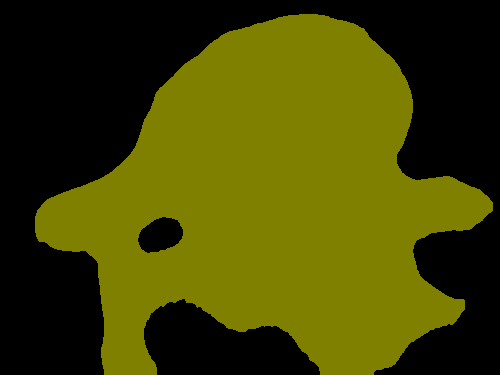} &
    \includegraphics[height=0.1\linewidth]{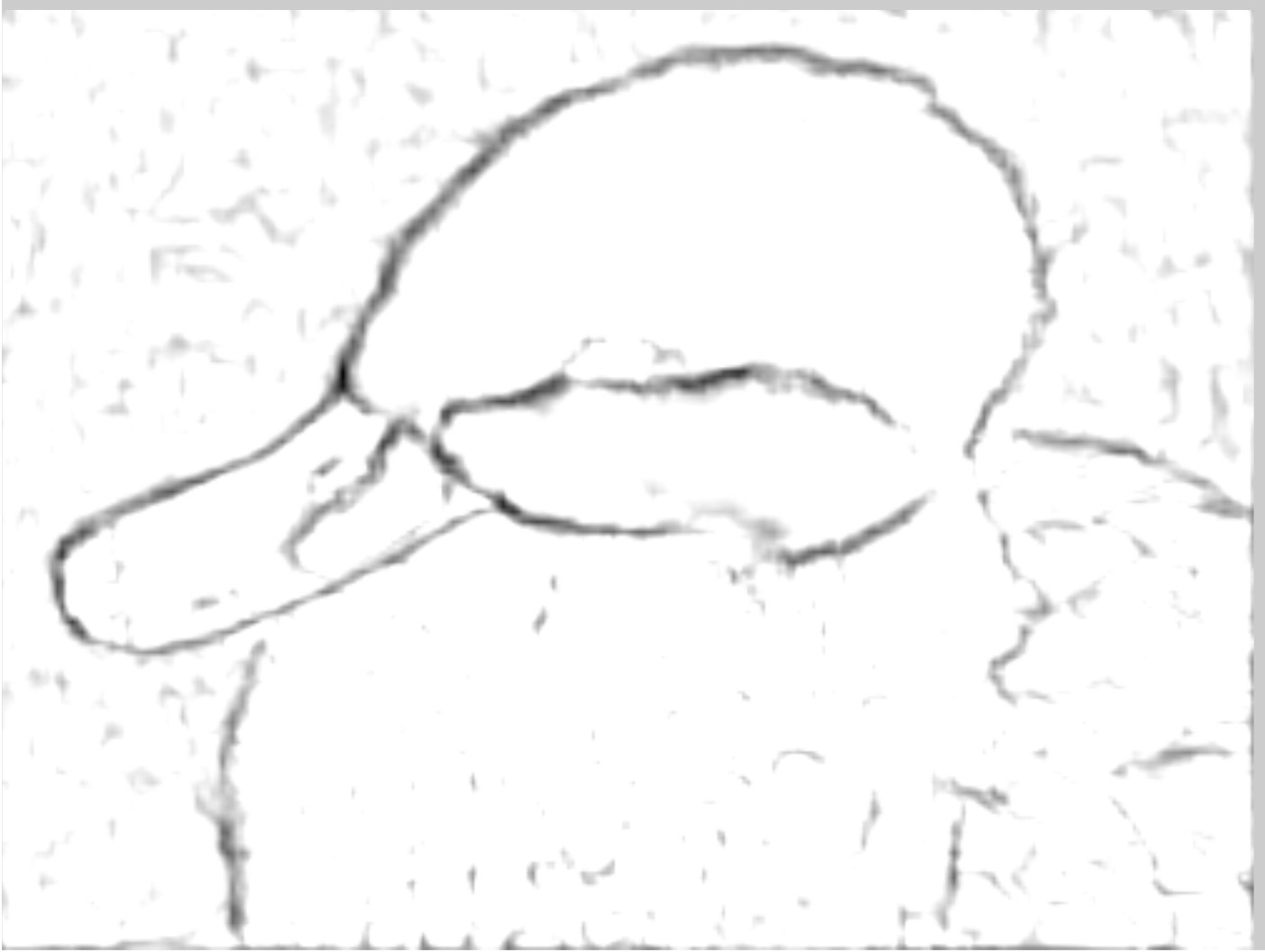} &
    \includegraphics[height=0.1\linewidth]{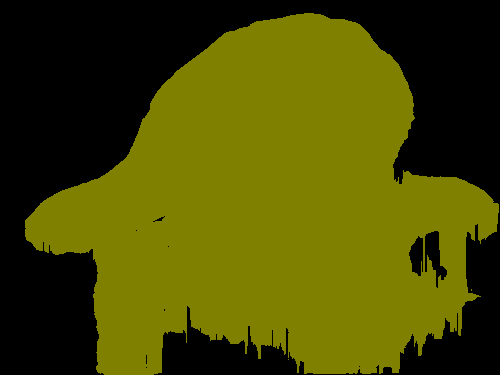} \\

    \includegraphics[height=0.096\linewidth]{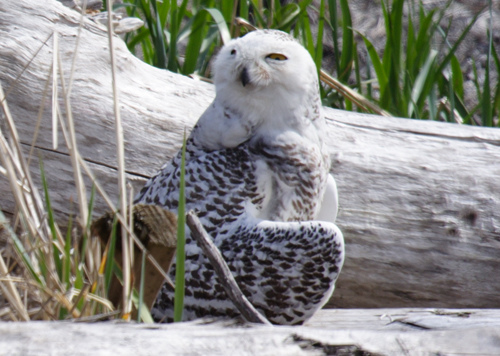} &
    \includegraphics[height=0.096\linewidth]{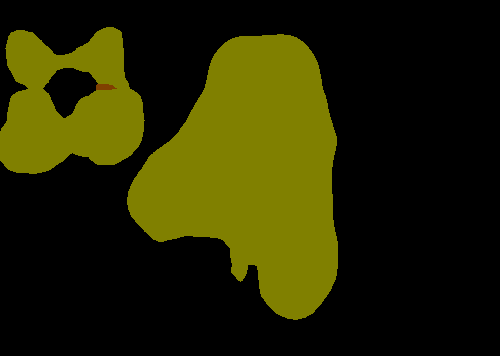} &
    \includegraphics[height=0.096\linewidth]{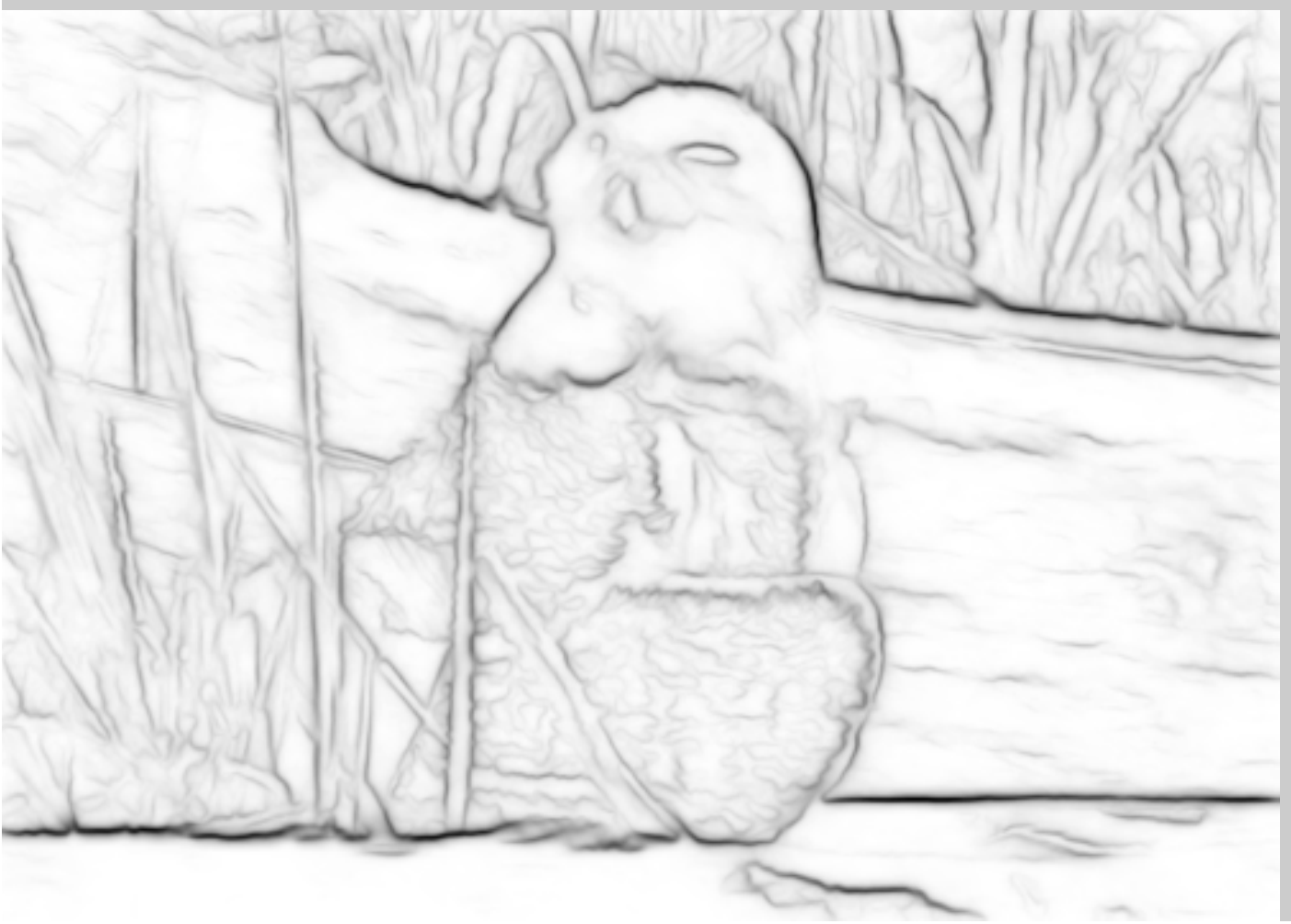} &
    \includegraphics[height=0.096\linewidth]{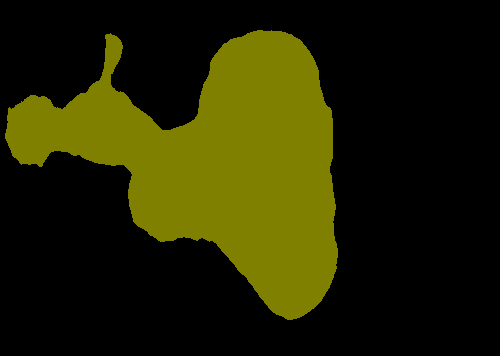} &
    \includegraphics[height=0.096\linewidth]{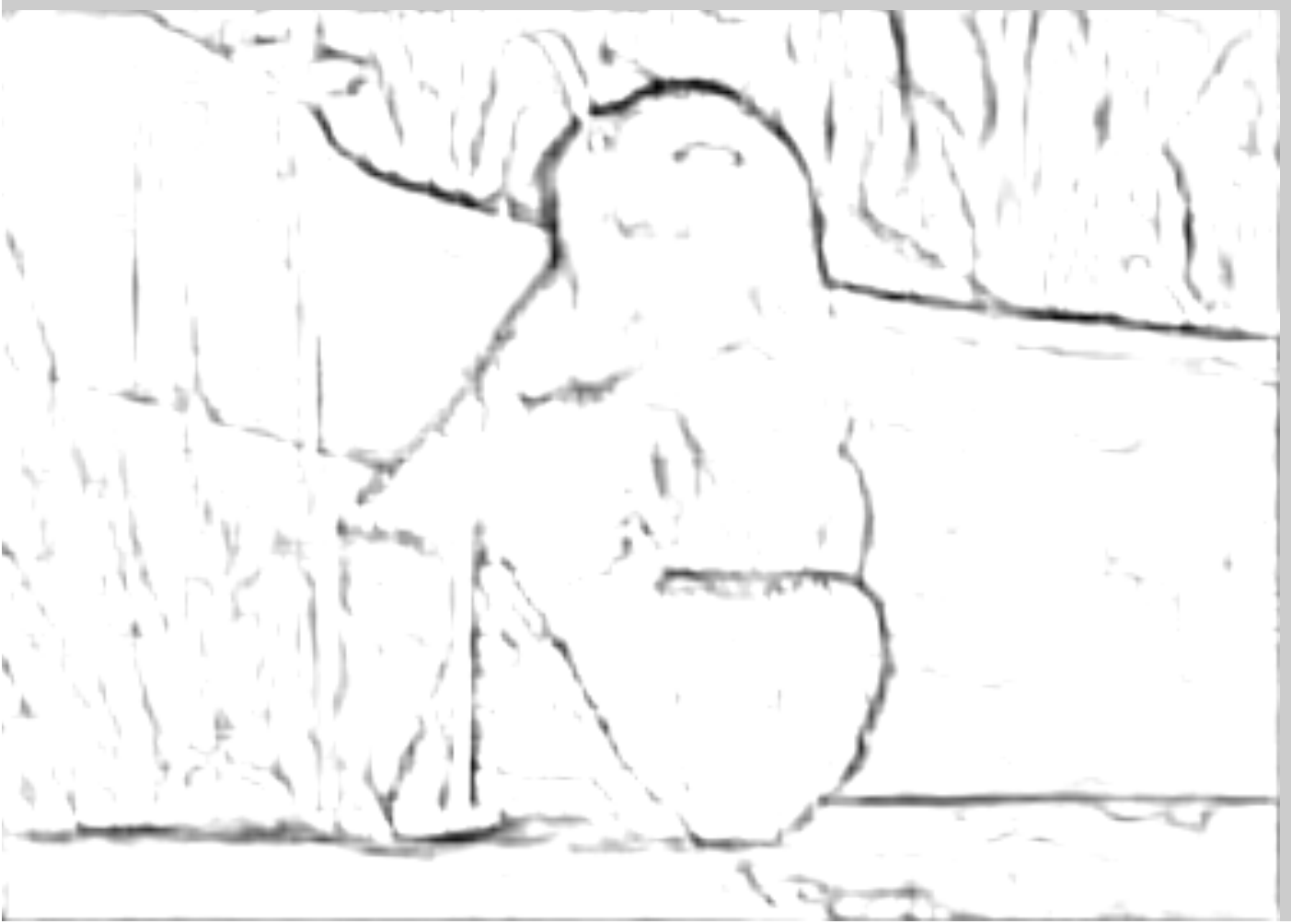} &
    \includegraphics[height=0.096\linewidth]{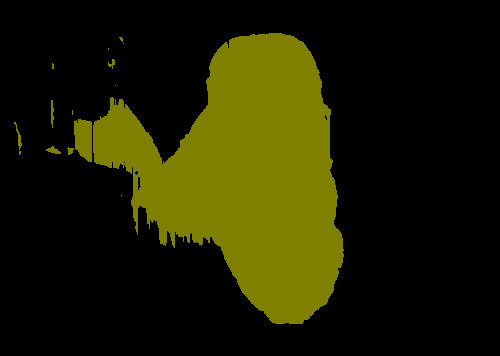} \\

    \includegraphics[height=0.12\linewidth]{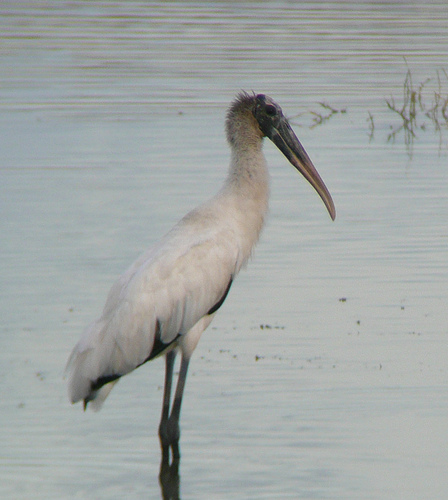} &
    \includegraphics[height=0.12\linewidth]{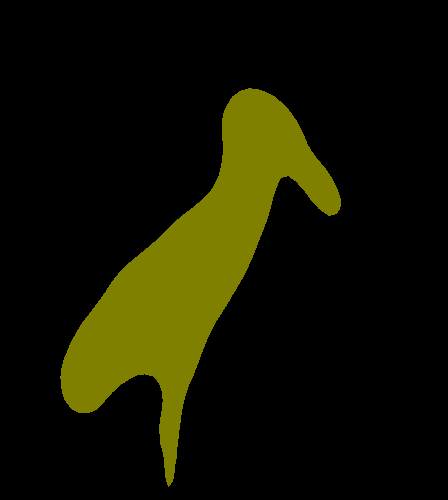} &
    \includegraphics[height=0.12\linewidth]{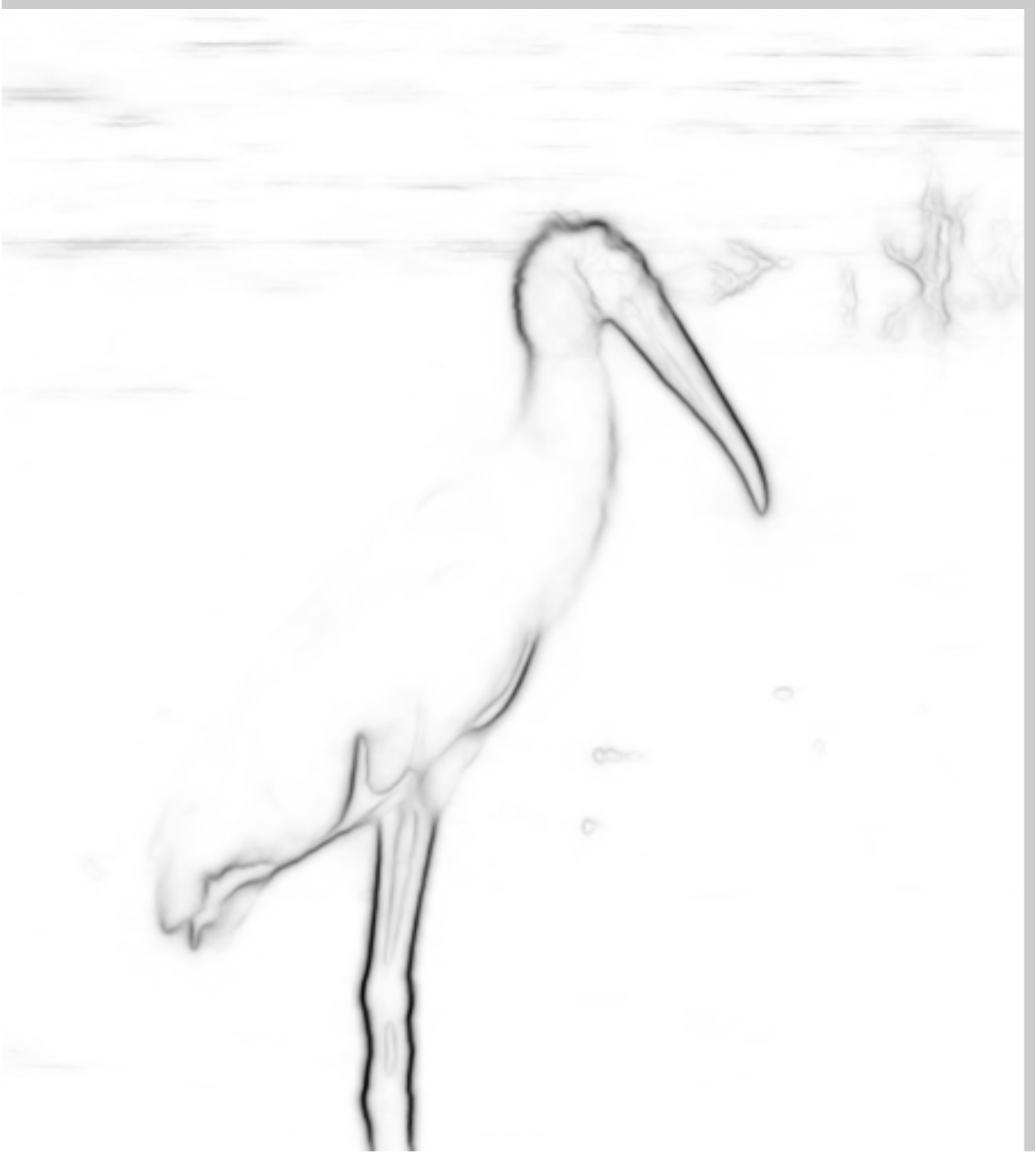} &
    \includegraphics[height=0.12\linewidth]{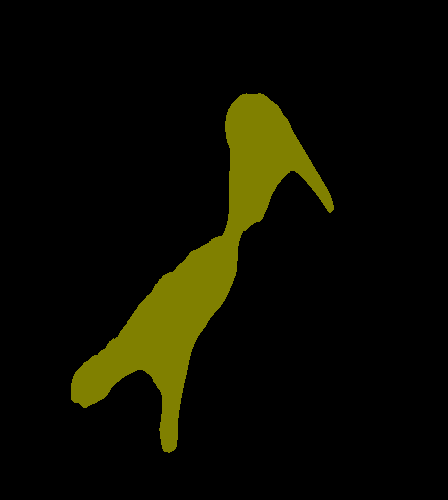} &
    \includegraphics[height=0.12\linewidth]{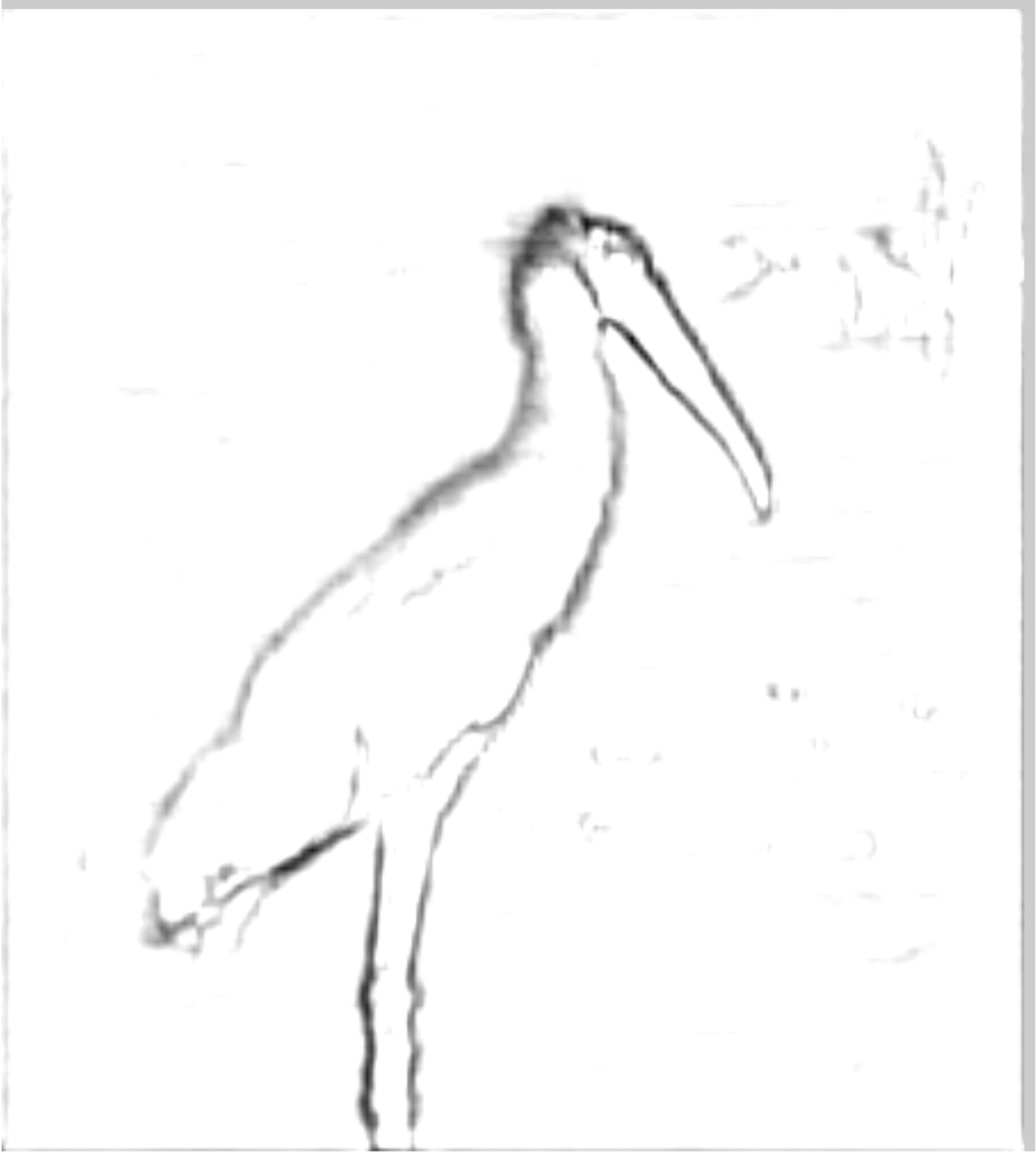} &
    \includegraphics[height=0.12\linewidth]{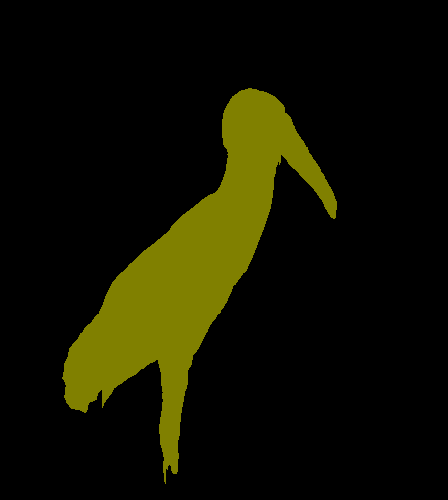} \\

    \includegraphics[height=0.09\linewidth]{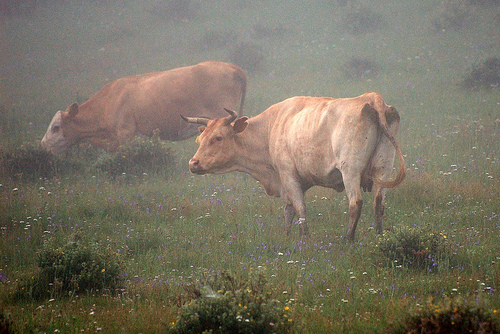} &
    \includegraphics[height=0.09\linewidth]{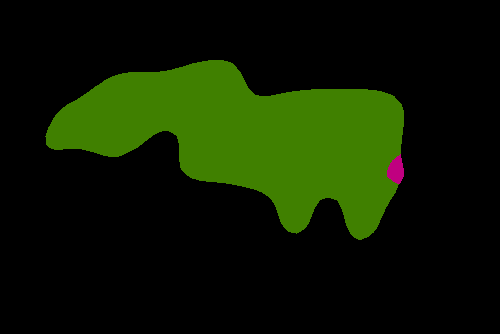} &
    \includegraphics[height=0.09\linewidth]{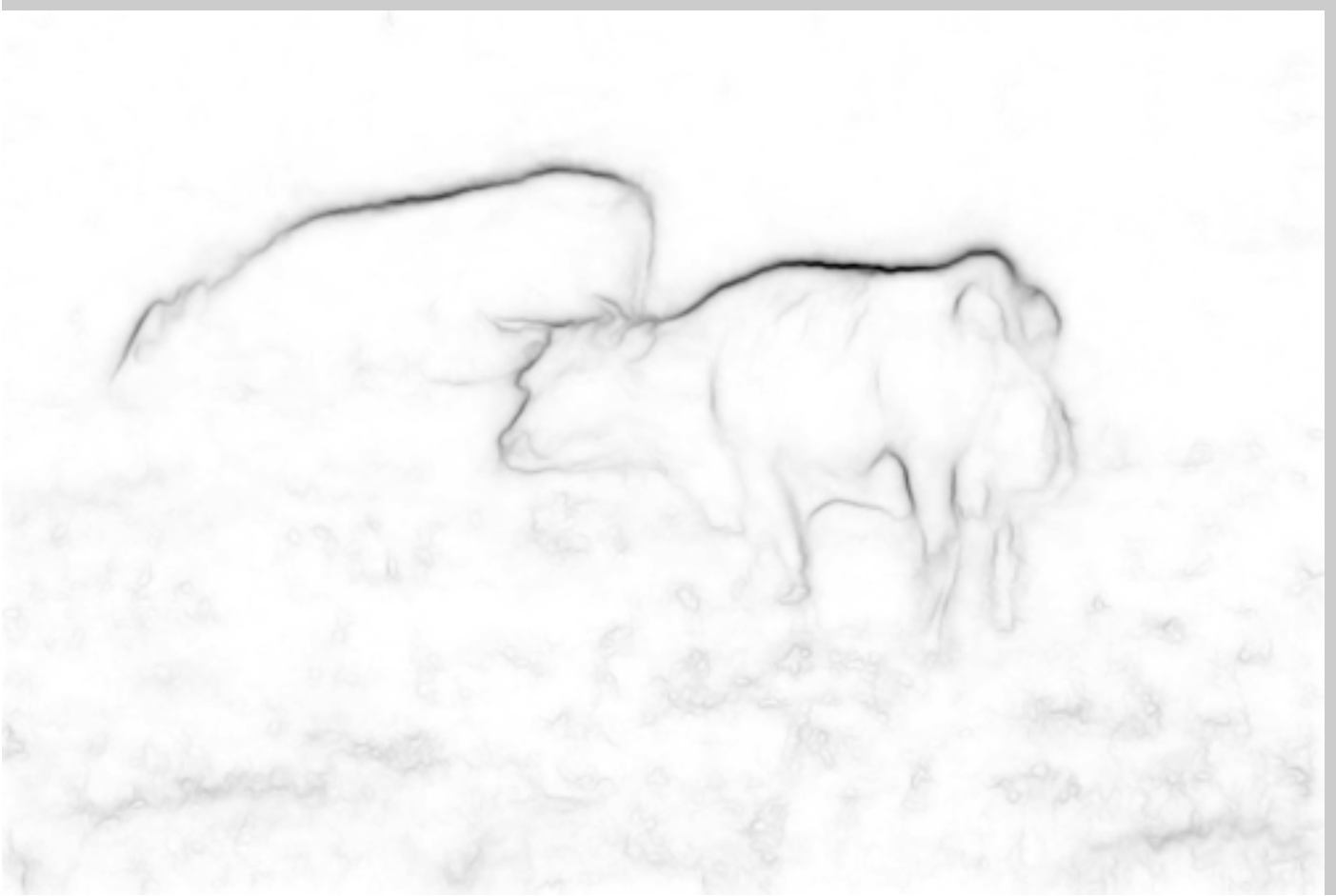} &
    \includegraphics[height=0.09\linewidth]{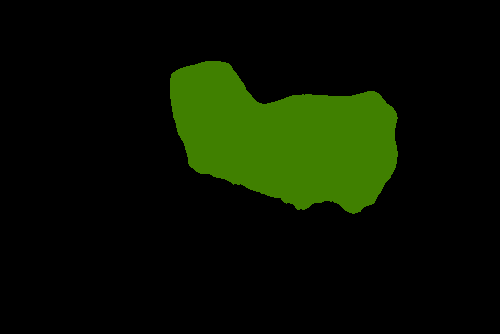} &
    \includegraphics[height=0.09\linewidth]{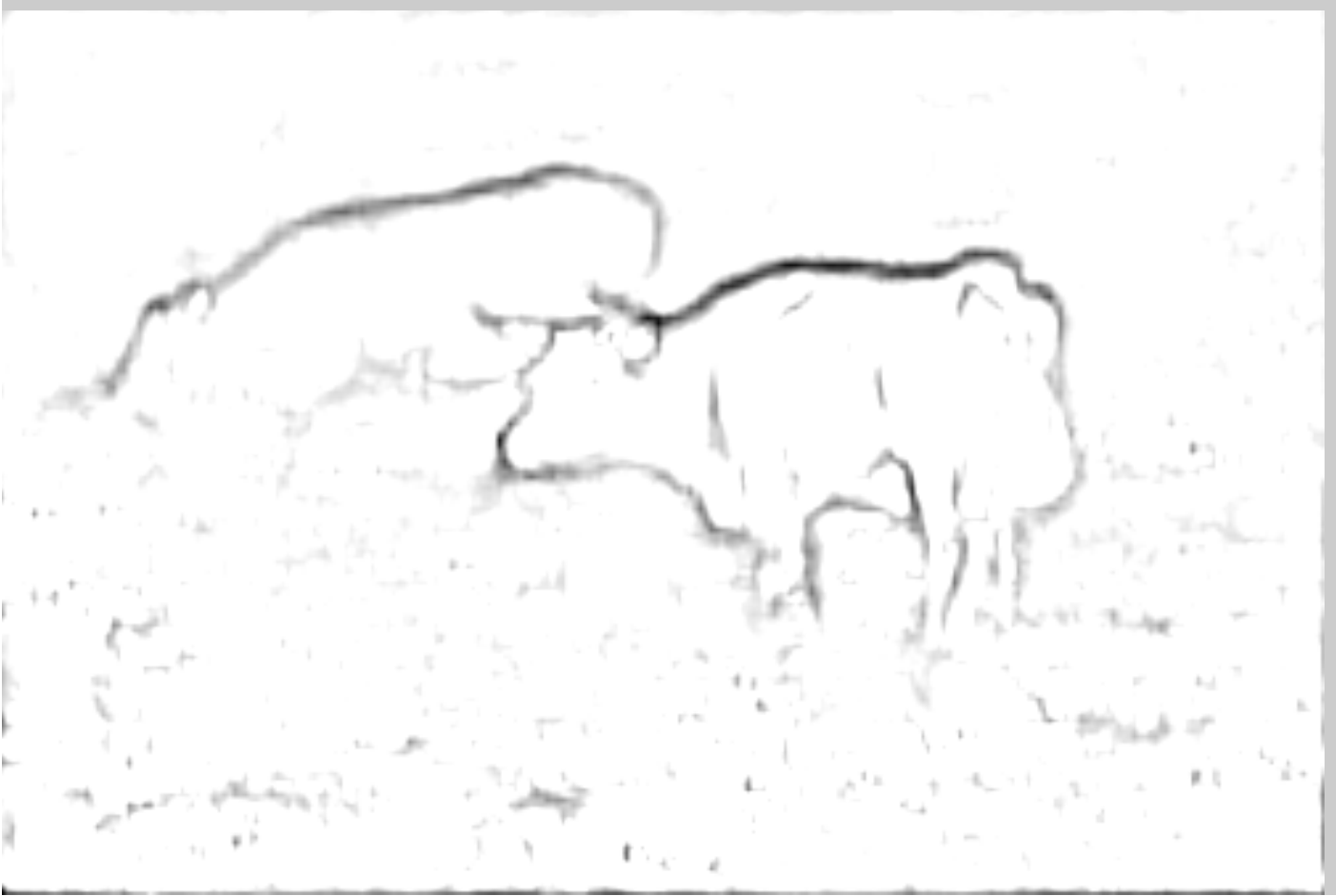} &
    \includegraphics[height=0.09\linewidth]{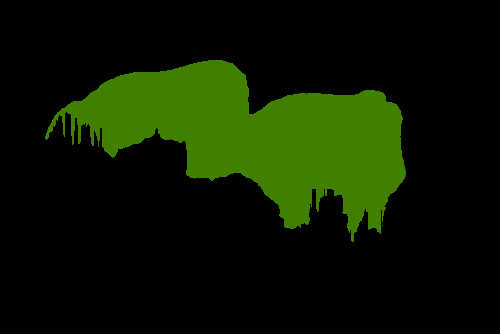} \\

\hline
    \includegraphics[height=0.10\linewidth]{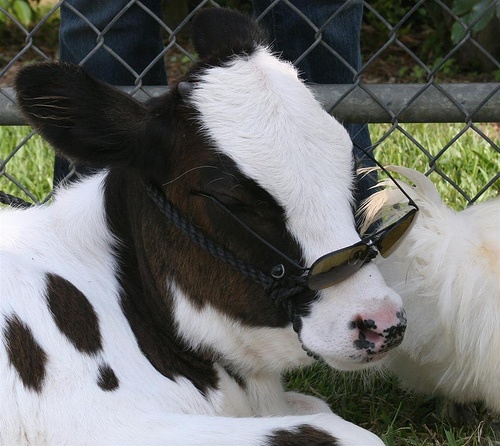} &
    \includegraphics[height=0.10\linewidth]{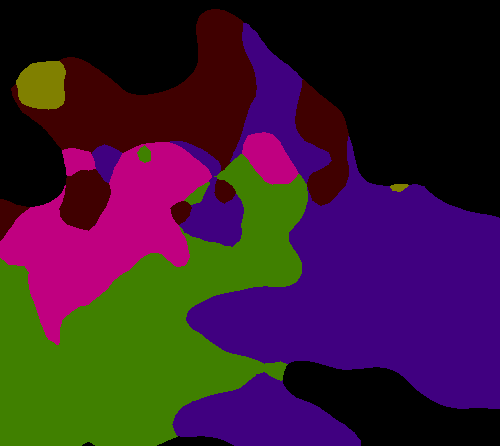} &
    \includegraphics[height=0.10\linewidth]{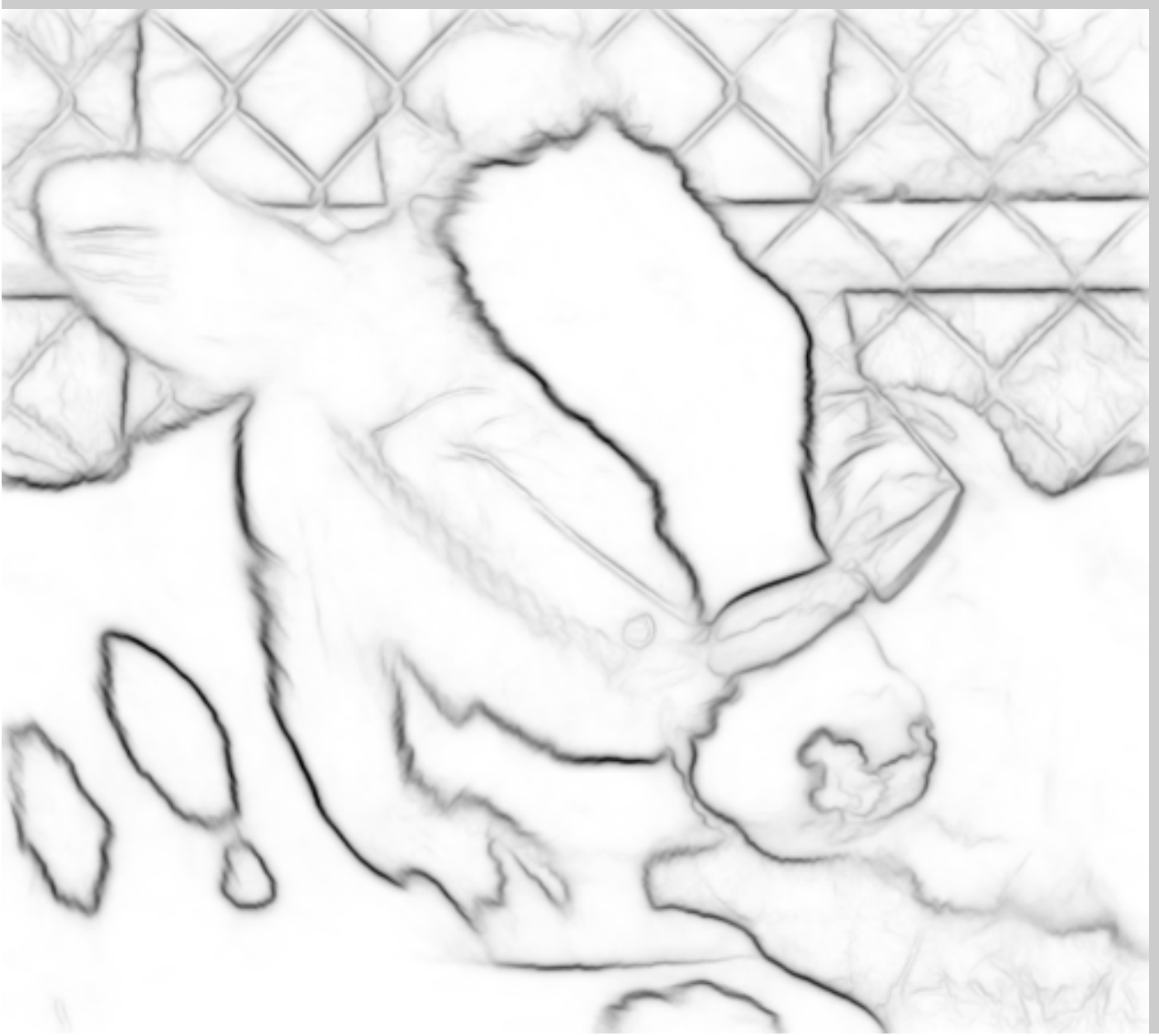} &
    \includegraphics[height=0.10\linewidth]{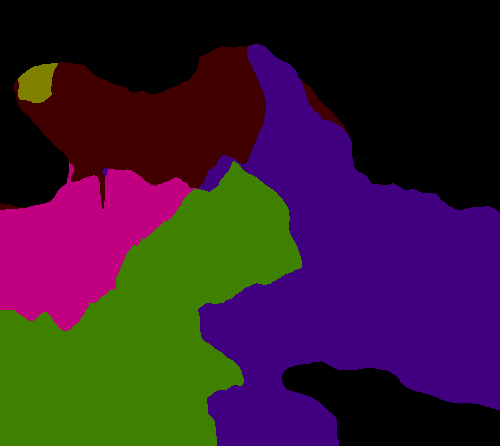} &
    \includegraphics[height=0.10\linewidth]{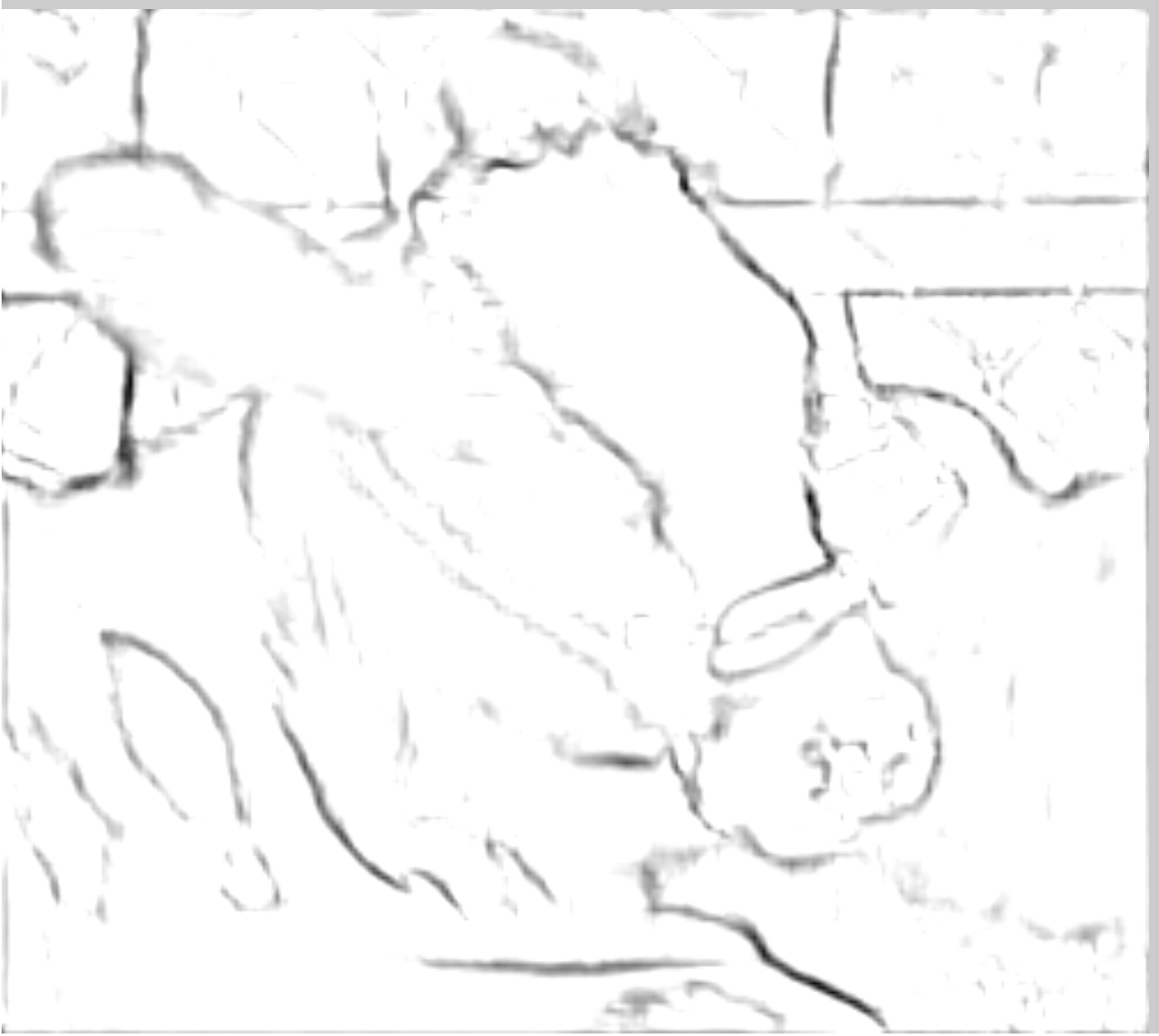} &
    \includegraphics[height=0.10\linewidth]{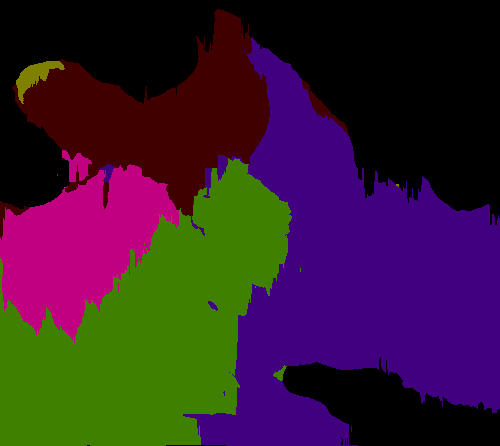} \\

    \includegraphics[height=0.12\linewidth]{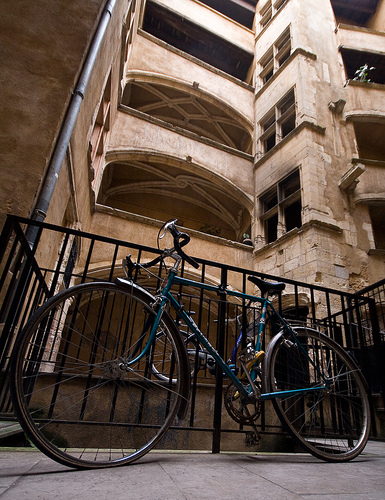} &
    \includegraphics[height=0.12\linewidth]{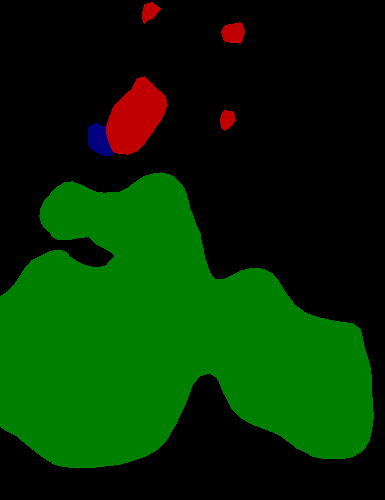} &
    \includegraphics[height=0.12\linewidth]{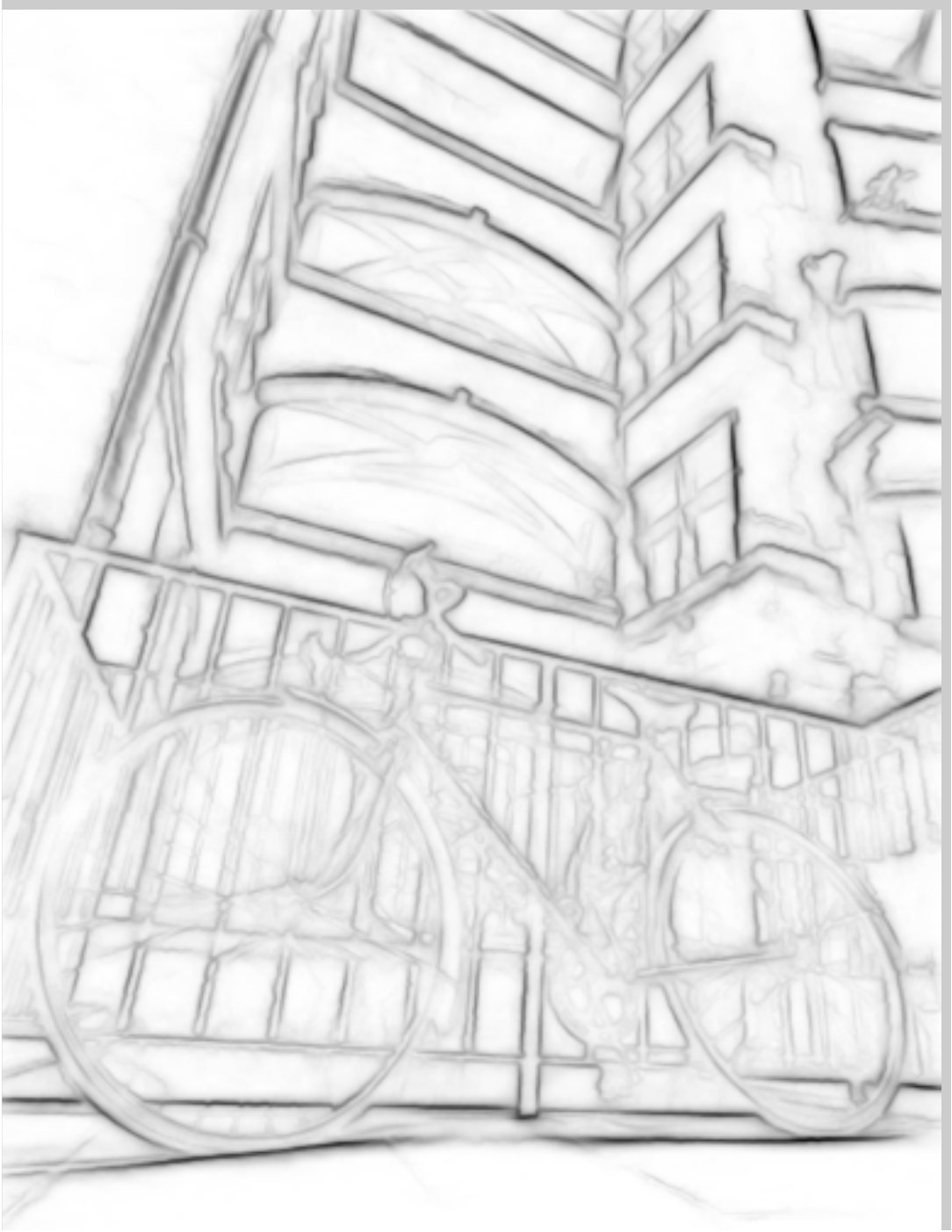} &
    \includegraphics[height=0.12\linewidth]{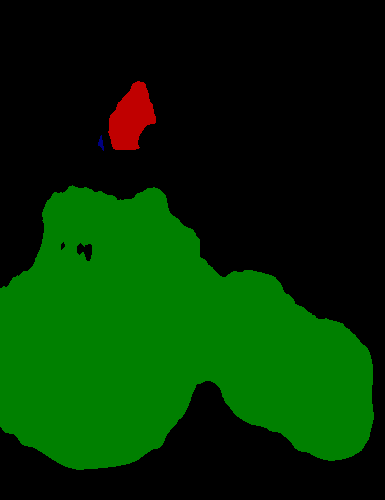} &
    \includegraphics[height=0.12\linewidth]{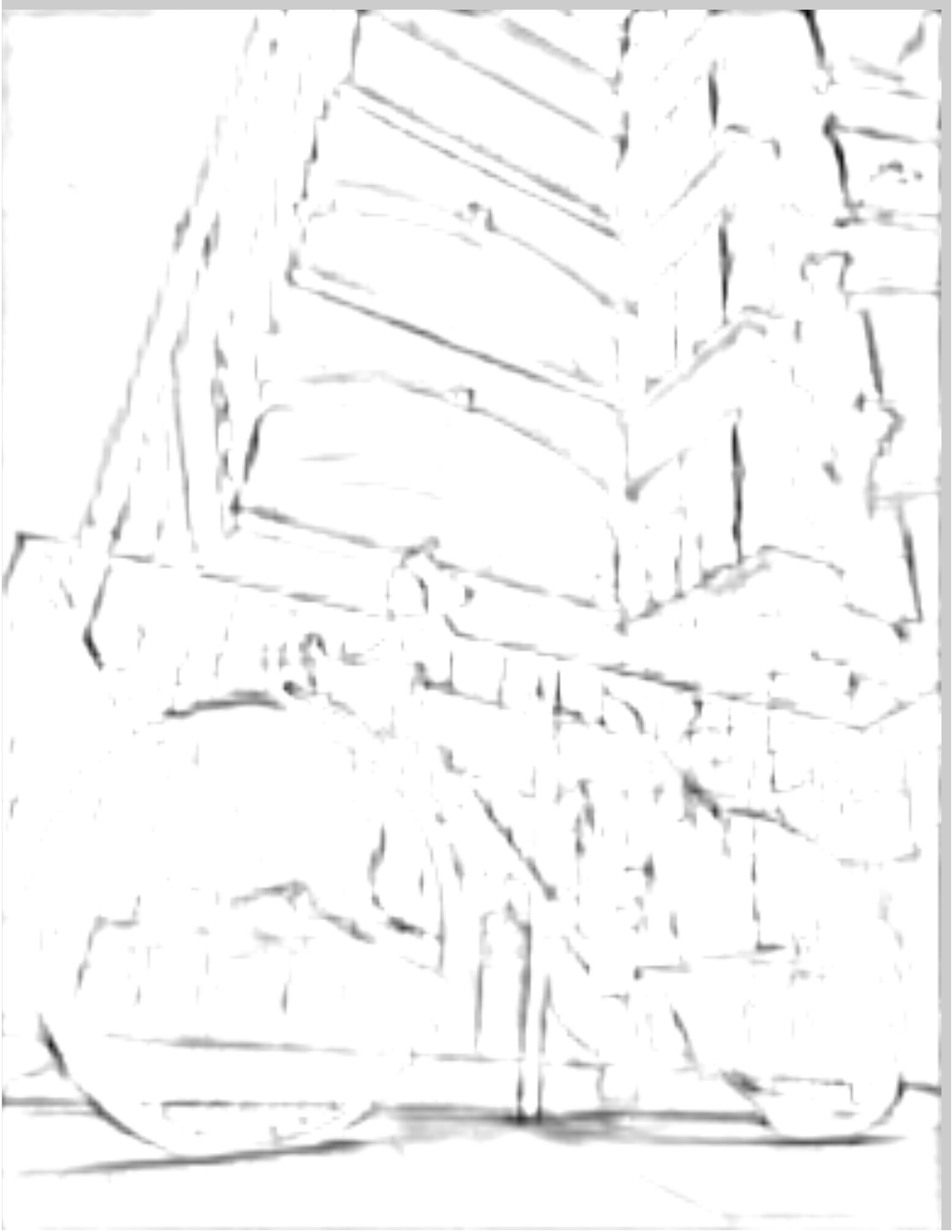} &
    \includegraphics[height=0.12\linewidth]{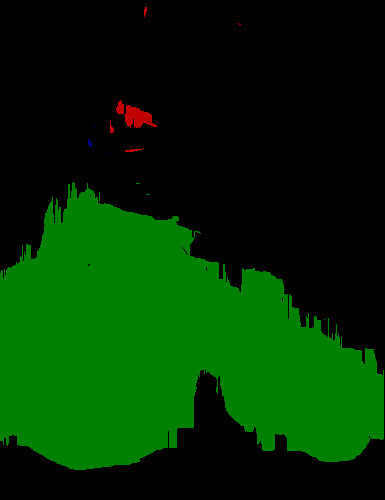} \\
    (a) Image &
    (b) Baseline &
    (c) SE &
    (d) DT-SE &
    (e) EdgeNet &
    (f) DT-EdgeNet \\
  \end{tabular}
  \caption{Visualizing results on VOC 2012 val set. For each row, we
    show (a) Image, (b) Baseline DeepLab segmentation
    result, (c) edges produced by Structured Edges, (d) segmentation
    result with Structured Edges, (e) edges generated by EdgeNet, and
    (f) segmentation result with EdgeNet. Note that our EdgeNet better
    captures the object boundaries and responds less to the background
    or object interior edges. For example, see the legs of left second
    person in the first image or the dog shapes in the second
    image. Two failure examples in the bottom.}
  \label{fig:pascal_voc12_results}  
\end{figure*}

\paragraph{EdgeNet on BSDS500} We further evaluate the edge detection
performance of our learned EdgeNet on the test set of BSDS500
\cite{amfm_pami2011}. We employ the standard metrics to evaluate edge
detection accuracy: fixed  contour  threshold (ODS F-score),
per-image  best  threshold (OIS F-score), and average precision
(AP). We also apply a standard non-maximal suppression technique to
the edge maps produced by EdgeNet for evaluation. Our method attains ODS=0.718, OIS=0.731, and AP=0.685. As shown in
\figref{fig:bsds500}, interestingly, our EdgeNet yields a
reasonably good performance (only $3\%$ worse than Structured Edges
\cite{dollar2013structured} in terms of ODS F-score), while our
EdgeNet is not trained on BSDS500 and there is no edge supervision
during training on PASCAL VOC 2012.


\paragraph{Comparison with dense CRF} Employing a fully-connected CRF
is an effective method to improve the segmentation performance. Our
best model (DT-EdgeNet) is $1.3\%$ and $1.0\%$ lower than
DeepLab-CRF on PASCAL VOC 2012 test set when the models are
pretrained with ImageNet or MS-COCO, respectively. However, our method
is many times faster in terms of computation time. To quantify
this, we time the inference computation on 50 PASCAL VOC 2012
validation images. As shown in \tabref{tab:time}, for CPU timing, on a
machine with Intel i7-4790K CPU, the well-optimized dense CRF
implementation \cite{krahenbuhl2011efficient} with 10 mean-field
iterations takes 830 ms/image, while our implementation of domain
transform with $K=3$ iterations (each iteration consists of separable
two-pass filterings across rows and columns) takes 180 ms/image (4.6
times faster). On a NVIDIA Tesla K40 GPU, our GPU implementation of
domain transform further reduces the average computation time to 25
ms/image. In our GPU implementation, the total computational cost
of the proposed method (EdgeNet+DT) is 26.2 ms/image, which amounts to
a modest overhead (about 18\%) compared to the 145 ms/image
required by DeepLab. Note there is no publicly available GPU
implementation of dense CRF inference yet.

\begin{table}
  \centering
  \addtolength{\tabcolsep}{2.5pt}
  \begin{tabular}{l | c }
    \toprule[0.2 em]
    Method & mIOU (\%) \\
    \toprule[0.2 em]
    DeepLab & 67.31 \\
    DeepLab-CRF & 71.01 \\
    \midrule
    DT-SE & 68.94 \\
    DT-EdgeNet & 69.96 \\
    DT-EdgeNet + DenseCRF & 71.77 \\
    \bottomrule[0.1 em]
  \end{tabular}
  \caption{Performance on PASCAL VOC 2012 val set. The models have
    been pretrained on {\it MS-COCO 2014} dataset.}
  \label{tab:models_valset_coco}
\end{table}


\begin{table}
  \centering
  \addtolength{\tabcolsep}{2.5pt}
  \begin{tabular}{l | c | c}
    \toprule[0.2 em]
    Method & {\small ImageNet} & COCO \\
    \toprule[0.2 em]
    DeepLab \cite{chen2014semantic, papandreou2015weakly} & 65.1 & 68.9\\
    DeepLab-CRF \cite{chen2014semantic, papandreou2015weakly} & 70.3 & 72.7\\
    \midrule
    DT-SE & 67.8 & 70.7\\
    DT-EdgeNet & 69.0 & 71.7\\
    DT-EdgeNet + DenseCRF & 71.2 & 73.6 \\
    \midrule
    DeepLab-CRF-Attention \cite{chen2015attention} & - & 75.7 \\
    DeepLab-CRF-Attention-DT & - & 76.3 \\
    \midrule \midrule
    CRF-RNN \cite{zheng2015conditional} & 72.0 & 74.7 \\
    BoxSup \cite{dai2015boxsup} & - & 75.2 \\
    {\small CentraleSuperBoundaries++} \cite{kokkinos2016pushing} & - & 76.0 \\
    DPN \cite{liu2015semantic} & 74.1 & 77.5 \\
    Adelaide\_Context \cite{lin2015efficient} & 75.3 & 77.8 \\
    \bottomrule[0.1 em]
  \end{tabular}
  \caption{mIOU (\%) on PASCAL VOC 2012 {\it test} set. We evaluate
    our models with two settings: the models are (1) pretrained with
    ImageNet, and (2) further pretrained with MS-COCO.}
  \label{tab:testset}
\end{table}


\begin{table}
  \centering
  \addtolength{\tabcolsep}{2.5pt}
  \scalebox{0.92}{
  \begin{tabular}{l | c | c}
    \toprule[0.2 em]
    Method & CPU time & GPU time \\
    \toprule[0.2 em]
    DeepLab & 5240  & 145 \\
    \midrule
    EdgeNet & 20 (0.4\%) & 1.2 (0.8\%) \\ 
    \midrule
    Dense CRF (10 iterations) & 830 (15.8\%) & - \\
    \midrule
    DT (3 iterations) & 180 (3.4\%) & 25 (17.2\%)\\
    \midrule \midrule
    CRF-RNN (CRF part) \cite{zheng2015conditional} & 1482 & - \\
    \bottomrule[0.1 em]
  \end{tabular}
  }
  \caption{Average inference time (ms/image). Number in parentheses 
    is the percentage \wrt the DeepLab computation. Note that EdgeNet computation time is improved by performing convolution first and then upsampling.}
  \label{tab:time}
\end{table}

\section{Conclusions}
We have presented an approach to learn edge maps useful for semantic
image segmentation in a unified system that is trained
discriminatively in an end-to-end fashion. The proposed method builds
on the domain transform, an edge-preserving filter traditionally used
for graphics applications. We show that backpropagating through the
domain transform allows us to learn an task-specific edge map
optimized for semantic segmentation. Filtering the raw semantic
segmentation maps produced by deep fully convolutional networks with
our learned domain transform leads to improved localization accuracy
near object boundaries. The resulting scheme is several times faster
than fully-connected CRFs that have been previously used for this
purpose.

\paragraph{Acknowledgments} 
This work wast partly supported by ARO 62250-CS and NIH Grant 5R01EY022247-03.

\appendix
\section*{Appendix}
The appendix contains: (1) Detailed quantitative results
for the proposed methods, showing per-class semantic segmentation IOU
on the PASCAL VOC 2012 test set. (2) Qualitative edge detection and
semantic segmentation results on additional images.

\section{Detailed quantitative image segmentation results}

We provide per-class semantic segmentation IOU on the PASCAL VOC 2012
test set. We compare with the DeepLab-LargeFOV and DeepLab-CRF-LargeFOV
baselines. In \tabref{tab:voc2012:imagenet} we show performance of
models that have only been pretrained on the Imagenet 2012 image
classification task \cite{ILSVRC15}, while in
\tabref{tab:voc2012:mscoco} we show performance of models that have
also been pretrained on the MS-COCO 2014 semantic segmentation task
\cite{lin2014microsoft}.

\begin{table*}[!h] 
\setlength{\tabcolsep}{3pt}
\resizebox{2.1\columnwidth}{!}{
\begin{tabular}{l||c||c*{20}{|c}}
\toprule[0.2 em]
Method         & mean & bkg &  aero & bike & bird & boat & bottle& bus & car  &  cat & chair& cow  &table & dog  & horse & mbike& person& plant&sheep& sofa &train & tv  \\
\midrule
DeepLab-LargeFOV \cite{chen2014semantic} & 65.1 & 90.7 & 74.7 & 34.0 & 74.3 & 57.1 & 62.0 & 82.6 & 75.5 & 79.1 & 26.2 & 65.7 & 55.8 & 73.0 & 68.0 & 78.6 & 76.2 & 50.6 & 73.9 & 45.5 & 66.6 & 57.1  \\
DeepLab-CRF-LargeFOV \cite{chen2014semantic} & 70.3 & 92.6 & 83.5 & 36.6 & 82.5 & 62.3 & 66.5 & 85.4 & 78.5 & 83.7 & 30.4 & 72.9 & 60.4 & 78.5 & 75.5 & 82.1 & 79.7 &  58.2 & 82.0 & 48.8 & 73.7 & 63.3 \\
\midrule
\href{http://host.robots.ox.ac.uk:8080/anonymous/HGBS30.html}{DT-SE} & 67.8 & 91.7 & 78.8 & 33.5 & 78.7 & 60.6 & 64.5 & 84.5 & 77.4 & 81.3 & 29.0 & 69.1 & 59.4 & 76.1 & 70.8 & 80.6 & 77.9 & 53.4 & 77.9 & 46.0 & 70.1 & 62.5 \\
\href{http://host.robots.ox.ac.uk:8080/anonymous/BTEHZF.html}{DT-EdgeNet} & 69.0 & 92.1 & 79.8 & 34.8 & 79.6 & 61.3 & 67.0 & 85.0 & 78.5 & 83.2 & 30.2 & 70.3 & 58.9 & 77.9 & 72.3 & 82.3 & 79.5 & 55.0 & 79.8 & 47.9 & 70.8 & 62.5 \\
\href{http://host.robots.ox.ac.uk:8080/anonymous/NGWRBG.html}{DT-EdgeNet + DenseCRF} & 71.2 &  92.8 & 83.6 & 35.8 & 82.4 & 63.1 & 68.9 & 86.2 & 79.6 & 84.7 & 31.8 & 74.2 & 61.1 & 79.6 & 76.6 & 83.2 & 80.9 & 58.3 & 82.6 & 49.1 & 74.8 & 65.1 \\
\bottomrule[0.1 em]
 \end{tabular}
}
\caption{Segmentation IOU on the PASCAL VOC 2012 test set, using the
  trainval set for training. Model only pretrained on the Imagenet
  image classification task.}
\label{tab:voc2012:imagenet}
\end{table*}

\begin{table*}[!h] 
\setlength{\tabcolsep}{3pt}
\resizebox{2.1\columnwidth}{!}{
\begin{tabular}{l||c||c*{20}{|c}}
\toprule[0.2 em]
Method         & mean & bkg &  aero & bike & bird & boat & bottle& bus & car  &  cat & chair& cow  &table & dog  & horse & mbike& person& plant&sheep& sofa &train & tv  \\
\midrule
DeepLab-COCO-LargeFOV \cite{papandreou2015weakly} & 68.9 & 92.1 & 81.6 & 35.4 & 81.4 & 60.1 & 65.9 & 84.3 & 79.3 & 81.8 & 28.4 &  71.2 & 59.0 & 75.3 & 72.6 & 81.5 & 80.1 & 53.5 & 78.8 & 50.8 & 72.7 & 60.3\\
DeepLab-CRF-COCO-LargeFOV \cite{papandreou2015weakly} & 72.7 & 93.4 & 89.1 & 38.3 & 88.1 & 63.3 & 69.7 & 87.1 & 83.1 & 85.0 & 29.3 & 76.5 & 56.5 & 79.8 & 77.9 & 85.8 & 82.4 & 57.4 & 84.3 & 54.9 & 80.5 & 64.1  \\
\midrule
\href{http://host.robots.ox.ac.uk:8080/anonymous/C5G8AZ.html}{DT-SE} & 70.7 & 92.6 & 83.8 & 35.0 & 85.5 & 61.9 & 67.6 & 85.4 & 80.3 & 84.4 & 30.2 & 73.6 & 60.4 & 77.8 & 74.8 & 82.3 & 81.0 & 54.9 & 81.2 & 52.3 & 75.5 & 64.1 \\
\href{http://host.robots.ox.ac.uk:8080/anonymous/JTIVRP.html}{DT-EdgeNet} & 71.7 & 93.0 & 85.6 & 36.0 & 86.4 & 63.0 & 69.3 & 86.0 & 81.2 & 85.9 & 30.7 & 75.1 & 60.8 & 79.3 & 76.1& 83.2 & 82.0 & 56.2 & 82.8 & 53.3 & 75.9 & 64.4 \\
\href{http://host.robots.ox.ac.uk:8080/anonymous/ZKKJ0Q.html}{DT-EdgeNet + DenseCRF} & 73.6 & 93.5 & 88.3 & 37.0 & 89.8 & 63.6 & 70.3 & 87.3 & 82.0 & 87.6 & 31.1 & 79.0 & 61.9 & 81.6 & 80.4 & 84.5 & 83.3 & 58.4 & 86.1 & 55.9 & 78.2 & 65.4 \\
\href{http://host.robots.ox.ac.uk:8080/anonymous/G4P5MB.html}{DeepLab-CRF-Attention-DT} & 76.3 & 94.3 & 93.2 & 41.7 & 88.0 & 61.7 & 74.9 & 92.9 & 84.5 & 90.4 & 33.0 & 82.8 & 63.2 & 84.5 & 85.0 & 87.2 & 85.7 & 60.5 & 87.7 & 57.8 & 84.3 & 68.2 \\
\bottomrule[0.1 em]
 \end{tabular}
}
\caption{Segmentation IOU on the PASCAL VOC 2012 test set, using the
  trainval set for training. Model pretrained on both the Imagenet
  image classification task and the MS-COCO semantic segmentation task.}
 \label{tab:voc2012:mscoco}
\end{table*}

\section{Qualitative edge detection and image segmentation results}

\begin{figure*}[p]
  \centering
  \begin{tabular}{c c c c c c}
    \includegraphics[height=0.145\linewidth]{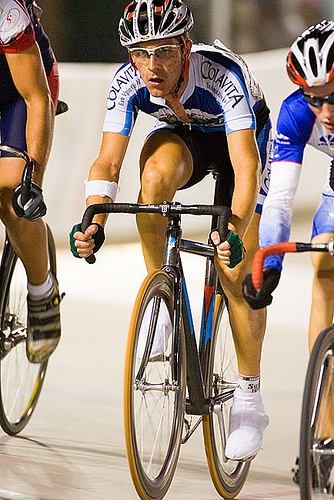} &
    \includegraphics[height=0.145\linewidth]{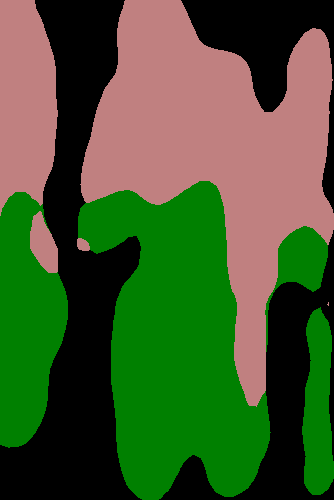} &
    \includegraphics[height=0.145\linewidth]{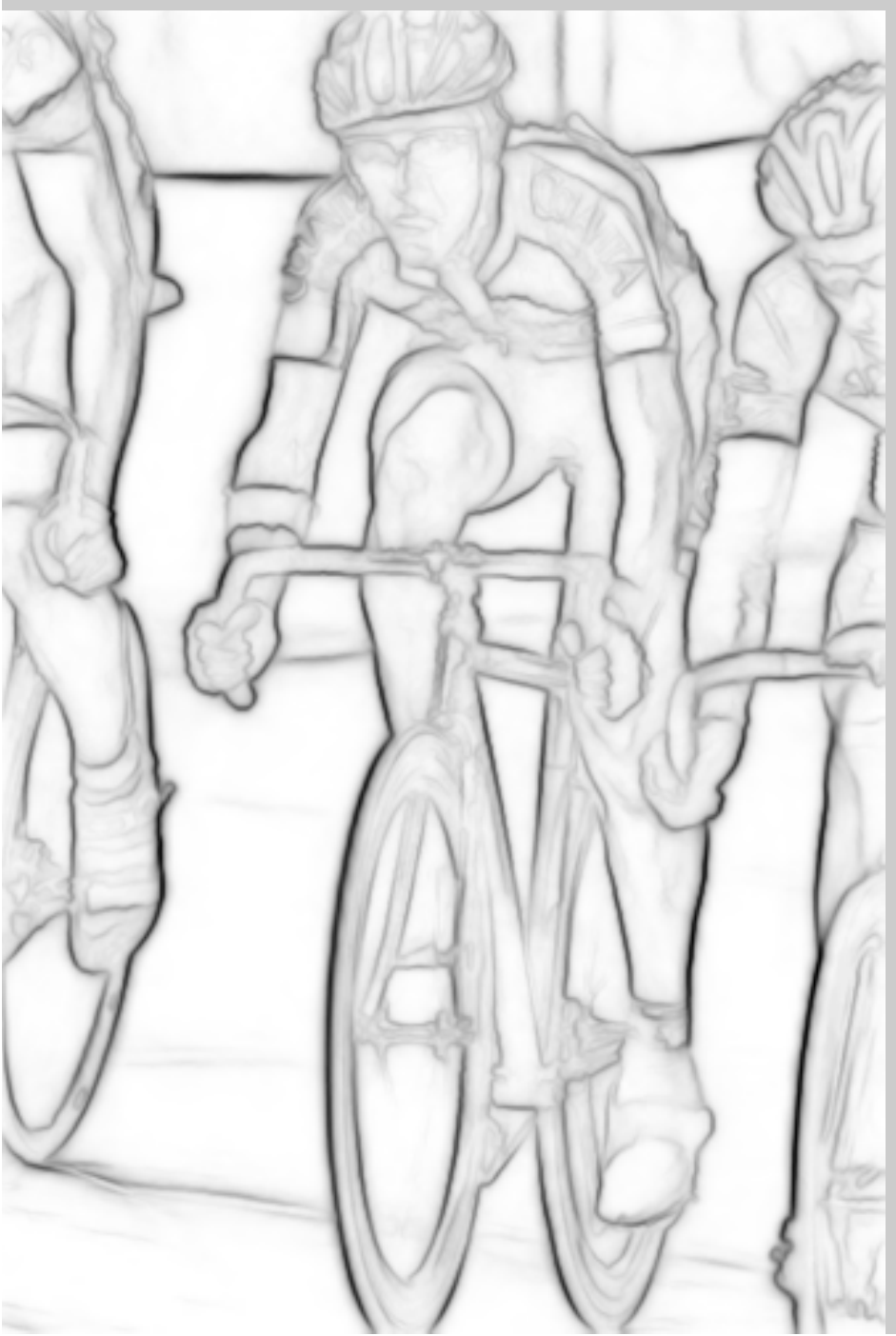} &
    \includegraphics[height=0.145\linewidth]{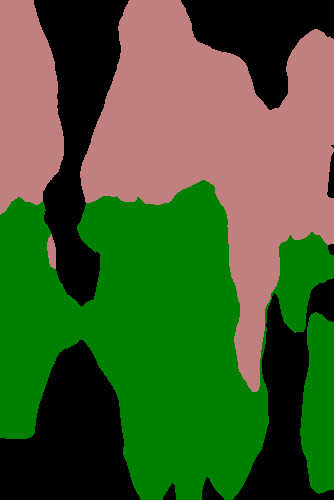} &
    \includegraphics[height=0.145\linewidth]{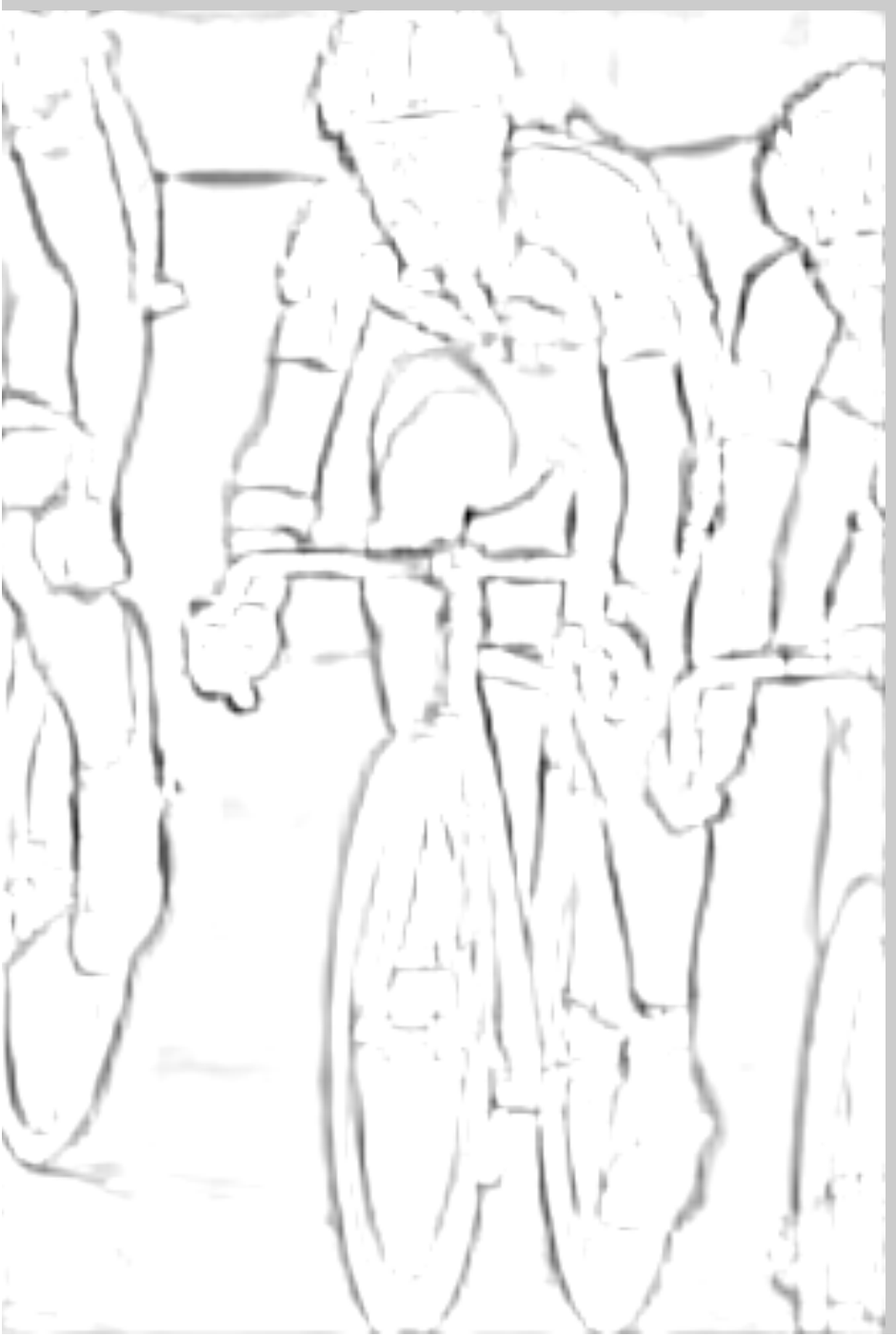} &
    \includegraphics[height=0.145\linewidth]{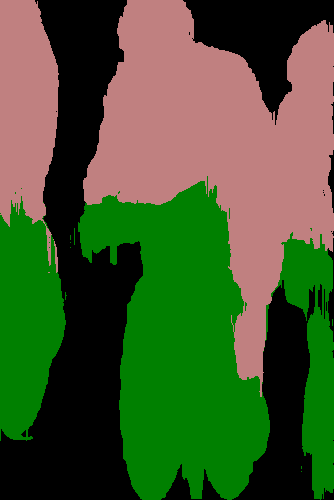} \\

    \includegraphics[height=0.13\linewidth]{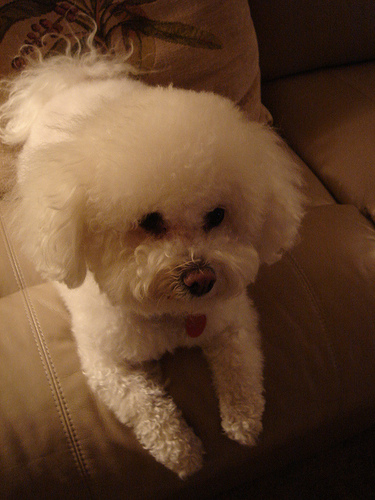} &
    \includegraphics[height=0.13\linewidth]{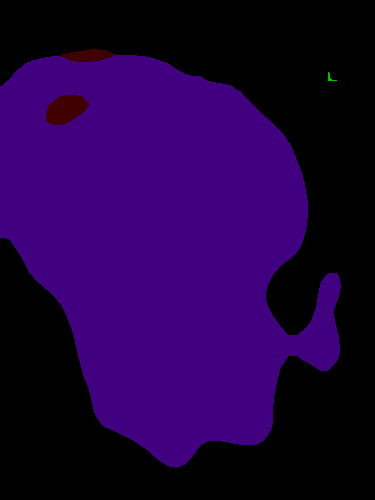} &
    \includegraphics[height=0.13\linewidth]{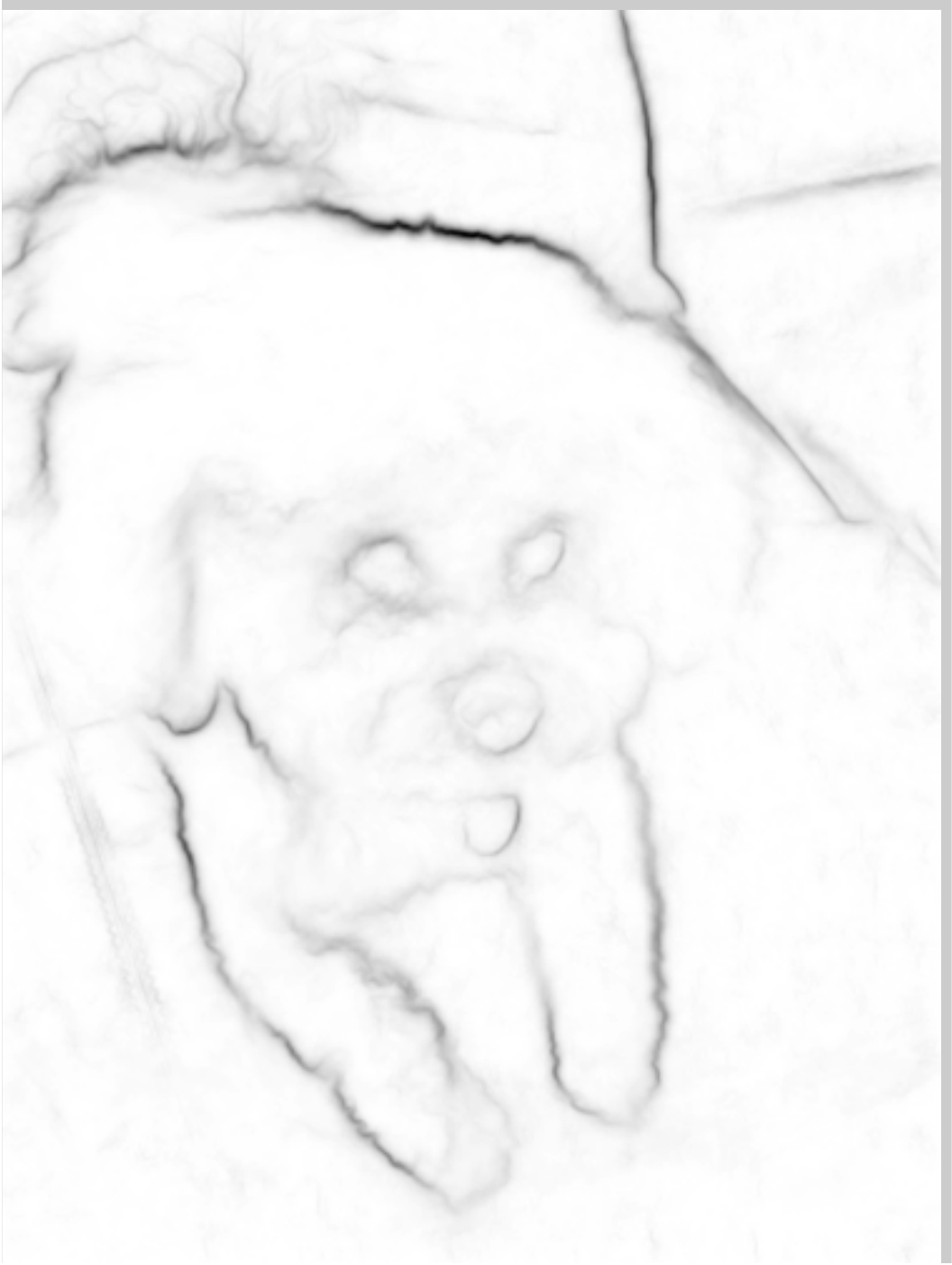} &
    \includegraphics[height=0.13\linewidth]{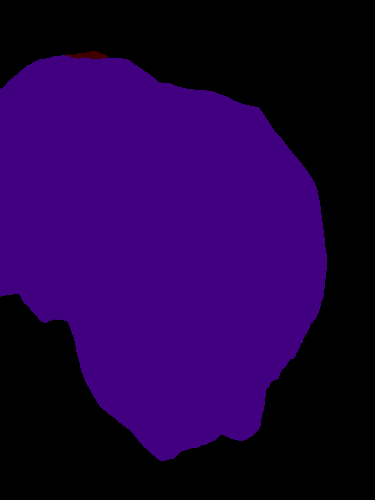} &
    \includegraphics[height=0.13\linewidth]{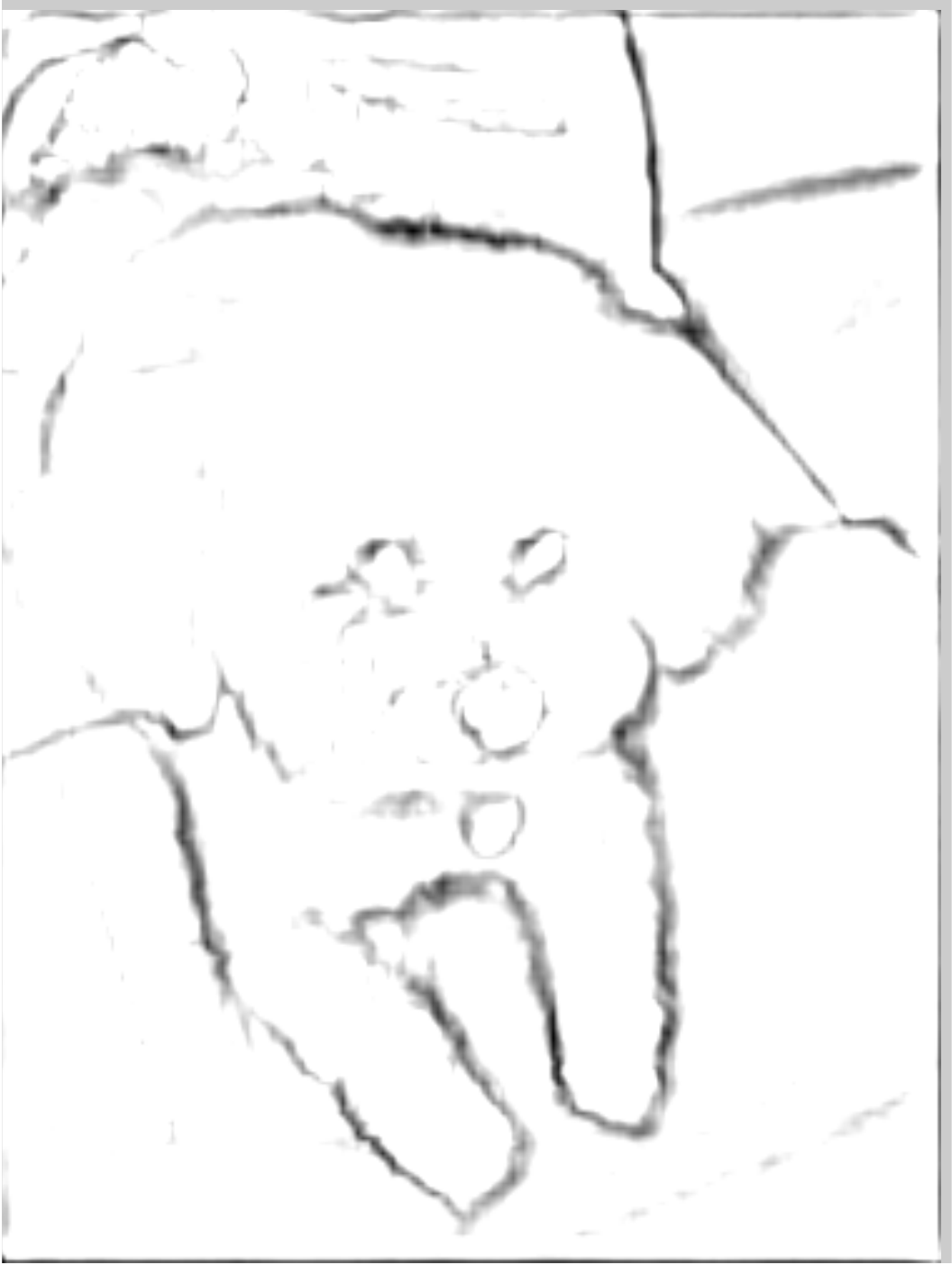} &
    \includegraphics[height=0.13\linewidth]{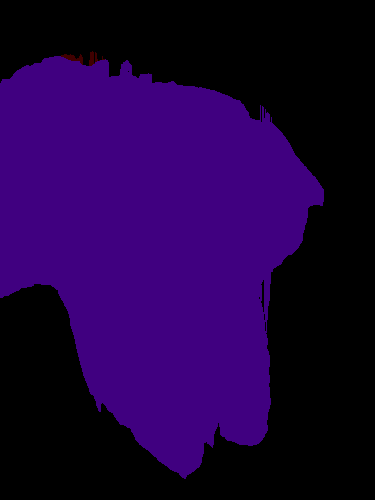} \\

    \includegraphics[height=0.09\linewidth]{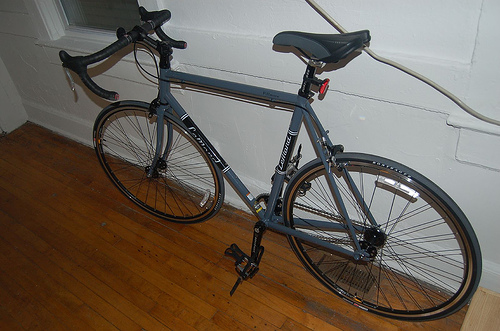} &
    \includegraphics[height=0.09\linewidth]{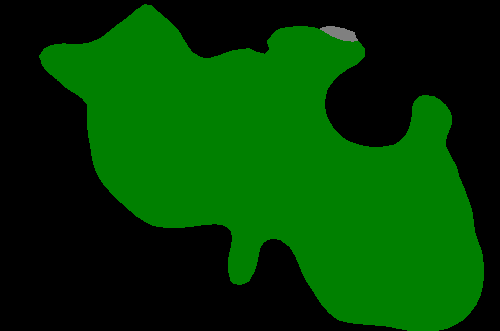} &
    \includegraphics[height=0.09\linewidth]{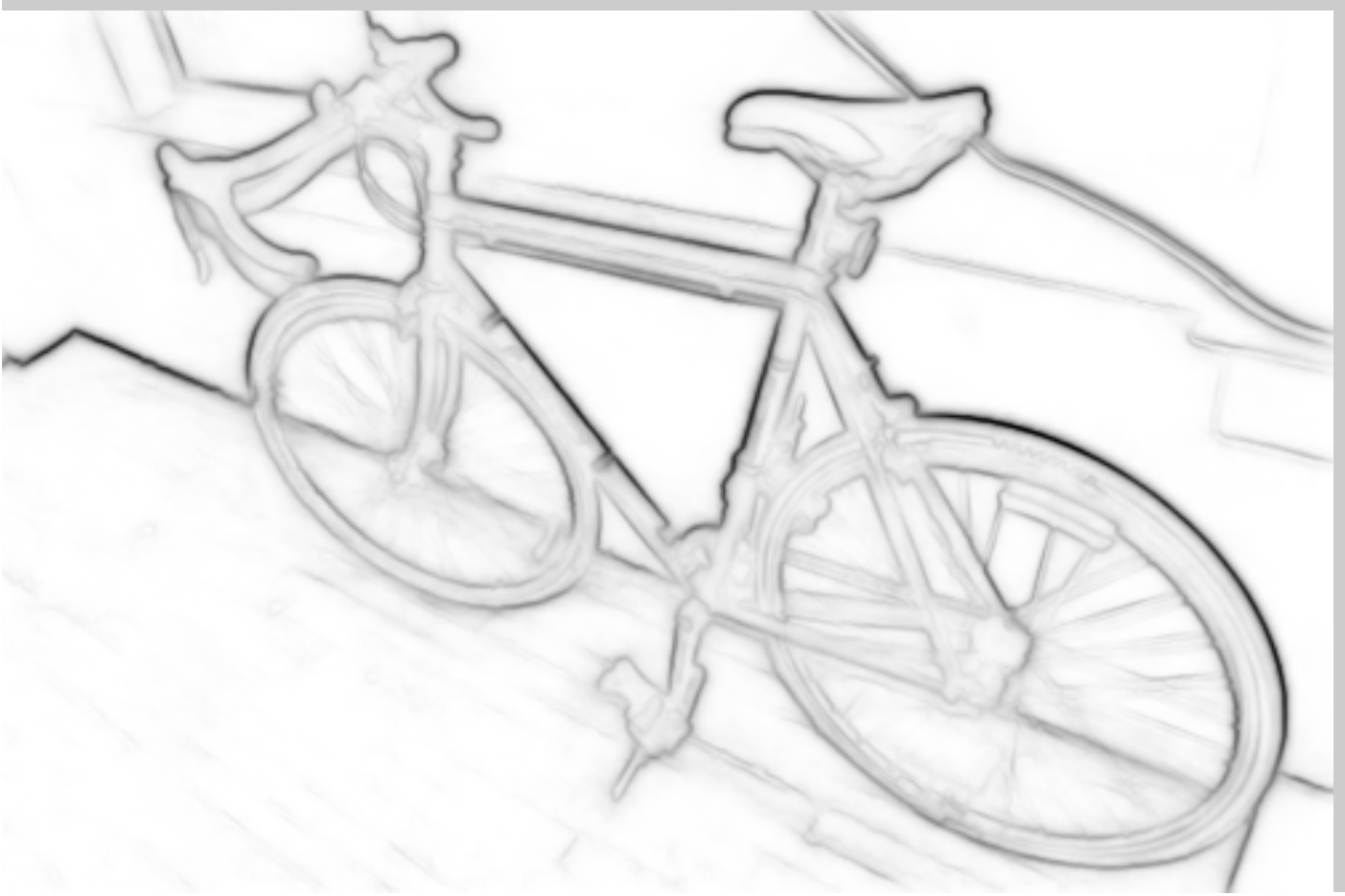} &
    \includegraphics[height=0.09\linewidth]{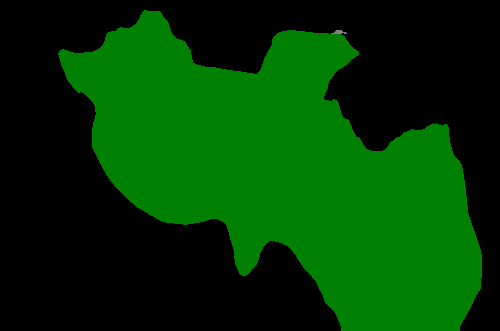} &
    \includegraphics[height=0.09\linewidth]{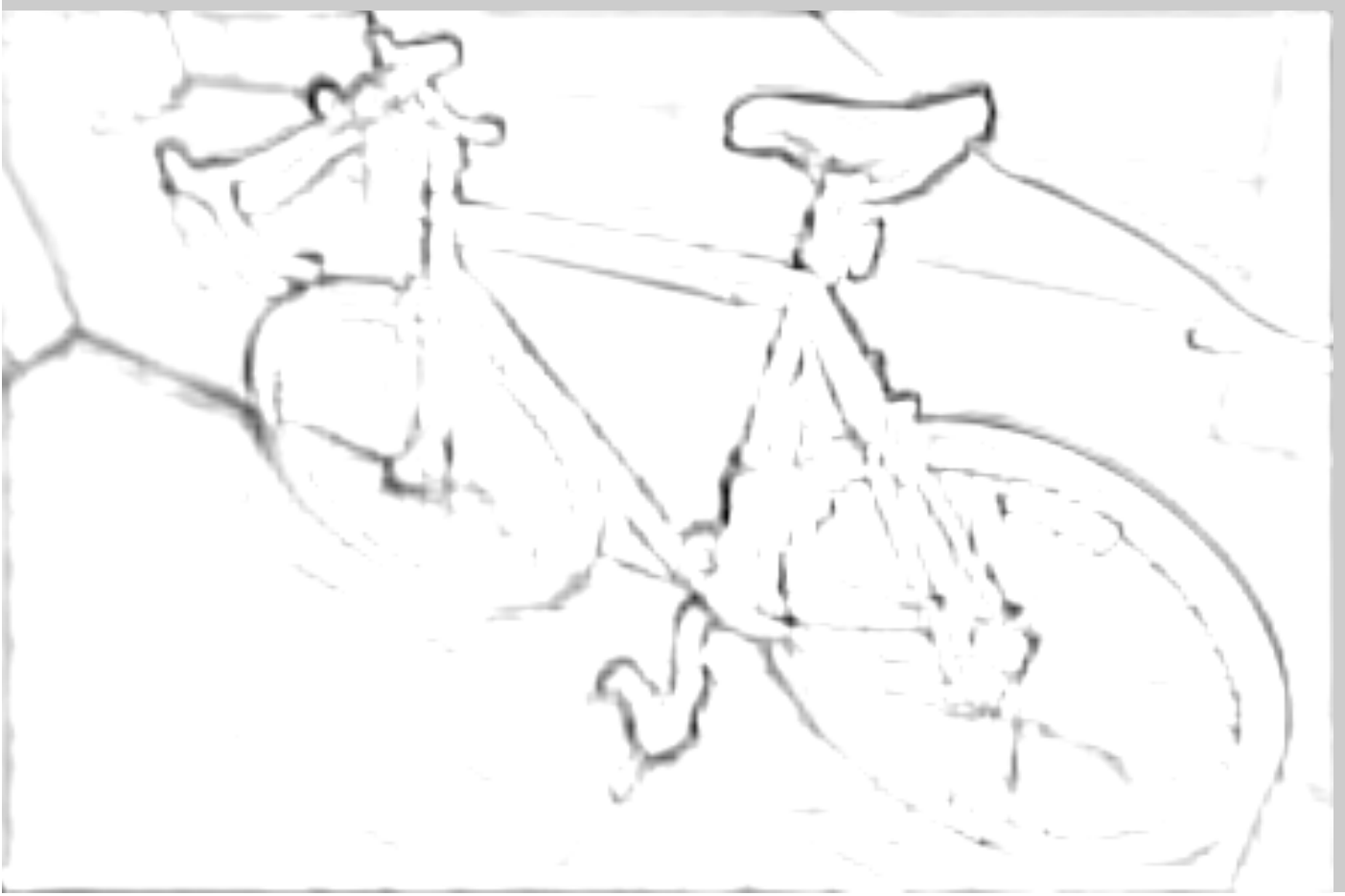} &
    \includegraphics[height=0.09\linewidth]{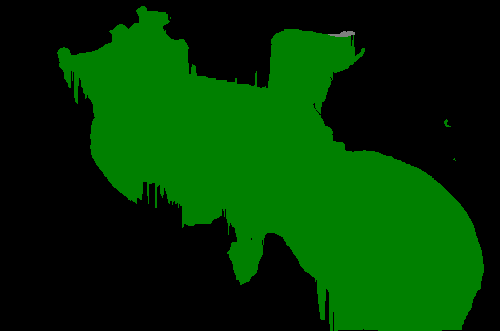} \\

    \includegraphics[height=0.1\linewidth]{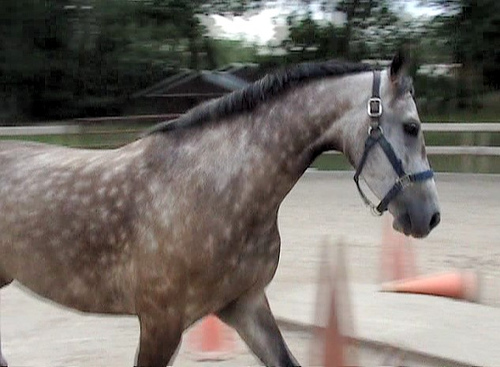} &
    \includegraphics[height=0.1\linewidth]{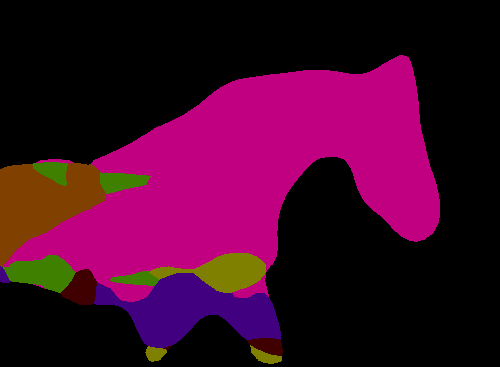} &
    \includegraphics[height=0.1\linewidth]{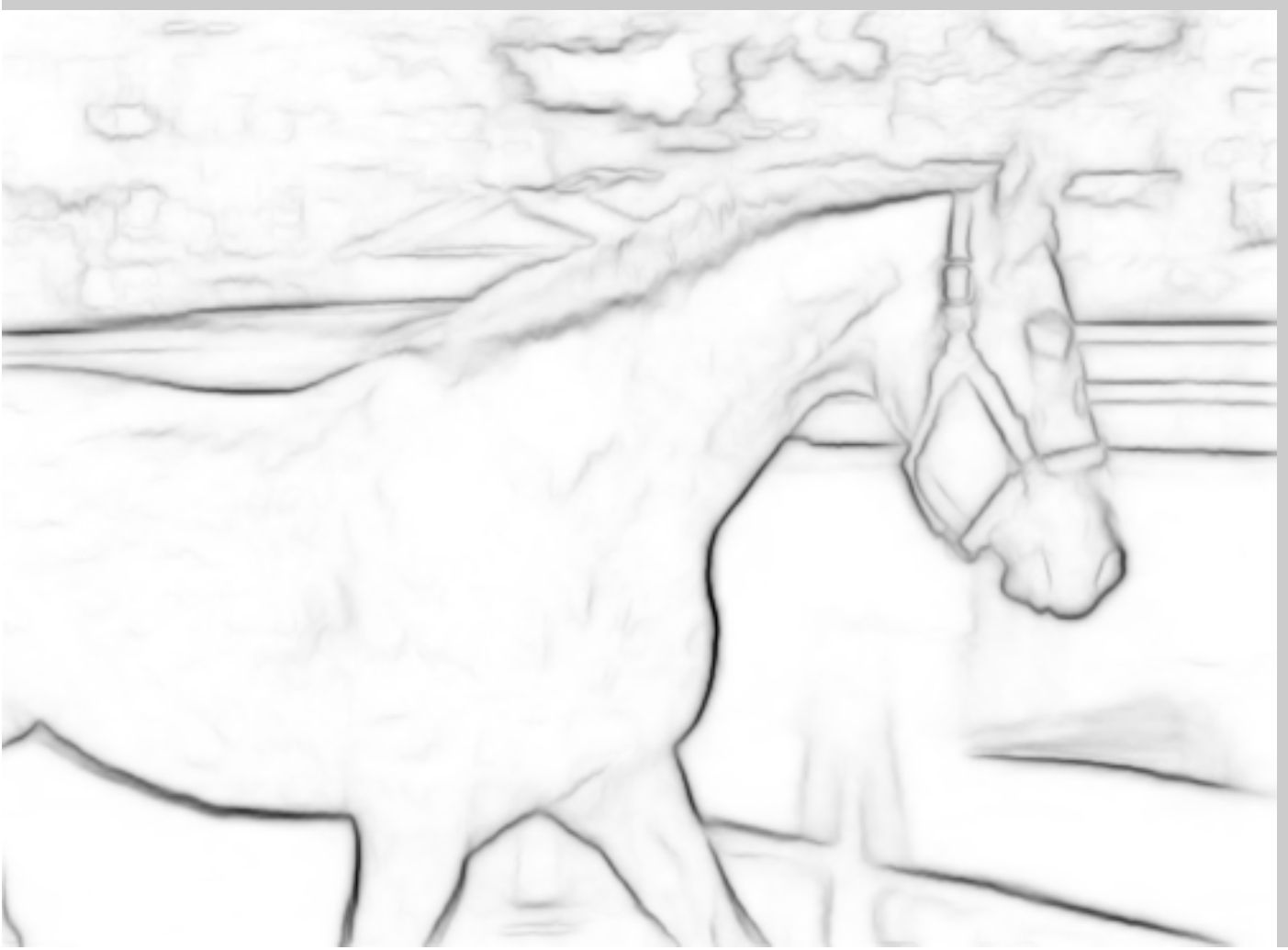} &
    \includegraphics[height=0.1\linewidth]{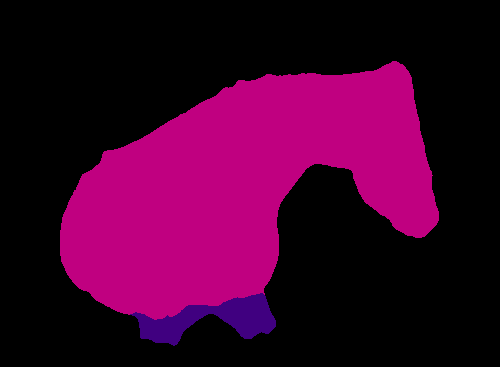} &
    \includegraphics[height=0.1\linewidth]{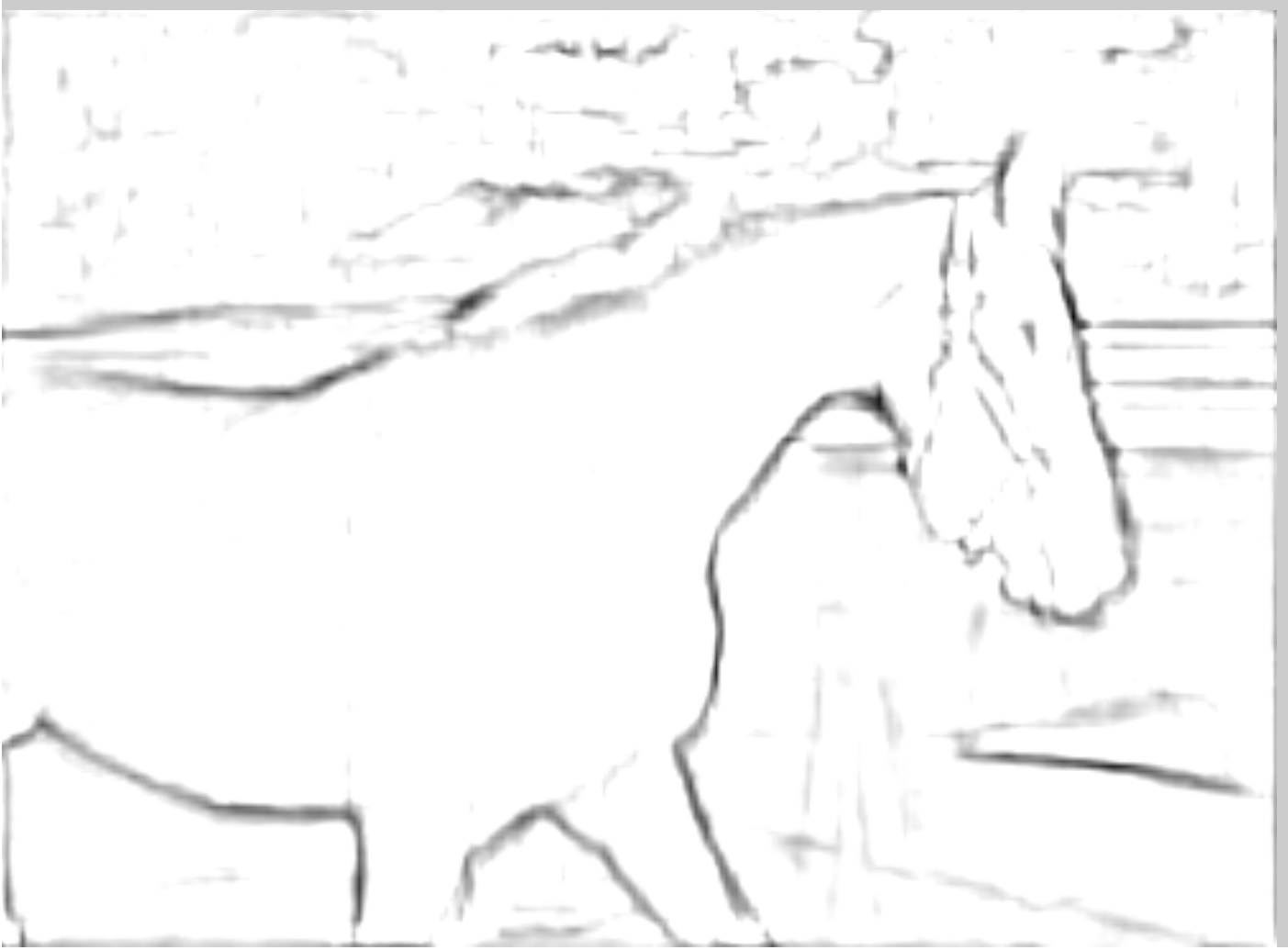} &
    \includegraphics[height=0.1\linewidth]{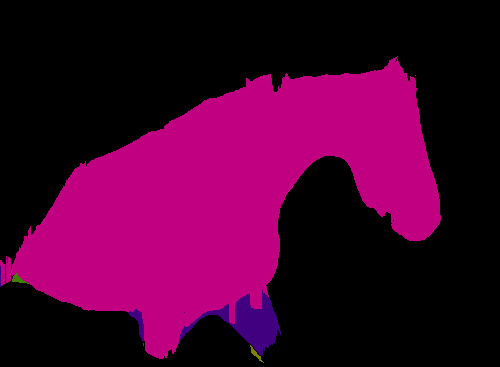} \\

    \includegraphics[height=0.102\linewidth]{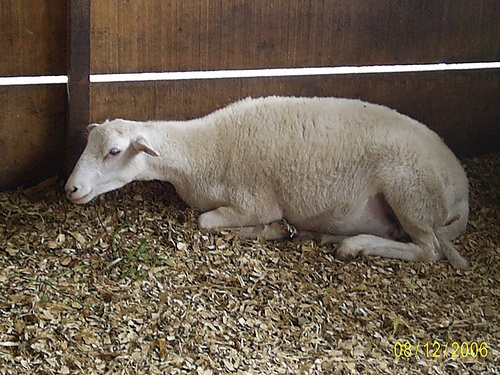} &
    \includegraphics[height=0.102\linewidth]{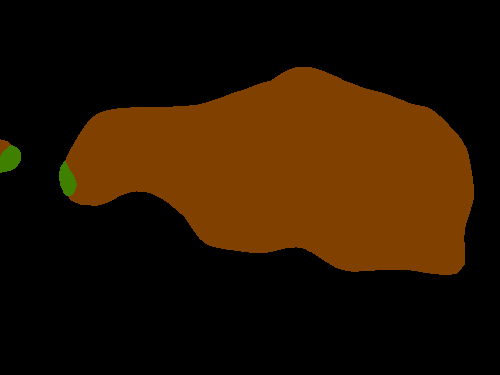} &
    \includegraphics[height=0.102\linewidth]{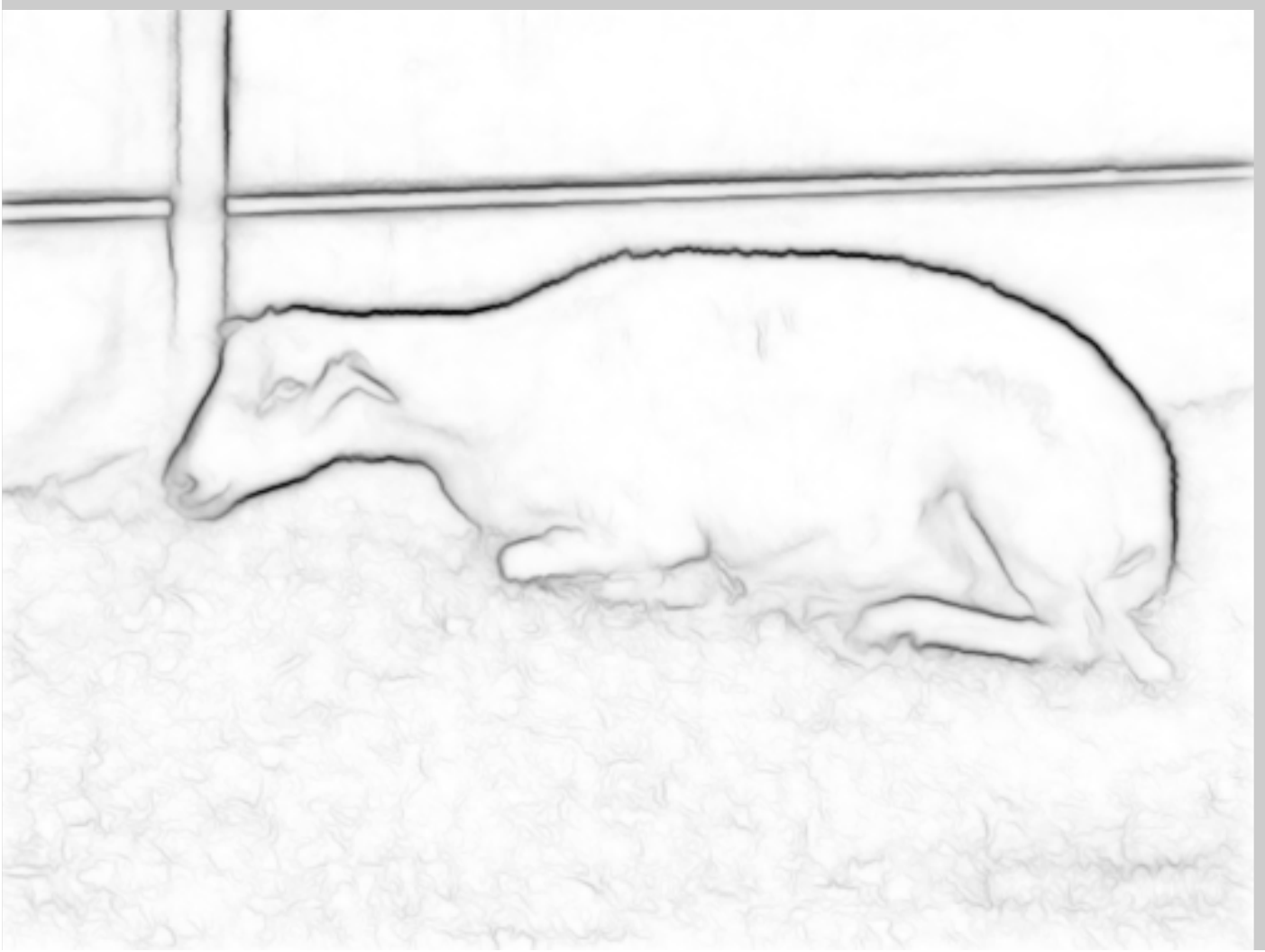} &
    \includegraphics[height=0.102\linewidth]{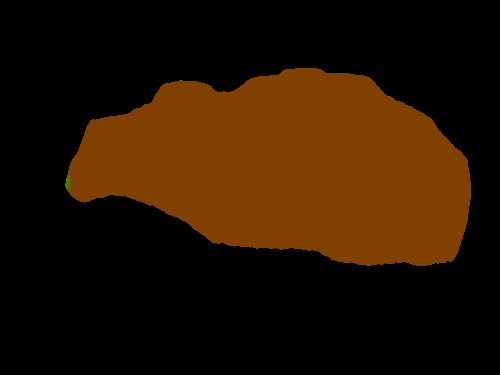} &
    \includegraphics[height=0.102\linewidth]{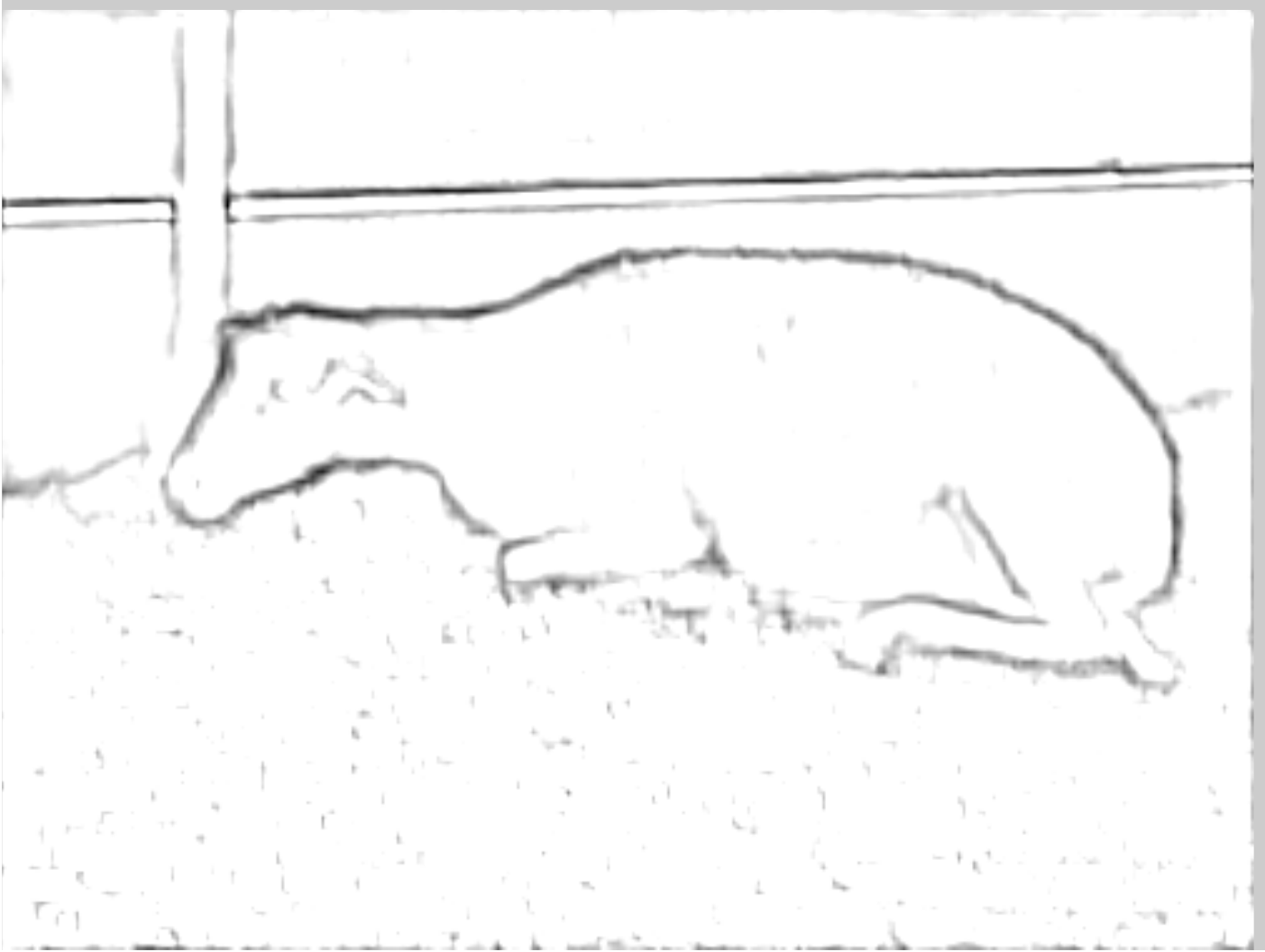} &
    \includegraphics[height=0.102\linewidth]{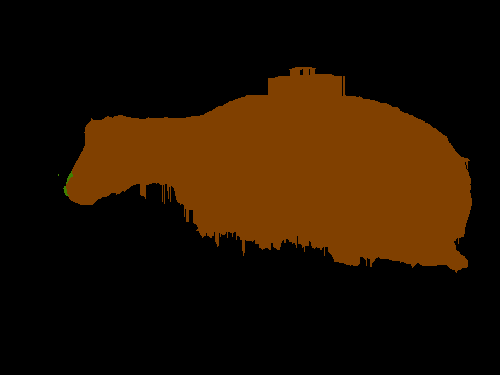} \\

    \includegraphics[height=0.092\linewidth]{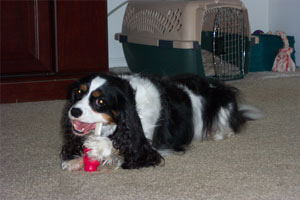} &
    \includegraphics[height=0.092\linewidth]{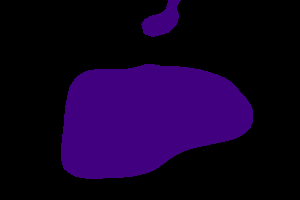} &
    \includegraphics[height=0.092\linewidth]{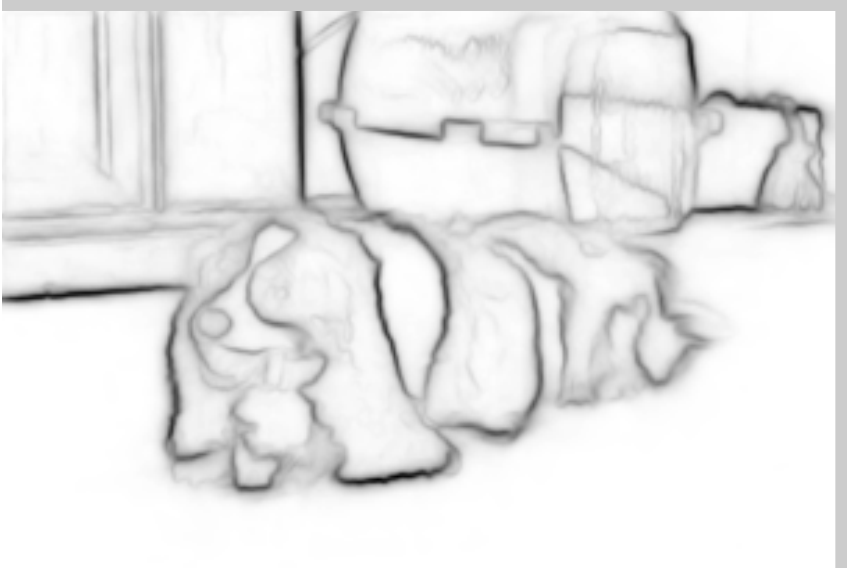} &
    \includegraphics[height=0.092\linewidth]{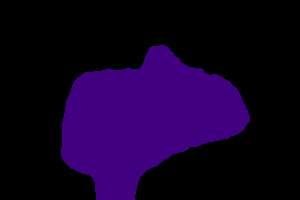} &
    \includegraphics[height=0.092\linewidth]{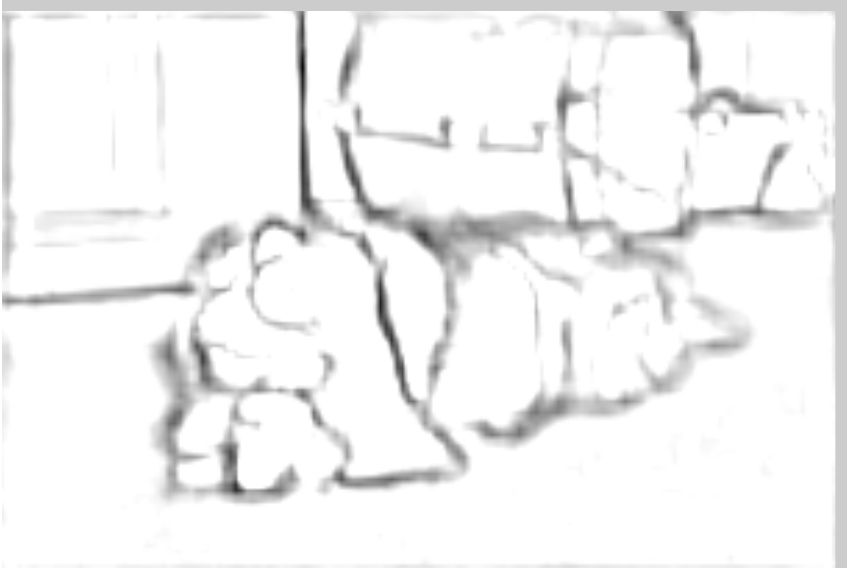} &
    \includegraphics[height=0.092\linewidth]{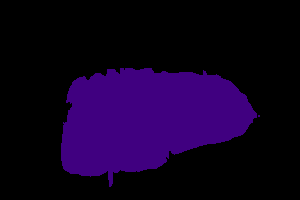} \\

    \includegraphics[height=0.104\linewidth]{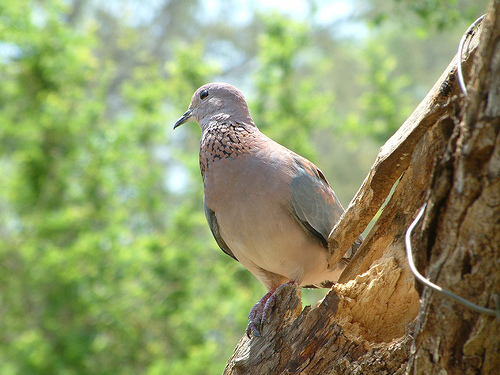} &
    \includegraphics[height=0.104\linewidth]{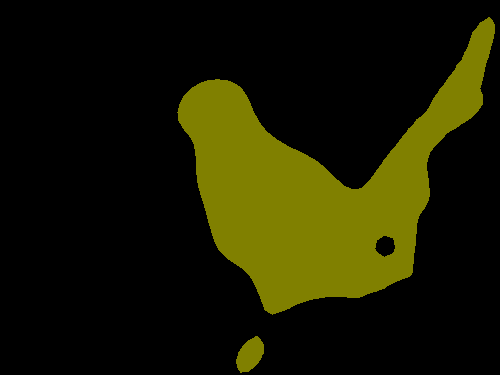} &
    \includegraphics[height=0.104\linewidth]{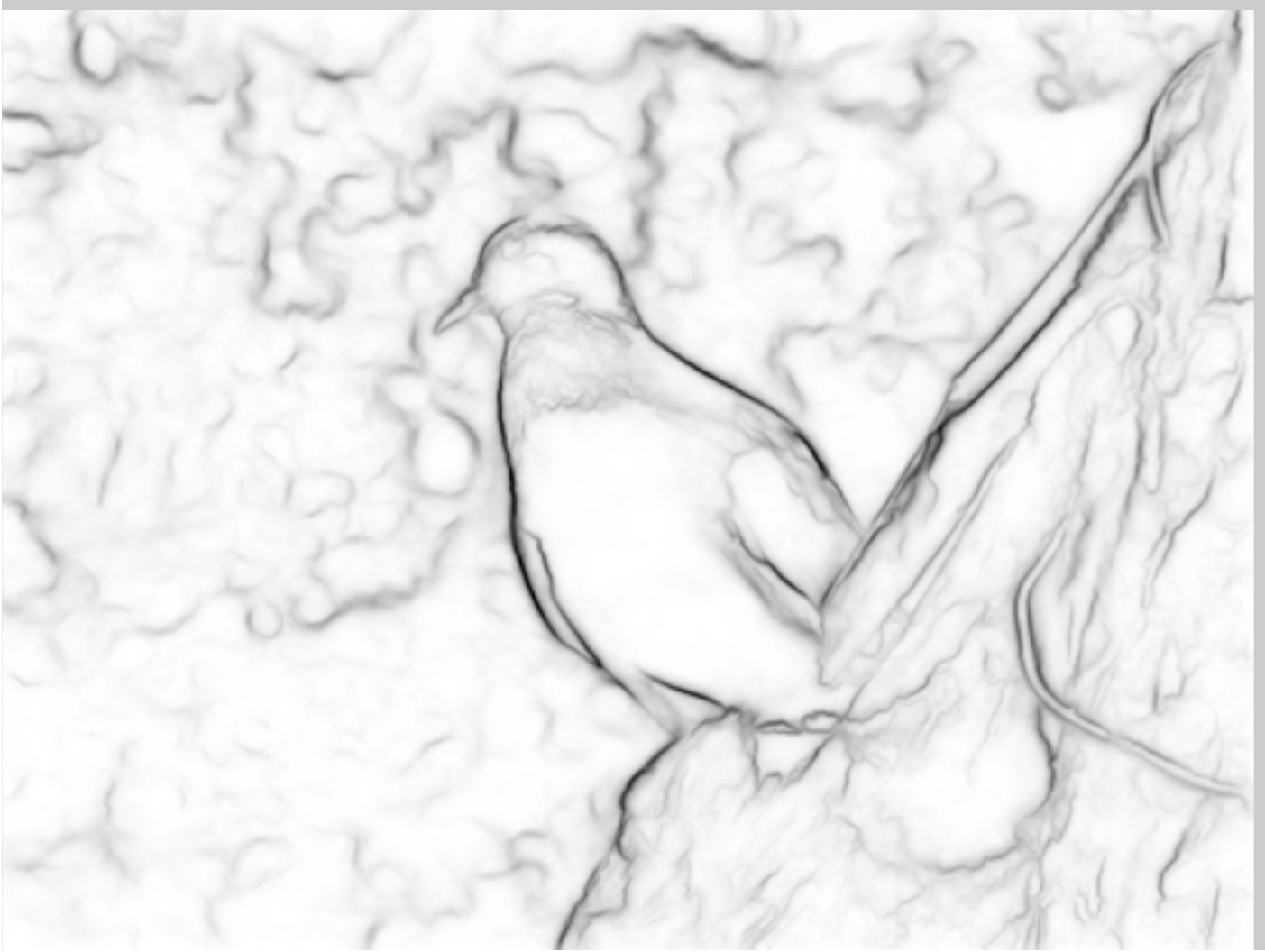} &
    \includegraphics[height=0.104\linewidth]{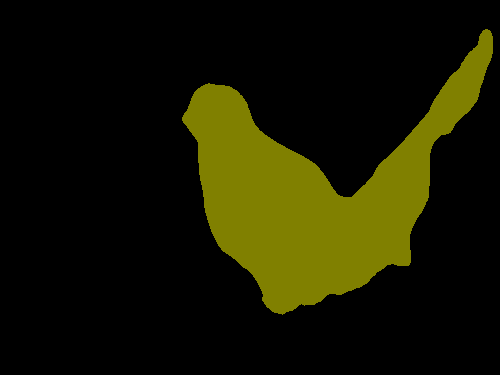} &
    \includegraphics[height=0.104\linewidth]{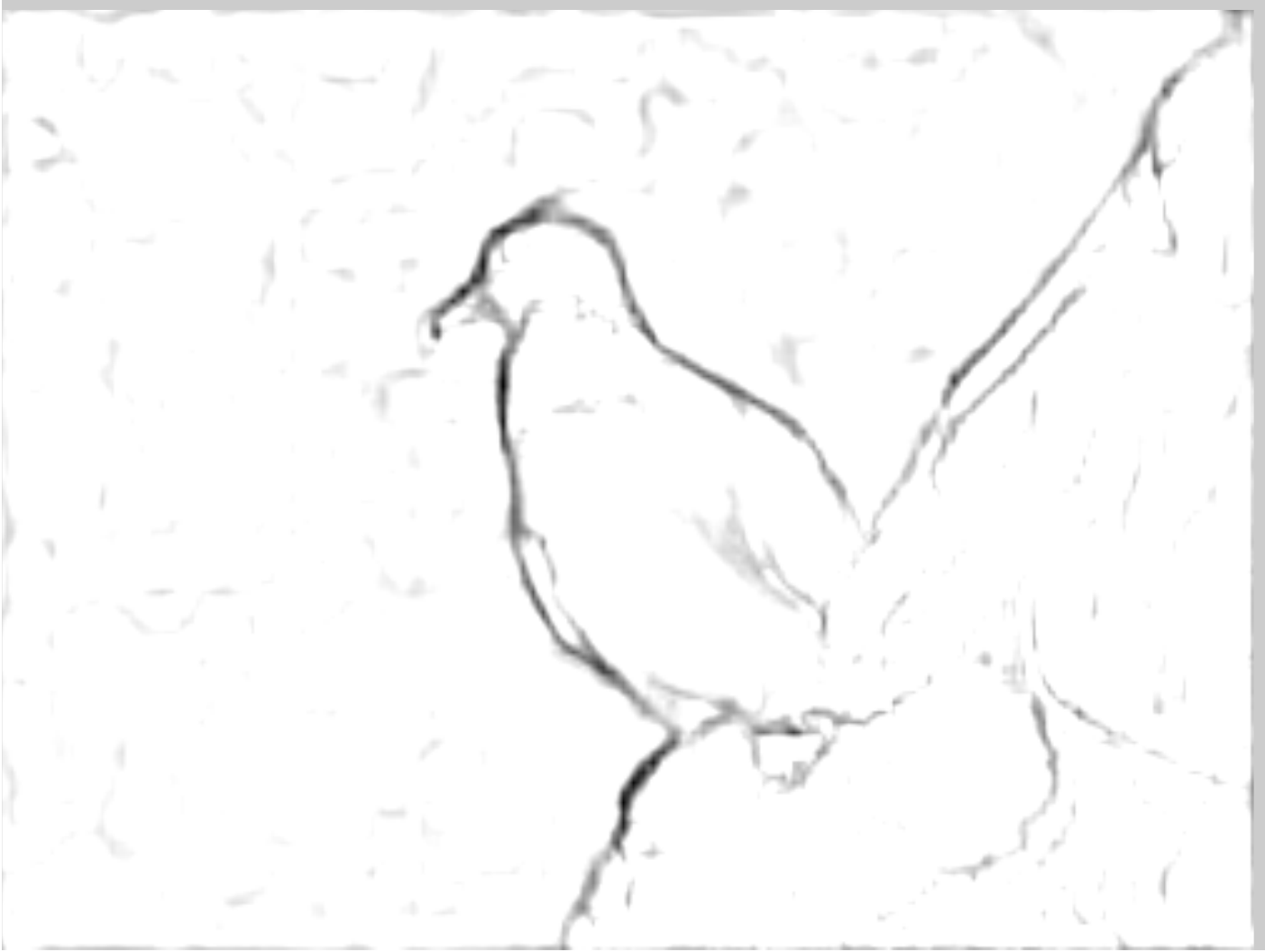} &
    \includegraphics[height=0.104\linewidth]{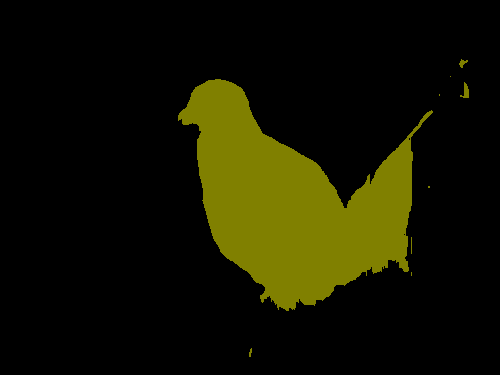} \\

    \includegraphics[height=0.104\linewidth]{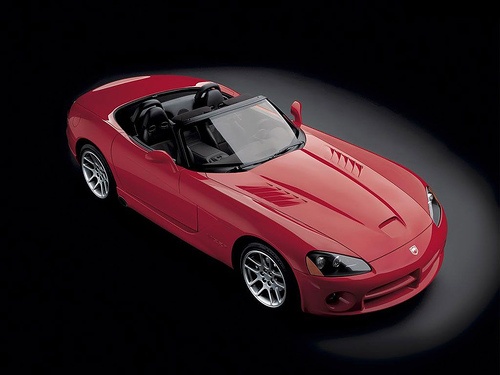} &
    \includegraphics[height=0.104\linewidth]{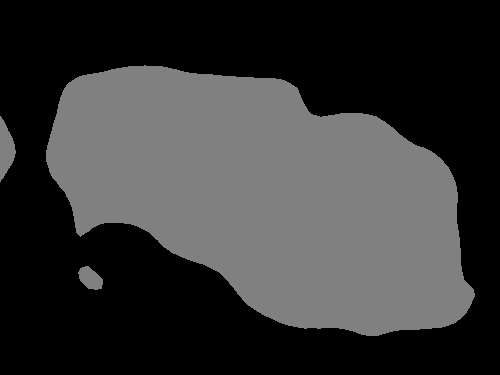} &
    \includegraphics[height=0.104\linewidth]{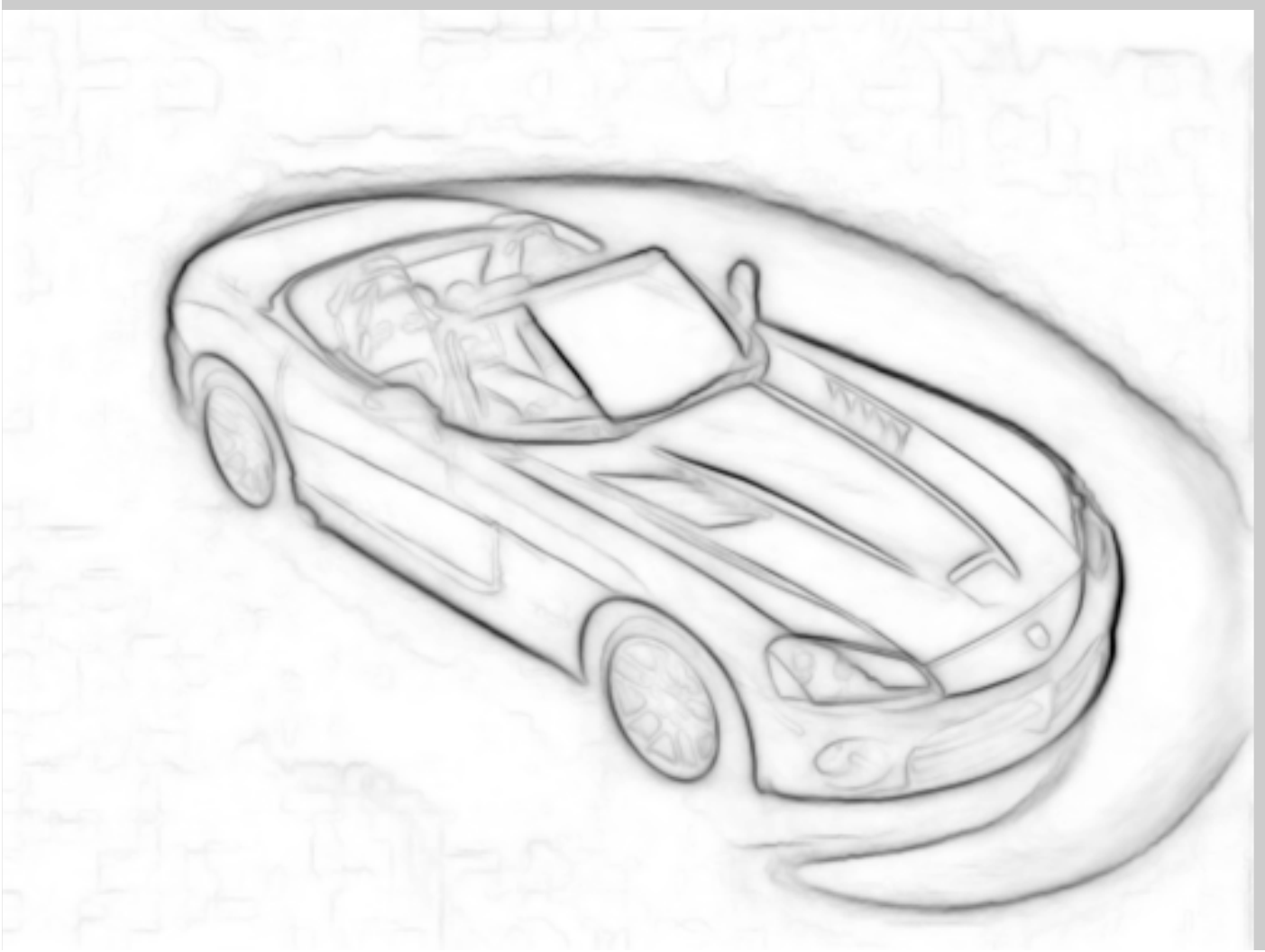} &
    \includegraphics[height=0.104\linewidth]{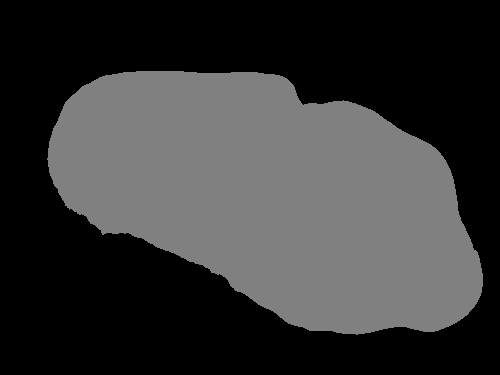} &
    \includegraphics[height=0.104\linewidth]{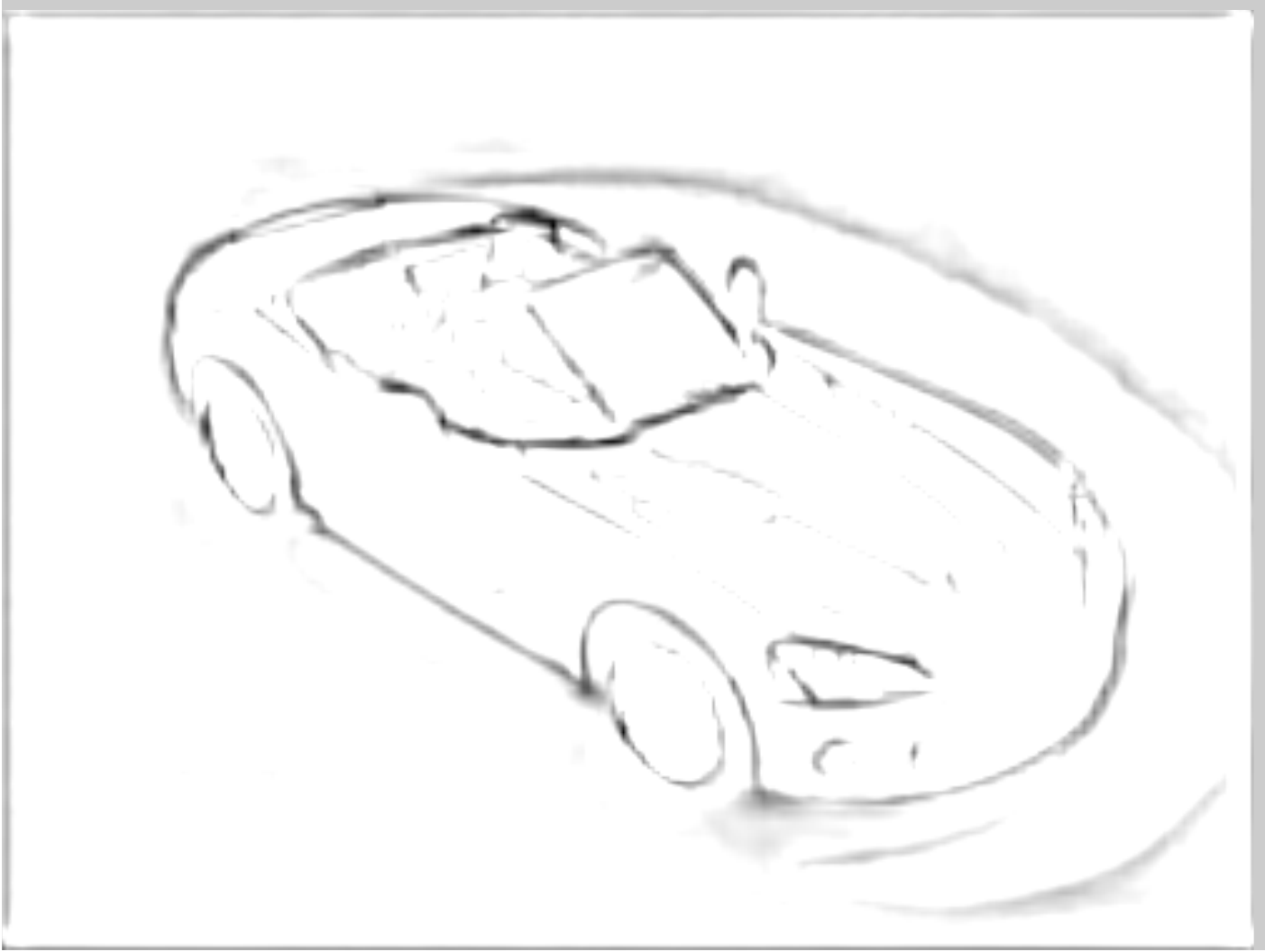} &
    \includegraphics[height=0.104\linewidth]{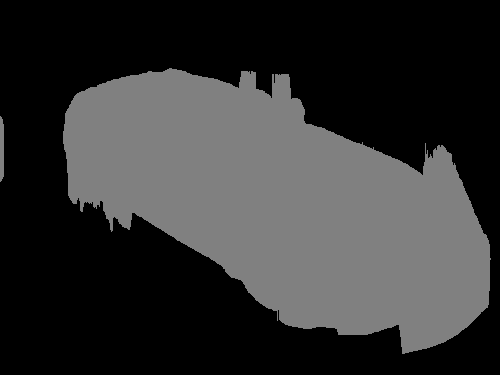} \\

    \includegraphics[width=0.14\linewidth]{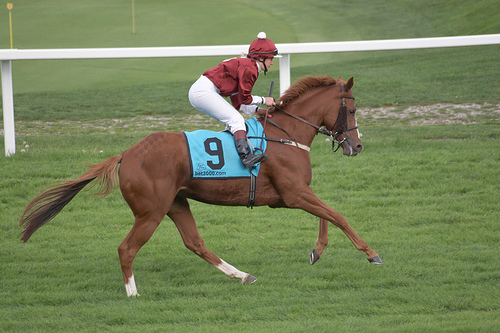} &
    \includegraphics[width=0.14\linewidth]{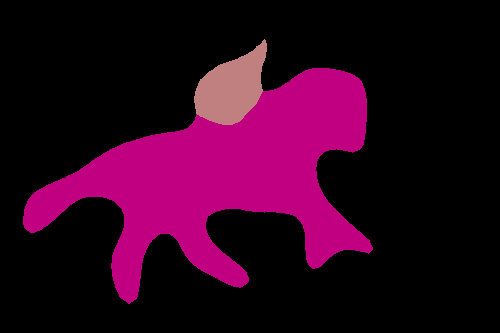} &
    \includegraphics[width=0.14\linewidth]{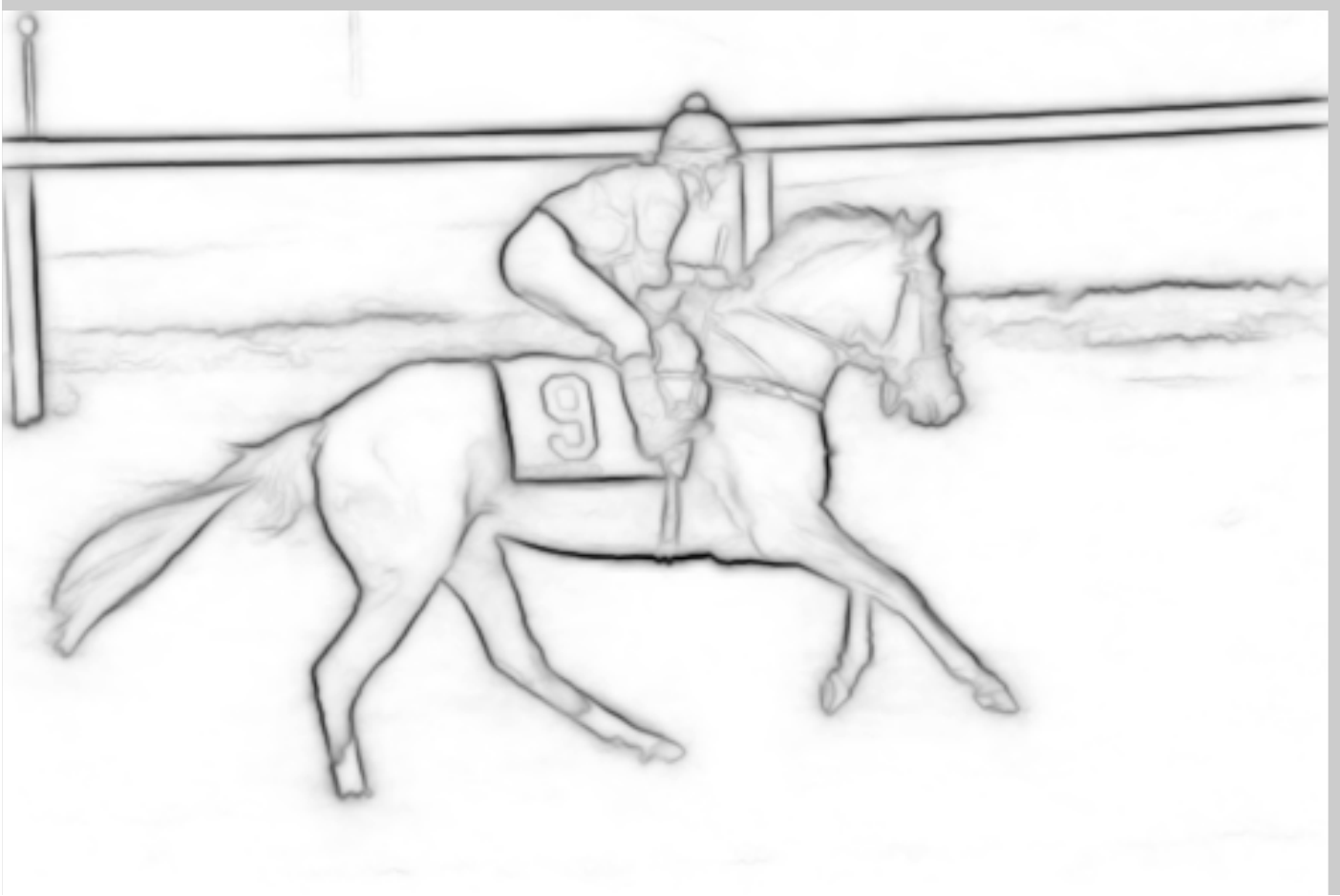} &
    \includegraphics[width=0.14\linewidth]{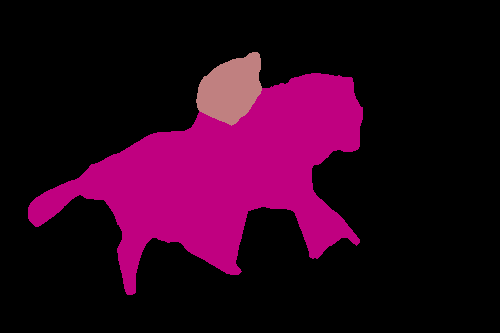} &
    \includegraphics[width=0.14\linewidth]{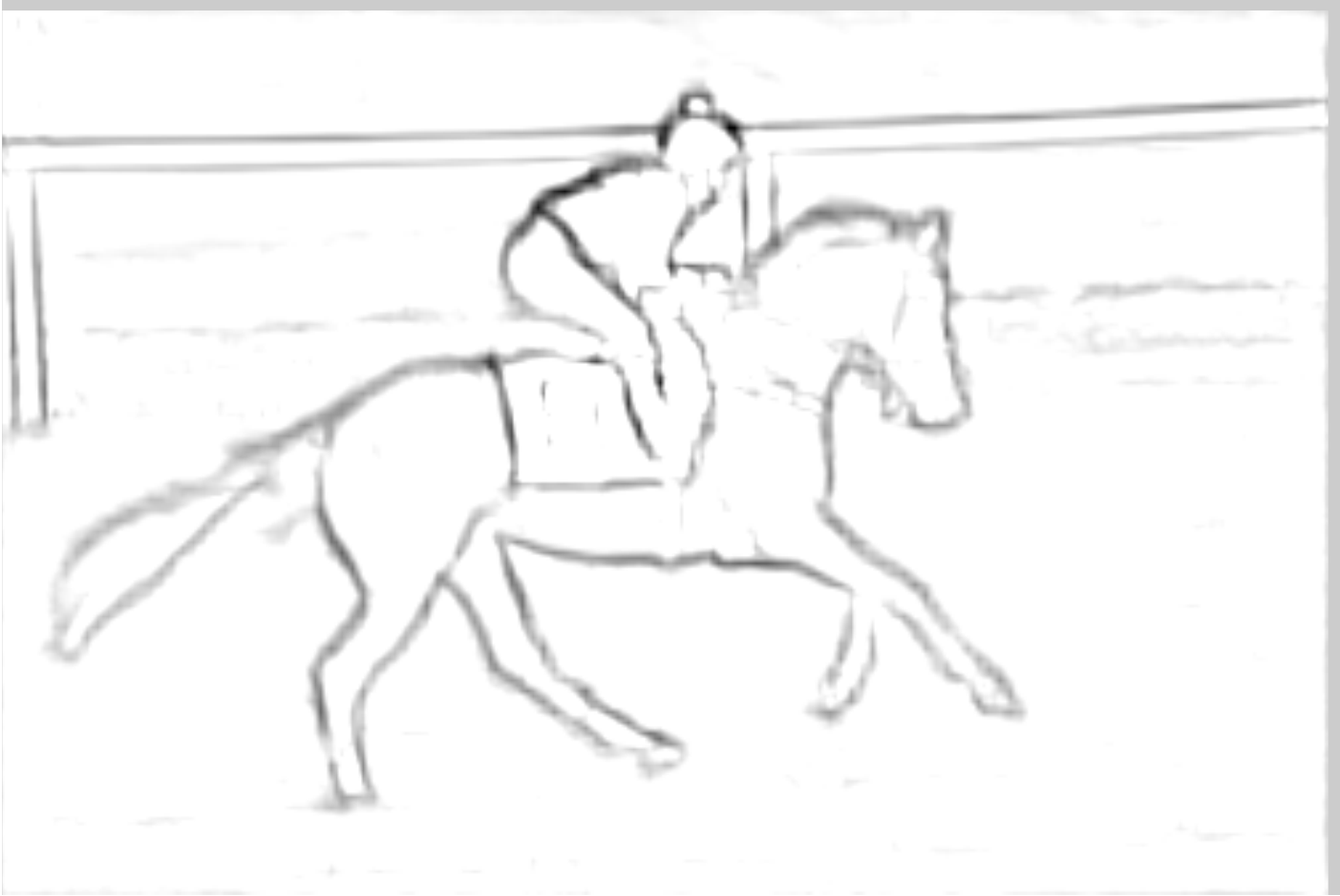} &
    \includegraphics[width=0.14\linewidth]{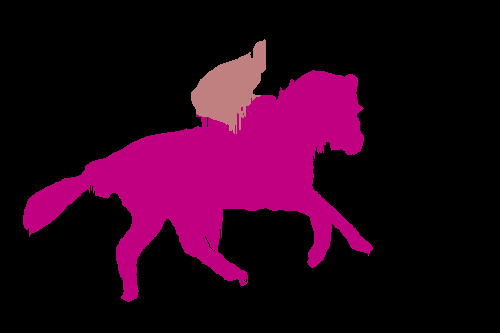} \\

    \includegraphics[width=0.14\linewidth]{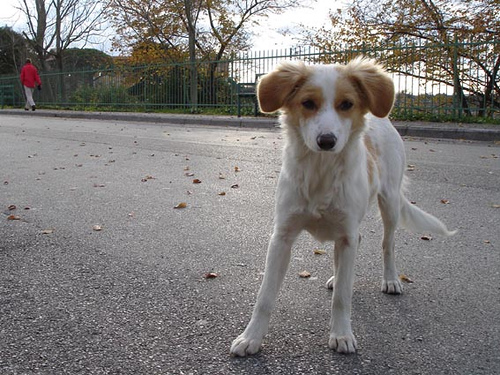} &
    \includegraphics[width=0.14\linewidth]{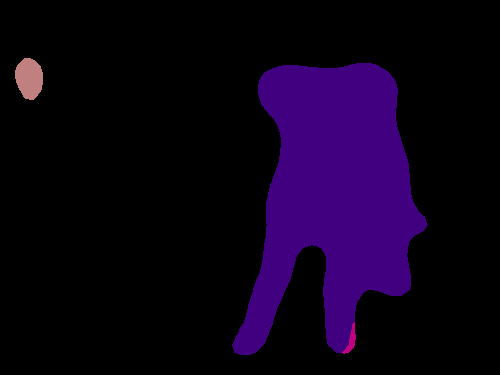} &
    \includegraphics[width=0.14\linewidth]{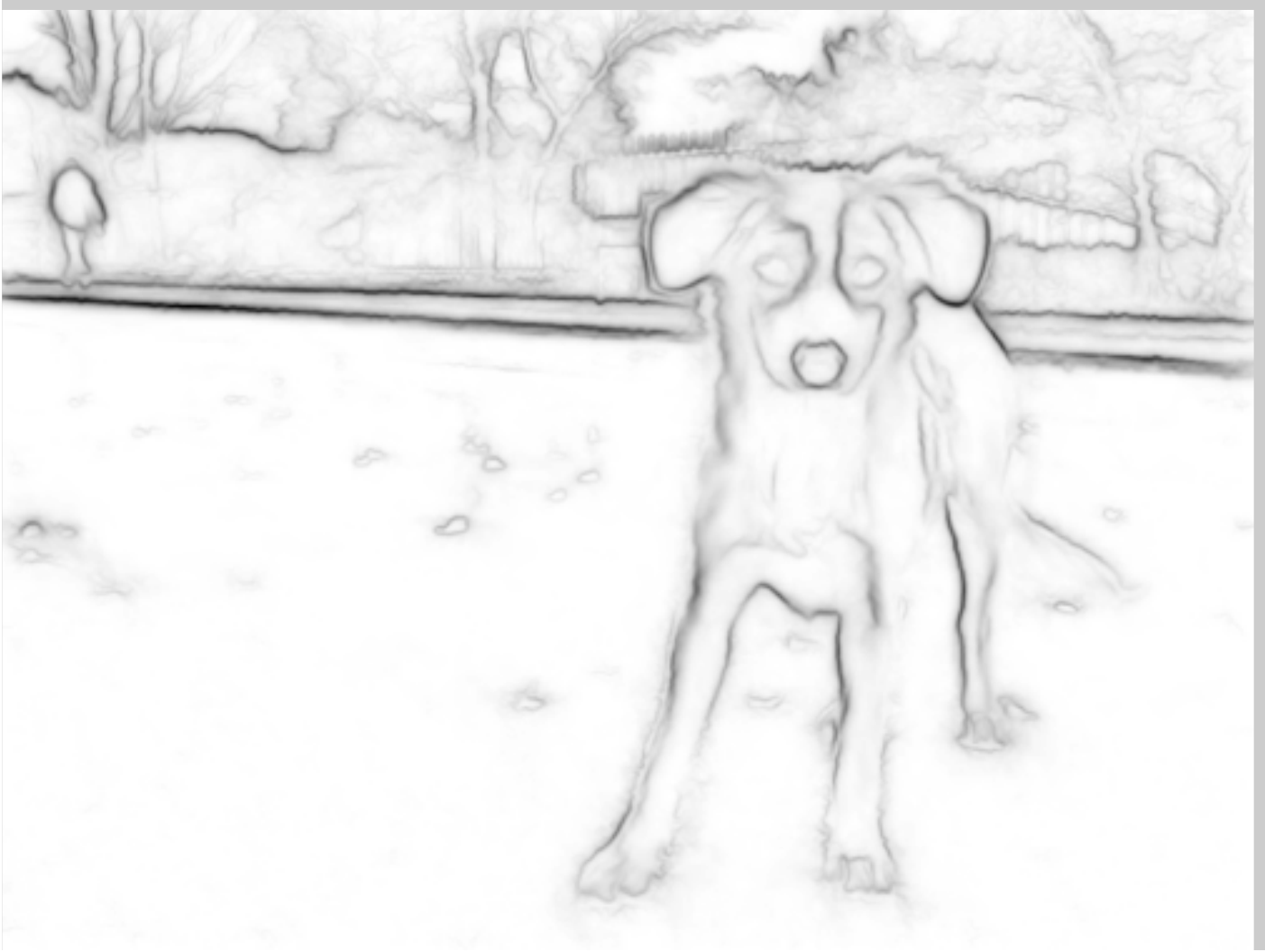} &
    \includegraphics[width=0.14\linewidth]{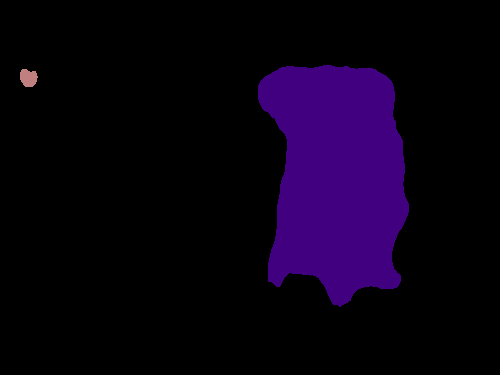} &
    \includegraphics[width=0.14\linewidth]{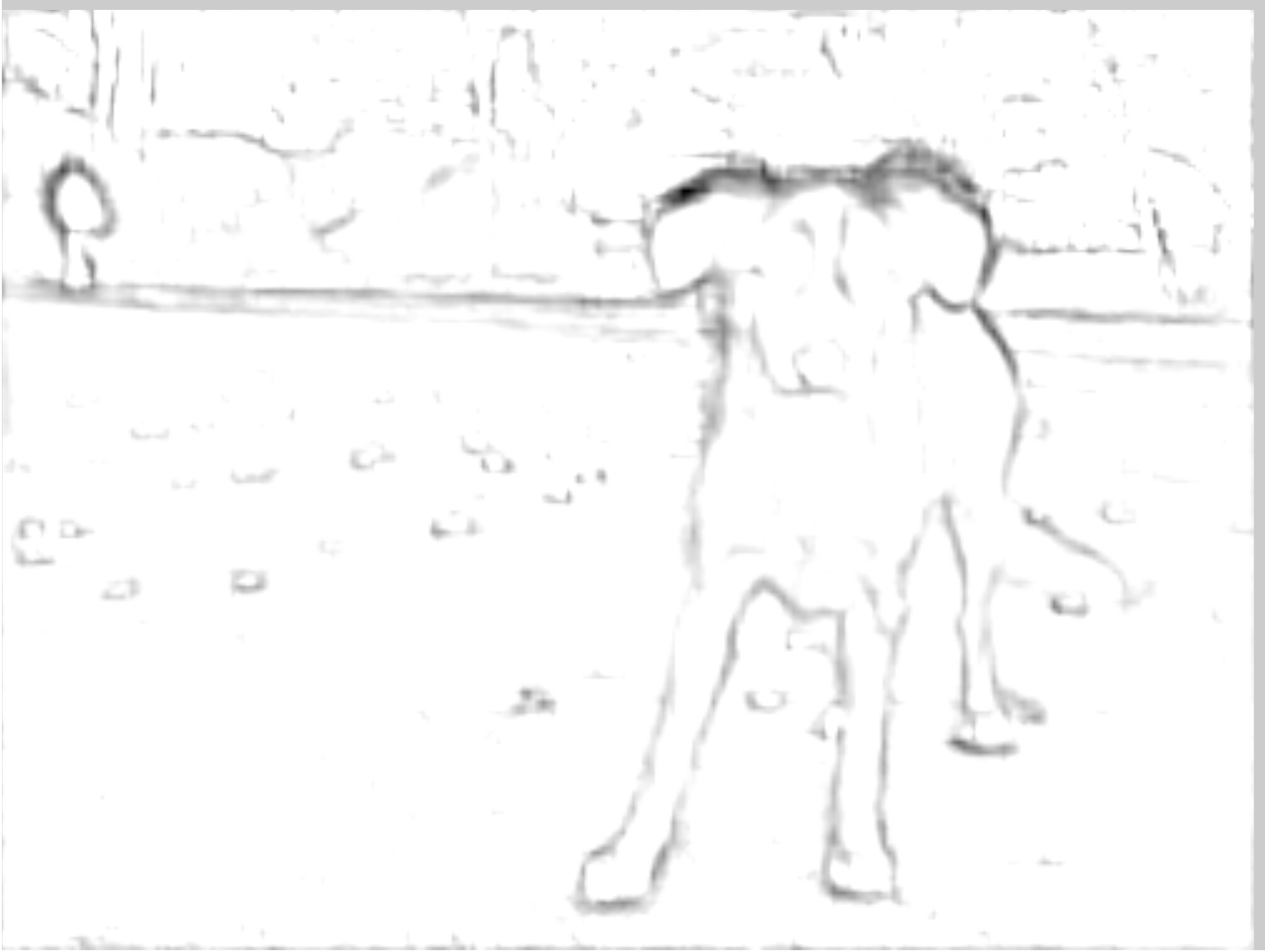} &
    \includegraphics[width=0.14\linewidth]{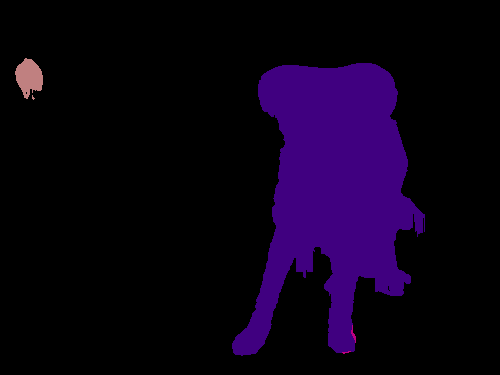} \\

    \includegraphics[width=0.14\linewidth]{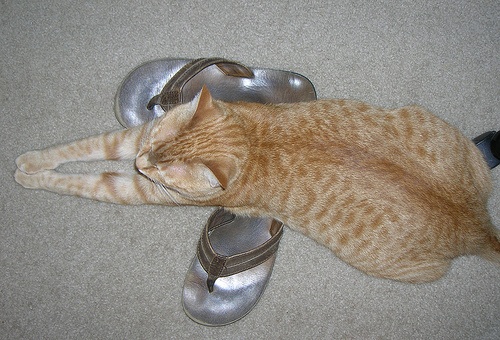} &
    \includegraphics[width=0.14\linewidth]{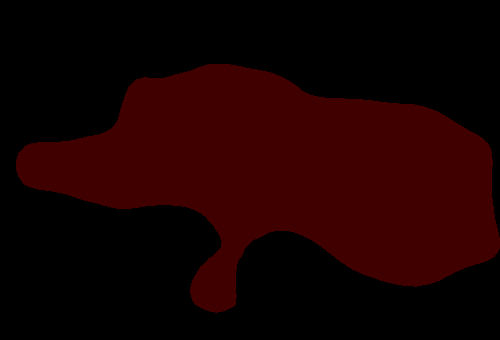} &
    \includegraphics[width=0.14\linewidth]{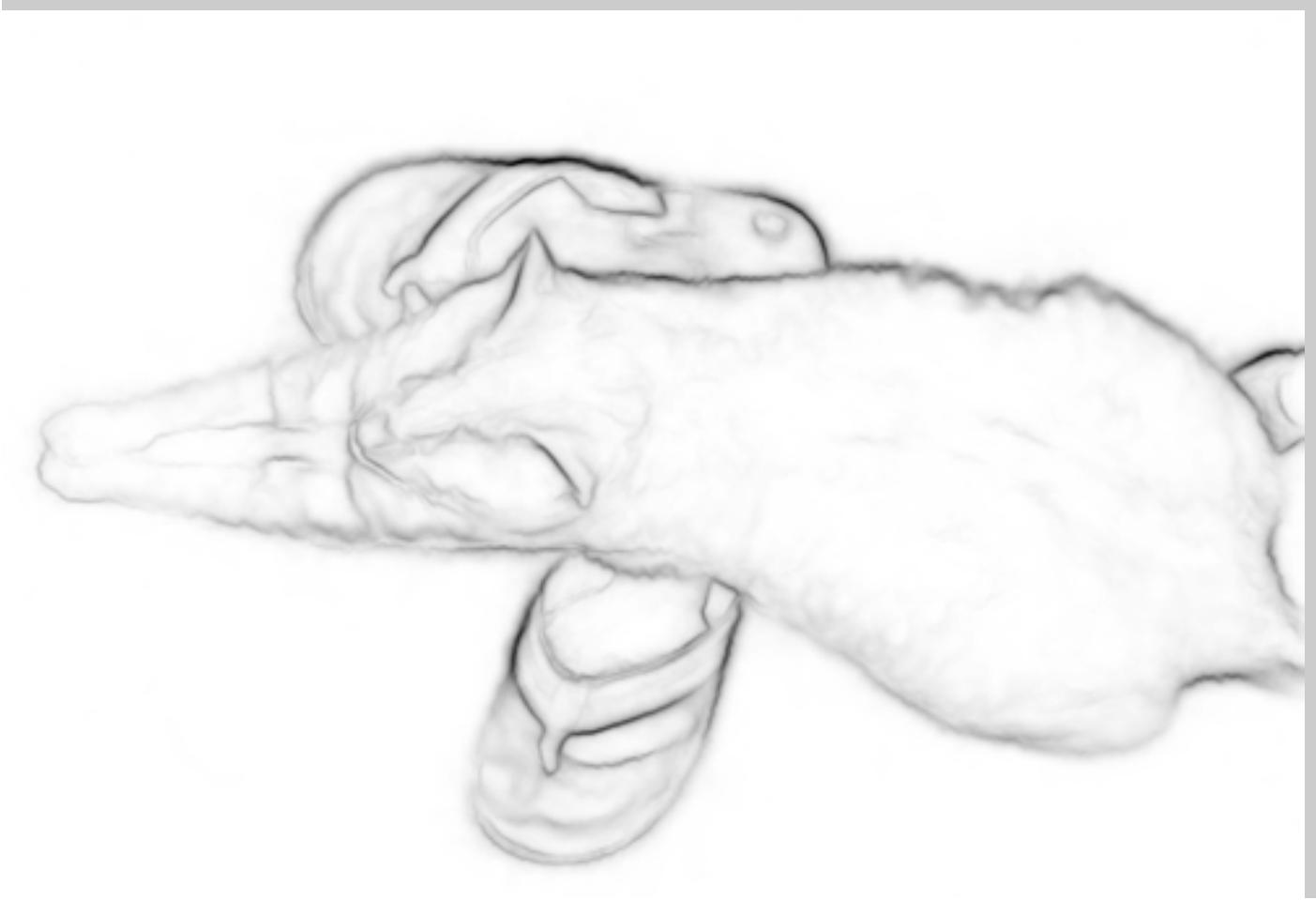} &
    \includegraphics[width=0.14\linewidth]{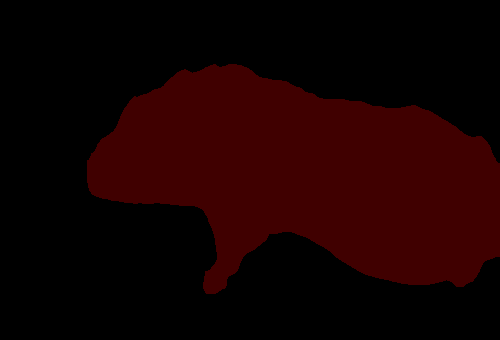} &
    \includegraphics[width=0.14\linewidth]{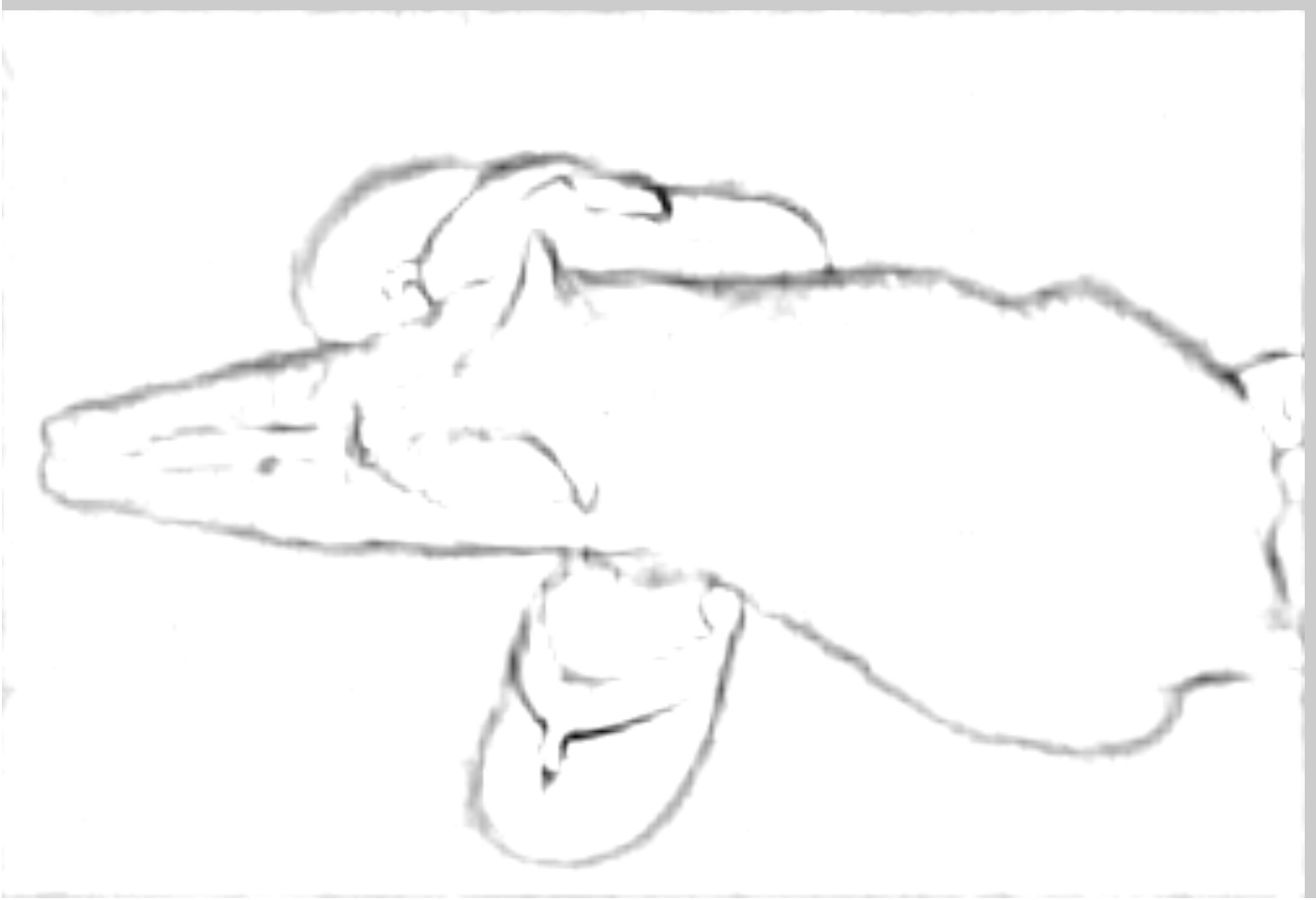} &
    \includegraphics[width=0.14\linewidth]{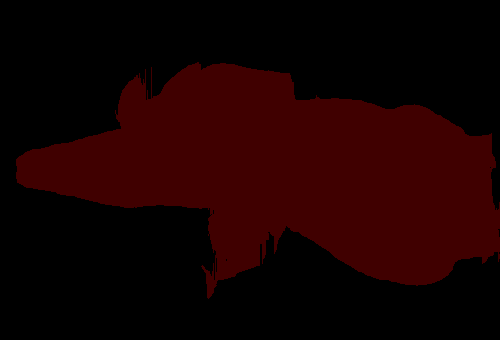} \\
    (a) Image &
    (b) Baseline &
    (c) SE &
    (d) DT-SE &
    (e) EdgeNet &
    (f) DT-EdgeNet \\
  \end{tabular}
  \caption{Visualizing results on VOC 2012 val set. For each row, we
    show (a) Image, (b) Baseline (DeepLab-LargeFOV) segmentation
    result, (c) edges produced by Structured Edges, (d) segmentation
    result with Structured Edges, (e) edges generated by EdgeNet, and
    (f) segmentation result with EdgeNet. Similar to Fig.~(9) of main
    paper.}
  \label{fig:pascal_voc12_seg_1}
\end{figure*}

\begin{figure*}[p]
  \centering
  \begin{tabular}{c c c c c c}
    \includegraphics[height=0.13\linewidth]{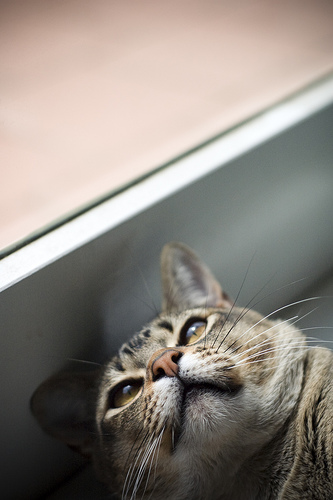} &
    \includegraphics[height=0.13\linewidth]{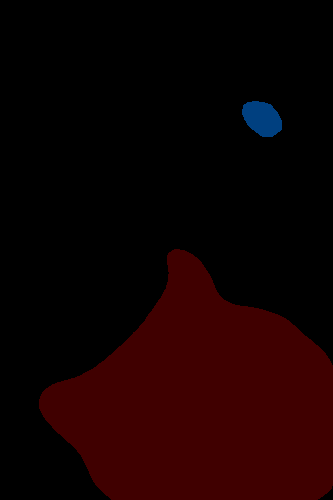} &
    \includegraphics[height=0.13\linewidth]{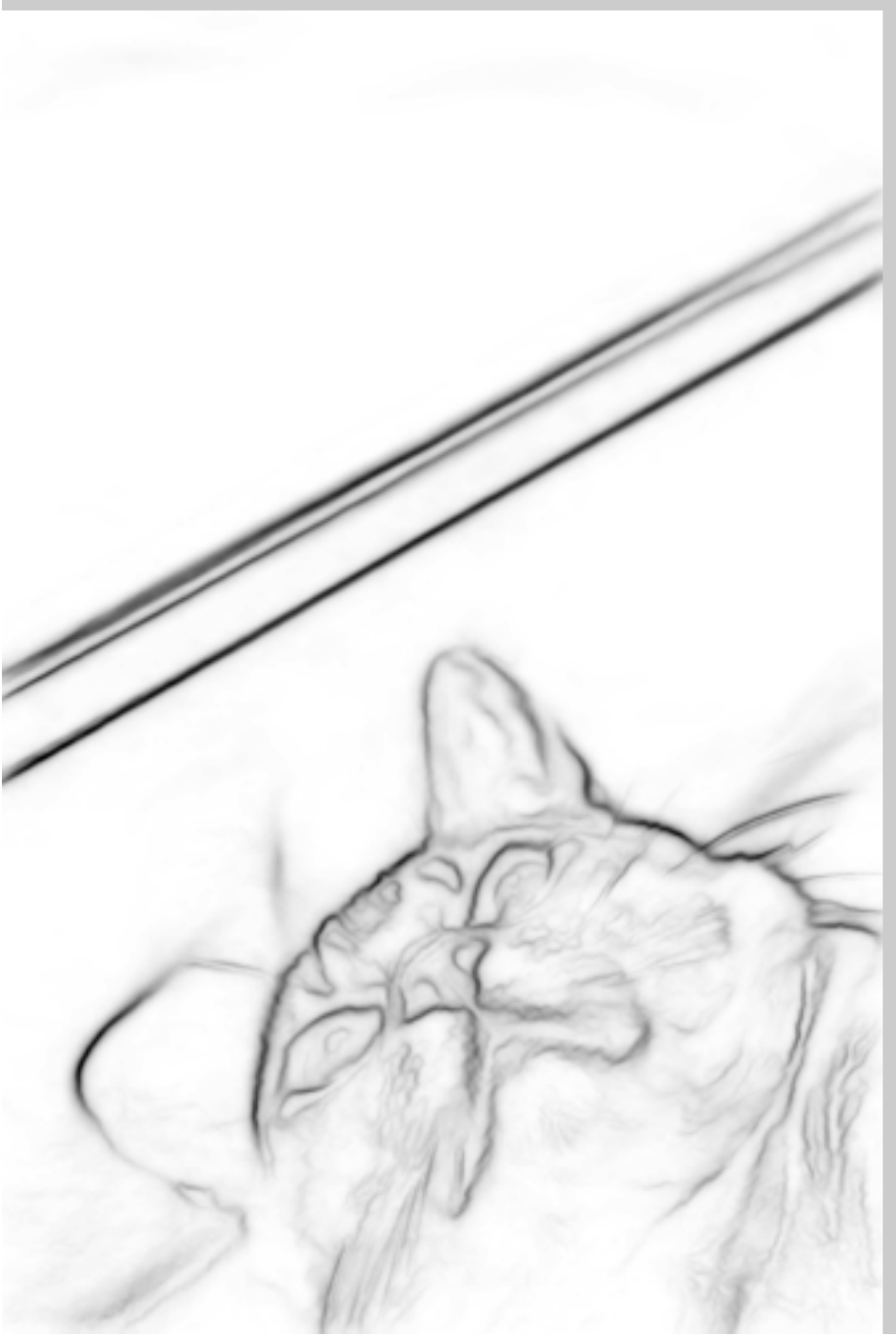} &
    \includegraphics[height=0.13\linewidth]{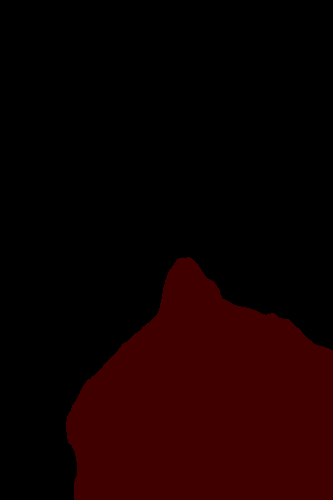} &
    \includegraphics[height=0.13\linewidth]{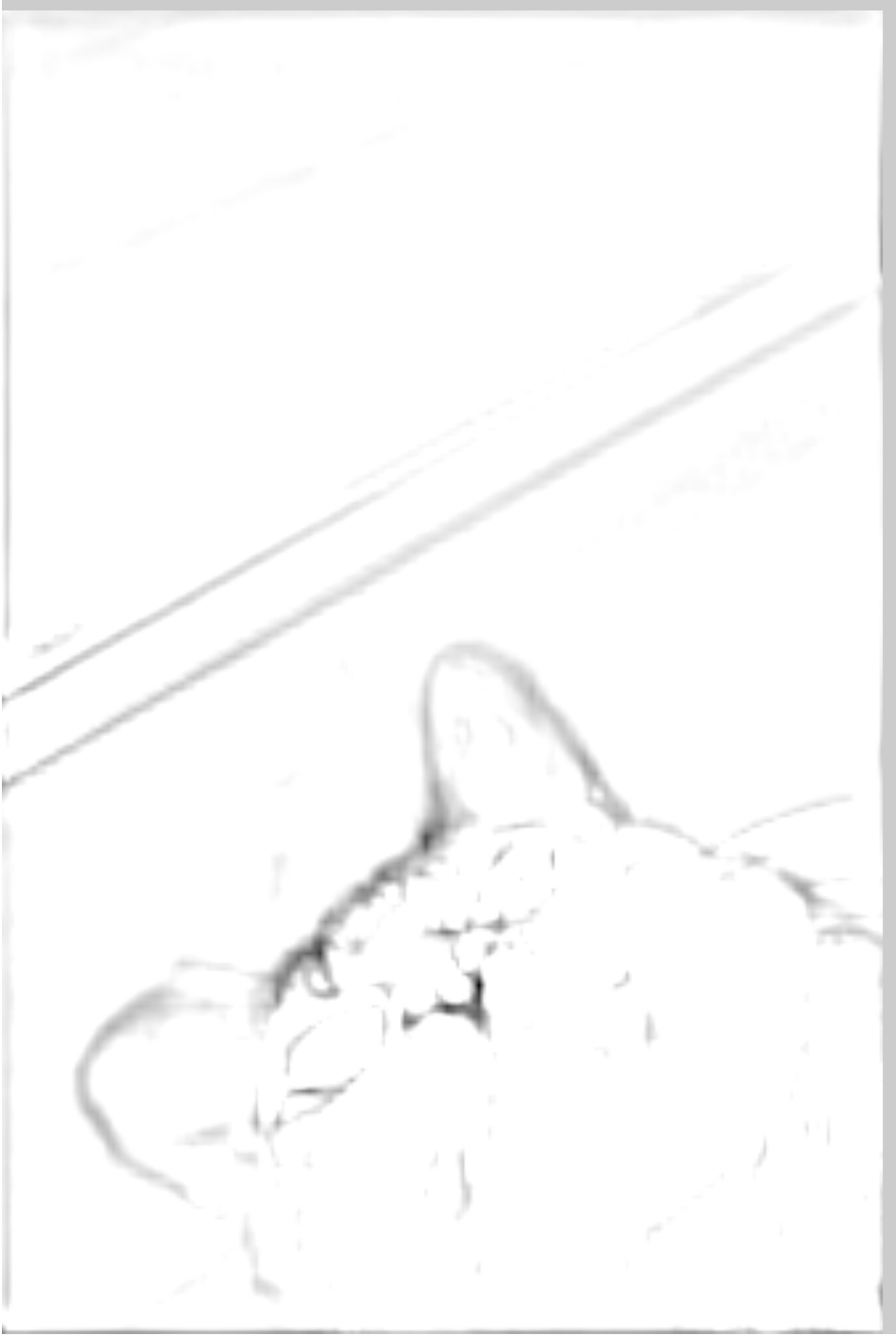} &
    \includegraphics[height=0.13\linewidth]{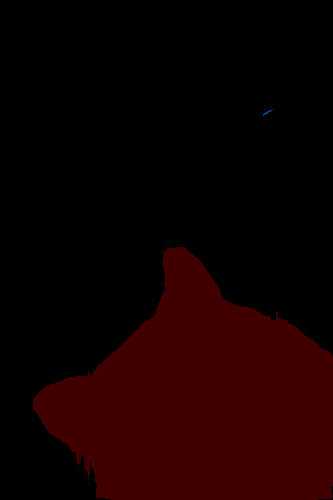} \\

    \includegraphics[height=0.13\linewidth]{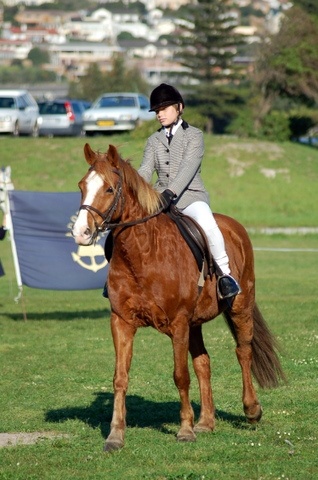} &
    \includegraphics[height=0.13\linewidth]{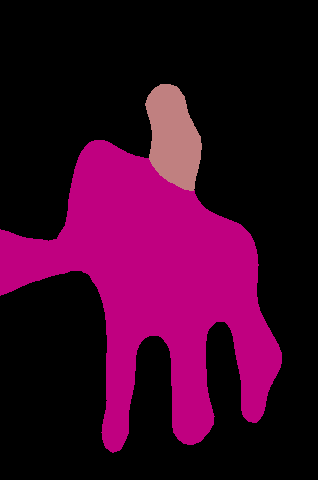} &
    \includegraphics[height=0.13\linewidth]{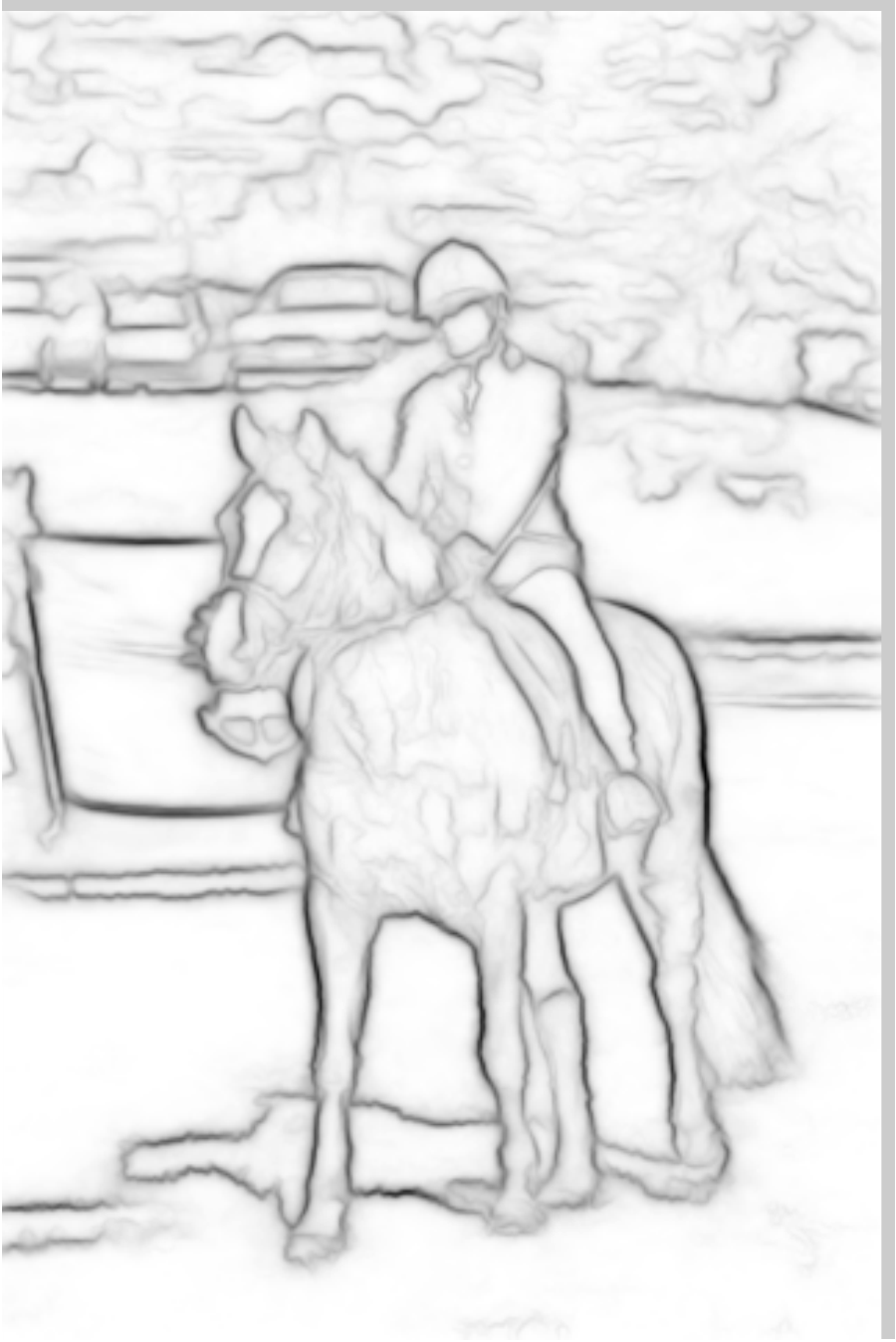} &
    \includegraphics[height=0.13\linewidth]{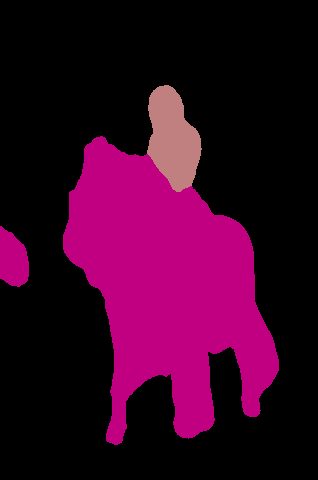} &
    \includegraphics[height=0.13\linewidth]{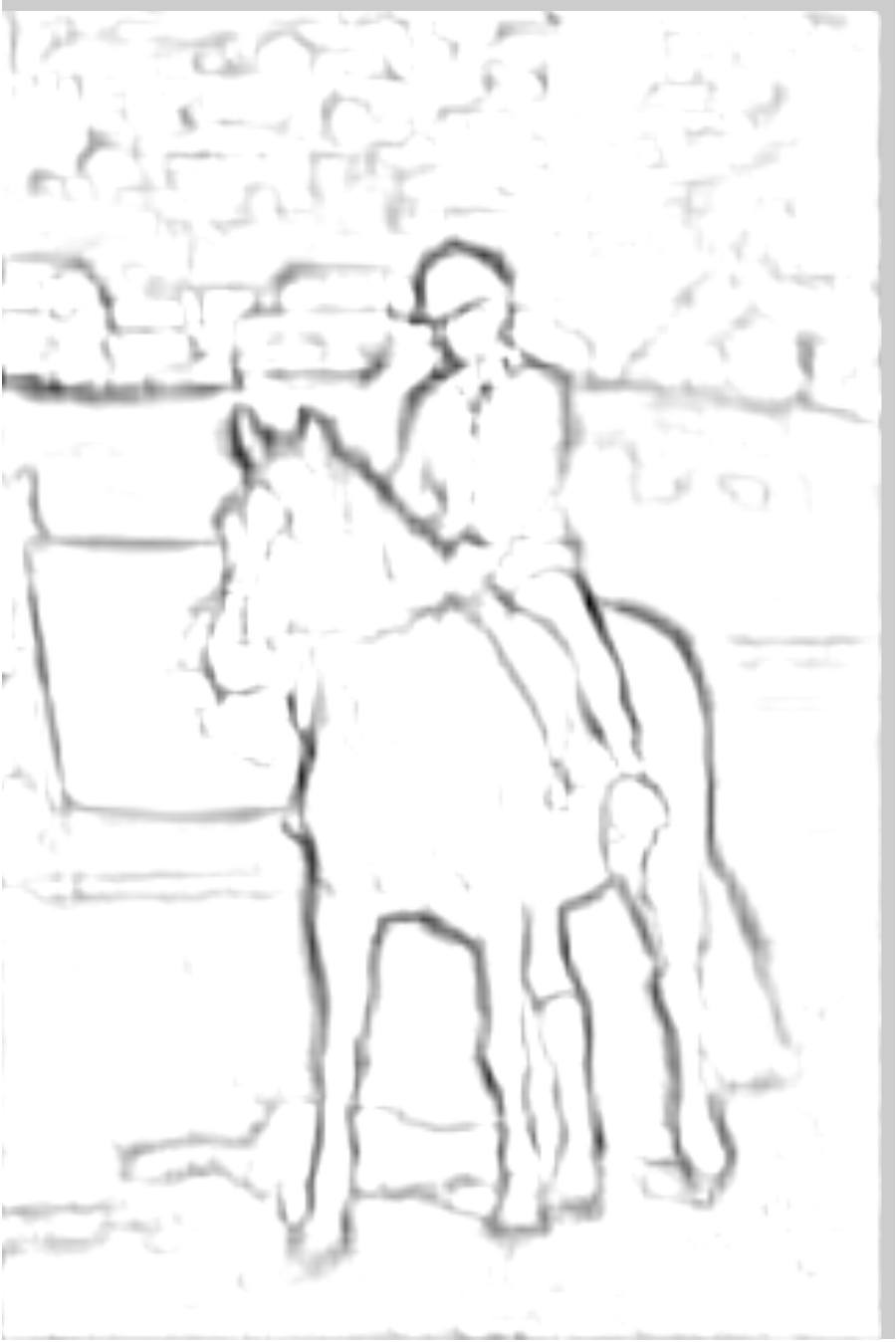} &
    \includegraphics[height=0.13\linewidth]{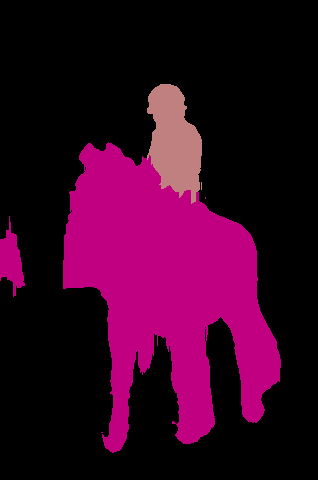} \\

    \includegraphics[width=0.14\linewidth]{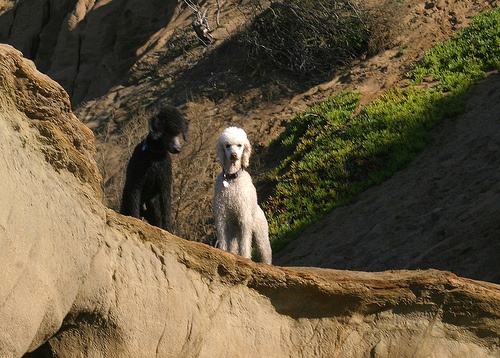} &
    \includegraphics[width=0.14\linewidth]{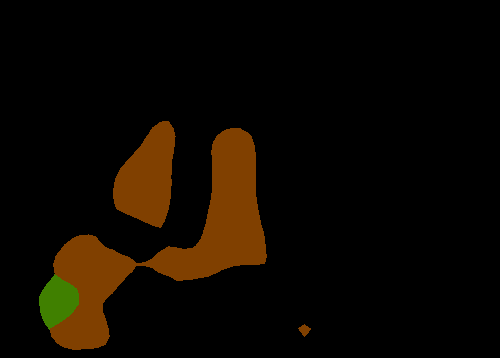} &
    \includegraphics[width=0.14\linewidth]{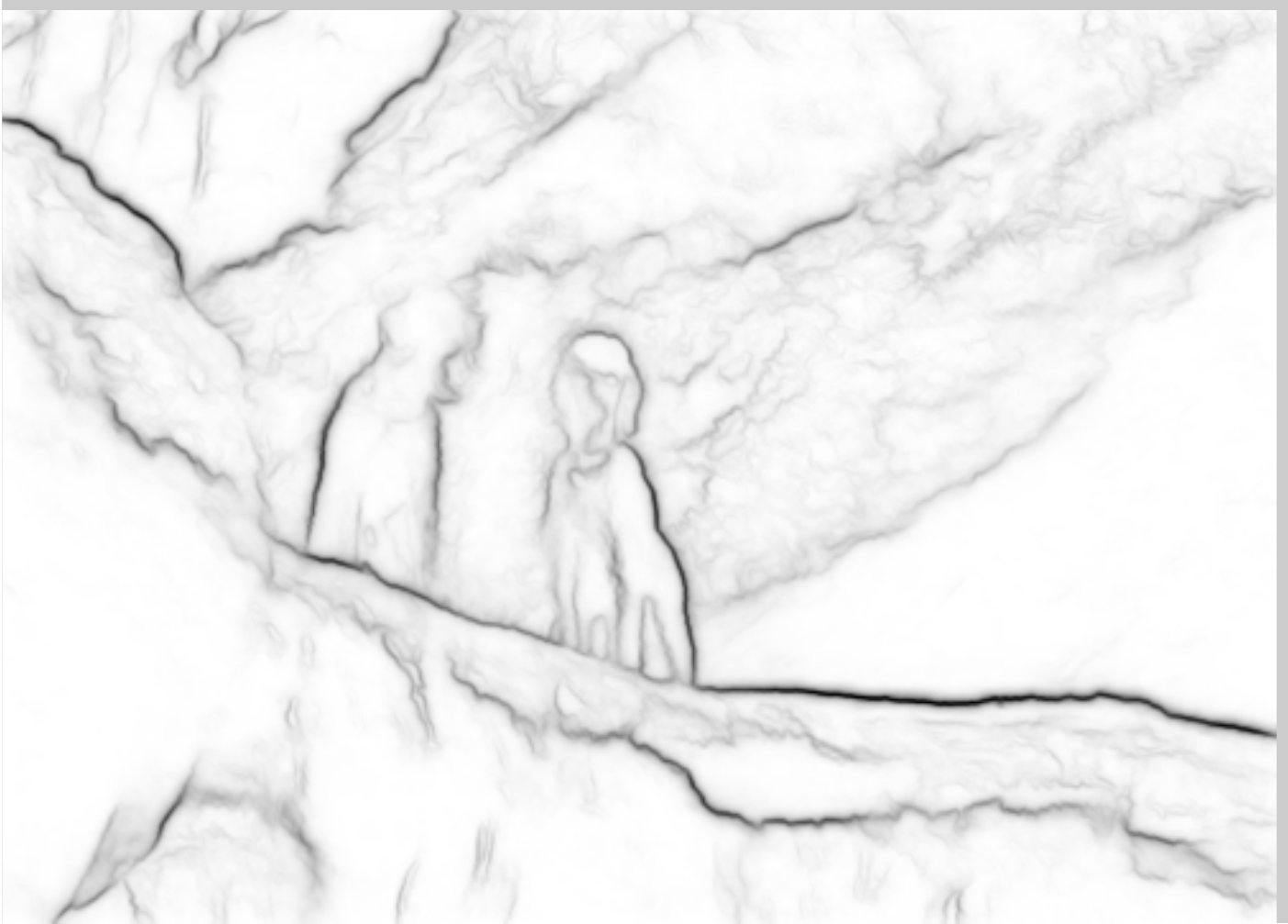} &
    \includegraphics[width=0.14\linewidth]{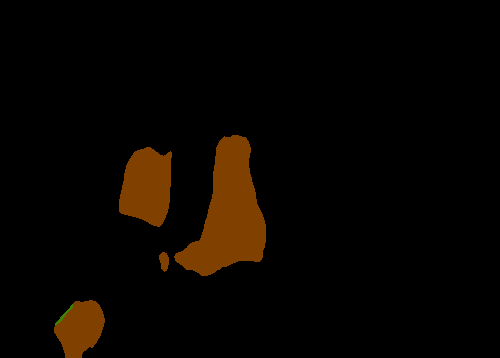} &
    \includegraphics[width=0.14\linewidth]{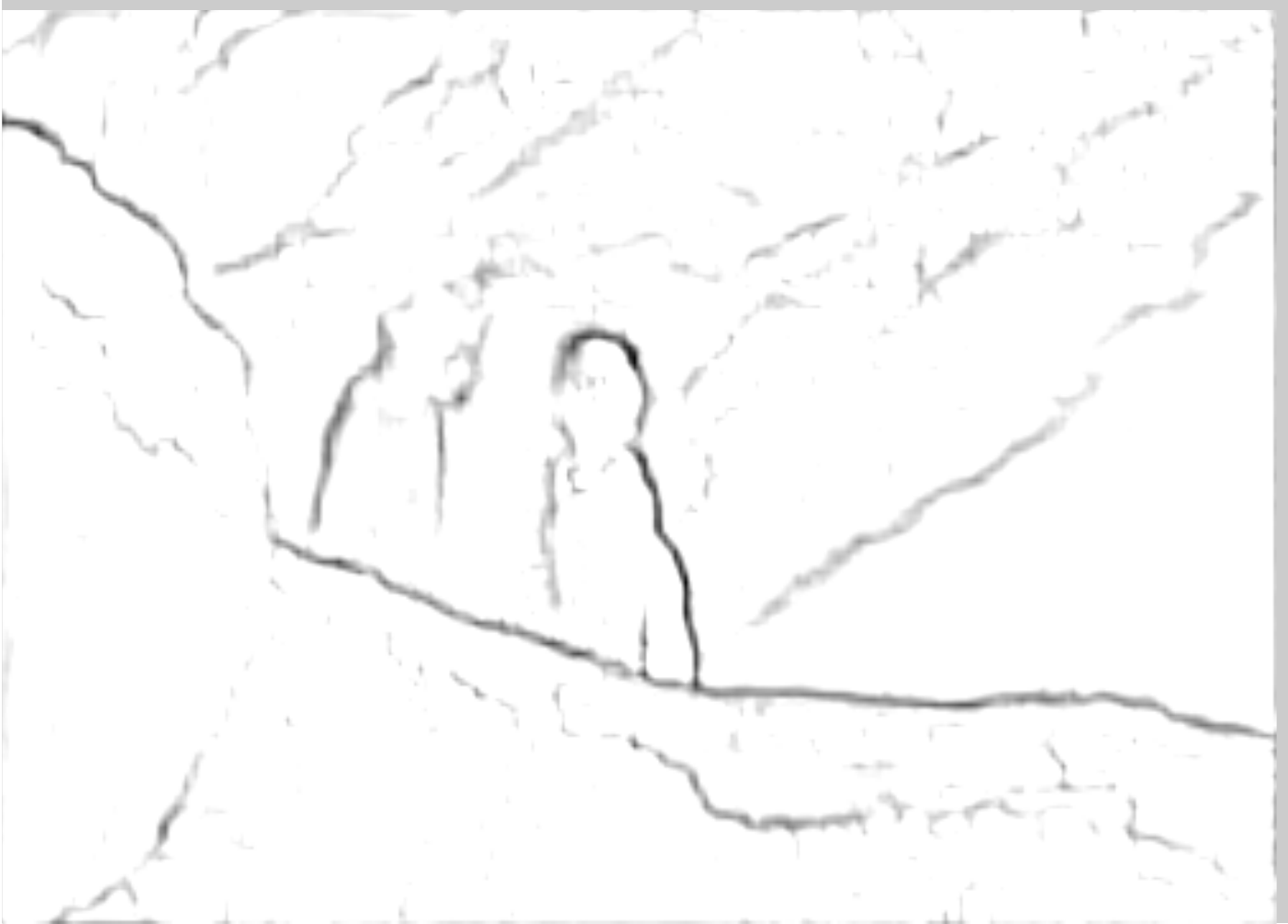} &
    \includegraphics[width=0.14\linewidth]{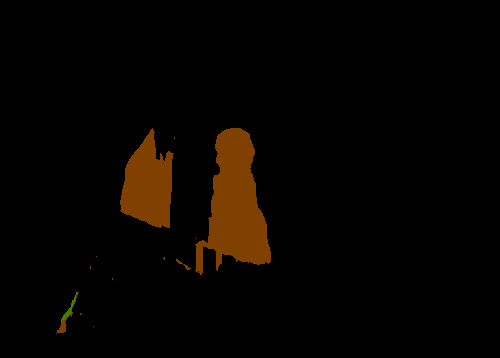} \\

    \includegraphics[width=0.14\linewidth]{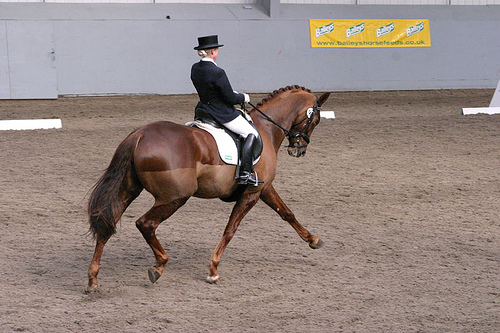} &
    \includegraphics[width=0.14\linewidth]{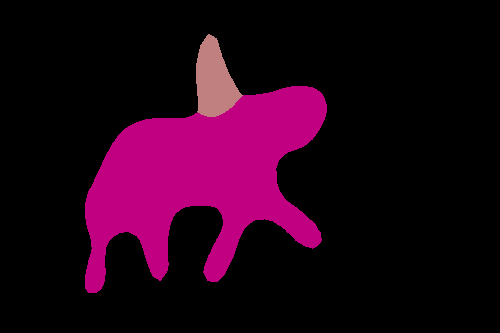} &
    \includegraphics[width=0.14\linewidth]{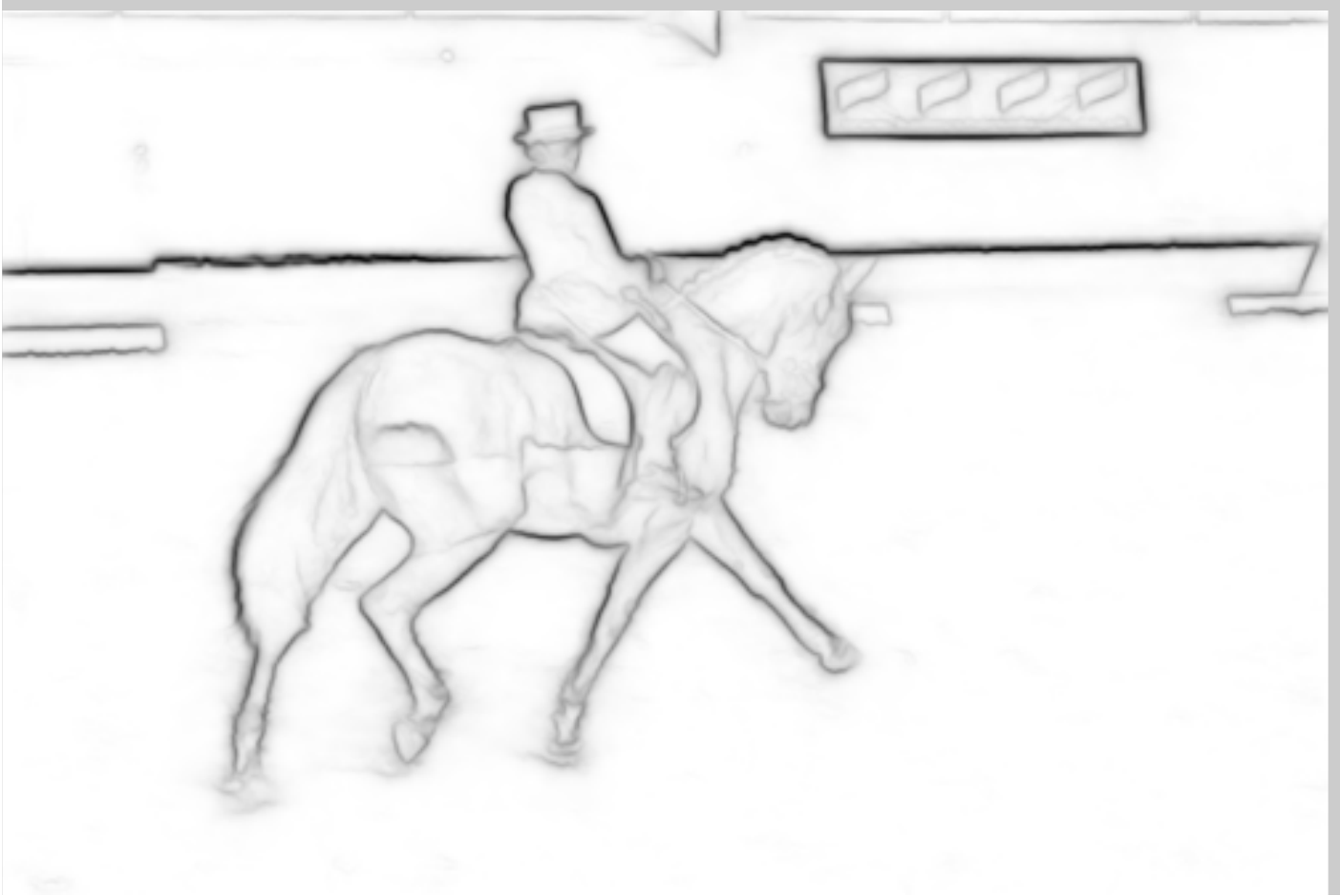} &
    \includegraphics[width=0.14\linewidth]{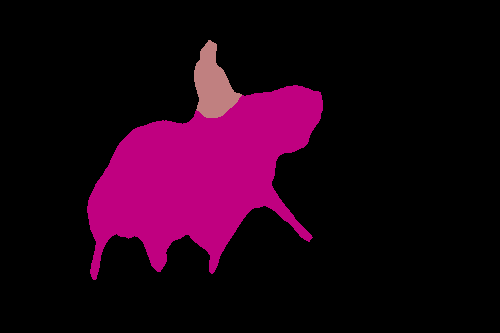} &
    \includegraphics[width=0.14\linewidth]{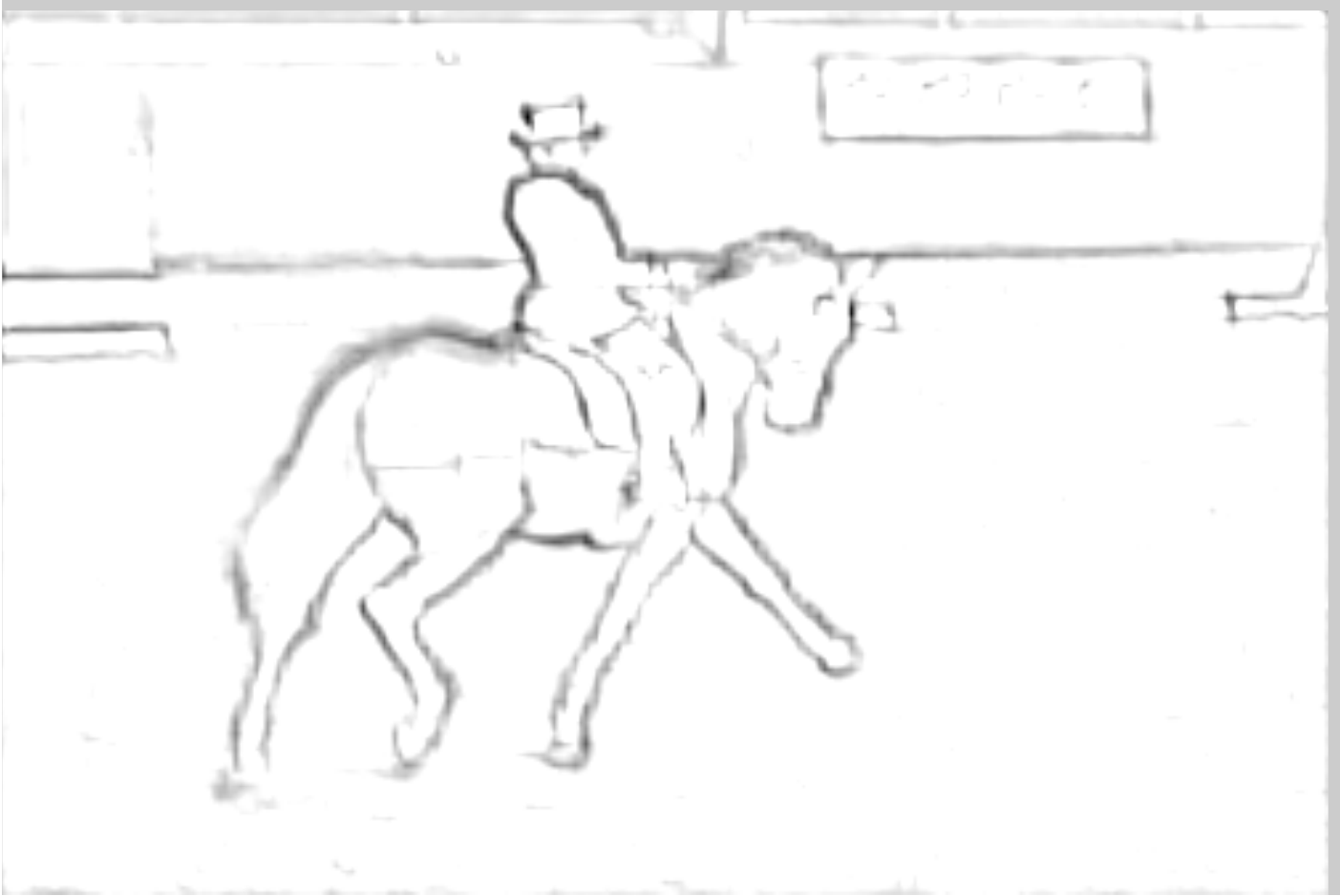} &
    \includegraphics[width=0.14\linewidth]{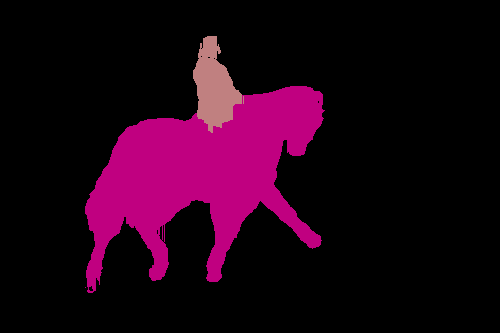} \\

    \includegraphics[width=0.14\linewidth]{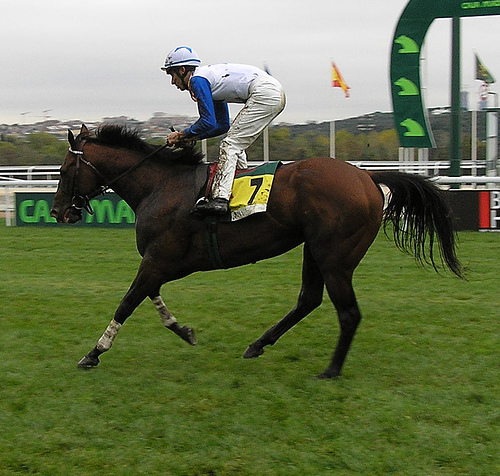} &
    \includegraphics[width=0.14\linewidth]{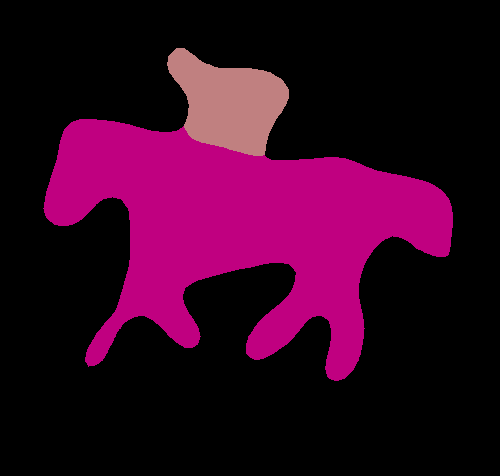} &
    \includegraphics[width=0.14\linewidth]{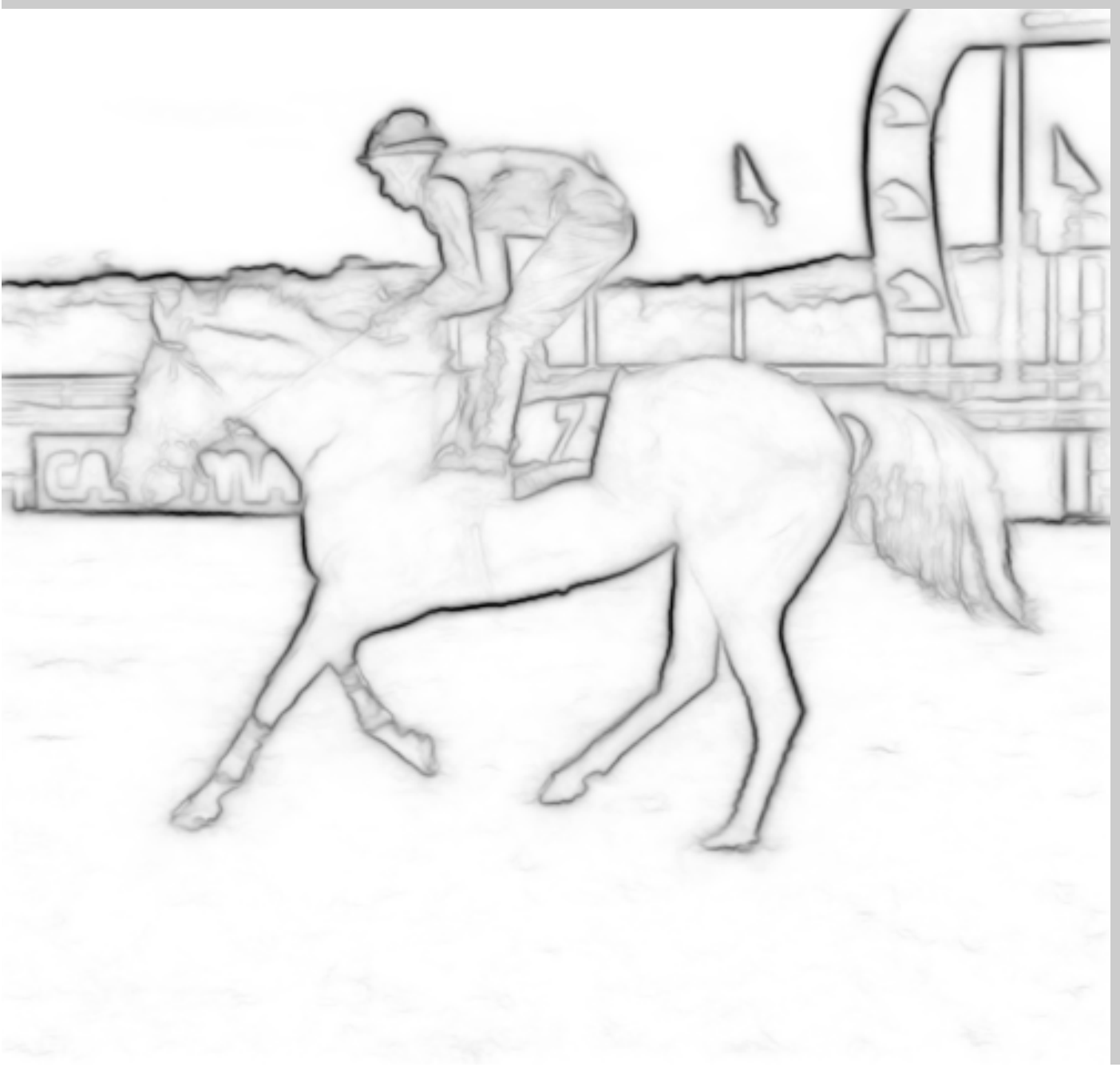} &
    \includegraphics[width=0.14\linewidth]{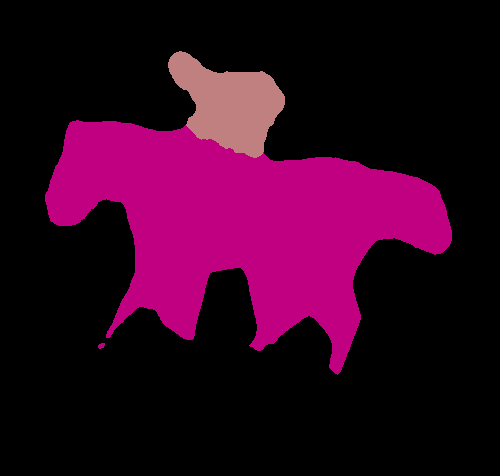} &
    \includegraphics[width=0.14\linewidth]{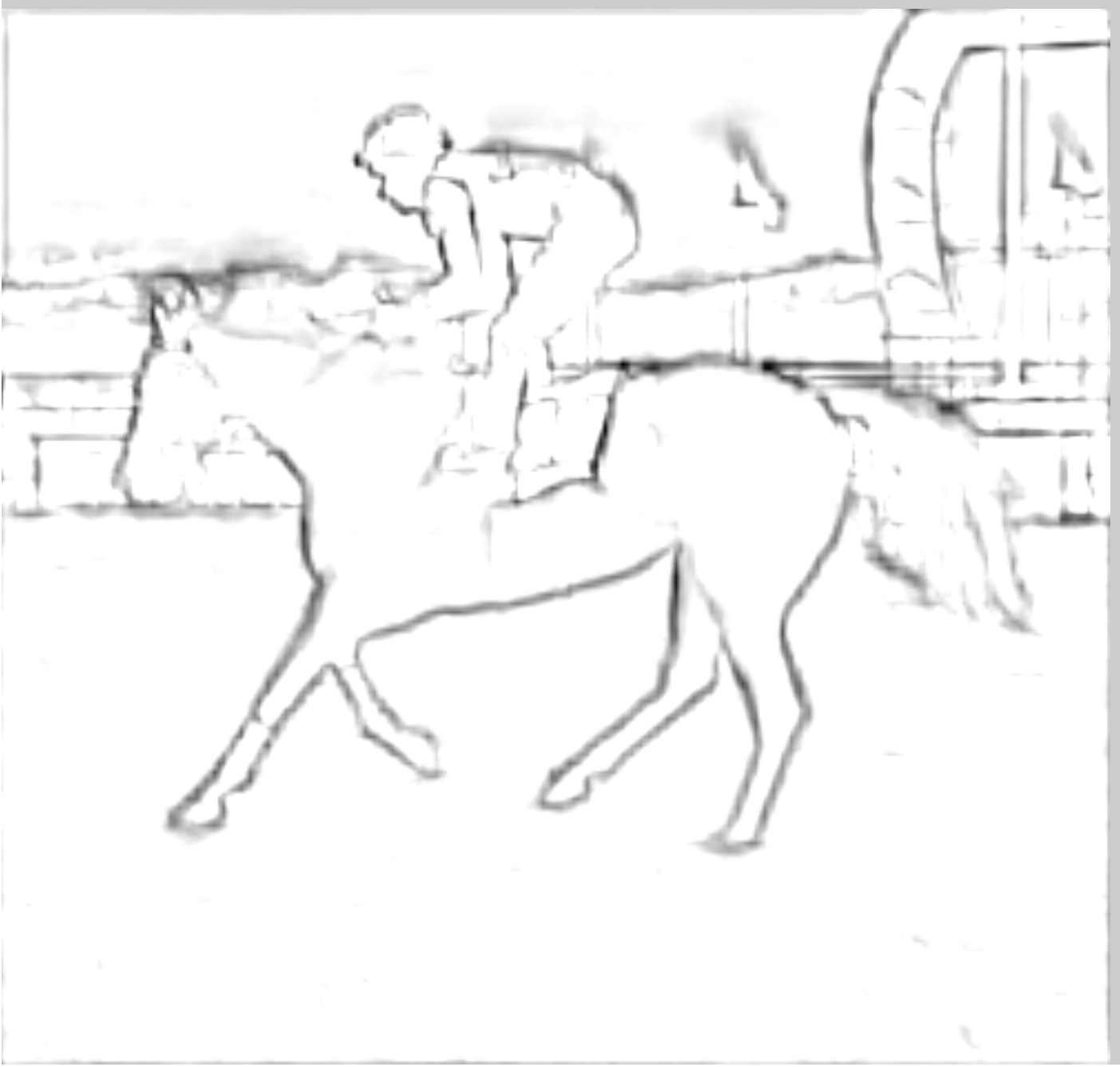} &
    \includegraphics[width=0.14\linewidth]{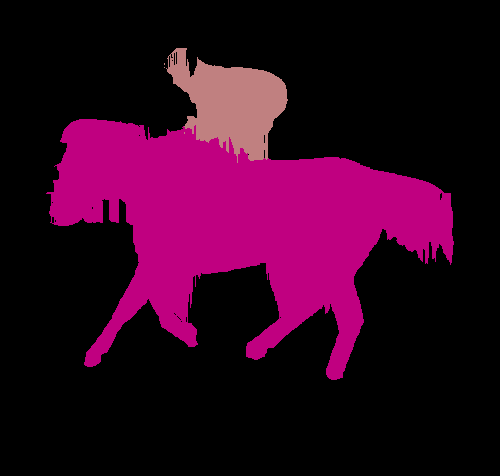} \\

    \includegraphics[height=0.11\linewidth]{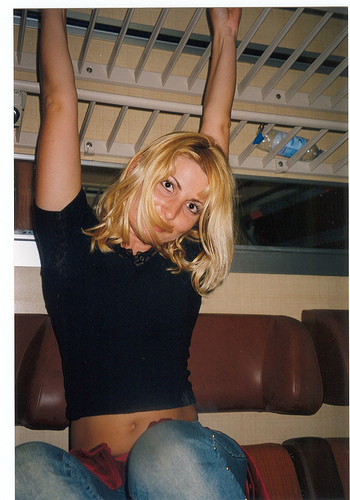} &
    \includegraphics[height=0.11\linewidth]{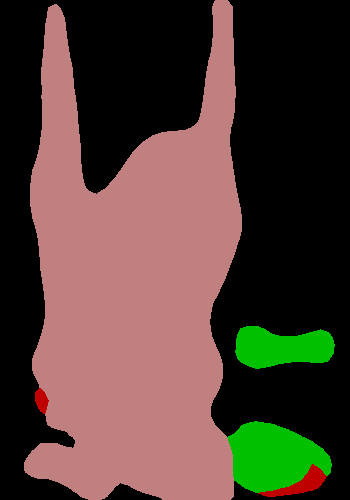} &
    \includegraphics[height=0.11\linewidth]{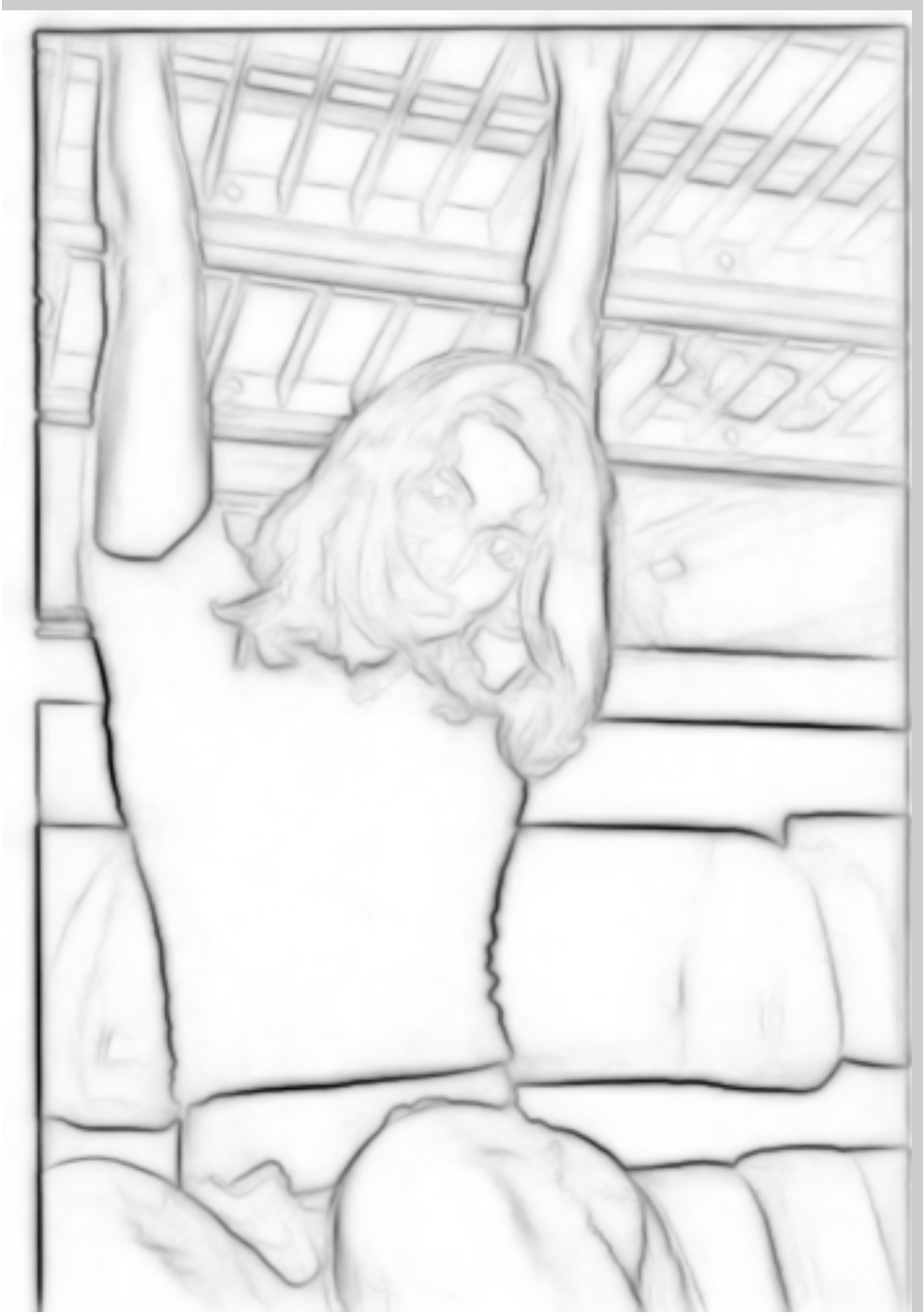} &
    \includegraphics[height=0.11\linewidth]{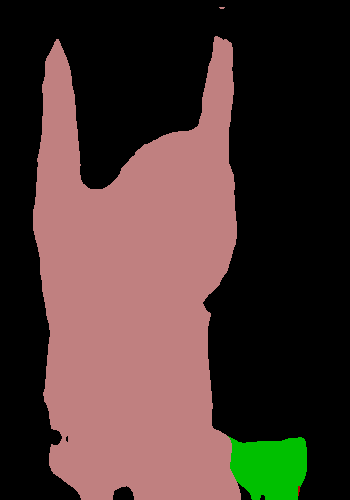} &
    \includegraphics[height=0.11\linewidth]{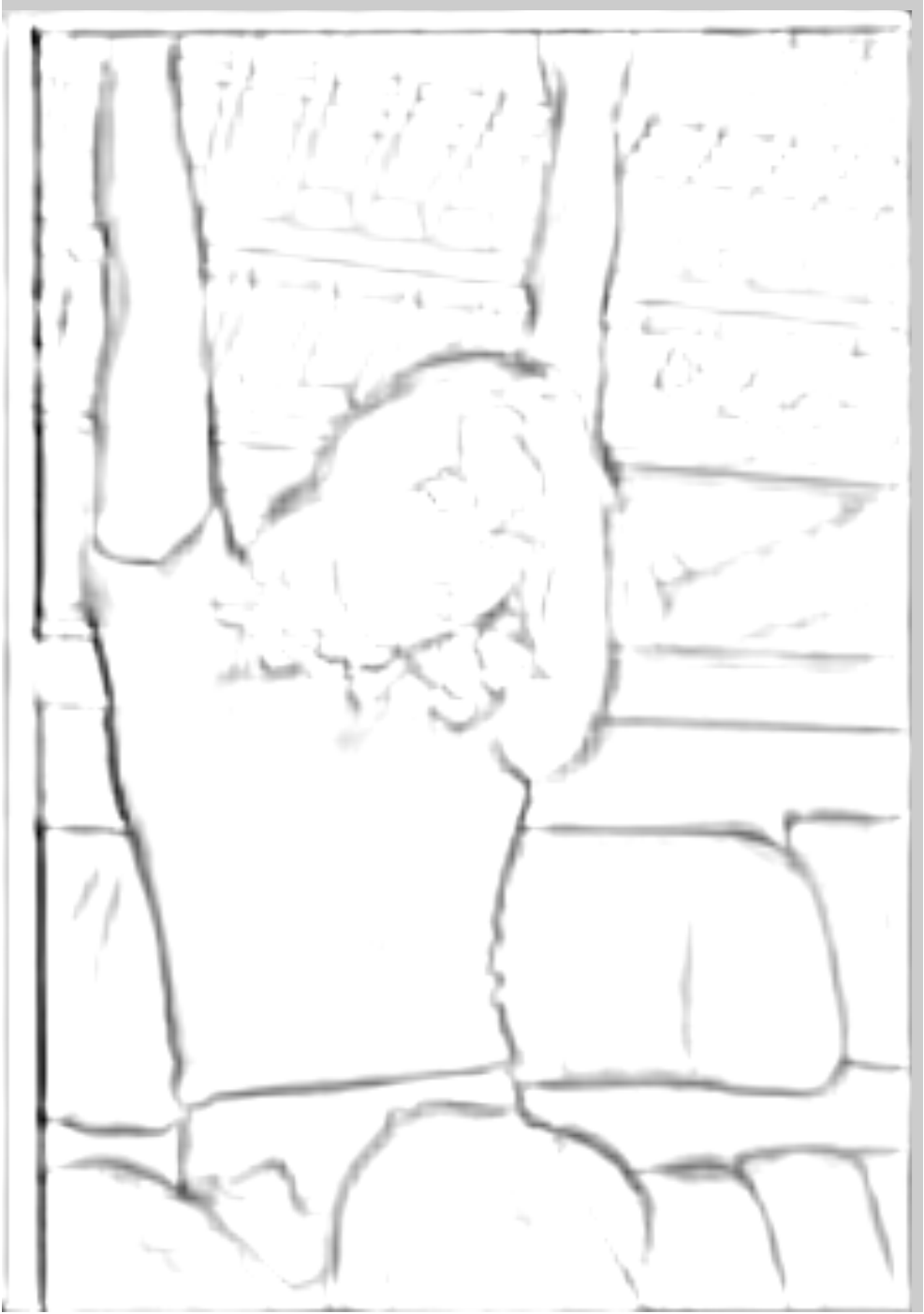} &
    \includegraphics[height=0.11\linewidth]{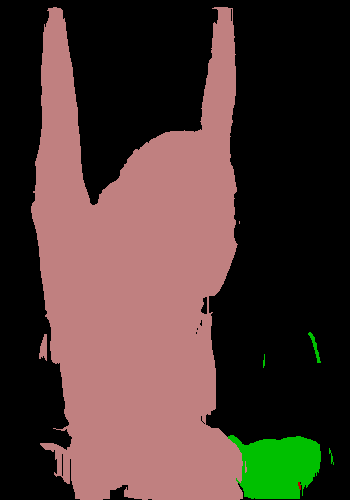} \\

    \includegraphics[height=0.11\linewidth]{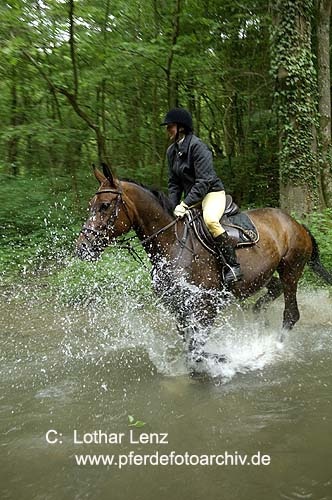} &
    \includegraphics[height=0.11\linewidth]{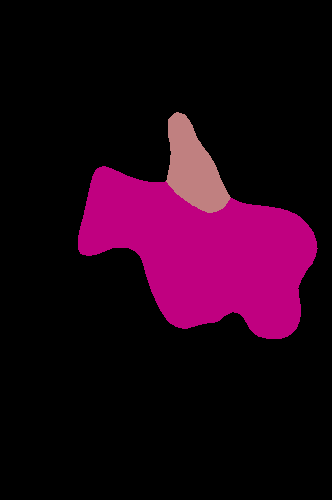} &
    \includegraphics[height=0.11\linewidth]{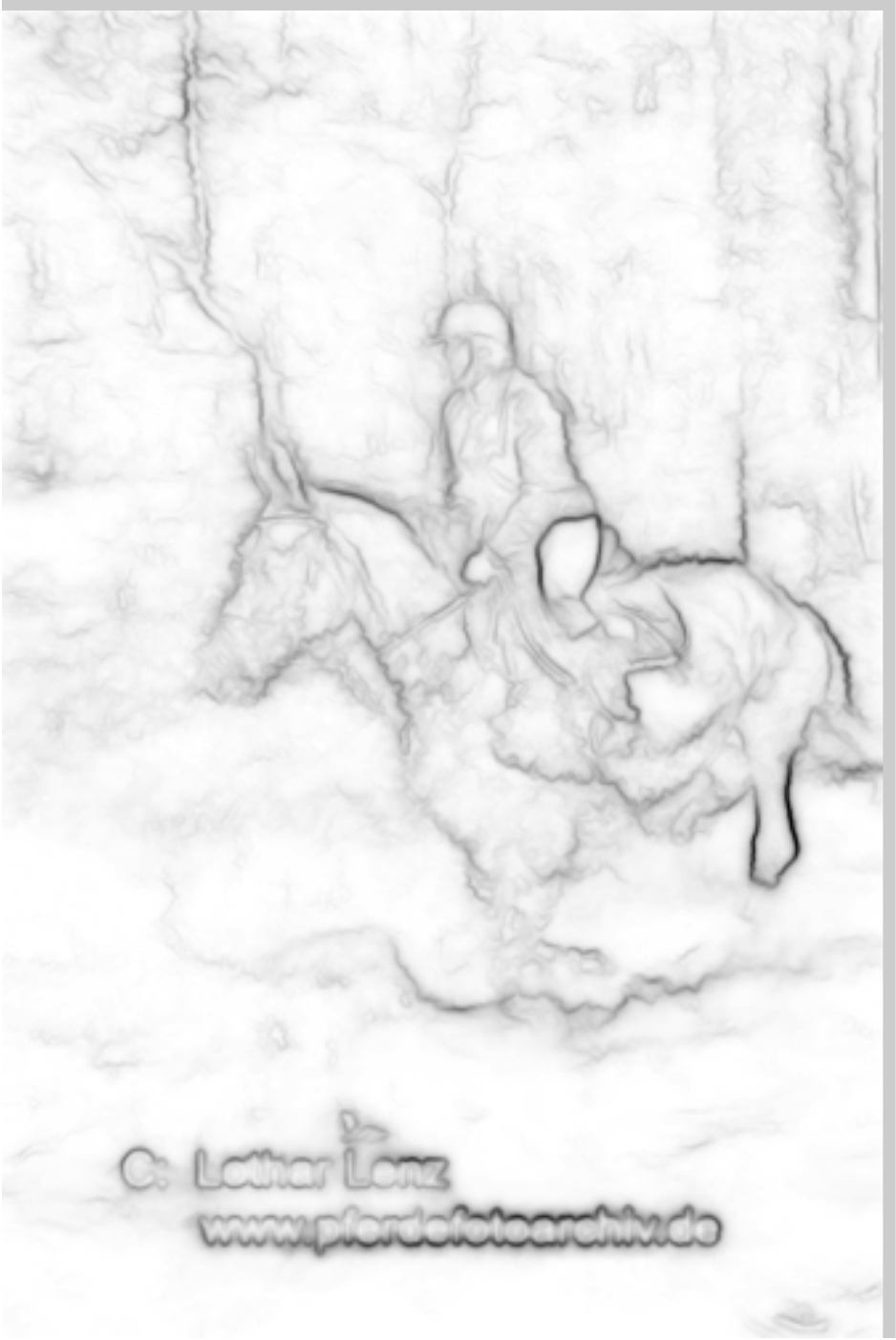} &
    \includegraphics[height=0.11\linewidth]{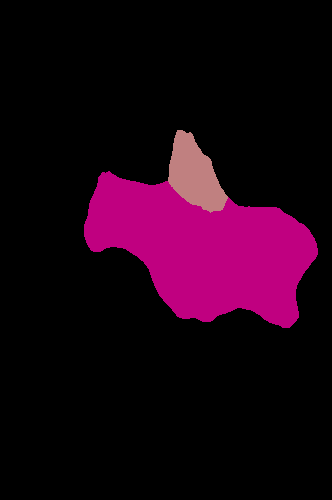} &
    \includegraphics[height=0.11\linewidth]{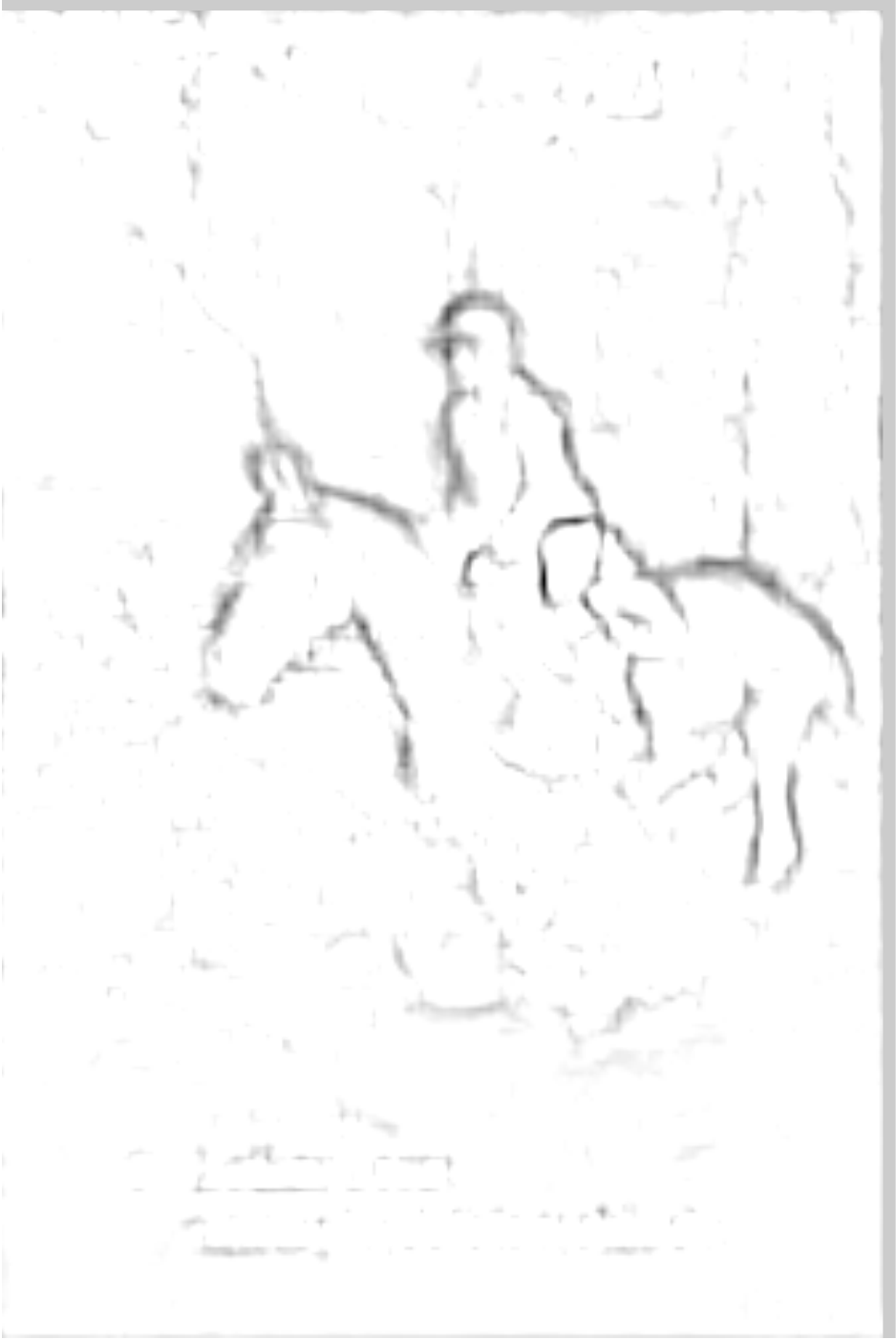} &
    \includegraphics[height=0.11\linewidth]{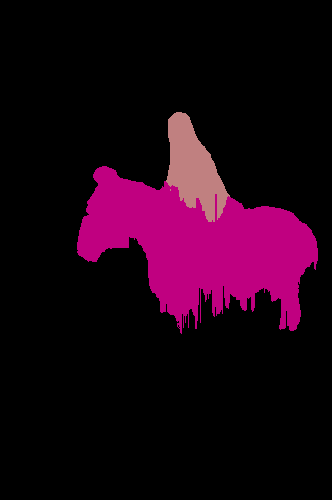} \\

    \includegraphics[width=0.14\linewidth]{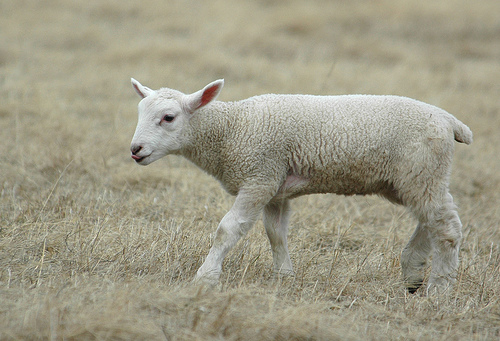} &
    \includegraphics[width=0.14\linewidth]{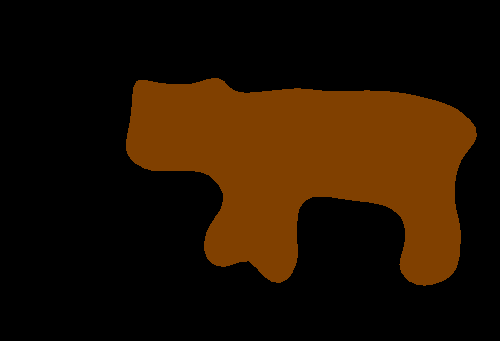} &
    \includegraphics[width=0.14\linewidth]{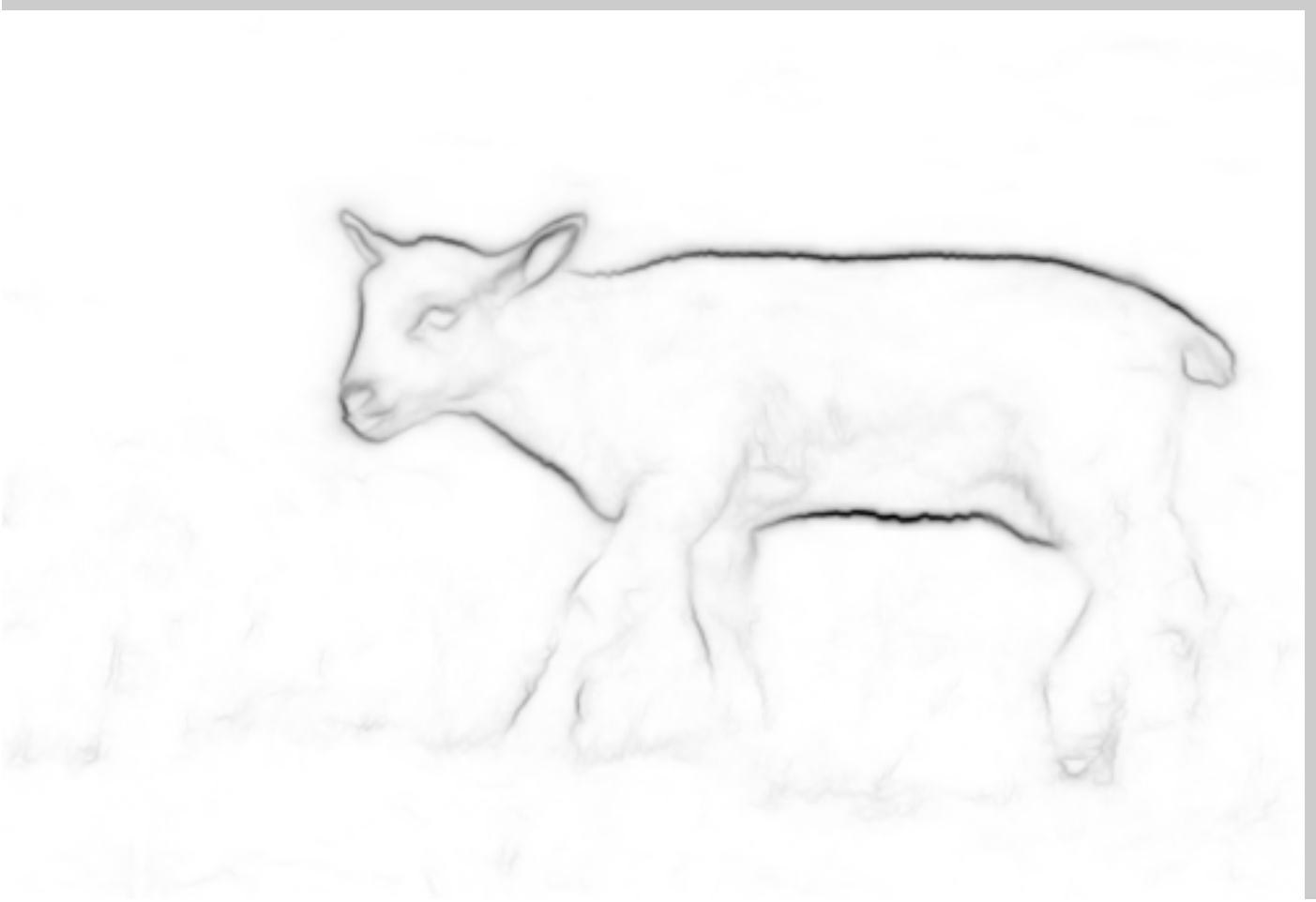} &
    \includegraphics[width=0.14\linewidth]{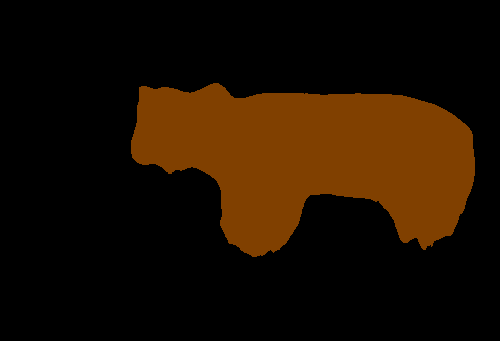} &
    \includegraphics[width=0.14\linewidth]{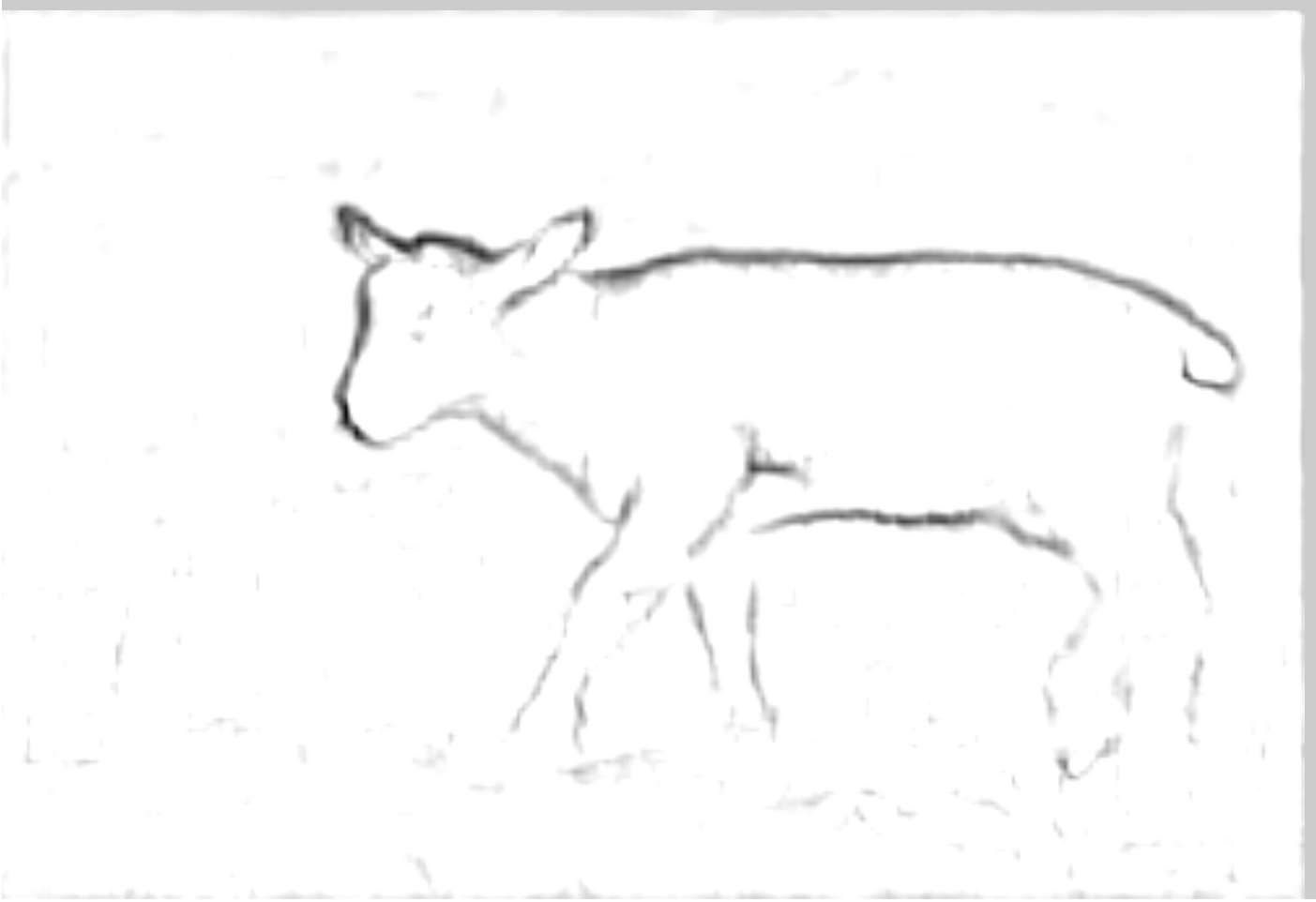} &
    \includegraphics[width=0.14\linewidth]{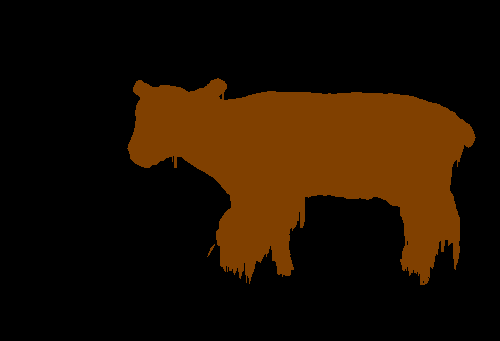} \\

    \includegraphics[width=0.14\linewidth]{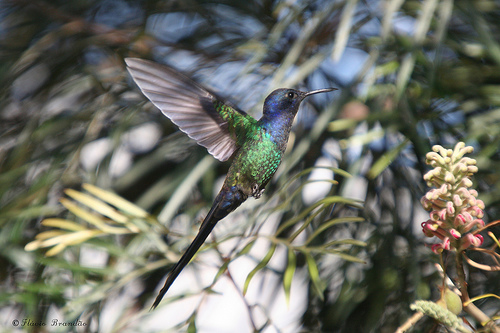} &
    \includegraphics[width=0.14\linewidth]{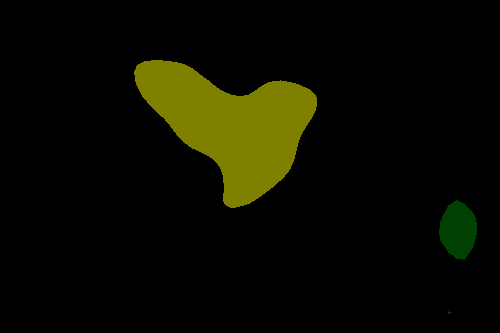} &
    \includegraphics[width=0.14\linewidth]{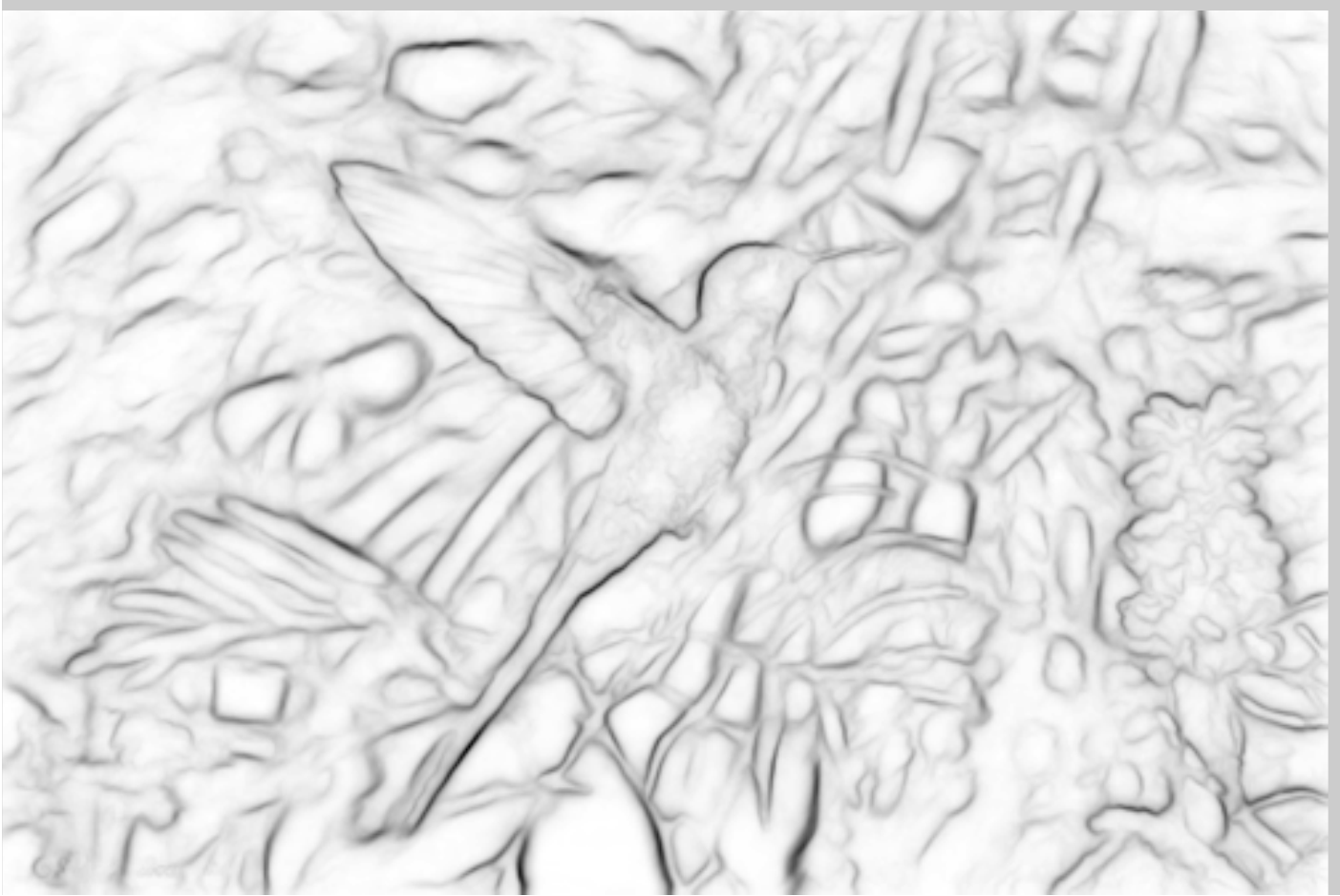} &
    \includegraphics[width=0.14\linewidth]{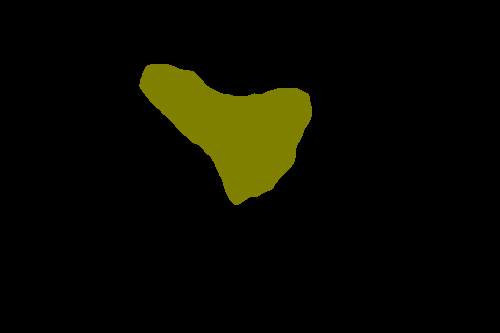} &
    \includegraphics[width=0.14\linewidth]{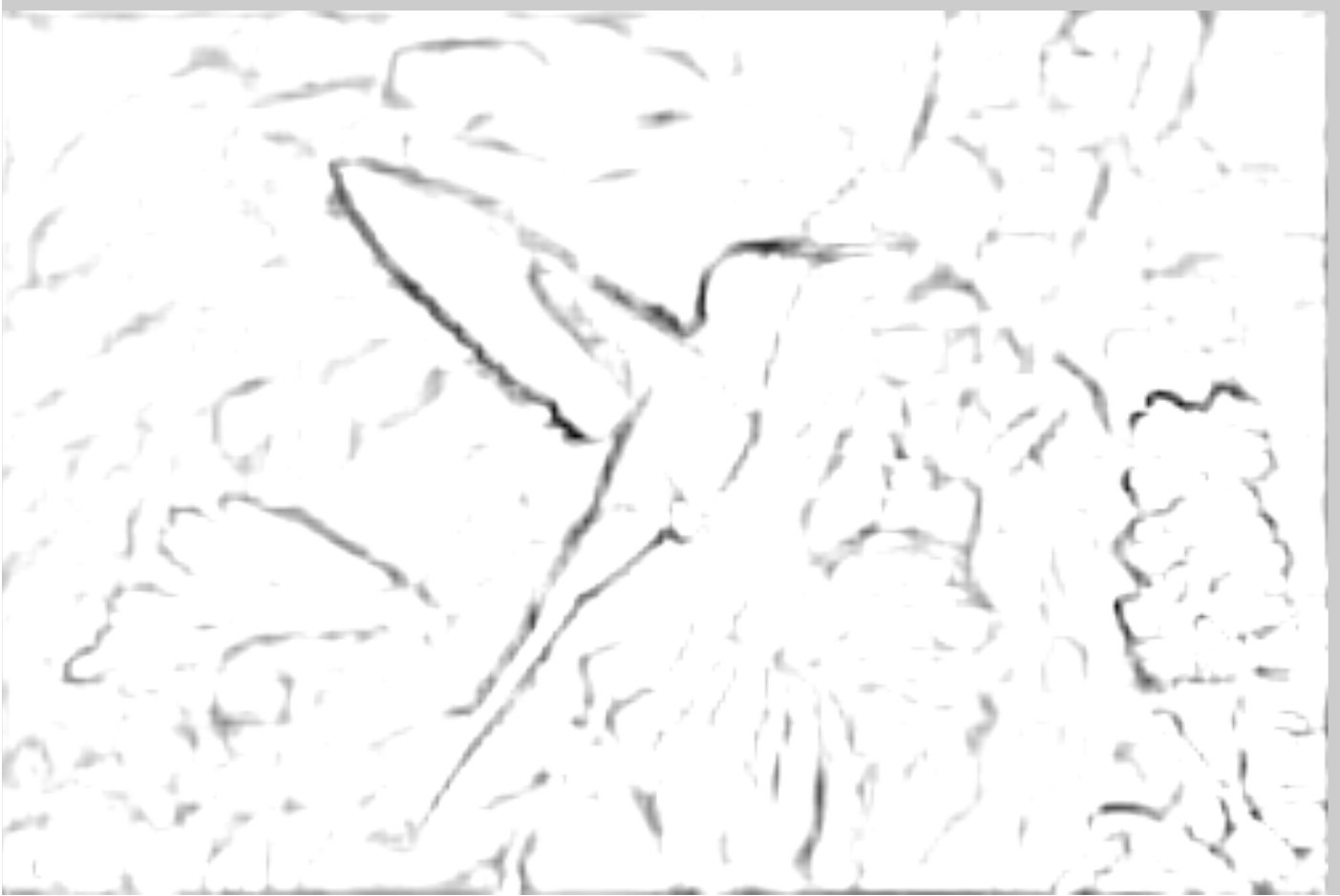} &
    \includegraphics[width=0.14\linewidth]{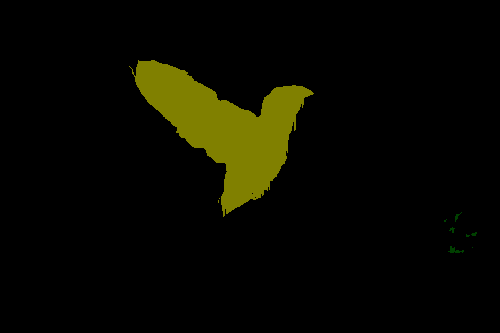} \\

    \includegraphics[width=0.14\linewidth]{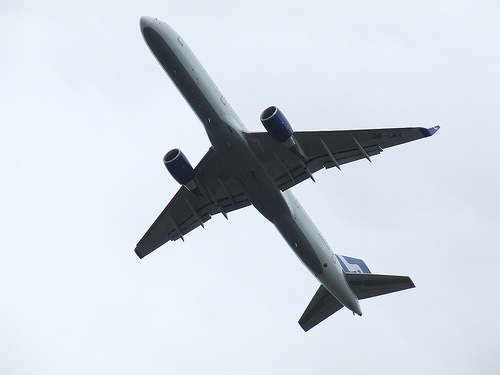} &
    \includegraphics[width=0.14\linewidth]{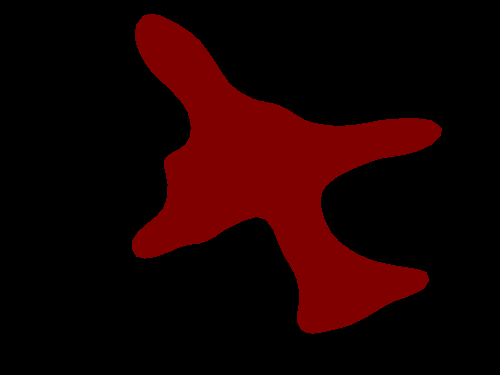} &
    \includegraphics[width=0.14\linewidth]{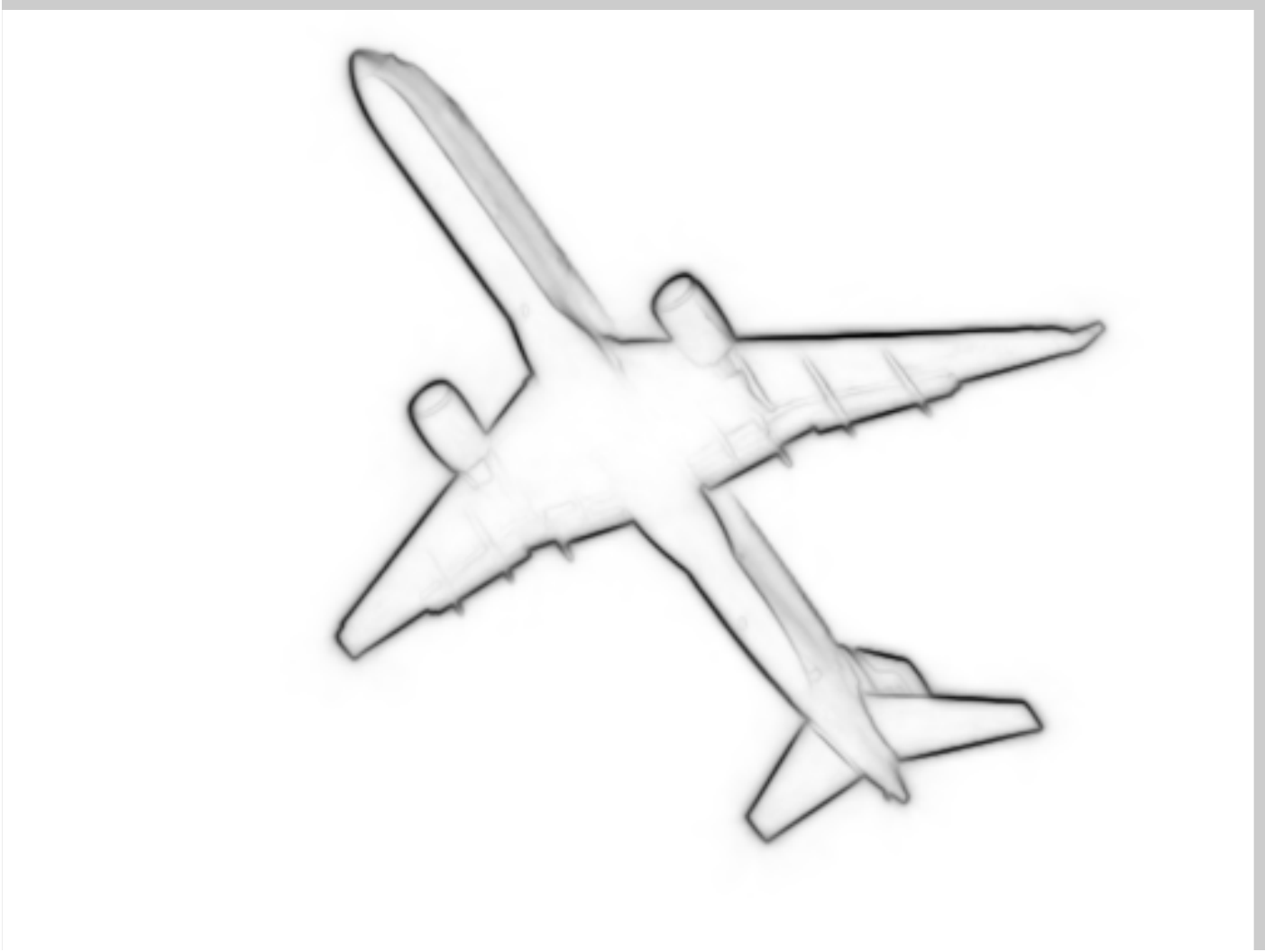} &
    \includegraphics[width=0.14\linewidth]{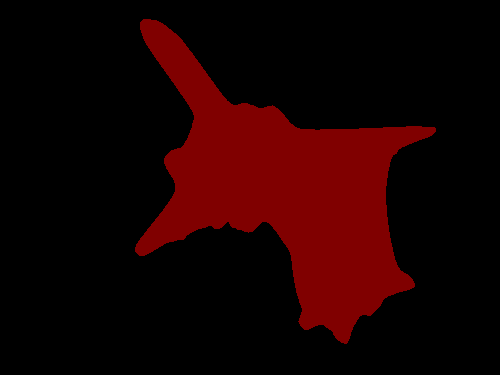} &
    \includegraphics[width=0.14\linewidth]{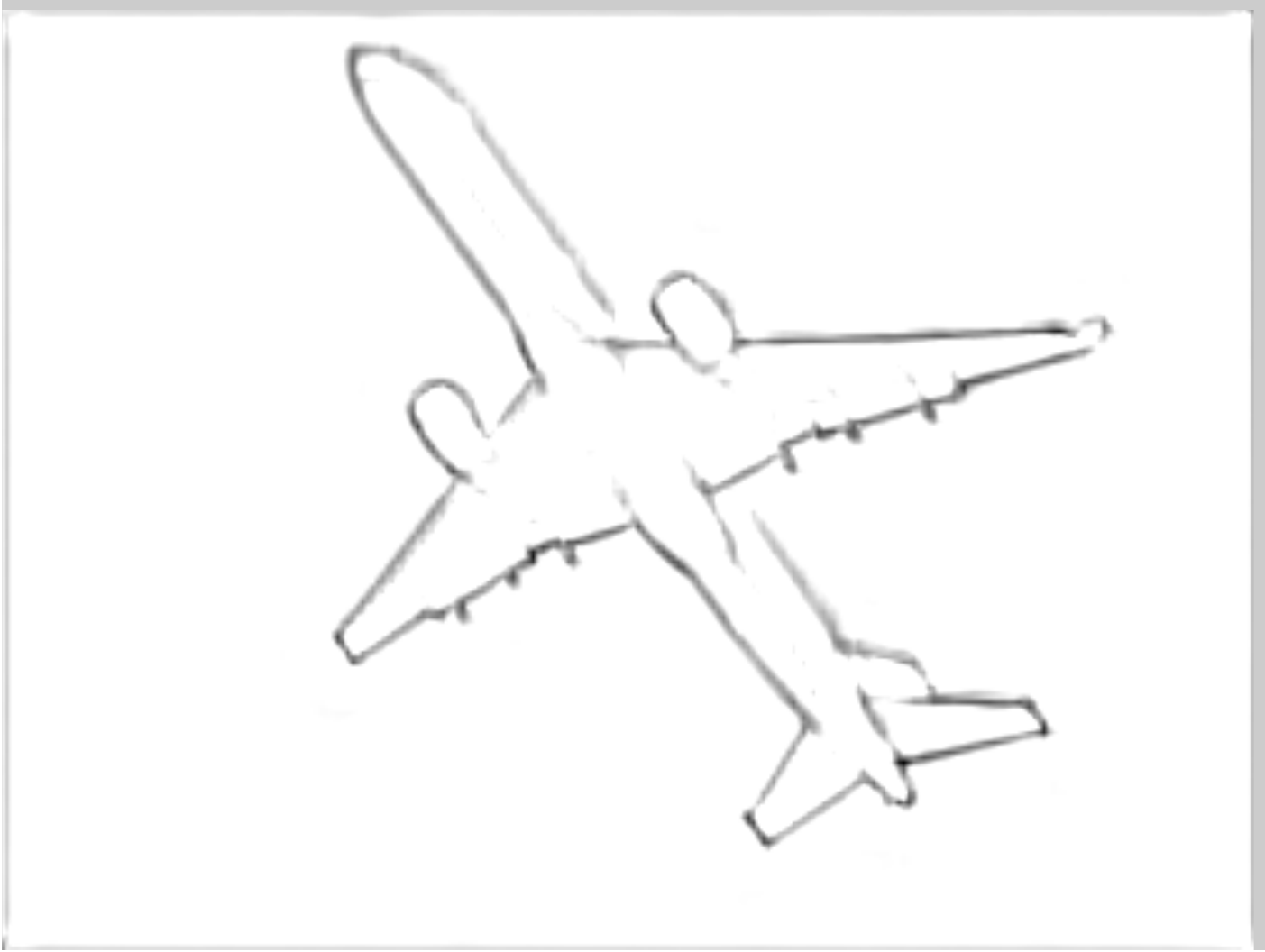} &
    \includegraphics[width=0.14\linewidth]{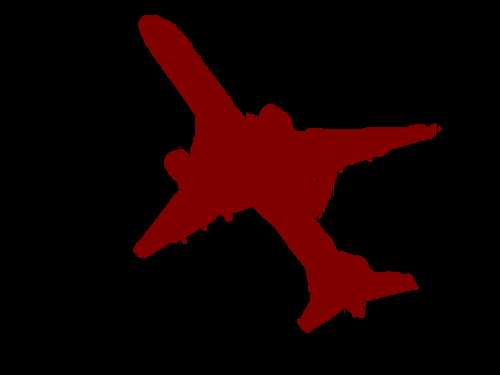} \\

    \includegraphics[width=0.14\linewidth]{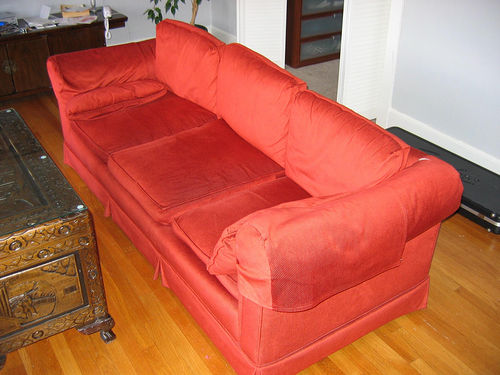} &
    \includegraphics[width=0.14\linewidth]{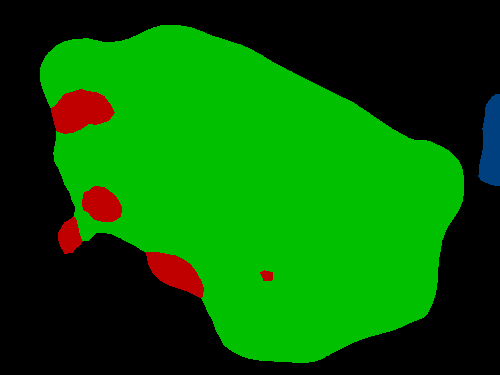} &
    \includegraphics[width=0.14\linewidth]{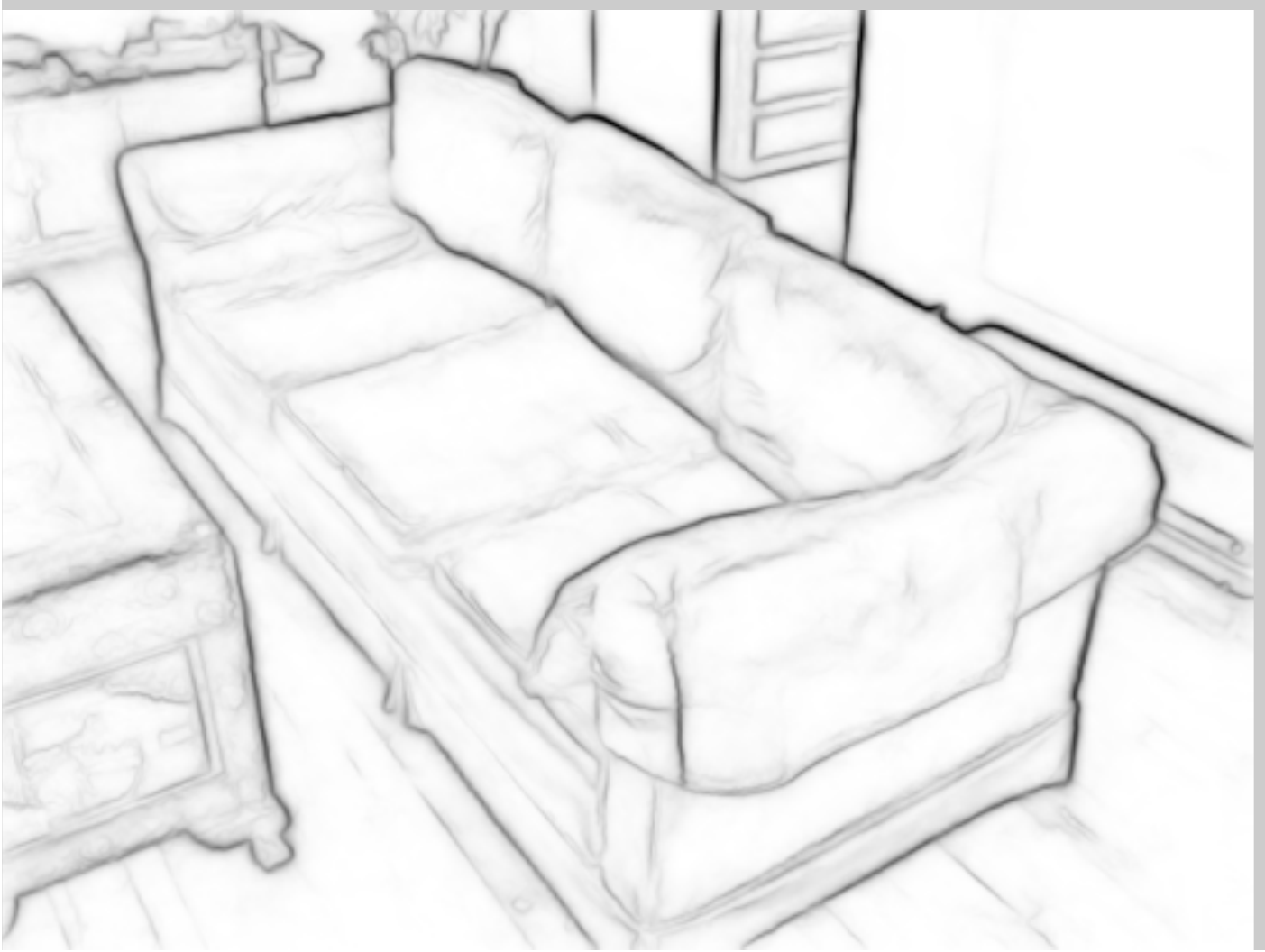} &
    \includegraphics[width=0.14\linewidth]{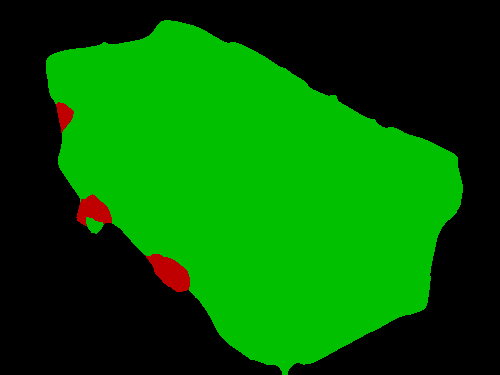} &
    \includegraphics[width=0.14\linewidth]{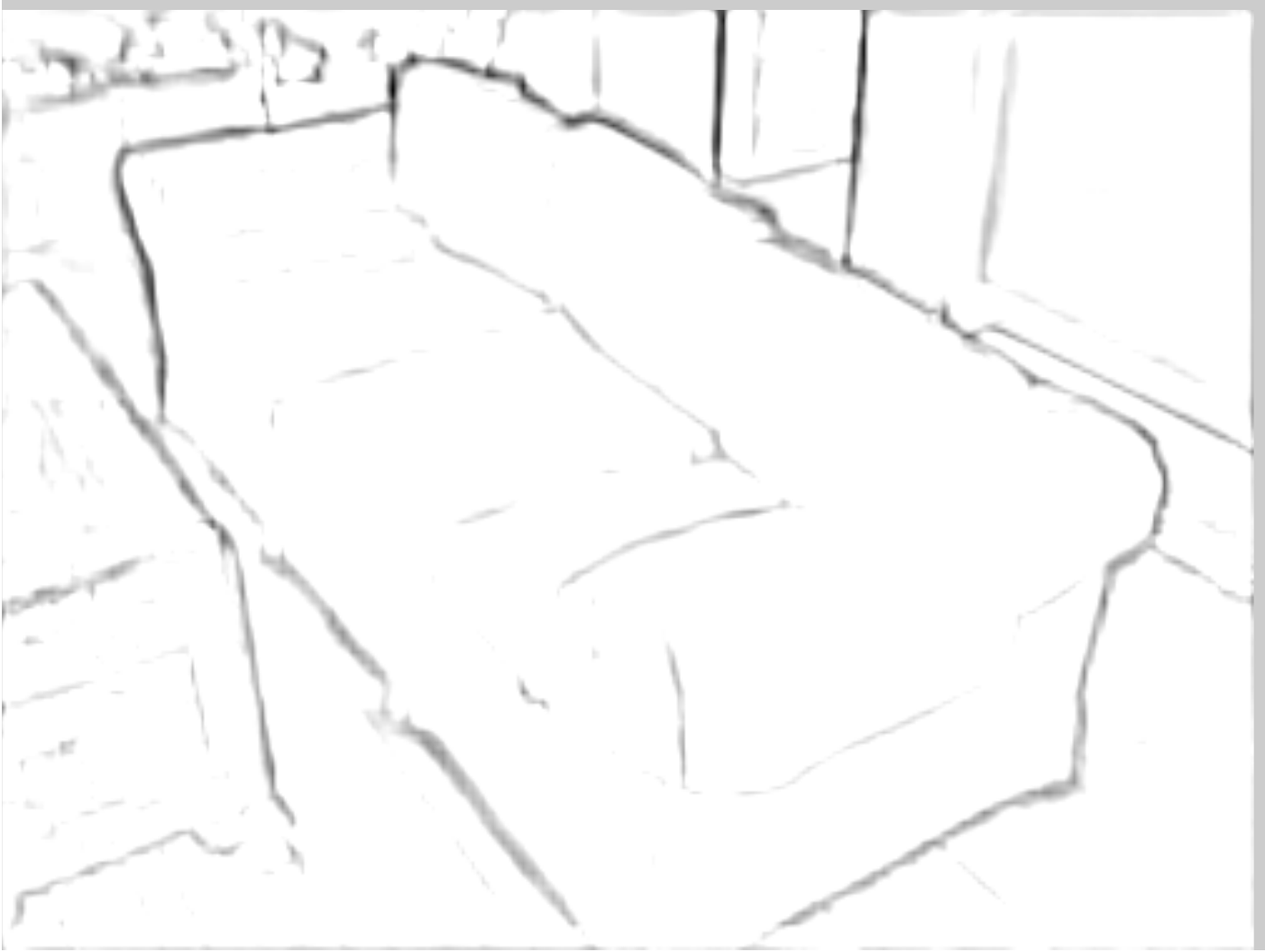} &
    \includegraphics[width=0.14\linewidth]{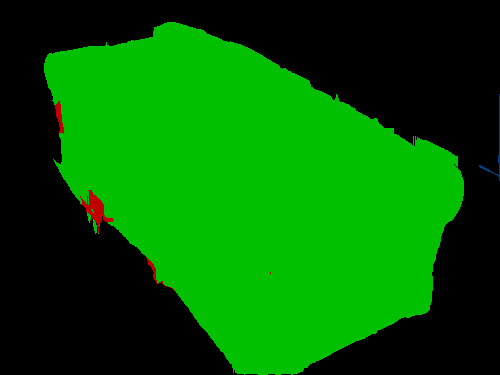} \\

    (a) Image &
    (b) Baseline &
    (c) SE &
    (d) DT-SE &
    (e) EdgeNet &
    (f) DT-EdgeNet \\
  \end{tabular}
  \caption{Visualizing results on VOC 2012 val set. Continued from
    \figref{fig:pascal_voc12_seg_1}.}
  \label{fig:pascal_voc12_seg_2}
\end{figure*}

We show additional edge detection and semantic segmentation results on
PASCAL VOC 2012 val set in Figs.~\ref{fig:pascal_voc12_seg_1}
and~\ref{fig:pascal_voc12_seg_2}. We compare results obtained with the
proposed domain transform when using our learned EdgeNet edges \vs the
SE edges of \cite{dollar2013structured}.

\clearpage
\newpage

\bibliographystyle{ieee}
\bibliography{egbib}

\bibliographystyle{ieee}
\bibliography{egbib}

\end{document}